\documentclass[10pt,twocolumn,letterpaper]{article}

\usepackage{cvpr}      % To produce the REVIEW version
\usepackage{graphicx}
\usepackage{amsmath}
\usepackage{amssymb}
\usepackage{booktabs}
\usepackage{float}
\usepackage{diagbox}
\usepackage{multirow}
\usepackage{enumitem}
\usepackage{tabularx}
\usepackage{microtype}
\usepackage{soul}

\makeatletter
\@namedef{ver@everyshi.sty}{}
\makeatother
\usepackage{tikz}
\usepackage{makecell}
\usepackage[pagebackref=true,breaklinks=true,letterpaper=true,colorlinks,bookmarks=false]{hyperref}

\usepackage[capitalize]{cleveref}
\crefname{section}{Sec.}{Secs.}
\Crefname{section}{Section}{Sections}
\Crefname{table}{Table}{Tables}
\crefname{table}{Tab.}{Tabs.}

\newcommand{\ignore}[1]{}

\newcommand{\paraheading}[1]{\vspace{-1.1em}\paragraph{#1.}}
\newcommand{\topparaheading}[1]{\vspace{-0.2em}\paragraph{#1.}}
\newcommand{\capvspace}{\vspace{-2mm}}
\newcommand{\figbottomvspace}{\vspace{-3.5mm}}
\newcommand{\tabcapvspace}{\vspace{-4.5mm}}
\newcommand{\tabbottomvspace}{\vspace{-4.5mm}}
\newcommand{\subcapfont}[1]{\vspace{-3.5mm}\\{\fontsize{6.5}{7.8}\selectfont #1}}

\def\etal{\textit{et al}.}
\def\ie{\textit{i.e.}}
\def\eg{\textit{e.g.}}

\DeclareMathOperator*{\cat}{Cat}

\DeclareMathOperator*{\hm}{HM}
\DeclareMathOperator*{\dec}{Dec}

\newcommand{\str}{g}
\newcommand{\apr}{h}

\definecolor{Gray}{gray}{0.85}
\definecolor{White}{gray}{1.0}
\newcolumntype{a}{>{\columncolor{Gray}}c}
\newcolumntype{z}{>{\columncolor{White}}c}

\def\cvprPaperID{3935} % *** Enter the CVPR Paper ID here
\def\confName{CVPR}
\def\confYear{2022}

\begin{document}

\title{High-resolution Face Swapping via Latent Semantics Disentanglement}%\vspace{-2mm}}

% \author{Yangyang Xu$^{1}$,
% Bailin Deng$^{2}$,
% Junle Wang$^{3}$,
% Yanqing Jing$^{3}$,
% Jia Pan$^{4}$,
% Shengfeng He$^{1}$\thanks{Corresponding author (hesfe@scut.edu.cn).}\\
% \normalsize{$^1$ School of Computer Science and Engineering, South China University of Technology}\\
% \normalsize{$^2$  School of Computer Science and Informatics, Cardiff University}\\
% \normalsize{$^3$ Interactive Entertainment Group, Tencent Inc}\\
% \normalsize{$^4$ Department of Computer Science, The University of HongKong}\\
% }

\author{Yangyang Xu$^{1,3}$ \quad
Bailin Deng$^{2}$\quad
Junle Wang$^{3}$\quad
Yanqing Jing$^{3}$\quad
Jia Pan$^{4}$\quad
Shengfeng He$^{1}$\thanks{Corresponding author (hesfe@scut.edu.cn).}\\
% {$^1$ School of Computer Science and Engineering, South China University of Technology}\\
% {$^2$  School of Computer Science and Informatics, Cardiff University}\\
% {$^3$ Interactive Entertainment Group, Tencent Inc}\\
% {$^4$ Department of Computer Science, The University of HongKong}\\
{$^1$South China University of Technology} \quad {$^2$Cardiff University}\\
{$^3$Tencent} \quad {$^4$The University of Hong Kong}
}

\teaser{%\vspace{5mm}
    \centering
    \begin{subfigure}{.14\linewidth}
        \centering
        \includegraphics[width=\linewidth]{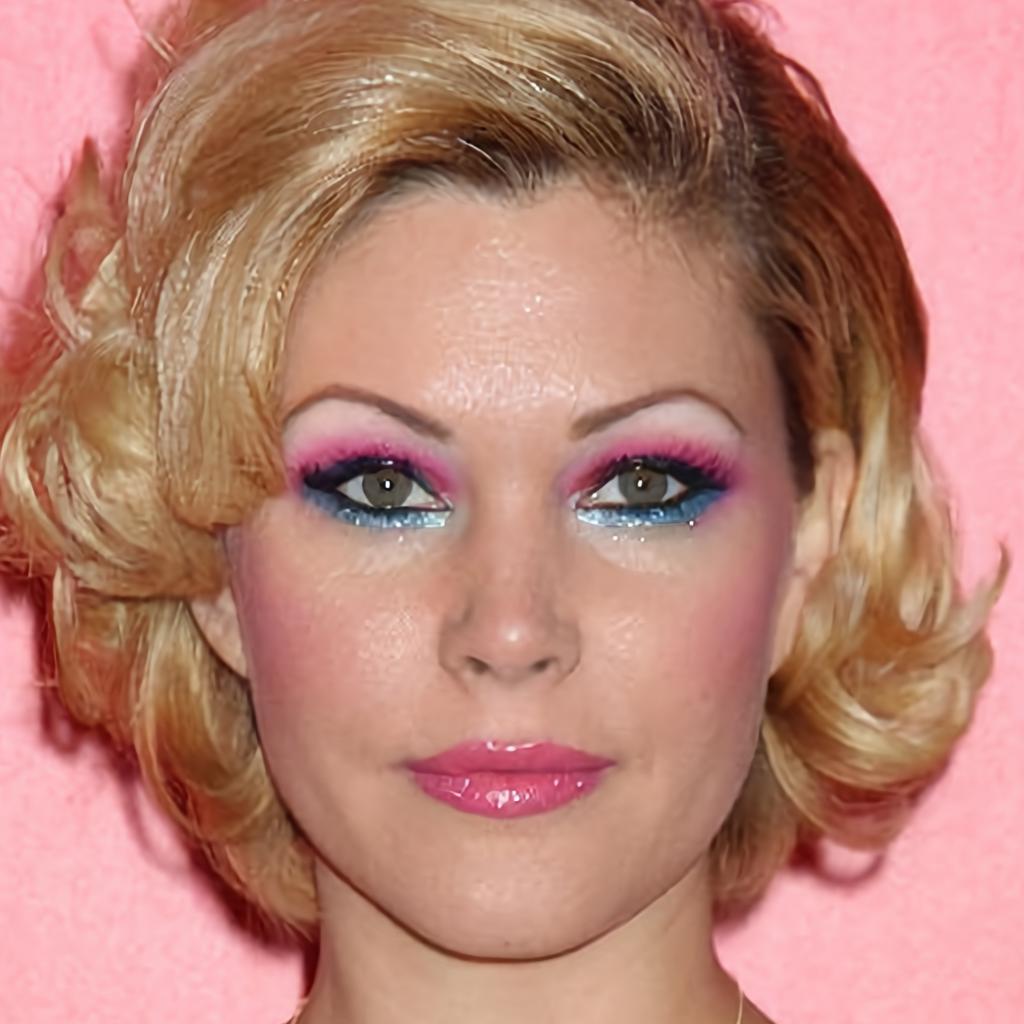}
        \includegraphics[width=\linewidth]{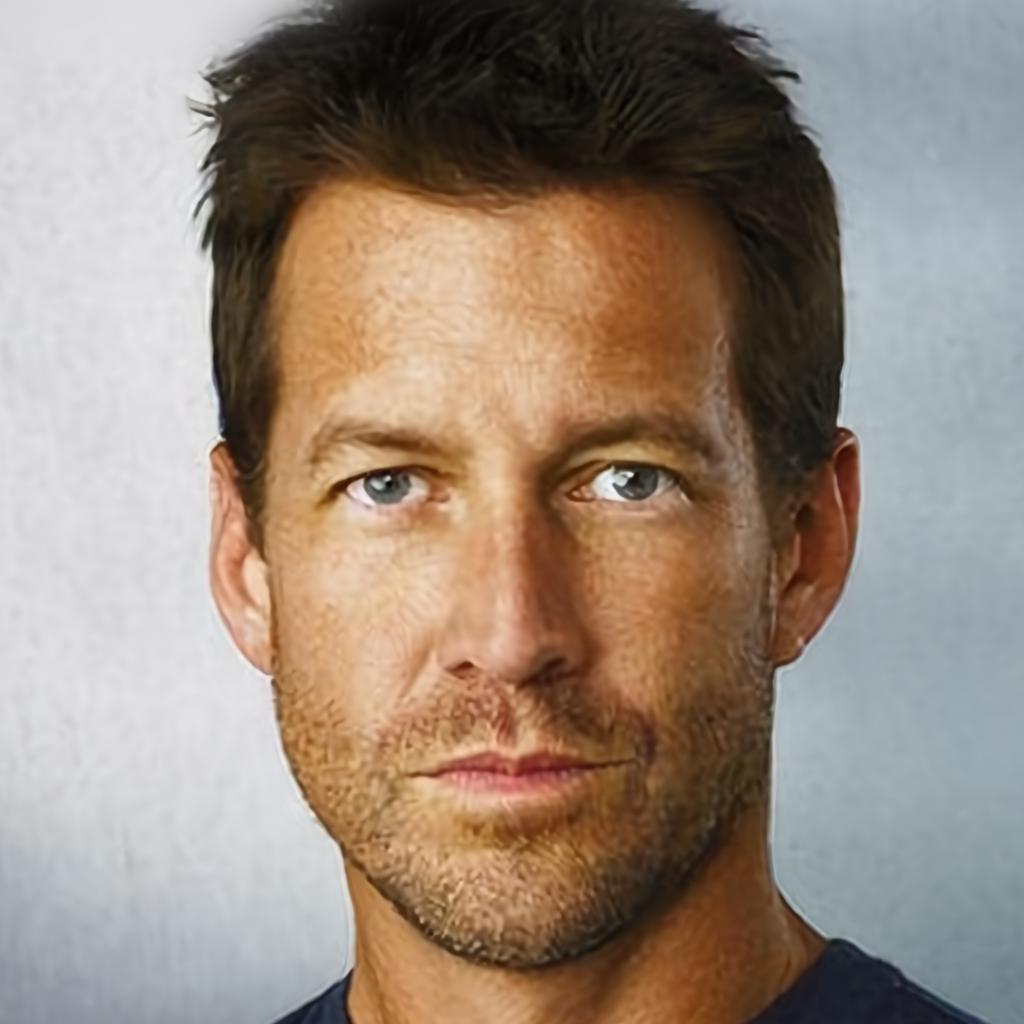}
        \caption{Target}\label{fig:teaser_a}
    \end{subfigure}
    \hspace{-1.5mm}
    \begin{subfigure}{.14\linewidth}
        \centering
        \includegraphics[width=\linewidth]{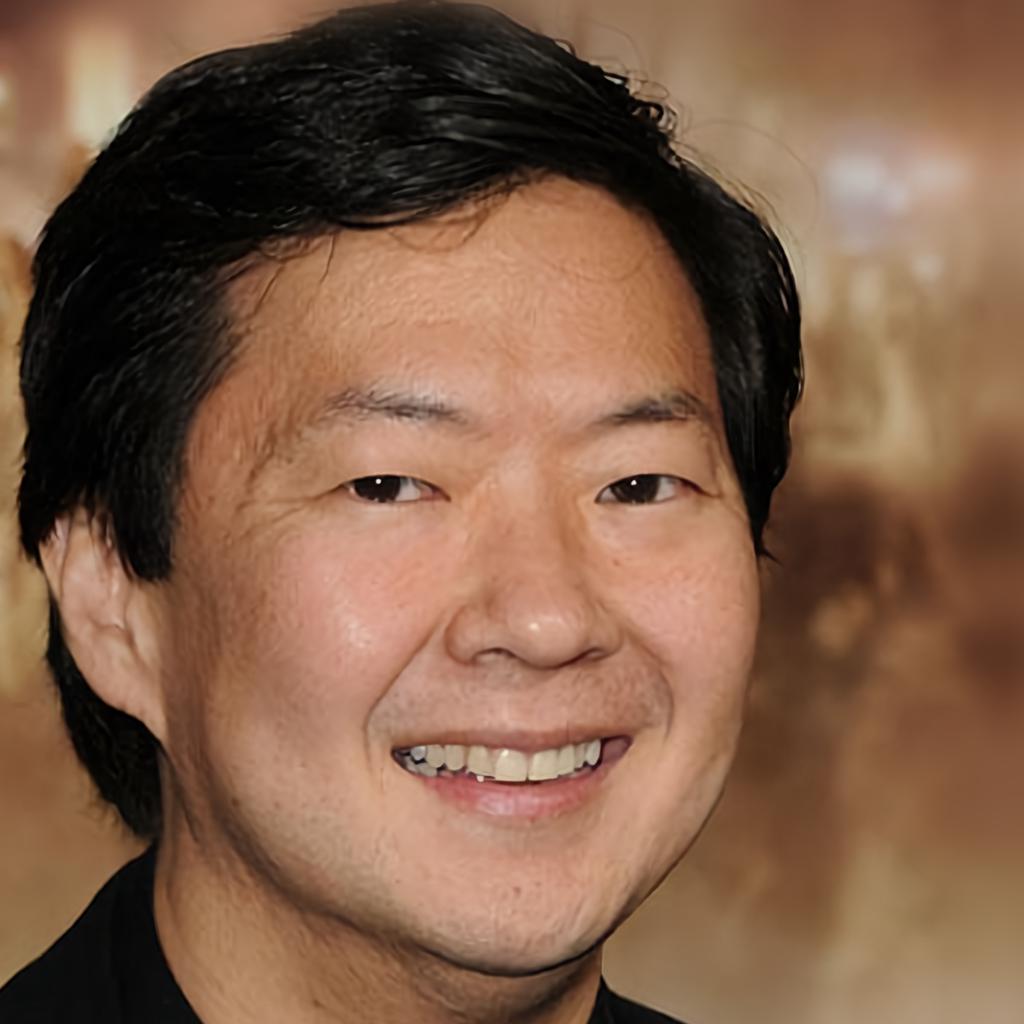}
        \includegraphics[width=\linewidth]{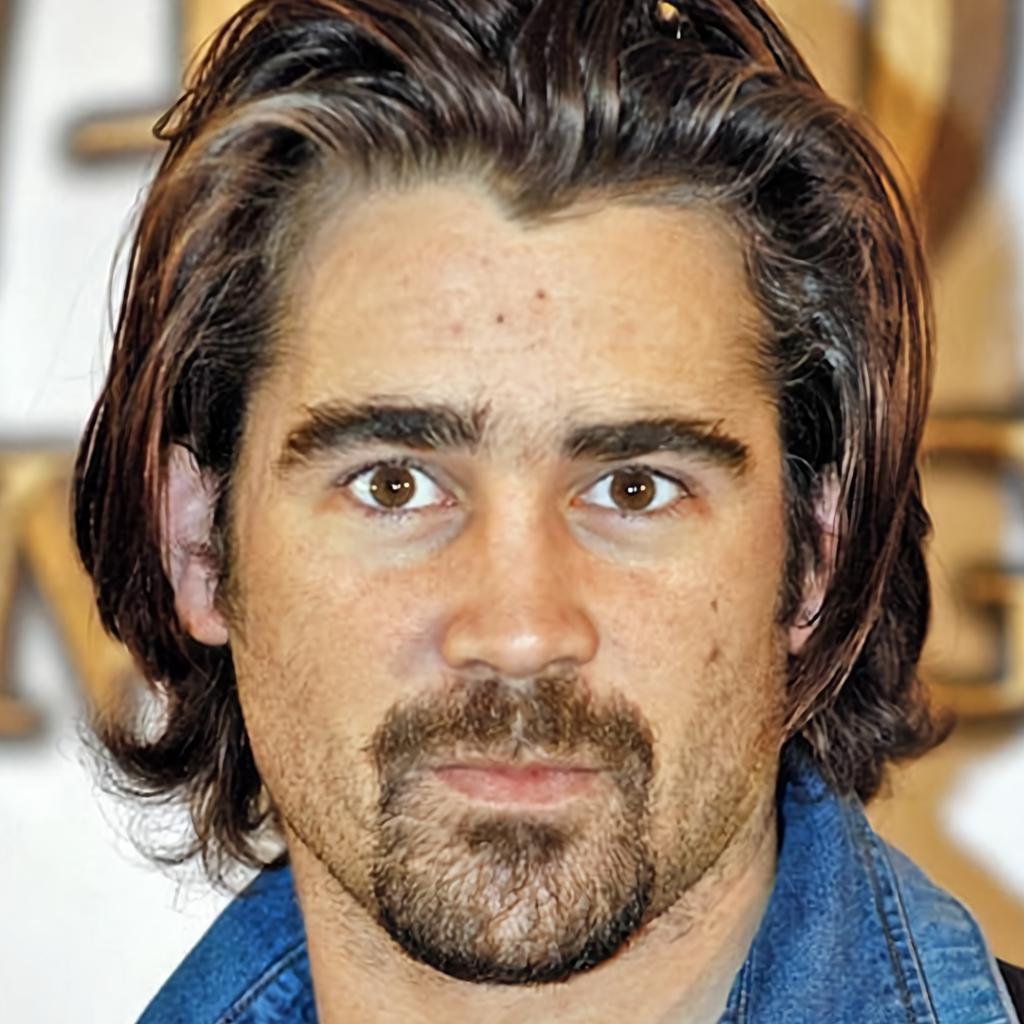}
        \caption{Source 1}\label{fig:teaser_b}
    \end{subfigure}
    \hspace{-1.5mm}
    \begin{subfigure}{.14\linewidth}
        \centering
        \includegraphics[width=\linewidth]{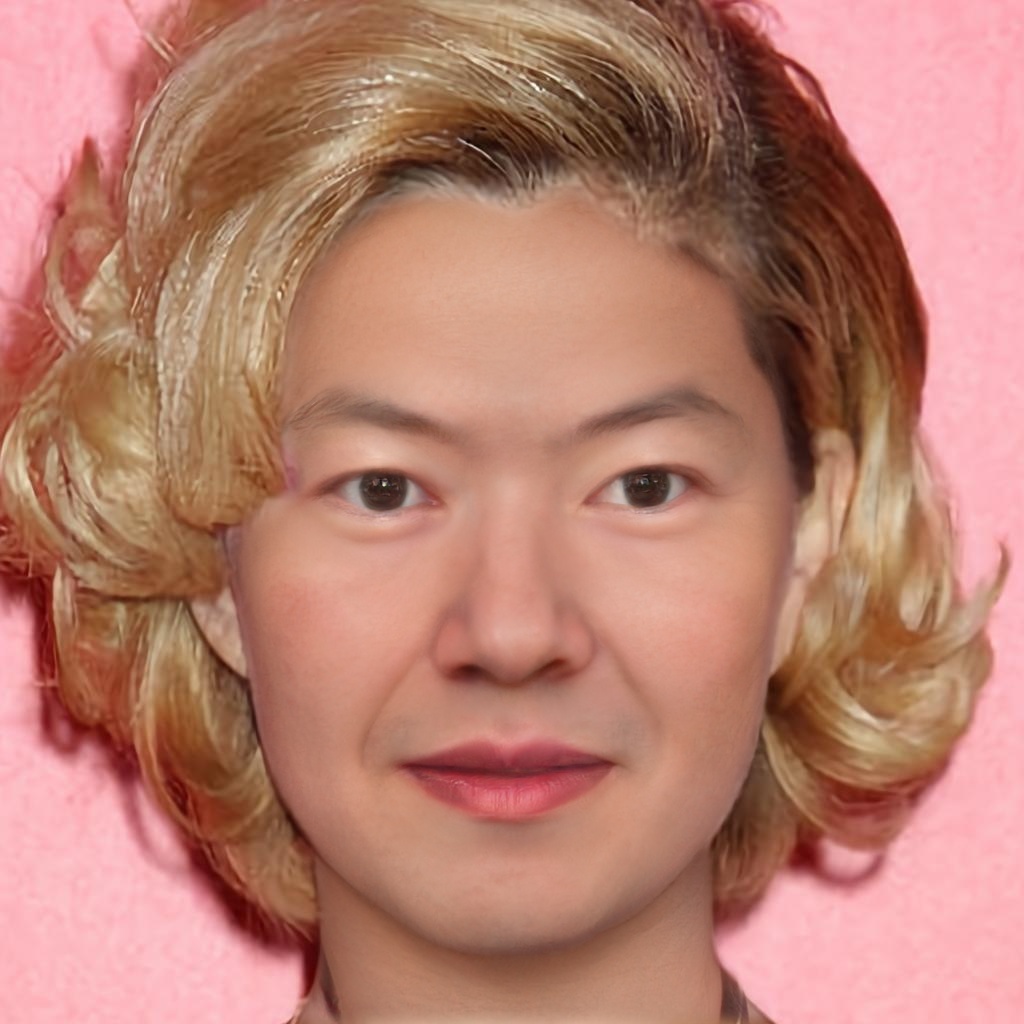}
        \includegraphics[width=\linewidth]{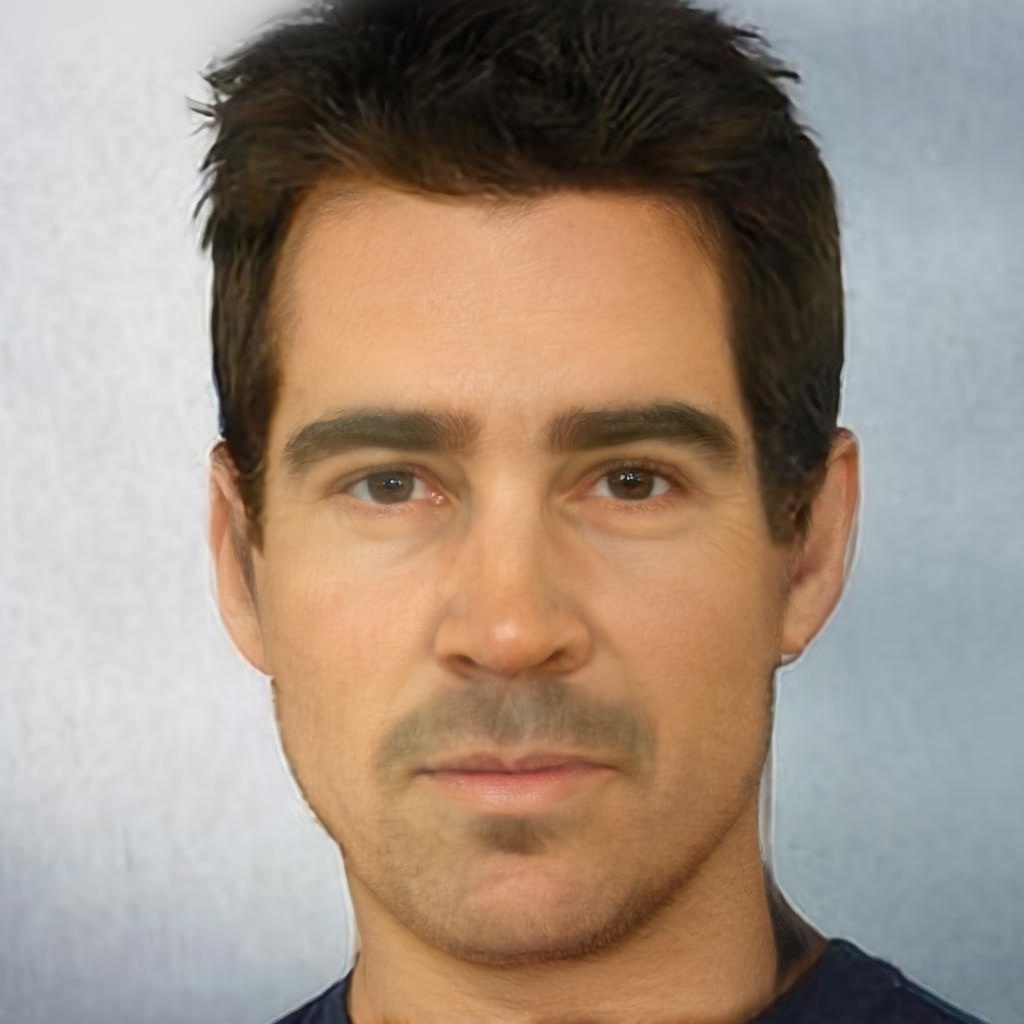}
        \caption{MegaFS~\cite{zhu2021one}}\label{fig:teaser_c}
    \end{subfigure}
    \hspace{-1.5mm}
    \begin{subfigure}{.14\linewidth}
        \centering
        \includegraphics[width=\linewidth]{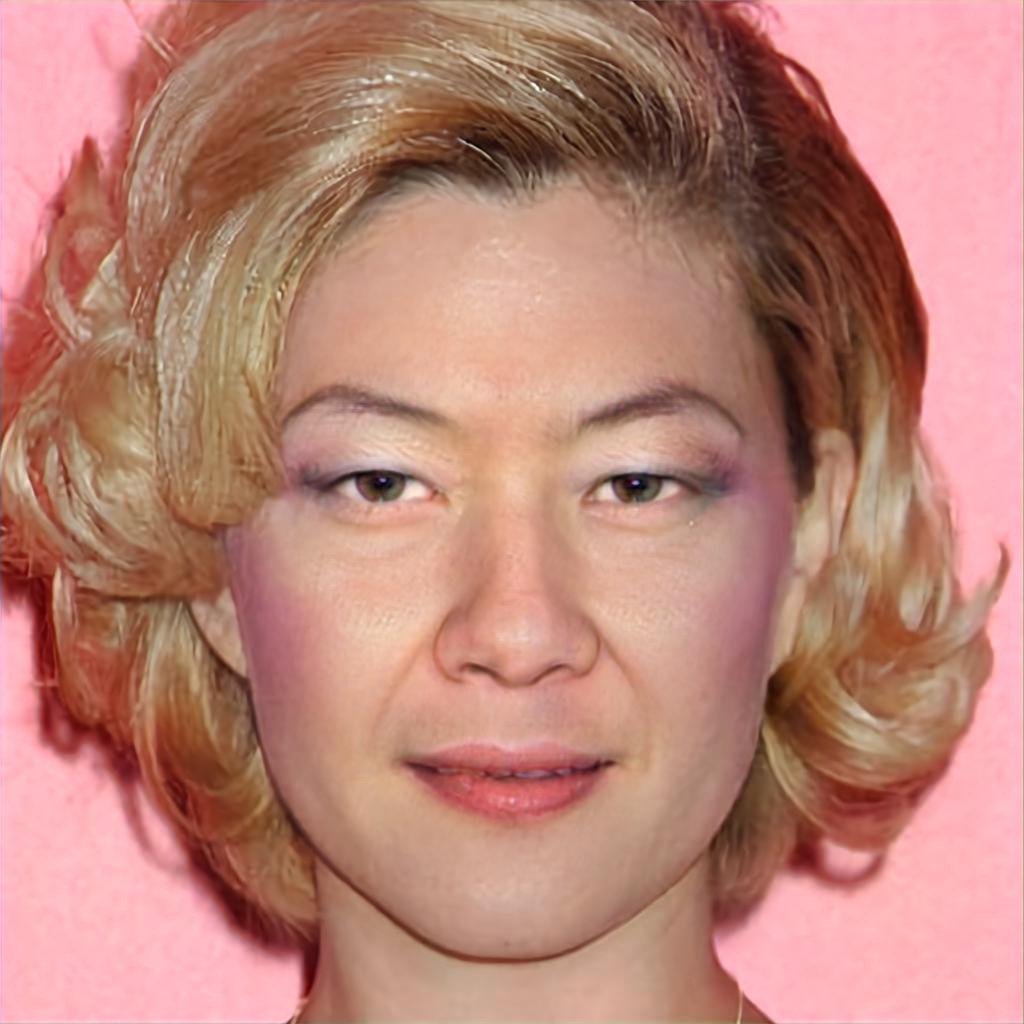}
        \includegraphics[width=\linewidth]{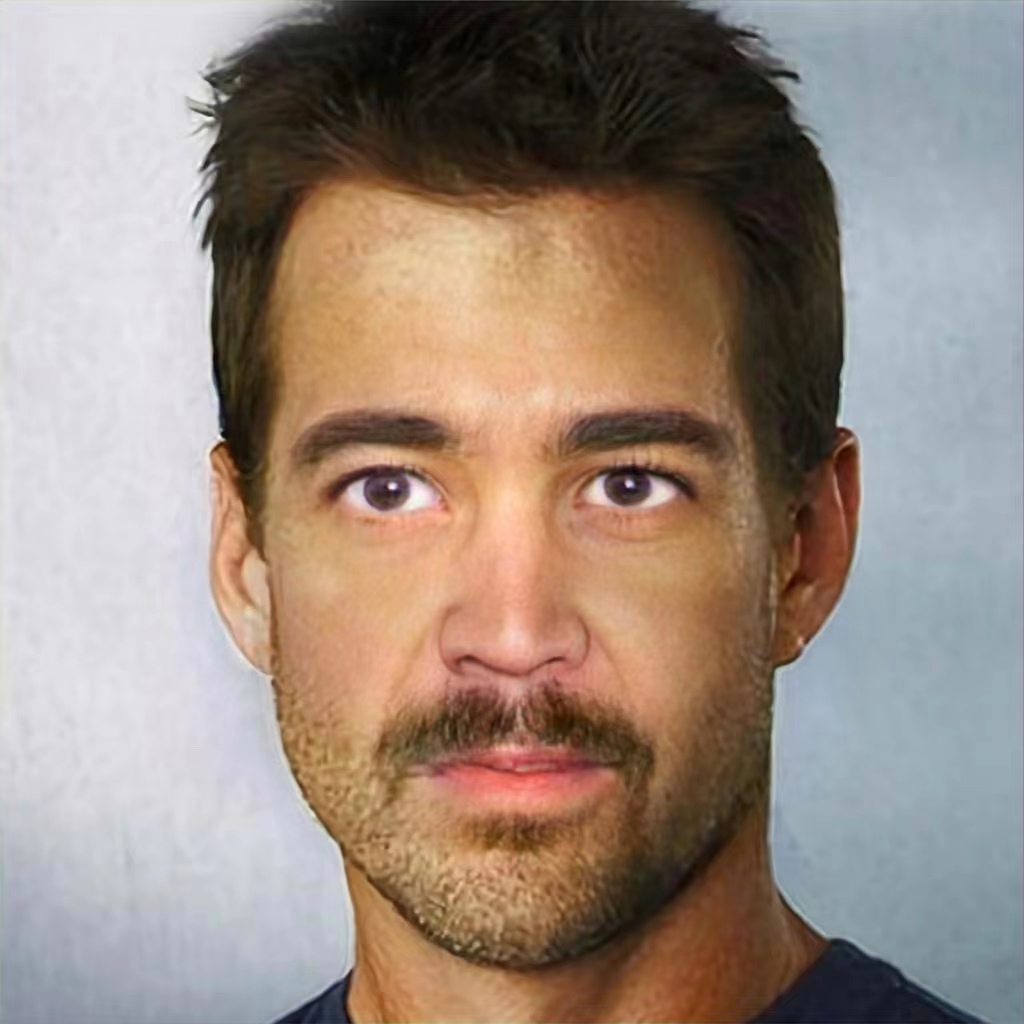}
        \caption{Ours}\label{fig:teaser_d}
    \end{subfigure}
    \hspace{-1.5mm}
    \begin{subfigure}{.14\linewidth}
        \centering
        \includegraphics[width=\linewidth]{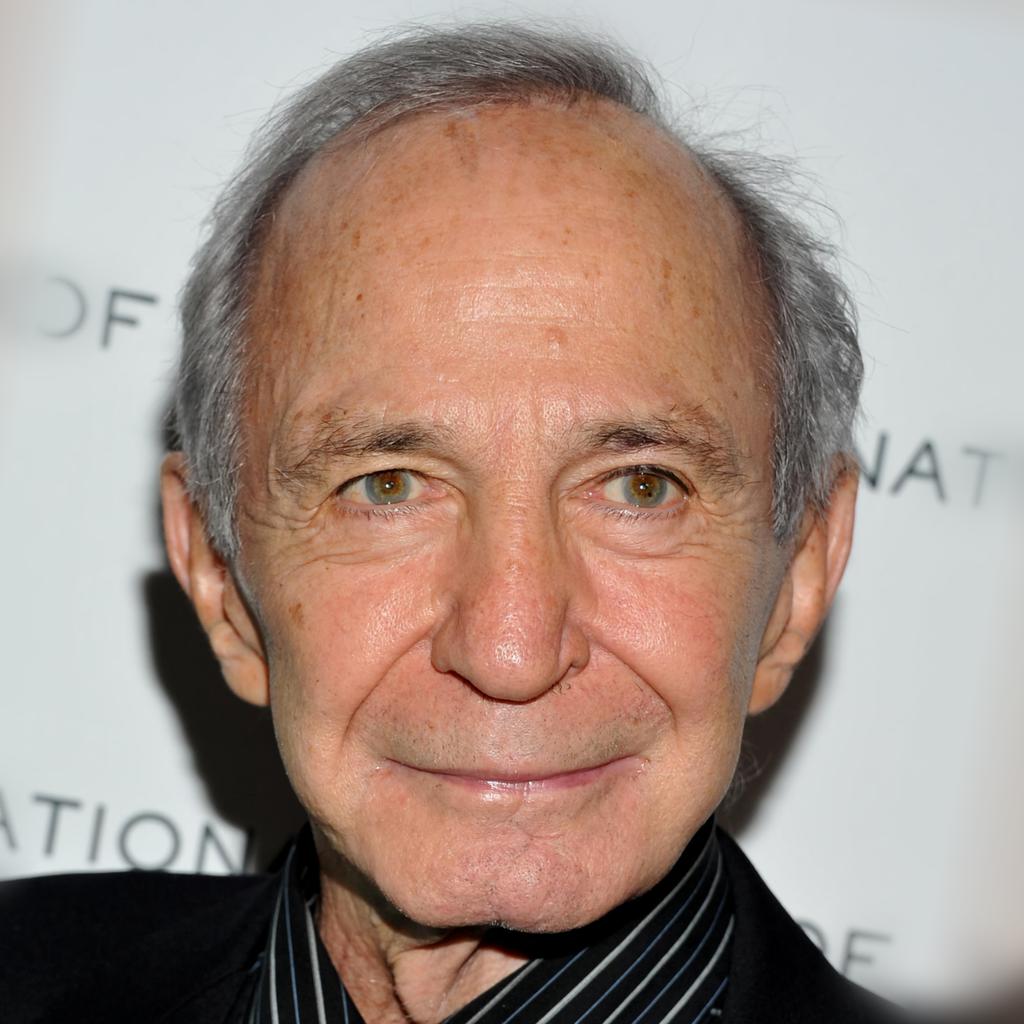}
        \includegraphics[width=\linewidth]{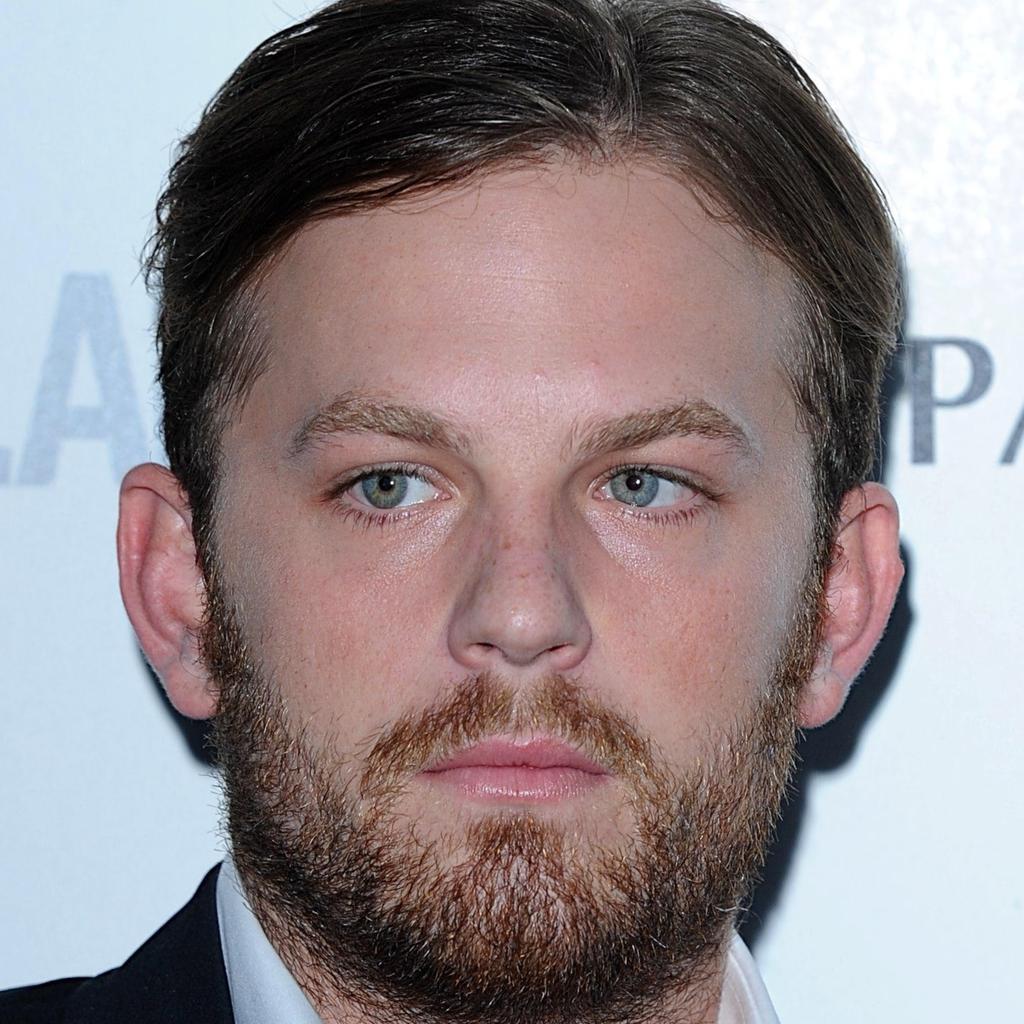}
        \caption{Source 2}\label{fig:teaser_e}
    \end{subfigure}
    \hspace{-1.5mm}
    \begin{subfigure}{.14\linewidth}
        \centering
        \includegraphics[width=\linewidth]{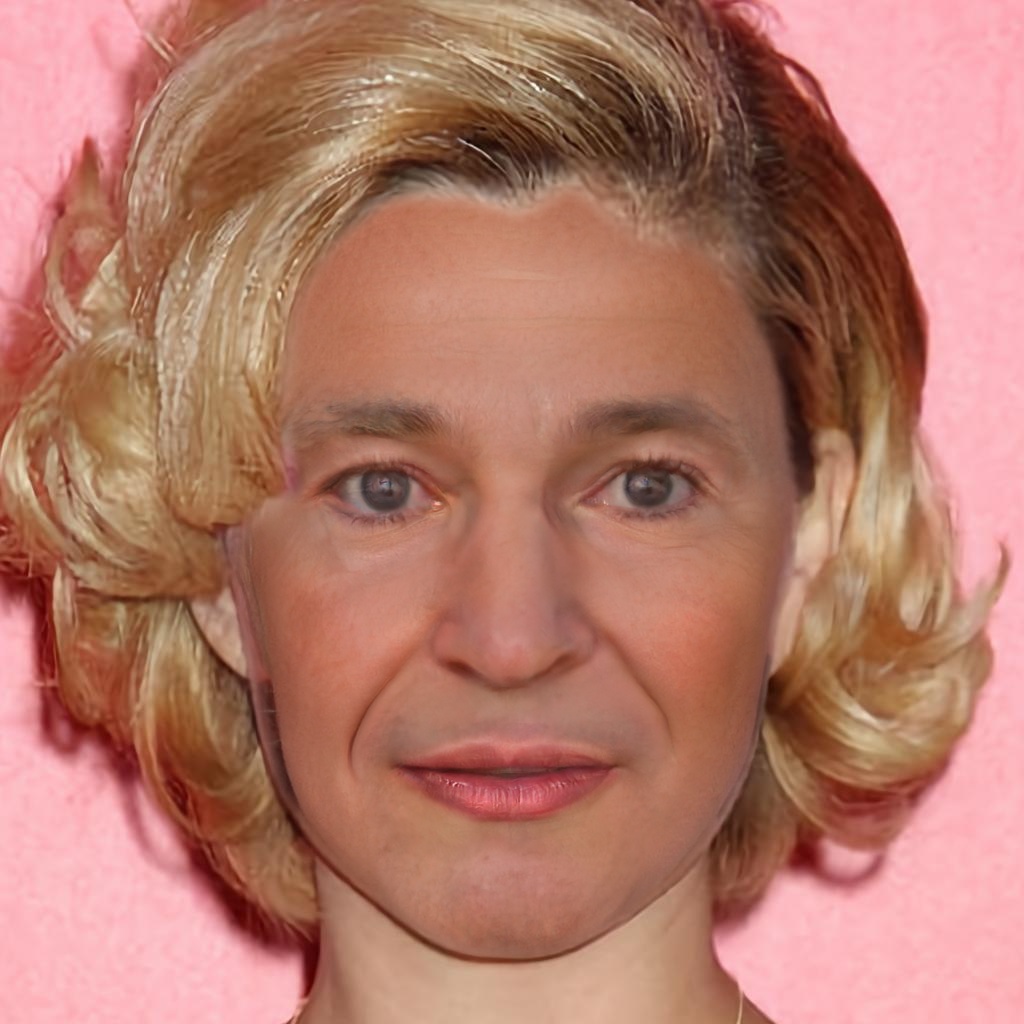}
        \includegraphics[width=\linewidth]{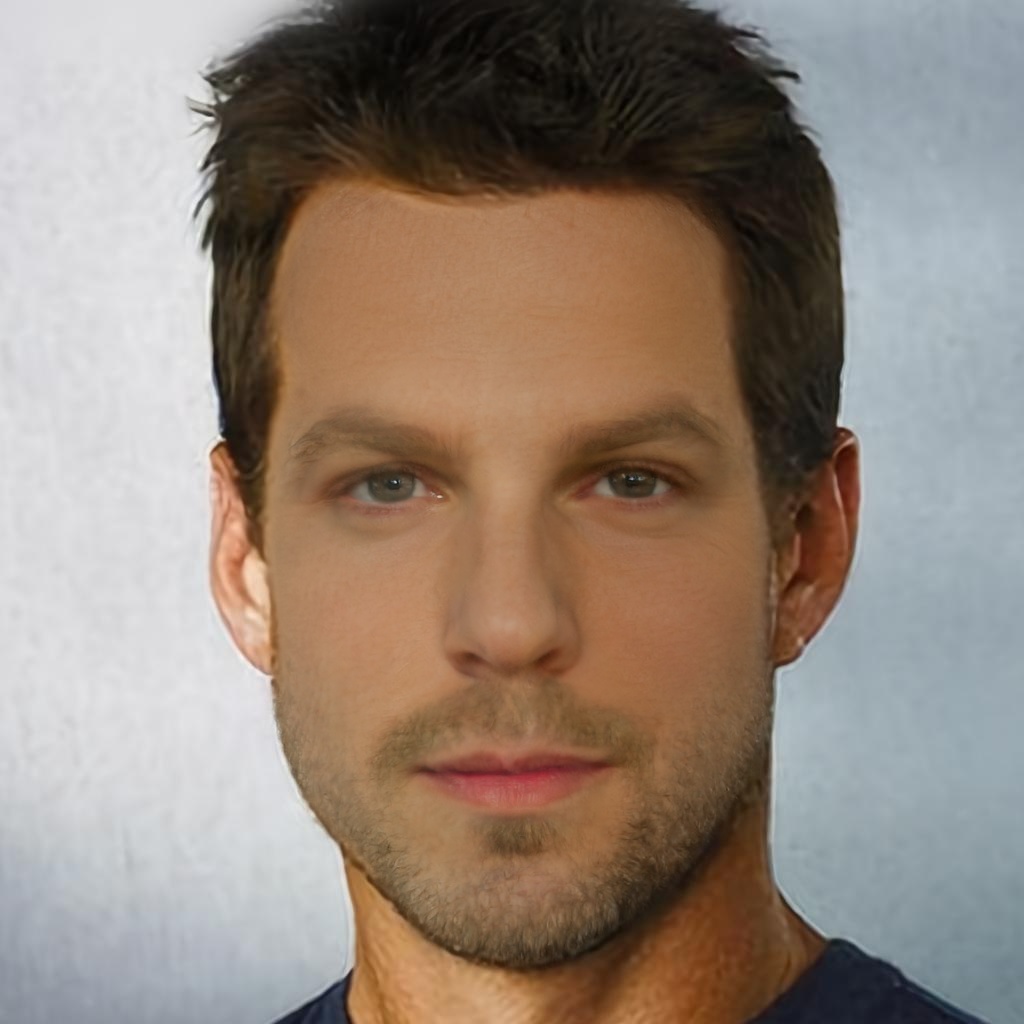}
        \caption{MegaFS~\cite{zhu2021one}}\label{fig:teaser_f}
    \end{subfigure}
    \hspace{-1.5mm}
    \begin{subfigure}{.14\linewidth}
        \centering
        \includegraphics[width=\linewidth]{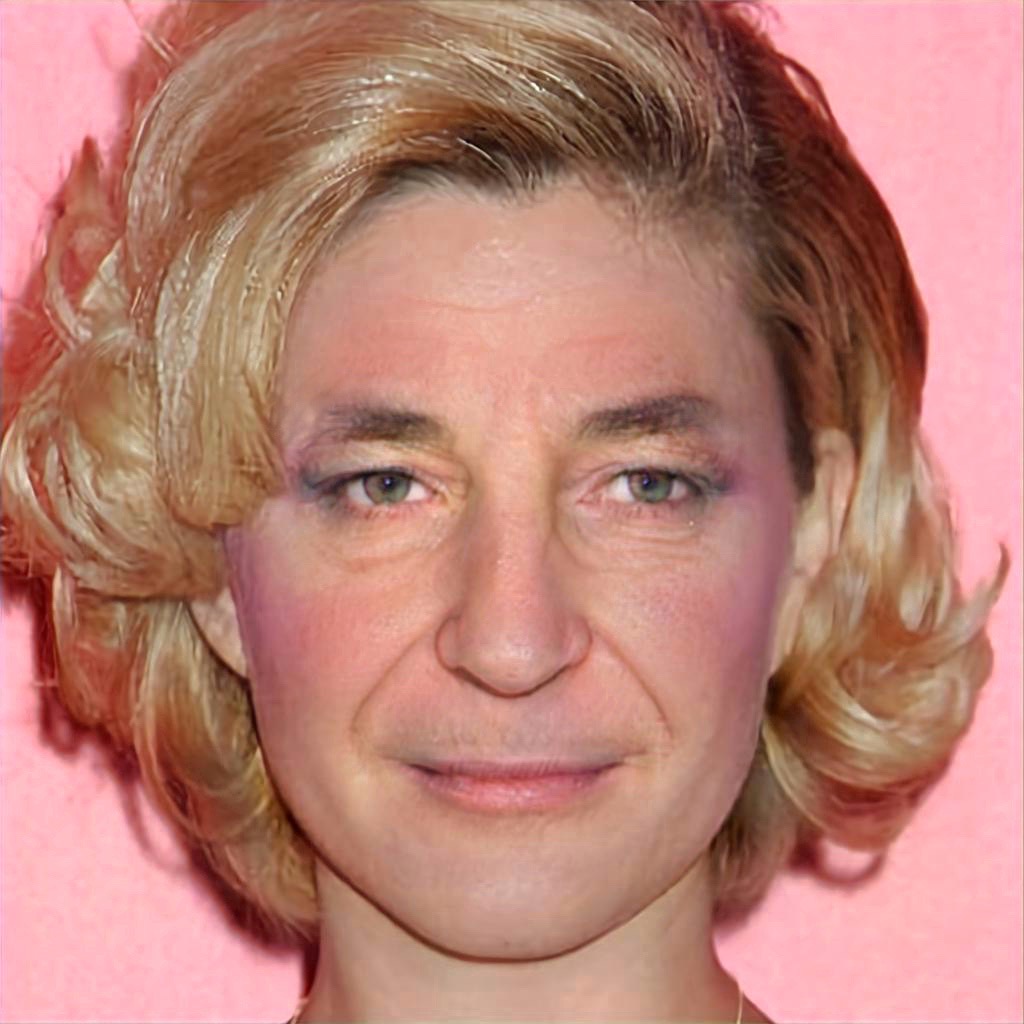}
        \includegraphics[width=\linewidth]{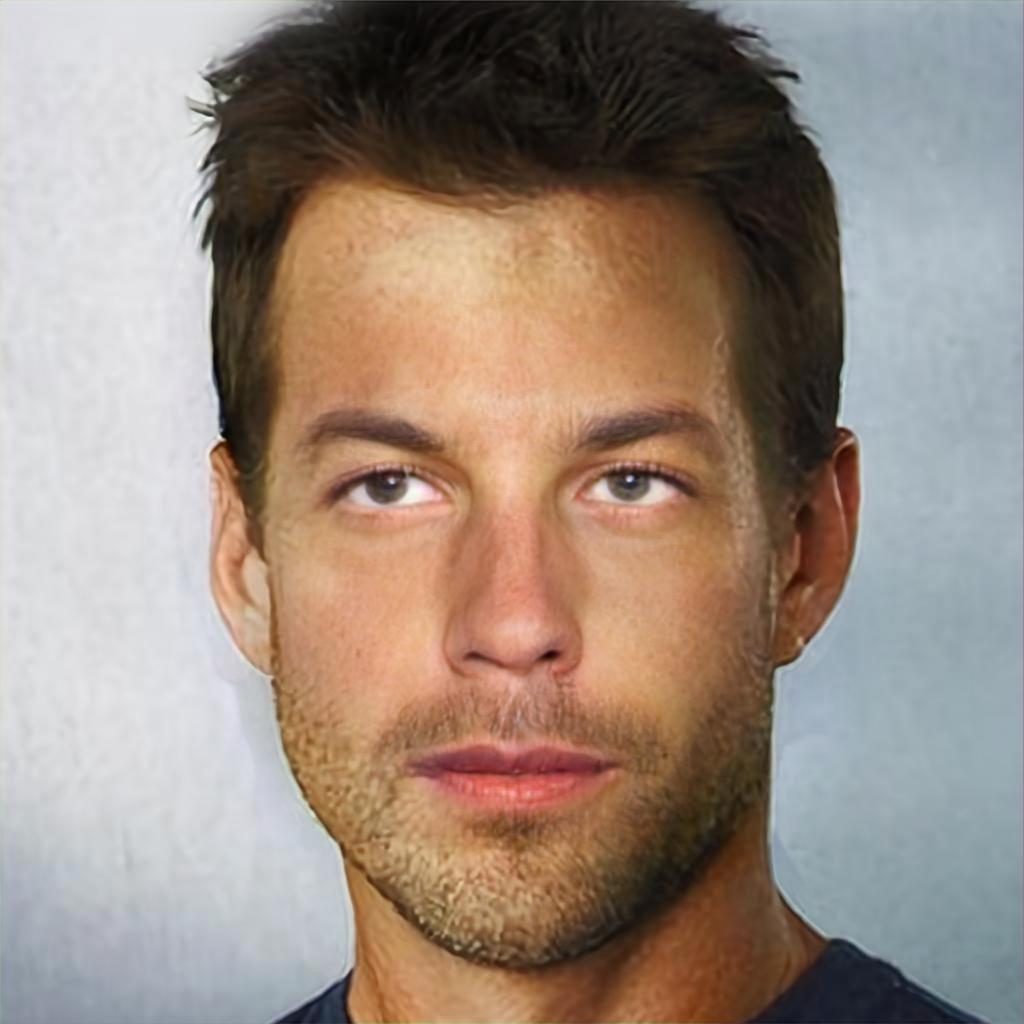}
        \caption{Ours}\label{fig:teaser_g}
    \end{subfigure}
  % Source    \    \               Target      \  \     MegaFS       \   \     Ours
  \capvspace
  \caption{We propose a high-resolution (1024$\times$1024) face swapping method by disentangling the latent semantics of a pre-trained GAN model. Compared with the existing high-resolution method MegaFS~\cite{zhu2021one}, our model can preserve target attributes (\eg, skin color and illumination), meanwhile transferring local details and identity from source faces.}
  \label{fig:teaser}
  \figbottomvspace
}
\maketitle
\begin{abstract}%\vspace{-2.5mm}
We present a novel high-resolution face swapping method using the inherent prior knowledge of a pre-trained GAN model. Although previous research can leverage generative priors to produce high-resolution results, their quality can suffer from the entangled semantics of  the latent space. We explicitly disentangle the latent semantics by utilizing the progressive nature of the generator, deriving structure attributes from the shallow layers and appearance attributes from the deeper ones. Identity and pose information within the structure attributes are further separated by introducing a landmark-driven structure transfer latent direction. The disentangled latent code produces rich generative features that incorporate feature blending to produce a plausible swapping result. We further extend our method to video face swapping by enforcing two spatio-temporal constraints on the latent space and the image space. Extensive experiments demonstrate that the proposed method outperforms state-of-the-art image/video face swapping methods in terms of hallucination quality and consistency. Code can be found at: \url{https://github.com/cnnlstm/FSLSD_HiRes}.

\end{abstract}
%%%%%%%%% BODY TEXT
% !TeX root = main.tex
%\vspace{-7mm}
\section{Introduction}
\label{intro.}
Face swapping aims at transferring the identity from a source face image to a target face image, while preserving attributes in the target image such as facial expression, head pose, illumination and background. It has received extensive attention in the computer vision and graphics community due to its wide range of potential applications, such as computer games, special effects and privacy protection~\cite{kemelmacher2016transfiguring,ross2010visual,rossler2019faceforensics,chen2020simswap}.

The main challenge for face swapping is to identify the highly entangled target facial attributes and source identity information for a natural-looking swapping.
Early works such as~\cite{bitouk2008face} replace the pixels of face regions and rely on similarities between the source and the target in pose and illumination. 3D-based approaches~\cite{cheng20093d,Lin2012Face,nirkin2018face} fit a 3D model to the faces and can handle large pose variation, although the fitting can be largely influenced by the environment and thus unstable. Other works introduce the generative adversarial networks (GANs) for hallucinating target attributes~\cite{li2020advancing,korshunova2017fast,natsume2018rsgan} due to its strong generative capability.

Though much progress has been made, many existing GAN-based approaches do not work well on high-resolution faces, due to the compressed representation of end-to-end frameworks~\cite{bao2018towards,natsume2018fsnet,li2020advancing}, the instability of adversarial training~\cite{arjovsky2017towards}, and the limitation in GPU memory size. Recently, Zhu~\etal~\cite{zhu2021one} utilize the inherent prior knowledge of a pre-trained high-resolution GAN model and propose MegaFS for high-resolution face swapping in the latent space of StyleGAN~\cite{karras2019style,karras2020analyzing}. It learns to assemble the inverted latent codes of the source and target images, and directly feed the fused code to a StyleGAN generator to produce the swapped result. However, since the identity and attributes are highly entangled in the latent space, assembling two latent codes without explicit guidance cannot guarantee the transfer of source identity and the preservation of target attributes simultaneously. Moreover, fine details embedded in the latent codes are easily diluted after the assembly, hence, their swapped face tends to have a blurred appearance with some loss of fine details (see Fig.~\ref{fig:teaser_c} and Fig.~\ref{fig:teaser_f} for examples).
To obtain disentangled semantics in the latent space, we argue that face features should be transferred in a class-specific manner. Intuitively, the structure attributes of a face, such as the facial shape, pose, and expression, should be treated differently than the appearance attributes such as illumination and skin tones. The swapped face should retain the source identity while hallucinating the appearance attributes of the target image. Such separate treatments would require proper disentanglement between the structure and appearance features.

In this paper, we delve into the latent semantics of StyleGAN. StyleGAN is a noise-to-image, coarse-to-fine generation process, and we leverage its progressive nature to disentangle key factors in swapping such as pose, expression, and appearance. 
Given the inverted source latent code, we decouple the structure (pose and expression) attributes from identity by deriving a structure transfer latent direction. The direction is determined by source and target landmarks, and serves as a latent space operation that transfers the structure. 
On the other hand, the appearance attributes are controlled in the deeper layers. Therefore, we regroup the target's appearance codes with the structure-transferred source code in the deep layers. In this way, the integrated latent code retains the identity attribute from the source, while having the appearance and structure of the target. This disentangled latent code is fed to the StyleGAN generator to produce generative features. This rich prior knowledge is aggregated with the target features in all scales, effectively eliminating the noticeable blending artifacts.

Furthermore, we extend our model to video face swapping, where we generate a swapped face video from a source face image and a target face video. Since directly applying our model to each video frame can lead to incoherence artifacts, we enforce two spatio-temporal constraints on the disentangled structure and appearance semantics. First, we require the inter-frame structural changes of the swapped faces to be consistent with those in the target faces. This is achieved by enforcing similarity between the swapped and target faces in terms of the latent offsets in the shallow layers (representing structural changes). Secondly, we adopt a linear assumption for the changes of image content between neighboring frames in the output video. These two constraints effectively ensure inter-frame coherence. Extensive experiments on several benchmarks for face swapping methods demonstrate superior performance against state-of-the-arts. As far as we are aware, this is the first feasible solution for high-resolution video face swapping.

In summary, our contributions are three-fold:
\begin{itemize}[leftmargin=*,nosep]
    \item We propose a novel framework for high-resolution face swapping. We disentangle the latent semantics of a pre-trained StyleGAN, enabling the transfer of source identity while preserving the appearance and structure of the target.
    \item We tailor two novel constraints to enforce the coherence of swapped face videos, including a code trajectory constraint that limits the offset between the latent codes of neighboring frames, and a flow trajectory constraint that works in the RGB space to guarantee the video smoothness.
    \item Experiments on several datasets demonstrate state-of-the-art results from our method, which can potentially serve as new high-resolution test cases for face forgery detection.
\end{itemize}

\section{{Related Work}}

\begin{figure*}[t]
	\centering
	\includegraphics[width=1\linewidth]{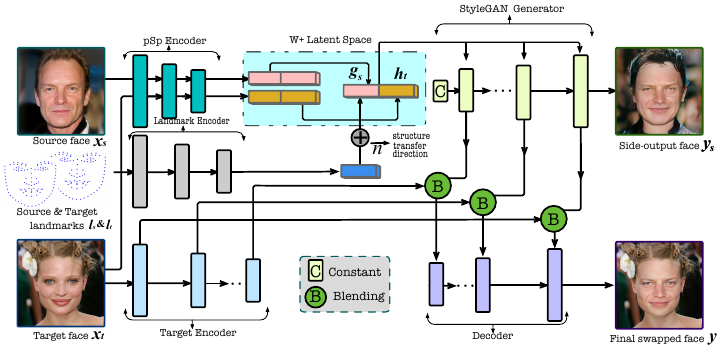}
	\vspace{-5mm}
	\caption{The pipeline of our disentangled high-resolution face swapping. We transfer different levels of attributes by three modules. We first invert both the source and target faces to the $W+$ latent space of the pre-trained StyleGAN $G$ by the pSp encoder, then learn a structure transfer direction by encoding the source and target landmarks. The appearance attributes are transferred by swapping the latter part of two latent codes. To eliminate the blending boundary, target features are aggregated with the generative features in a multi-scale manner.}
	\label{fig:overview}
	\figbottomvspace
\end{figure*}

% \subsection{Face Swapping}
\topparaheading{Face Swapping}
Face swapping has been an active research topic. Early works such as~\cite{bitouk2008face} can only handle subjects of the same pose. 3D-based approaches fit a 3D template to the source and target faces, to better handle large pose variations~\cite{cheng20093d,Lin2012Face,nirkin2018face}. Face2Face~\cite{thies2016face2face} fits a bilinear face model to the source and target faces to transfer the expression, with additional steps to synthesize realistic mouth appearance. However, 3D-based methods cannot handle appearance attributes like illumination and style.
Recently, many works introduce GANs for face swapping. RSGAN~\cite{natsume2018rsgan} handles face and hair regions in the latent spaces, and conducts swapping by replacing latent representations. Similarly, FSNet~\cite{natsume2018fsnet} encodes the face region and the non-face region to different latent codes. FSGAN~\cite{nirkin2019fsgan} performs face reenactment and swapping simultaneously. Recently, Faceshifter~\cite{li2020advancing} encodes the source identity and target attributes separately. However, none of them can swap face at the high resolution due to the unstable adversarial training, or the compressed representation of end-to-end frameworks.
Recently, Naruniec~\etal~\cite{naruniec2020high} swap faces at high resolutions but their model is subject-specific. MegaFS~\cite{zhu2021one} utilizes the prior knowledge of pre-trained StyleGAN~\cite{karras2020analyzing,karras2019style}. It inverts both source and target faces to the latent space, then designs a face transfer block to assemble the latent codes. However, the identity and attributes are entangled in the latent space, and assembling two latent codes without explicit guidance is insufficient. On the contrary, we propose to transfer attributes in a class-specific manner.
%\vspace{-5mm}

\paraheading{Generative Prior}
Generative models show great potential in synthesizing high-quality images~\cite{goodfellow2014generative,karras2018progressive,brock2018large,karras2020analyzing,karras2019style}. It has been shown that the latent space of a pre-trained GAN encodes rich semantic information. As a result, many works utilize the prior knowledge hidden in the generative model for semantic editing, image translation, super-resolution, and so on~\cite{gu2020image,pan2020exploiting,abdal2021styleflow,richardson2020encoding,yang2021discovering,xu2021continuity}. In particular, Gu~\etal~\cite{gu2020image} propose a GAN inversion approach for image colorization and semantic editing, while some other works \cite{menon2020pulse,chan2021glean,zhou2022pro} use the pre-trained StyleGAN for image super-resolution.
Recently, generative prior has also been introduced for face swapping. Besides MegaFS~\cite{zhu2021one}, Nitzan~\etal~\cite{nitzan2020face} transfer attributes from one face to another in the latent space by a fully-connected network.
Different from those works, we transfer the target attributes to the source in a more elaborated way. We decompose the attributes into structure attributes and appearance attributes, and take full usage of disentanglement and editability of the generative model. 
% !TeX root = main.tex
\section{Method}

\subsection{Overview}

Given two high-resolution face images, we aim to construct a face image with the identity of the source face $x_s$ and the attributes of the target face $x_t$ such as the pose, expression, illumination and background. To do so, we first construct a side-output swapped face using a StyleGAN generator, by blending the structure attributes of the source and the target in the latent space while reusing the appearance attributes of the target (Sec.~\ref{sec.att}). To further transfer the background of the target face, we use an encoder to generate multi-resolution features from the target image, and blend them with the corresponding features from the upsampling blocks of the StyleGAN generator. The blended features are fed into a decoder to synthesis the final swapped face image (Sec.~\ref{sec.Background}). An overview of our pipeline is provided in Fig.~\ref{fig:overview}. The networks are trained using a set of loss functions that enforce desirable properties such as the preservation of source identity and target appearance (Sec.~\ref{sec.loss}). We also extend our approach to video face swapping, using additional spatio-temporal constraints to enforce coherence (Sec.~\ref{sec.Video}).

\subsection{Class-Specific Attributes Transfer}
\label{sec.att}

A face image typically contains different classes of attributes. For example, the pose and expression are related to the face structure, while the lighting and colors are related to the appearance. Previous face swapping techniques, such as FaceShifter~\cite{li2020advancing} and MegaFS~\cite{zhu2021one}, do not explicitly distinguish different classes of attributes when transferring them to the output image.
We instead argue that it is beneficial to transfer structure and appearance attributes separately: the structure attributes of the output can be jointly determined from their counterparts in the source and target images, to obtain the same pose and expression as the target face while retaining the identity of the source face. Meanwhile, the appearance attributes of the target face can be directly reused to achieve similar facial appearance in the output.
%Such separate treatments require disentangling the structure and appearance attributes. To this end,

To this end, we note that StyleGAN~\cite{karras2019style,karras2020analyzing}, a state-of-the-art generative model, provides a suitable representation for such separate treatments.
In particular, to use the StyleGAN generator, it is common practice to encode an image in an extended latent space $W+$ where a latent code consists of multiple high-dimensional vectors, one for each input layer of StyleGAN~\cite{abdal2019image2stylegan,richardson2020encoding}. As noted in~\cite{richardson2020encoding,yang2021semantic}, different input layers of StyleGAN correspond to different levels of details.
Thus we treat the first $K$ vectors of the latent code, which correspond to the shallow layers, as the encoding for the structure attributes. The remaining vectors, which correspond to the deeper layers, are used for appearance. For $1024 \times 1024$ images, the latent code consists of 18 different 512-dimensional vectors, and we follow~\cite{richardson2020encoding} and choose the first 7 vectors as the structure part.
With such disentanglement, the structure part and the appearance part of the latent code can be transferred separately. Specifically, we first use the pre-trained pSp encoder~\cite{richardson2020encoding} to invert the source face $x_s$ and the target face $x_t$ to obtain their $W+$ latent codes $w_s = (g_s,h_s)$ and $w_t=(g_t,h_t)$ respectively, where $g_s,g_t$ are the structure parts and $h_s,h_t$ are the appearance parts. To construct the structure attributes of a swapped face that has the same identity of the source with the pose and expression of the target,
we note that they should be derived from the source structure attributes with modifications that take into account the pose and expression difference between the source and the target. Therefore, we compute the structure part $\widehat{g}_s$ of the latent code for a swapped face by applying a structure transfer latent direction $\overrightarrow{{n}}$ that is derived from the source and target structures:
\begin{equation}
    \widehat{\str}_{s} = {\str}_{s} + \overrightarrow{{n}}.
\label{eq:EditedStructureCode}
\end{equation}

To derive $\overrightarrow{{n}}$, we note that the face structure can be indicated by the facial landmarks. Thus we train a landmark encoder $E_{le}(\cdot,\cdot)$ that generates $\overrightarrow{{n}}$ from the heat-map encodings of the source landmarks $l_{s}$ and the target landmarks $l_{t}$:
\[
\overrightarrow{{n}} = E_{le}(l_{s},l_{t}).
\]

To transfer the appearance attributes from the target face to the swapped face, we directly reuse the appearance part $h_t$ of the target latent code. It is then re-integrated with $\widehat{\str}_{s}$ to form a latent code for the swapped face:
\begin{equation}
\widehat{w}_{s} = \cat(\widehat{\str}_{s},{\apr}_{t}),
\label{eq:InitSwapLatentCode}
\end{equation}
where $\cat(\cdot,\cdot)$ denotes the concatenation operator. The code is fed into a pre-trained StyleGAN generator to obtain a side-output swapped face $y_{s}$.

\subsection{Background Transfer}
\label{sec.Background}

For face swapping applications, the background of the target face also needs to be retained in the output image. This often cannot be guaranteed for the swapped face $y_s$ computed in Sec.~\ref{sec.att}. A common solution is to apply Poisson Blending as the post-process that blends the swapped inner-face with the target image. However, this can lead to unnatural appearance around the boundary of the inner-face.

To address this issue, we discard the side-output face $y_s$, but retain the features $F_s = \{f^0_s,f^1_s,...f^N_s\}$ produced by each upsampling block of the StyleGAN generator from the latent code $\widehat{w}_{s}$ in Eq.~\eqref{eq:InitSwapLatentCode}. We note that $F_s$ can be regarded as a representation for different levels of details from the side-output face image. Accordingly, we apply an encoder $E_t$ to the target face $x_t$ such that its layers generate corresponding features $F_t = \{f^0_t,f^1_s,...f^N_s\}$, where $f^i_t$ is of the same dimension as $f^i_s$ and represent the details of the target face image at the same resolution. We then aggregate each pair of corresponding features $(f^i_s, f^i_t)$ by replacing the components of $f^i_t$ for the inner-face region with their counterparts in $f^i_s$. All aggregated features are fed into a decoder to produce the final face image $y_f$, which can be written as:
\[
y_{f} = \dec(F_t,F_s,m_t),
\]
where $\dec(\cdot,\cdot,\cdot)$ denotes the decoder, and $m_t$ is the inner-face mask for the target face image. In this way, the decoder transfers the background in multi-level features, which not only eliminates the need for explicit background blending but also enables the code $\widehat{w}_{s}$ to focus on the facial region and facilitates attributes transfer.

\subsection{Loss Functions}
\label{sec.loss}
We tailor several losses that are applied on the side-output swapped face $y_{s}$ or the final face $y_{f}$ for efficient attributes transfer, as explained in the following. We also introduce a style-preservation loss on the final output to narrow the style gaps between the swapped one and the target.

\paraheading{Adversarial loss} We utilize an adversarial loss for the distribution alignment between the final swapped faces and  the real faces. In particular, we align the final face $y_{f}$ with the target face with the following loss:
\begin{equation*}
{L}_{adv} = \underset{y_{f} \sim P_{Y_{f}}}{\mathbb{E}} [-\log( D_{f}(y_{f}))],
%\label{eq7}
\end{equation*}
where $P_{Y_{f}}$ denotes the distribution of the final faces.
The discriminator $D_{f}$ that distinguishes the final face from the real target face $x_t$ is trained with a loss
\begin{equation*}
{L}_{D_{f}} = \underset{y_{f} \sim P_{Y_{f}}}{\mathbb{E}} [-\log( 1- D_{f}(y_{f})] + \underset{x_t \sim P_{X_t}}{\mathbb{E}} [-\log(D_{f}(x_t)] ,
%\label{eq8}
\end{equation*}
where $P_{X_{t}}$ denote the distribution of real faces.

\paraheading{Identity-preservation loss} To preserve the identity of the source face, we introduce an identity-preservation loss for the final face $y_f$ with the source face $x_s$:
\begin{equation*}
{L}_{id} = 1 - \texttt{cos}(\Phi_{id}(y_f),\Phi_{id}(x_s)),
%\label{eq12}
\end{equation*}
where $\Phi_{id}(\cdot)$ is the pre-trained ArcFace network for face recognition~\cite{deng2019arcface}, and $\texttt{cos}(\cdot, \cdot)$ denotes the cosine similarity.

\paraheading{Landmark-alignment Loss} Since we use facial landmarks to represent structure attributes, we introduce the following loss to align the landmarks of the side-output swapped face $y_s$, the final face $y_f$, and the target face $x_t$:
\begin{equation*}
{L}_{lmk} = \|E_{lmk}(y_s)-E_{lmk}(x_t)\|_2 + \|E_{lmk}(y_f)-E_{lmk}(x_t)\|_2,
%\label{eq13}
\end{equation*}
where $E_{lmk}(\cdot)$ is a pre-trained landmark estimator~\cite{wang2020deep} and $\|\cdot\|_2$ denotes the $\ell_2$-norm.

\paraheading{Reconstruction Loss} Intuitively, if the source face $x_s$ and the target face $x_t$ are the same image, the network should reconstruct this image for both the side-output swapped face $y_s$ and the final face $y_f$. Hence, we follow pSp~\cite{richardson2020encoding} and use pixel-wise similarity and perceptual similarity to define a reconstruction loss that penalize the deviation between $y_s, y_f$ and $x_t$ when $x_s = x_t$:
%\vspace{-2mm}
\begin{align*}
&{L}_{rec} = \nonumber\\
&
\begin{cases}
\|y_f - x_t\|_2 + \alpha \|F(y_f) - F(x_t)\|_2 & \\
~ + \|y_s - x_t\|_2 + \alpha \|F(y_s) - F(x_t)\|_2, & \text{if}~x_s = x_t,\\
0, &{\text{otherwise}},
\end{cases}
\end{align*}
% \vspace{-2mm}
where $F(\cdot)$ is the perceptual feature extractor, and $\alpha$ is a weight that balances the pixel-wise similarity and perceptual similarity terms. We set $\alpha=0.8$ in our experiment.

\paraheading{Style-transfer Loss} As mentioned in Sec.~\ref{sec.att}, we transfer the appearance attributes in the latent space. However, if the difference between the styles of the source and target faces is too large, simple latent code replacement may not reduce the style difference effectively. Inspired by BeautyGAN~\cite{li2018beautygan}, we create a guidance face image $\hm(y_f,x_t)$ through histogram mapping, and align the final face $y_f$ with the guidance face via the following loss:
\begin{equation*}
{L}_{st} = \|y_f - \hm(y_f,x_t) \|_2.
%\label{eq14}
\end{equation*}
Compared to simple latent code replacement, this provides a stronger guidance for the appearance attributes transfer.

\paraheading{Final Objective} The final loss function for training our model is a weighted combination of above losses:
\begin{equation}
{L}_{total} = \lambda_1 {L}_{adv} + \lambda_2 {L}_{id} + \lambda_3 {L}_{lmk} + \lambda_4 {L}_{rec} + \lambda_5 {L}_{st},
\label{eq14}
\end{equation}
where $\lambda_1, \lambda_2, \lambda_3, \lambda_4, \lambda_5$ are weights for the loss terms.

\subsection{Video Face Swapping}
\label{sec.Video}

Our method can be extended to video face swapping. Given a source face $x_s$ and a sequence of target faces with $M$ consecutive frames $S_t=\{x_{t}^0,x_{t}^1,...,x_{t}^{M-1}\}$, we would like to obtain a swapped face sequence $Y=\{y^0,y^1,...,y^{M-1}\}$.
Most of the existing works apply the image-based face swapping methods to each video frame separately, which can lead to incoherent results between neighboring frames and cause artifacts such as flickering. Such artifacts are particularly noticeable in high resolutions.
To address this issue, we need to enforce consistency between neighboring frames in terms of both the structure and the appearance, such that these attributes vary smoothly across the frames.
Existing temporal consistency works~\cite{lai2018learning,bonneel2015blind} only take the appearance consistency into account and cannot be used directly for face swapping. We propose two spatio-temporal constraints for the latent space and the image space respectively, to achieve consistency in both the structure and the appearance.

\paraheading{Code Trajectory Constraint}
Since the structure attributes of the target frames vary smoothly, we can enforce structure consistency for the swapped face frames by requiring them to have similar changes in structure attributes as the target.
To this end, we note that the offset between the structure parts of the latent codes in two neighboring frames can be can be regarded as an indication of the change in the structure attributes. Therefore, we use the following loss to enforce similar trajectories of the structure codes between the target video and the output video in the latent space:
%\vspace{-2mm}
\begin{equation*}
{L}_{ct} = \sum\nolimits_{k=1}^{M} \left\|(\widehat{\str}_{s}^{k} - \widehat{\str}_s^{k-1}) - ({\str}_{t}^{k} - {\str}_t^{k-1})\right\|_2,
% \label{eq15}
\end{equation*}
where ${\str}_{t}^k$ denotes the structure part of the latent code for the target frame $x_{t}^k$, and $\widehat{\str}_{s}^{k}$ denotes the structure code of the swapped face obtained from $x_{s}$ and $x_{t}^k$ using Eq.~\eqref{eq:EditedStructureCode}.

\begin{figure*}[t]
   \centering
    % \rotatebox[origin=c]{90}{Ours \hspace{15mm} InD \hspace{15mm} pSp \hspace{15mm} I2S}
    \captionsetup[subfigure]{labelformat=empty,justification=centering}
    \begin{subfigure}{.12\linewidth}
     \centering
     \includegraphics[width=\linewidth]{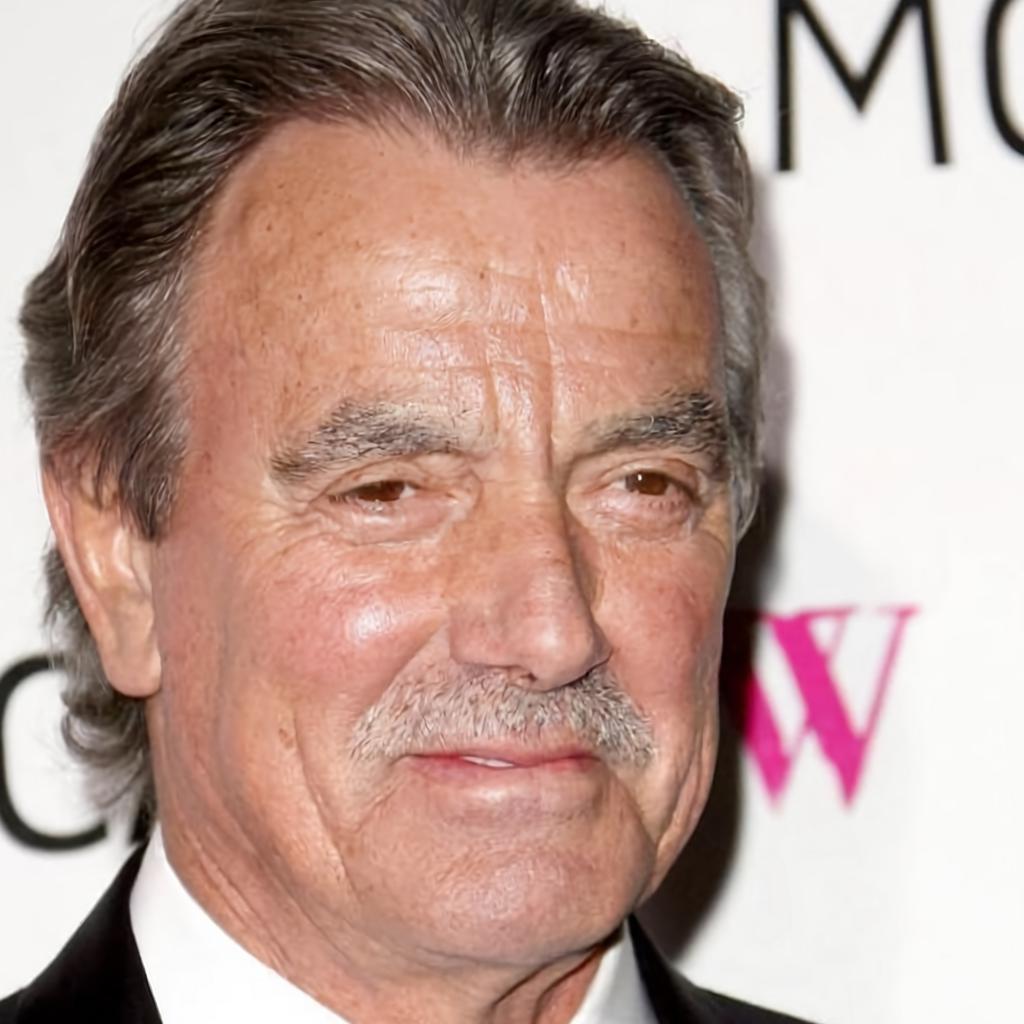}
     \includegraphics[width=\linewidth]{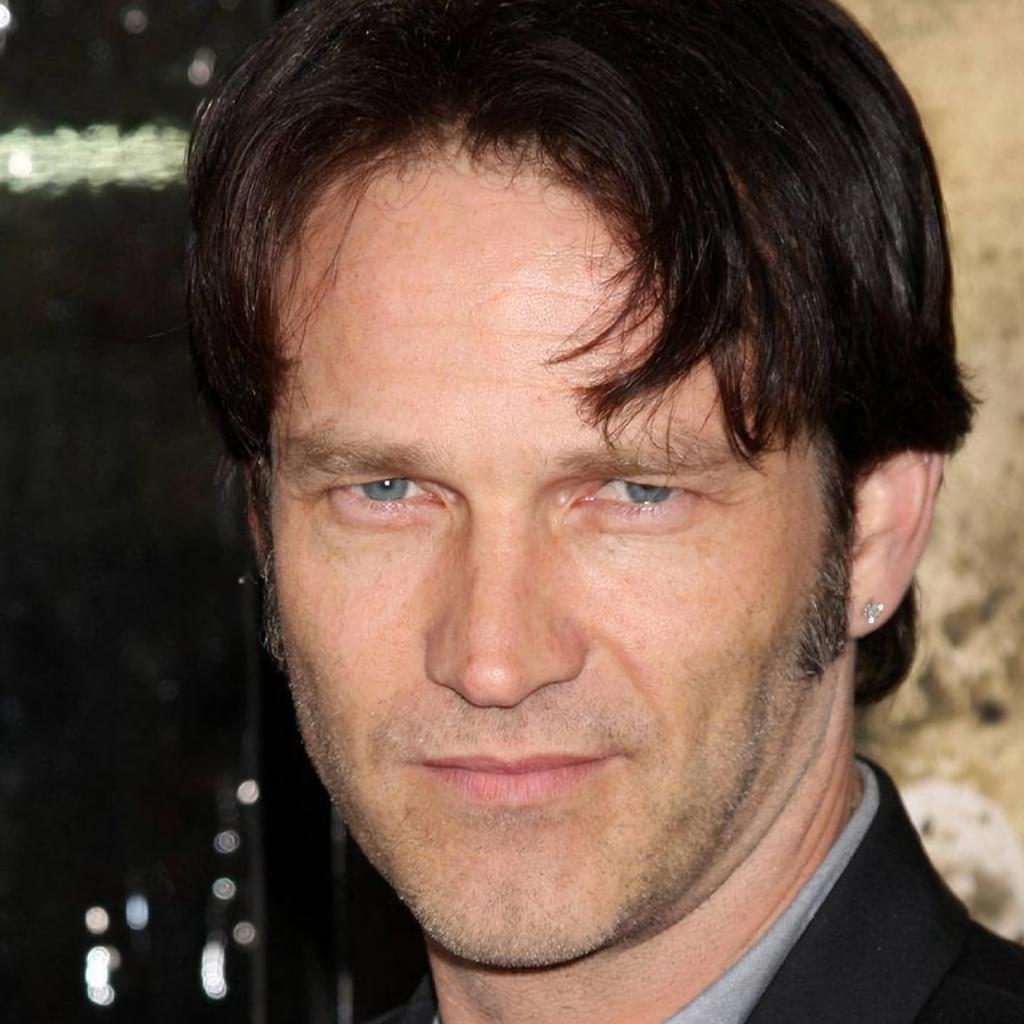}
     \includegraphics[width=\linewidth]{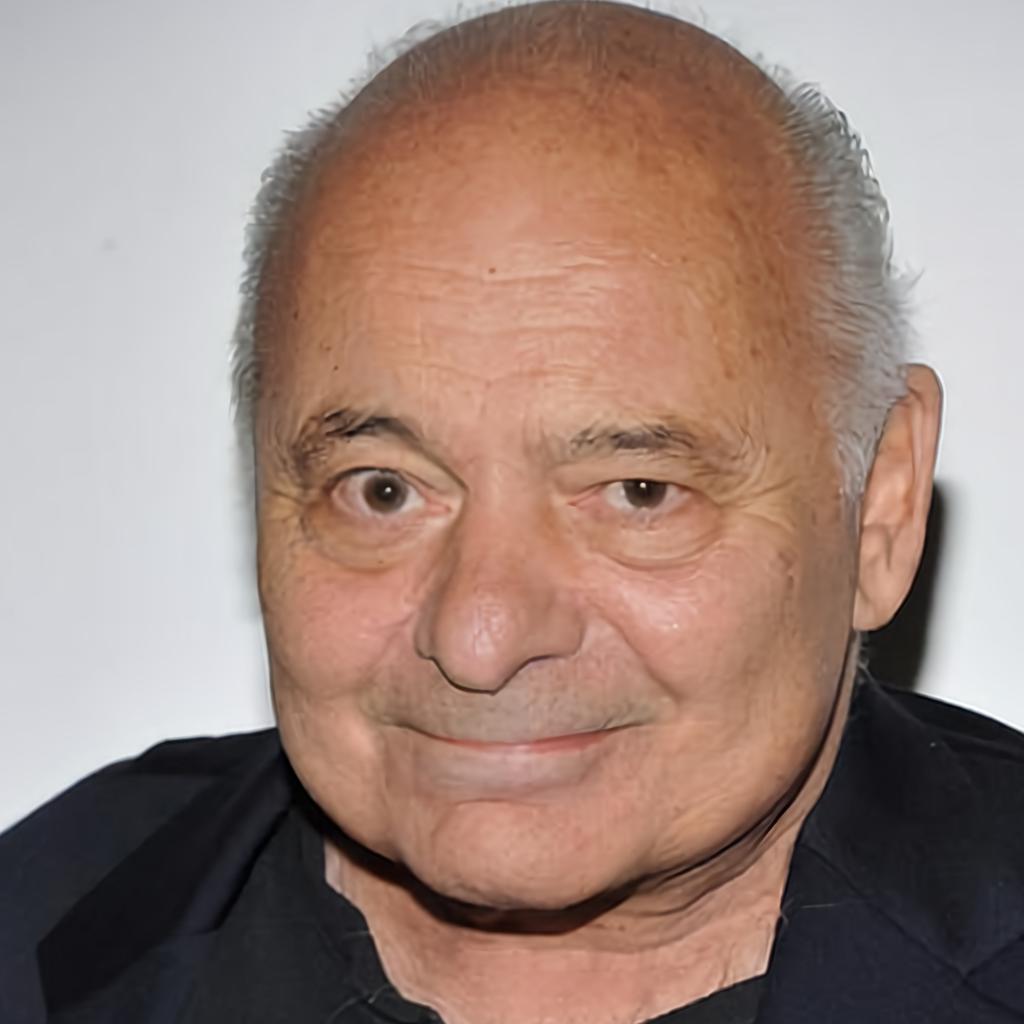}
     \includegraphics[width=\linewidth]{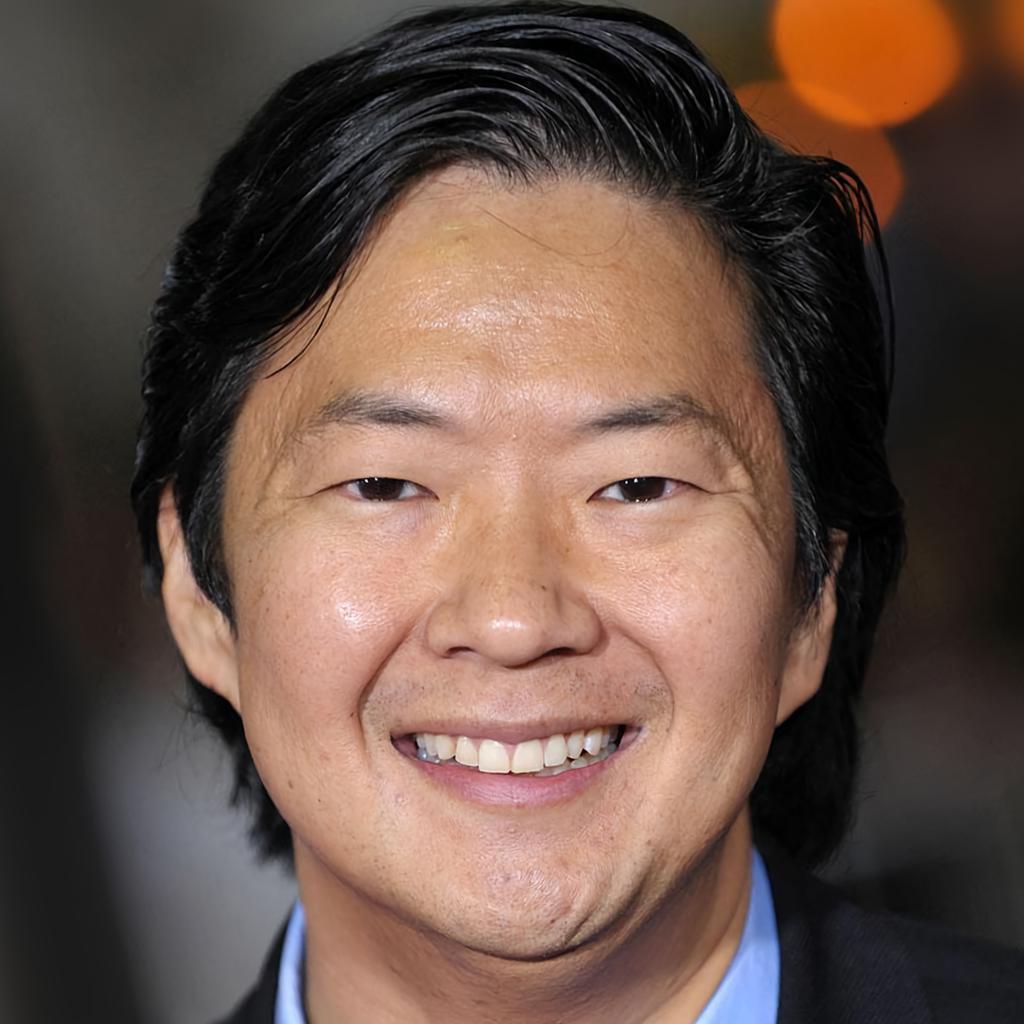}
     \caption{Source}
     \end{subfigure}
     \hspace{-2mm}
     \begin{subfigure}{.12\linewidth}
      \centering
     \includegraphics[width=\linewidth]{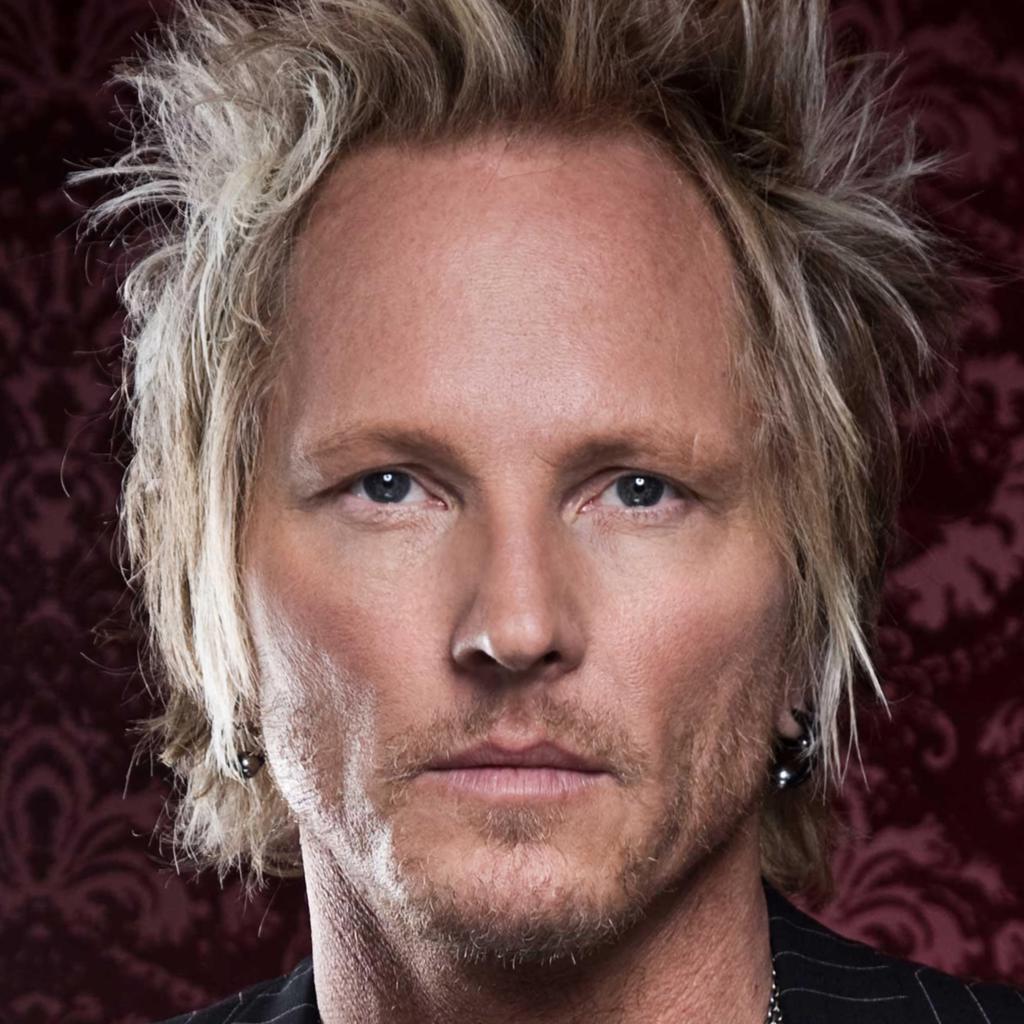}
     \includegraphics[width=\linewidth]{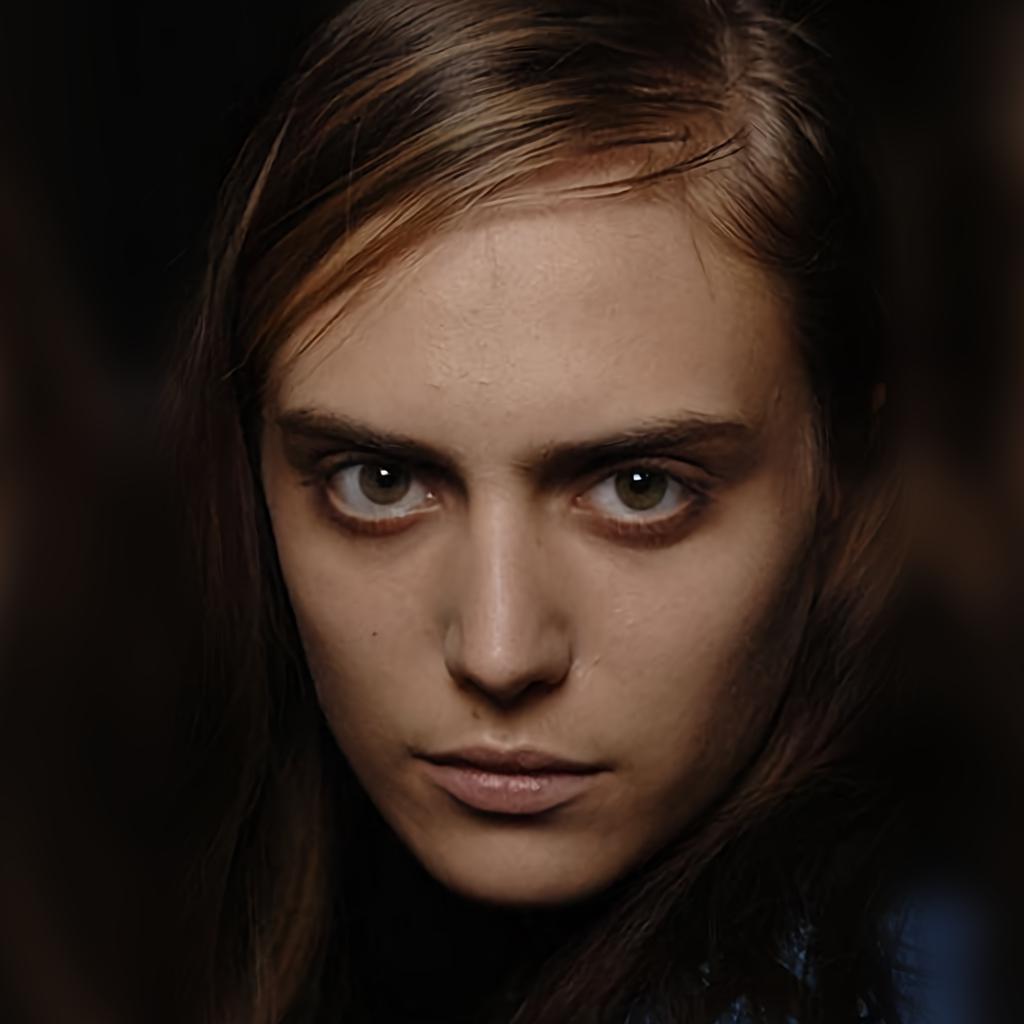}
     \includegraphics[width=\linewidth]{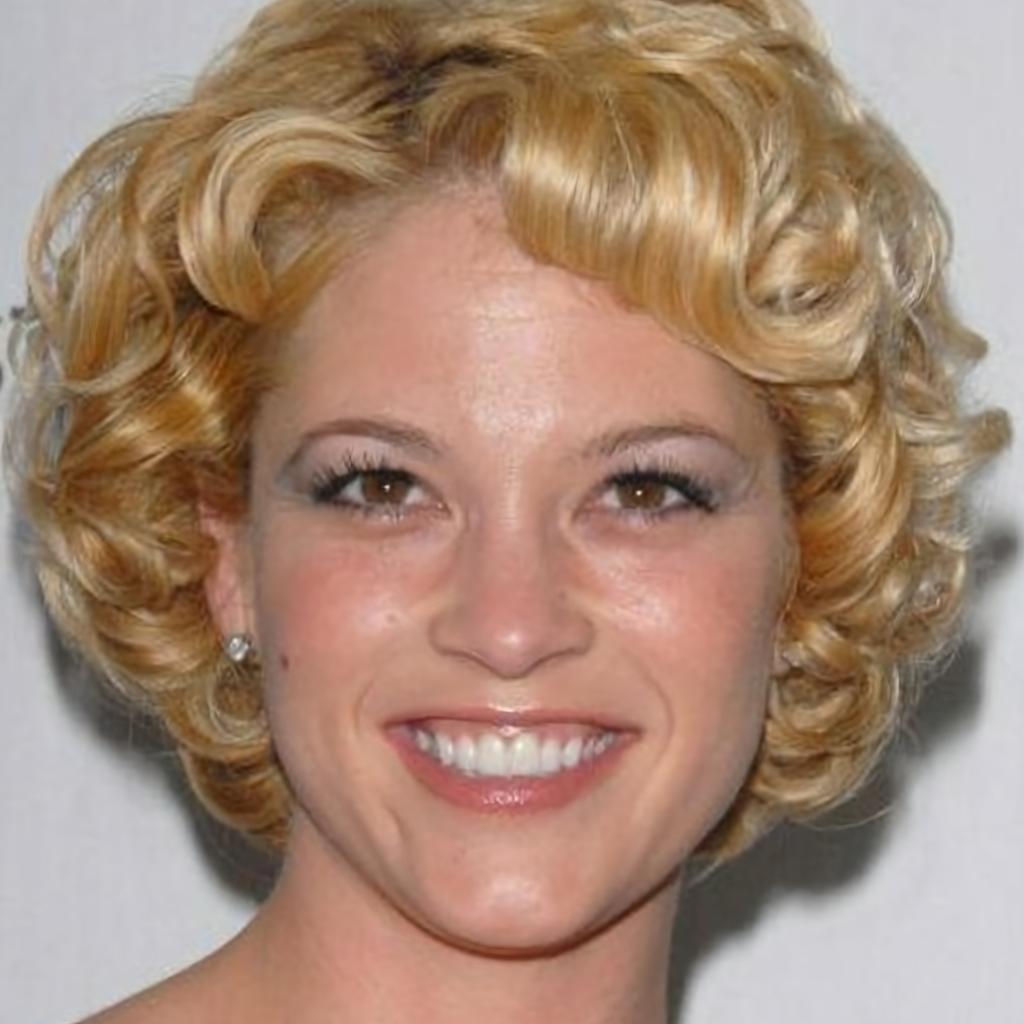}
     \includegraphics[width=\linewidth]{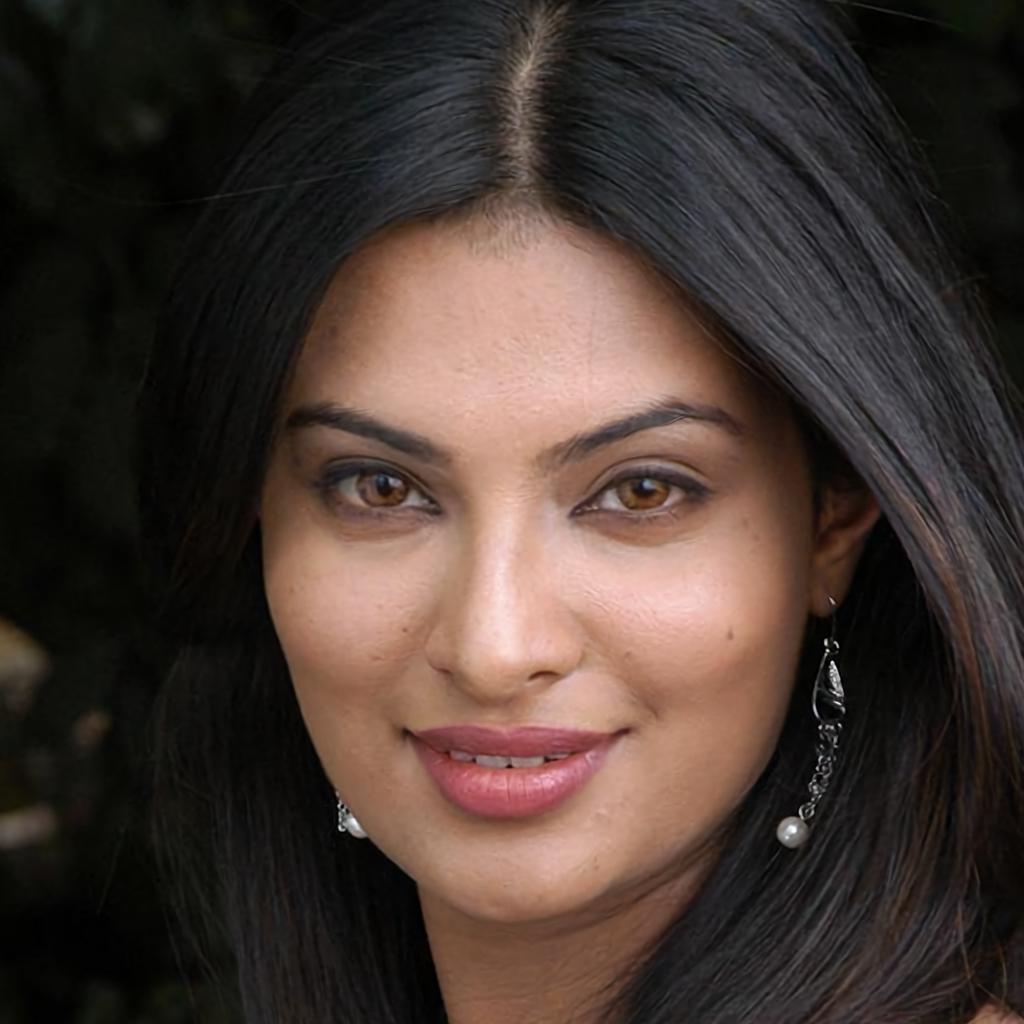}
     \caption{Target}
     \end{subfigure}
     \hspace{-2mm}
     \begin{subfigure}{.12\linewidth}
     \centering
     \includegraphics[width=\linewidth]{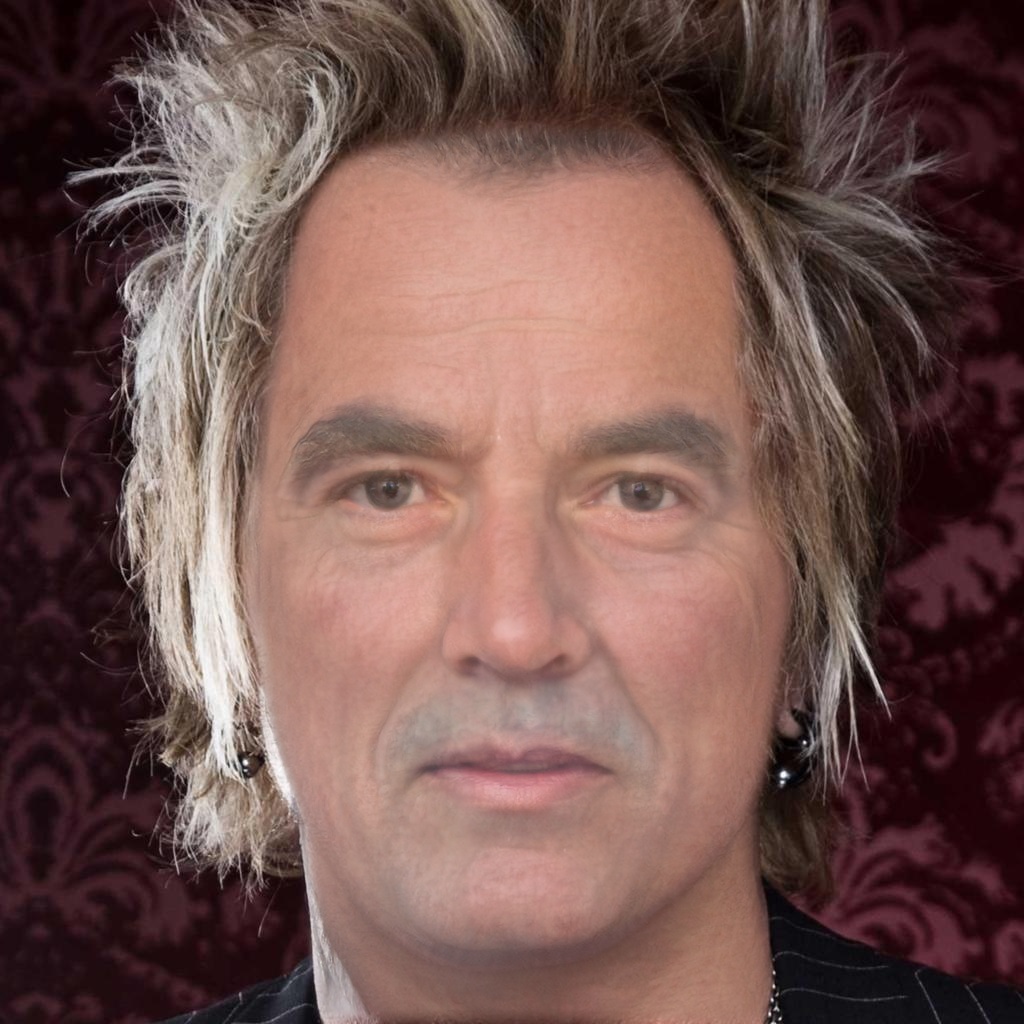}
     \includegraphics[width=\linewidth]{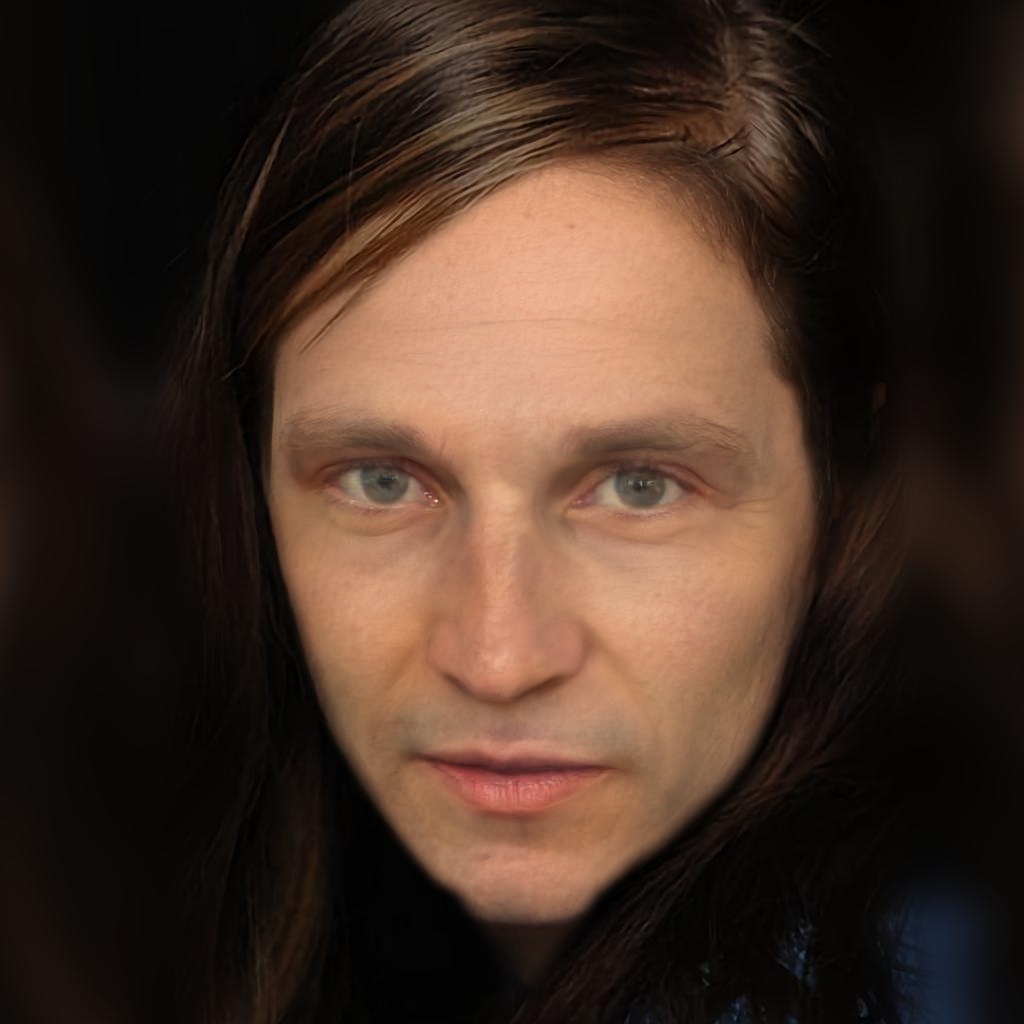}
     \includegraphics[width=\linewidth]{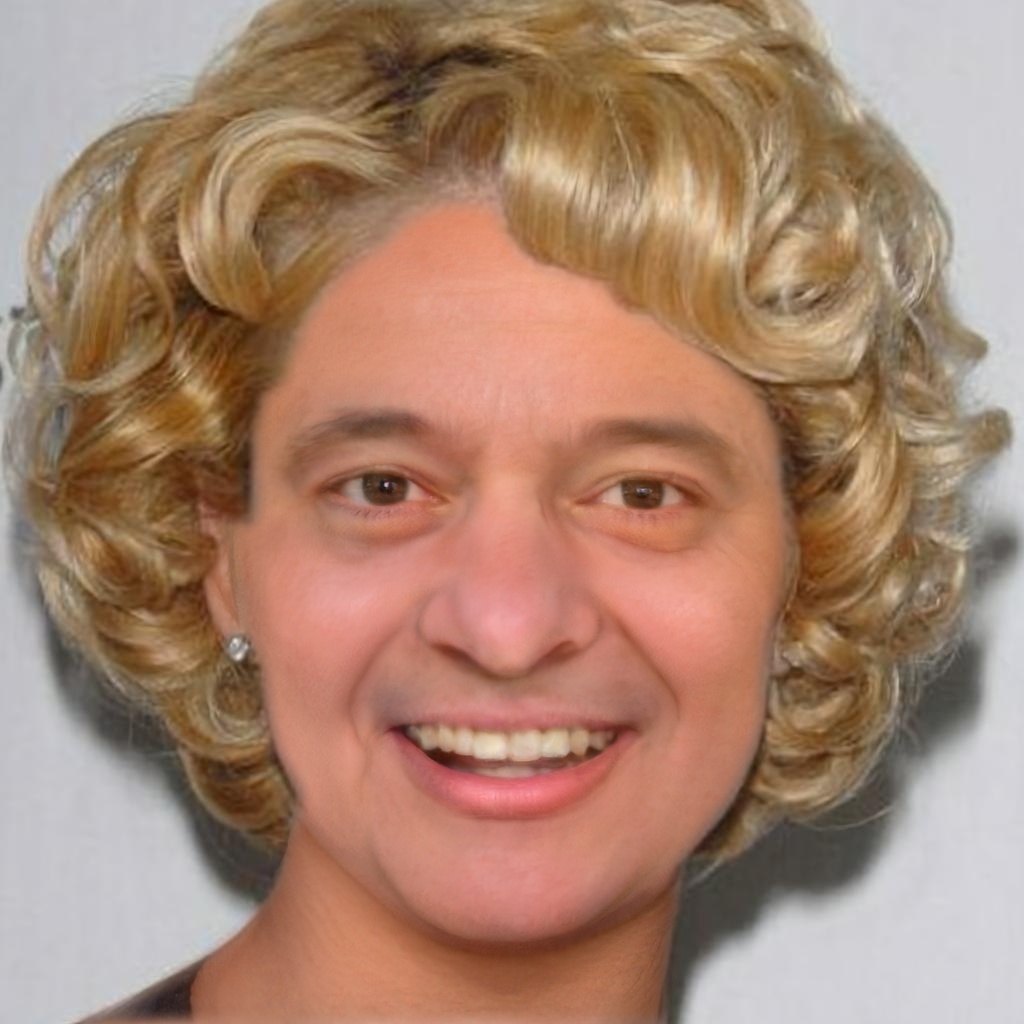}
     \includegraphics[width=\linewidth]{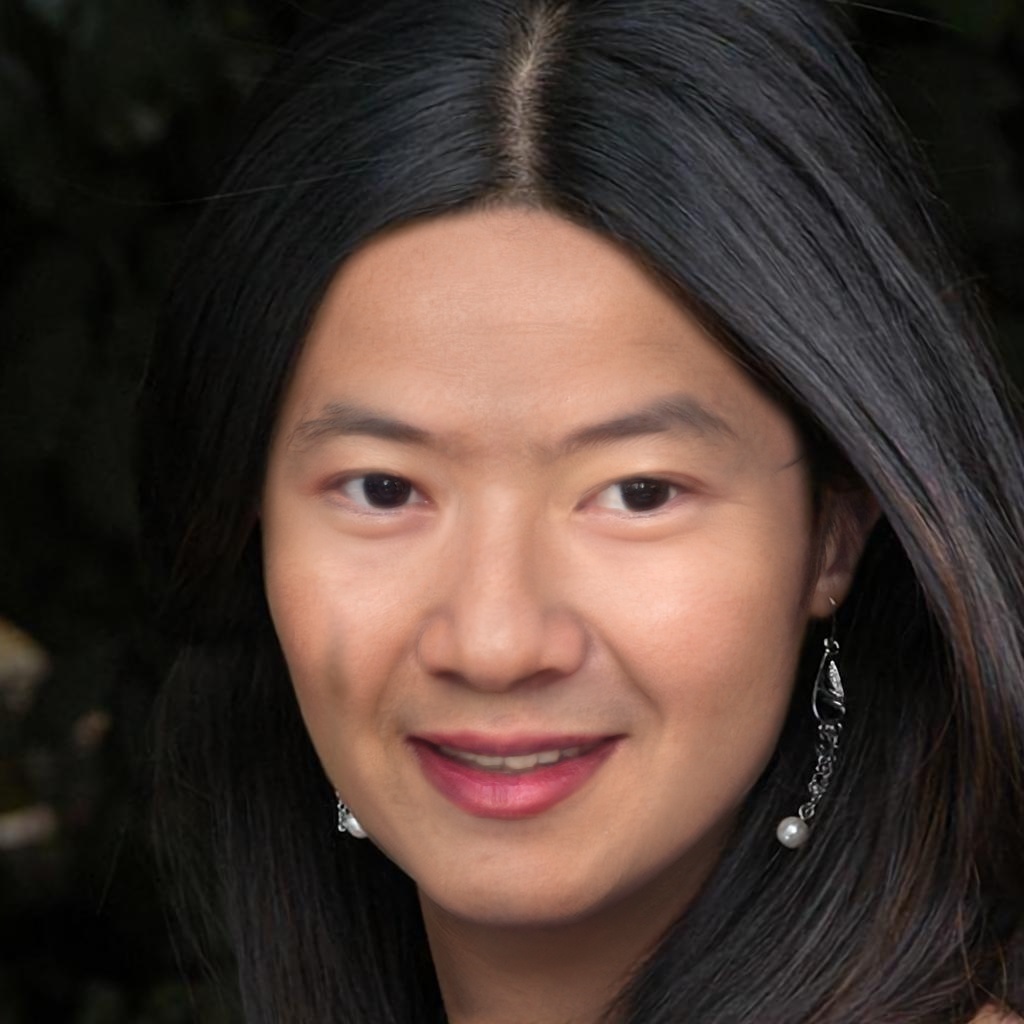}
     \caption{MegaFS}
     \end{subfigure}
     \hspace{-2mm}
     \begin{subfigure}{.12\linewidth}
     \centering
     \includegraphics[width=\linewidth]{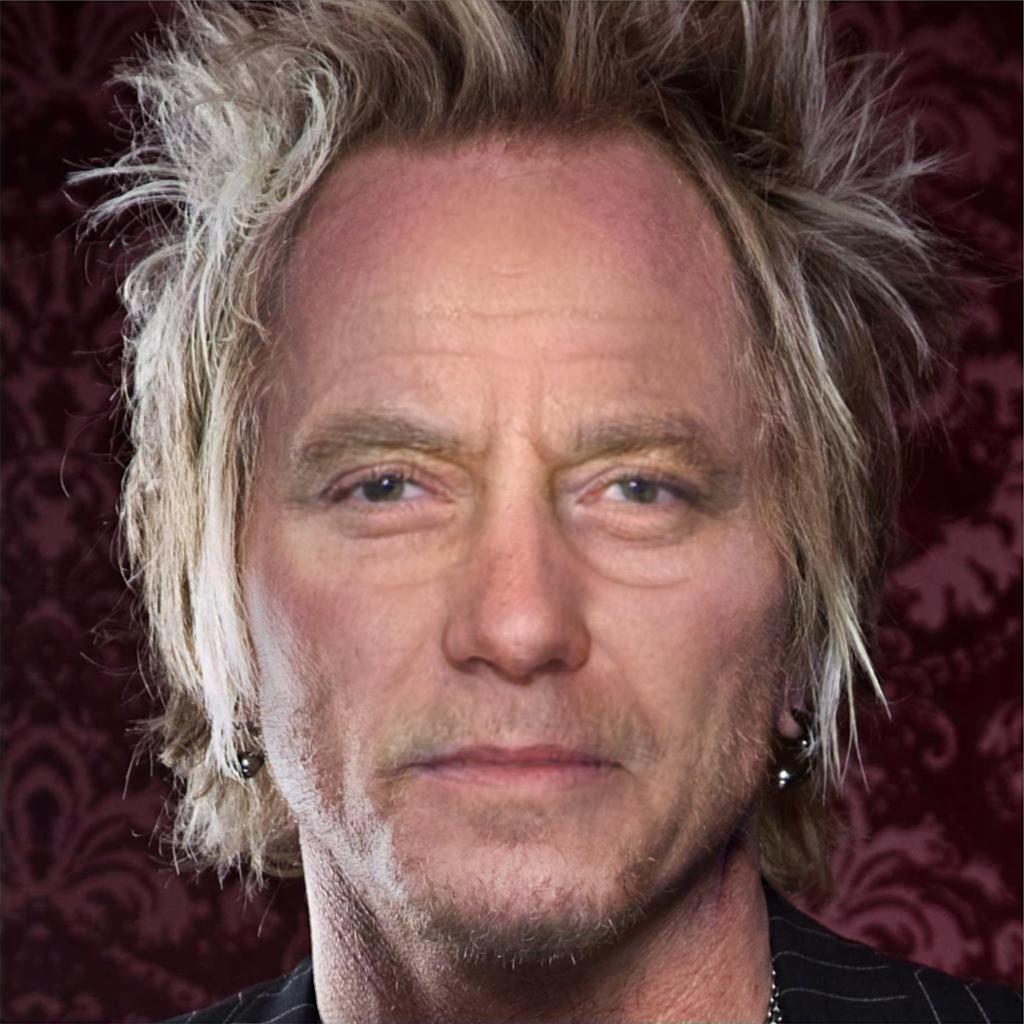}
     \includegraphics[width=\linewidth]{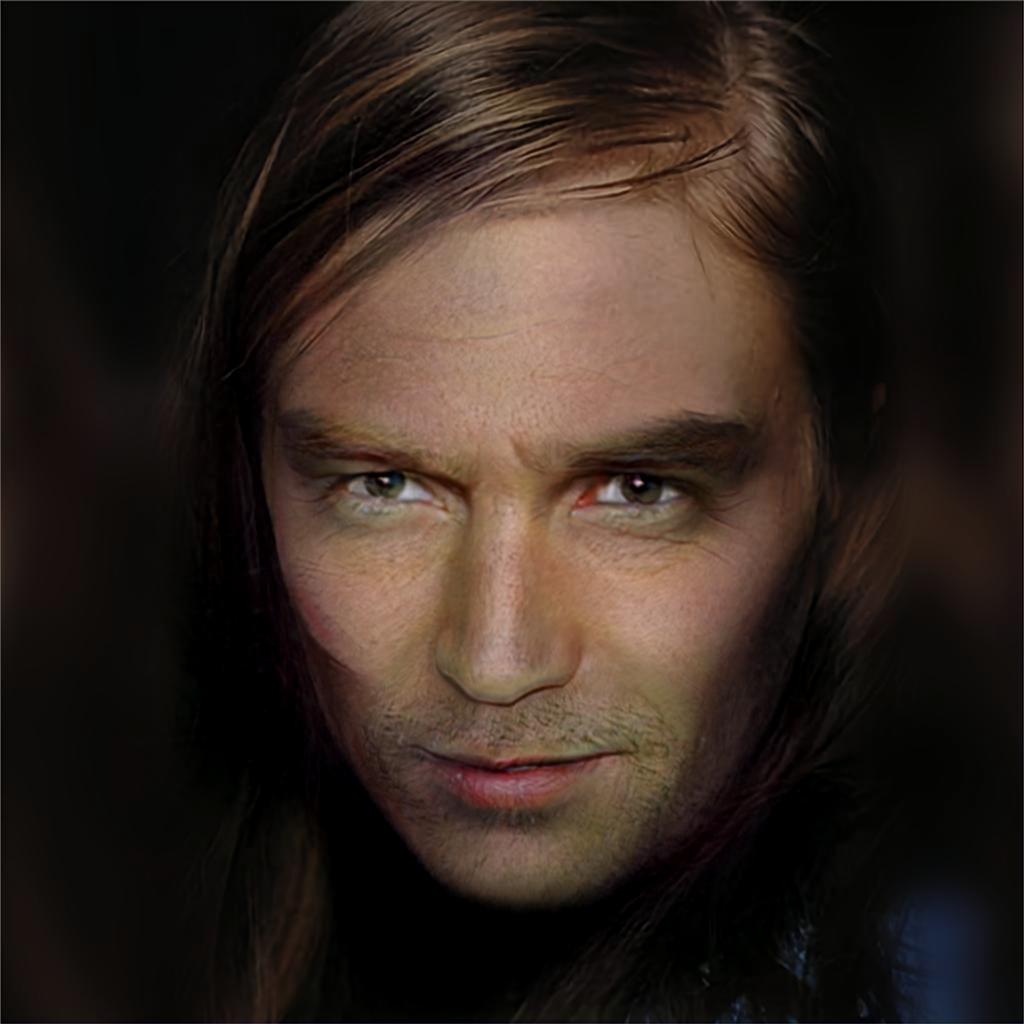}
     \includegraphics[width=\linewidth]{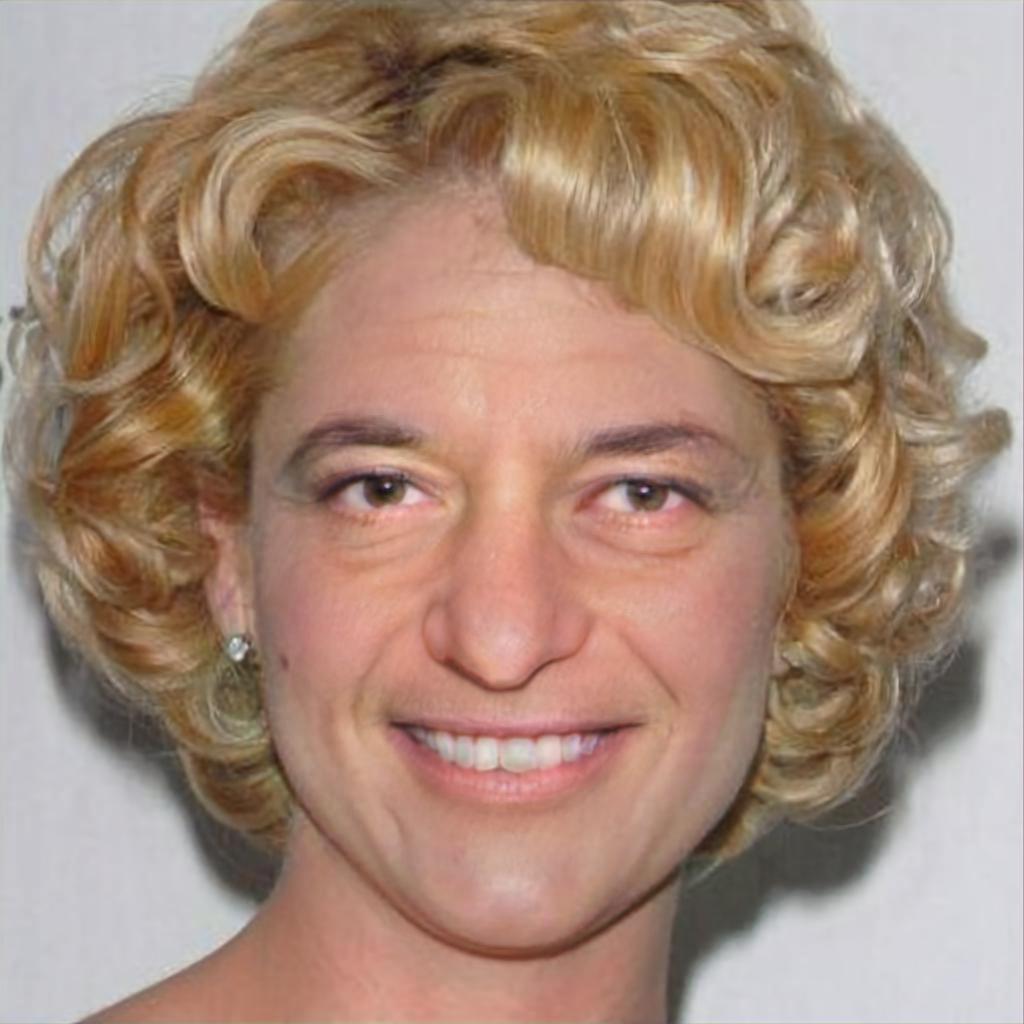}
     \includegraphics[width=\linewidth]{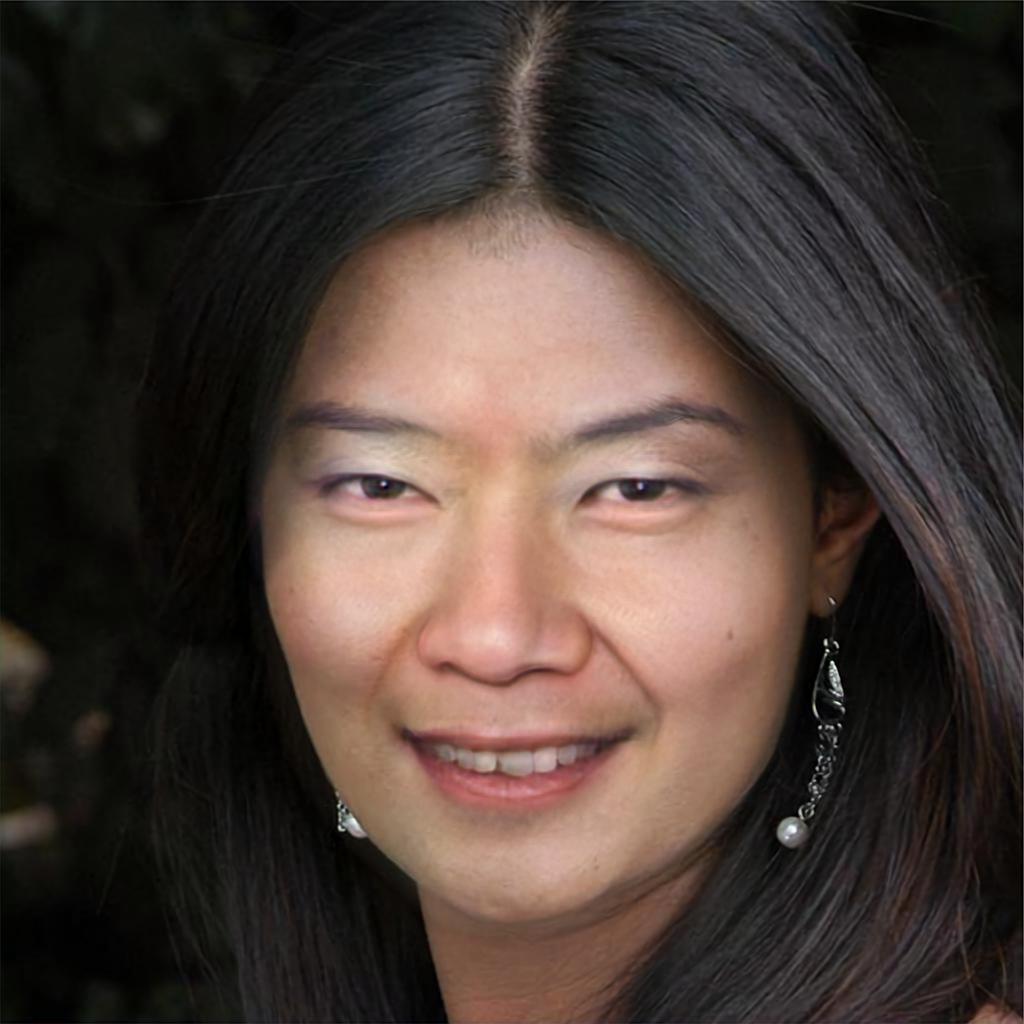}
     \caption{Ours}
     \end{subfigure}
     % \hspace{-2mm}
     \begin{subfigure}{.12\linewidth}
     \centering
     \includegraphics[width=\linewidth]{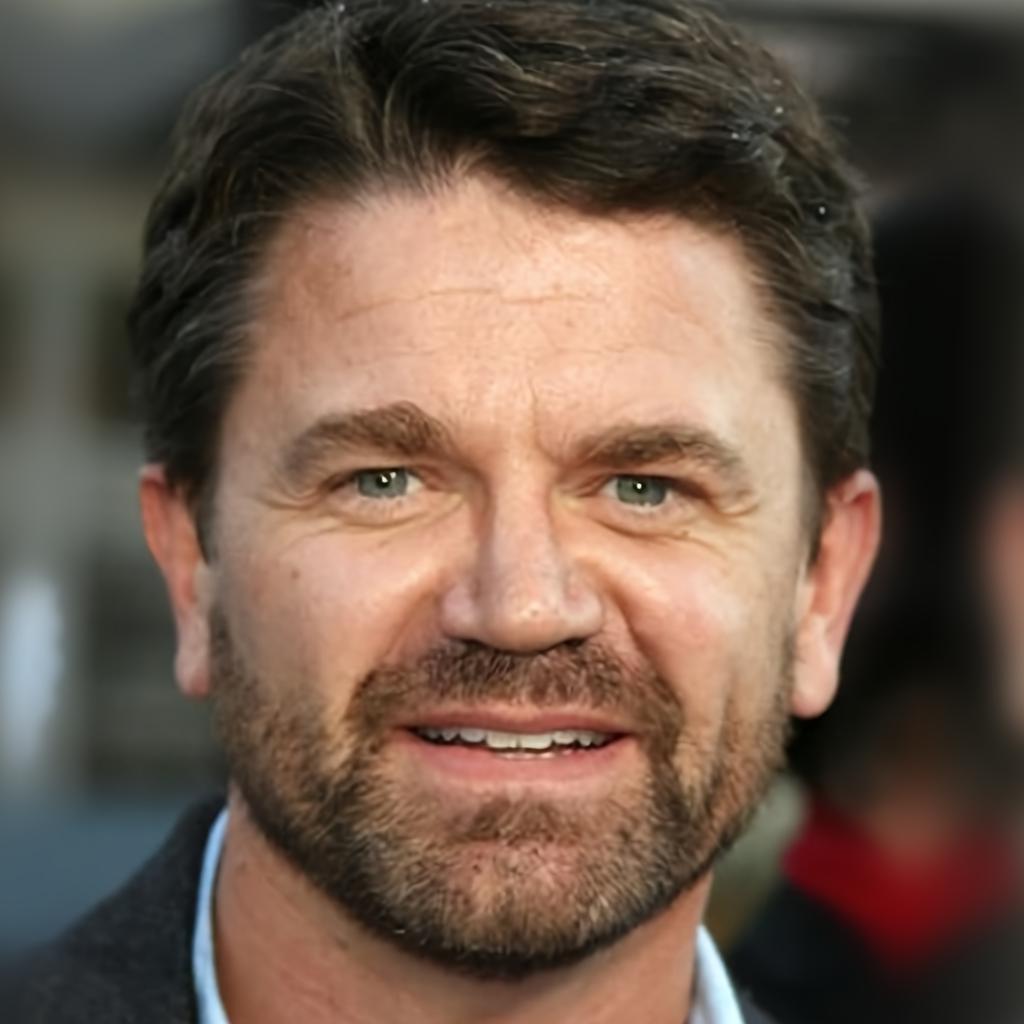}
     \includegraphics[width=\linewidth]{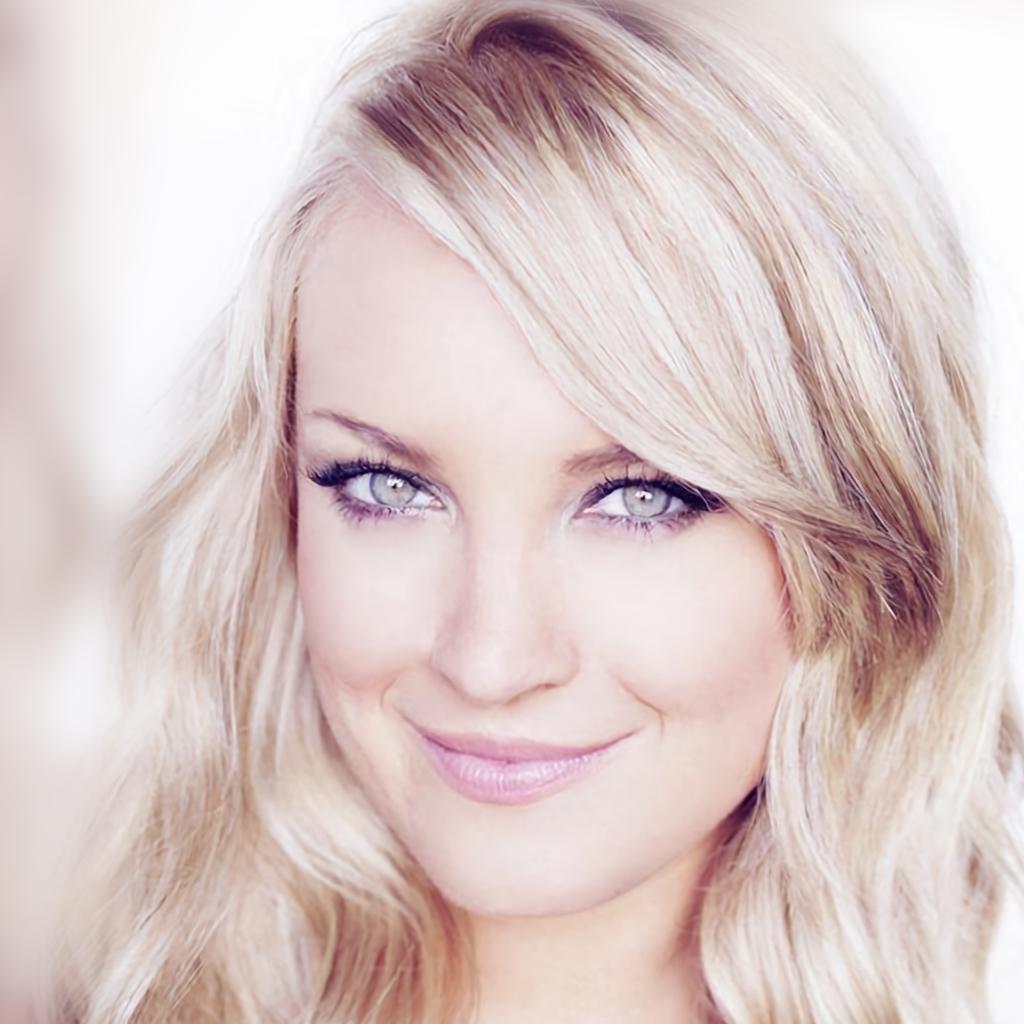}
     \includegraphics[width=\linewidth]{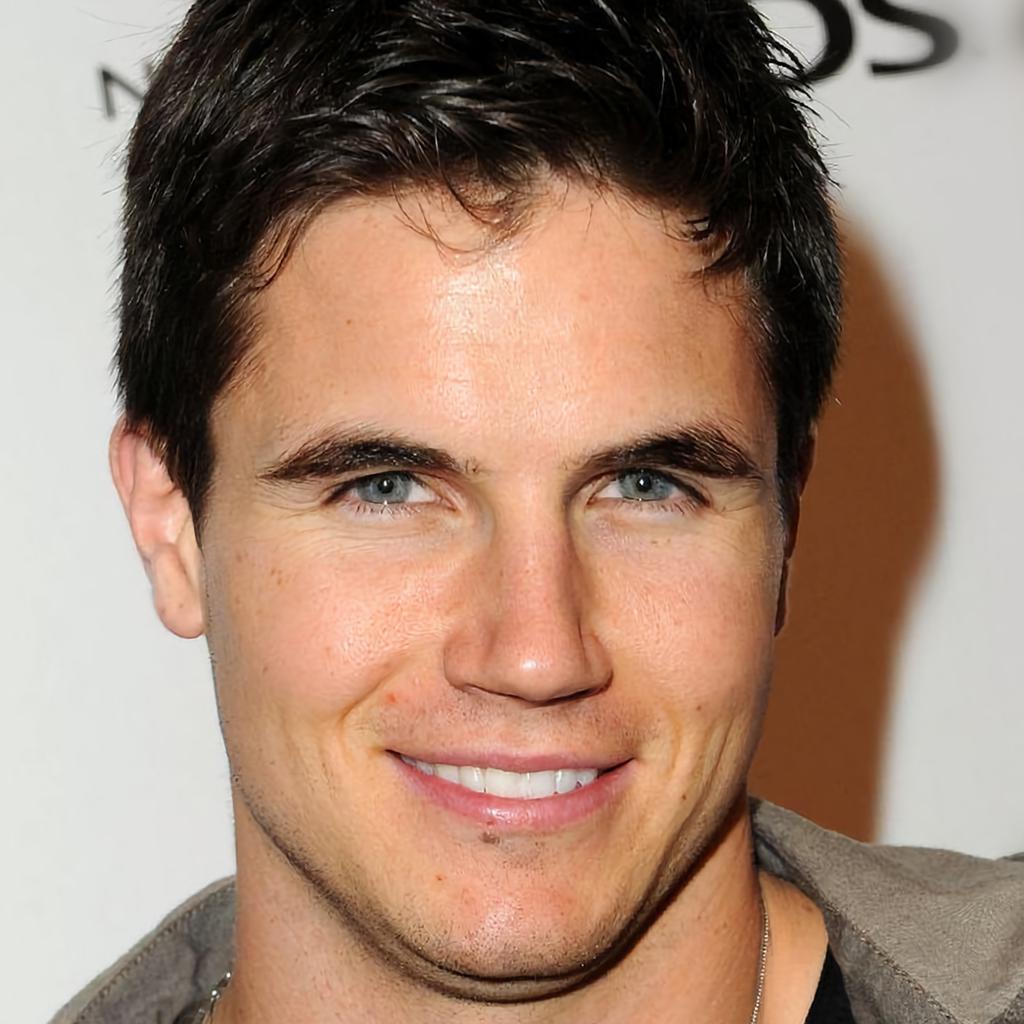}
     \includegraphics[width=\linewidth]{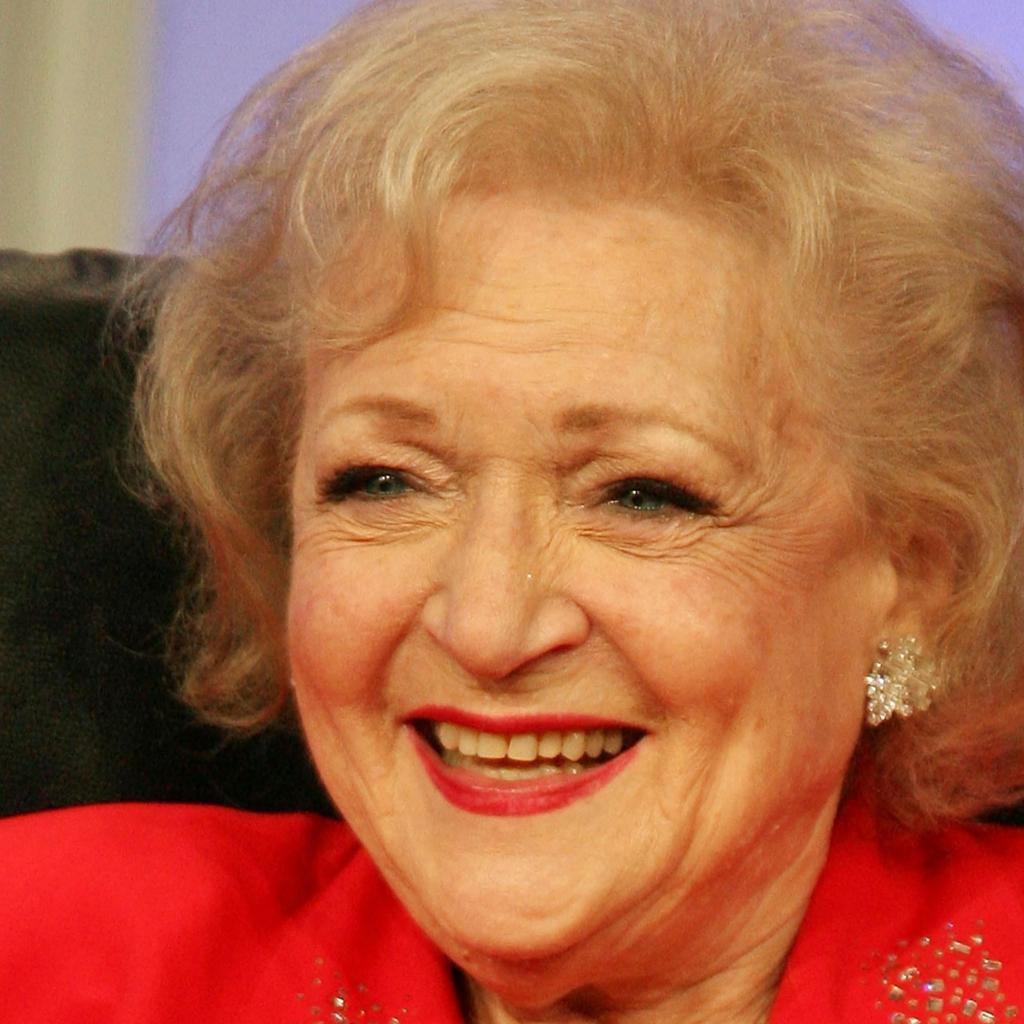}
     \caption{Source}
     \end{subfigure}
     \hspace{-2mm}
     \begin{subfigure}{.12\linewidth}
     \centering
     \includegraphics[width=\linewidth]{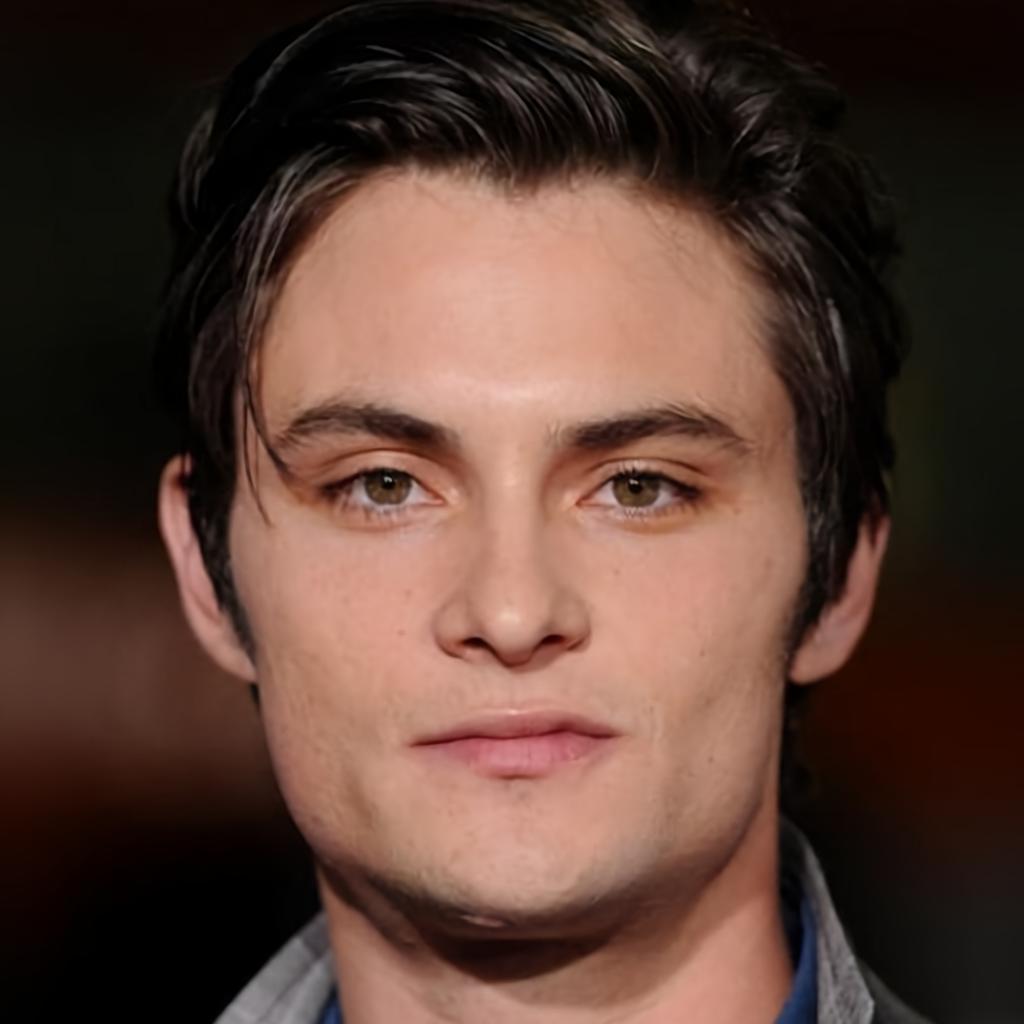}
     \includegraphics[width=\linewidth]{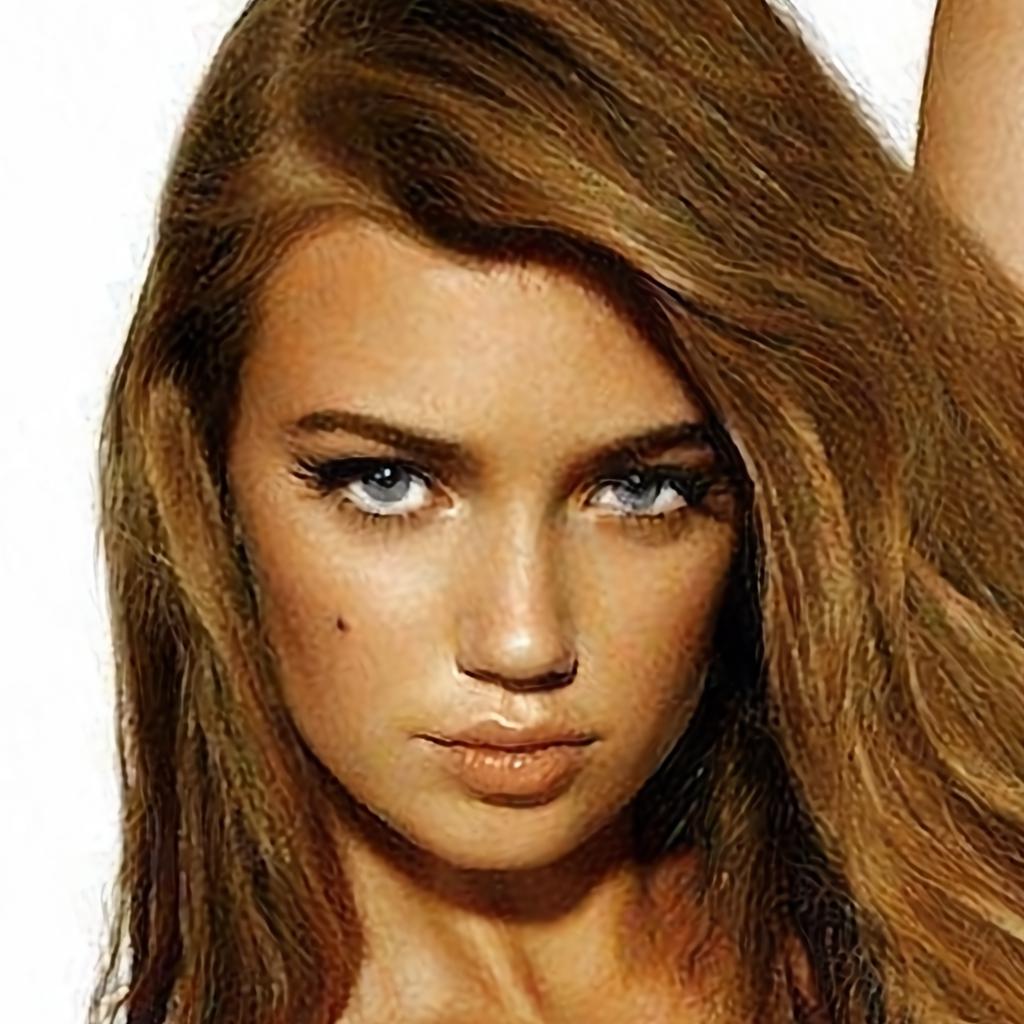}
     \includegraphics[width=\linewidth]{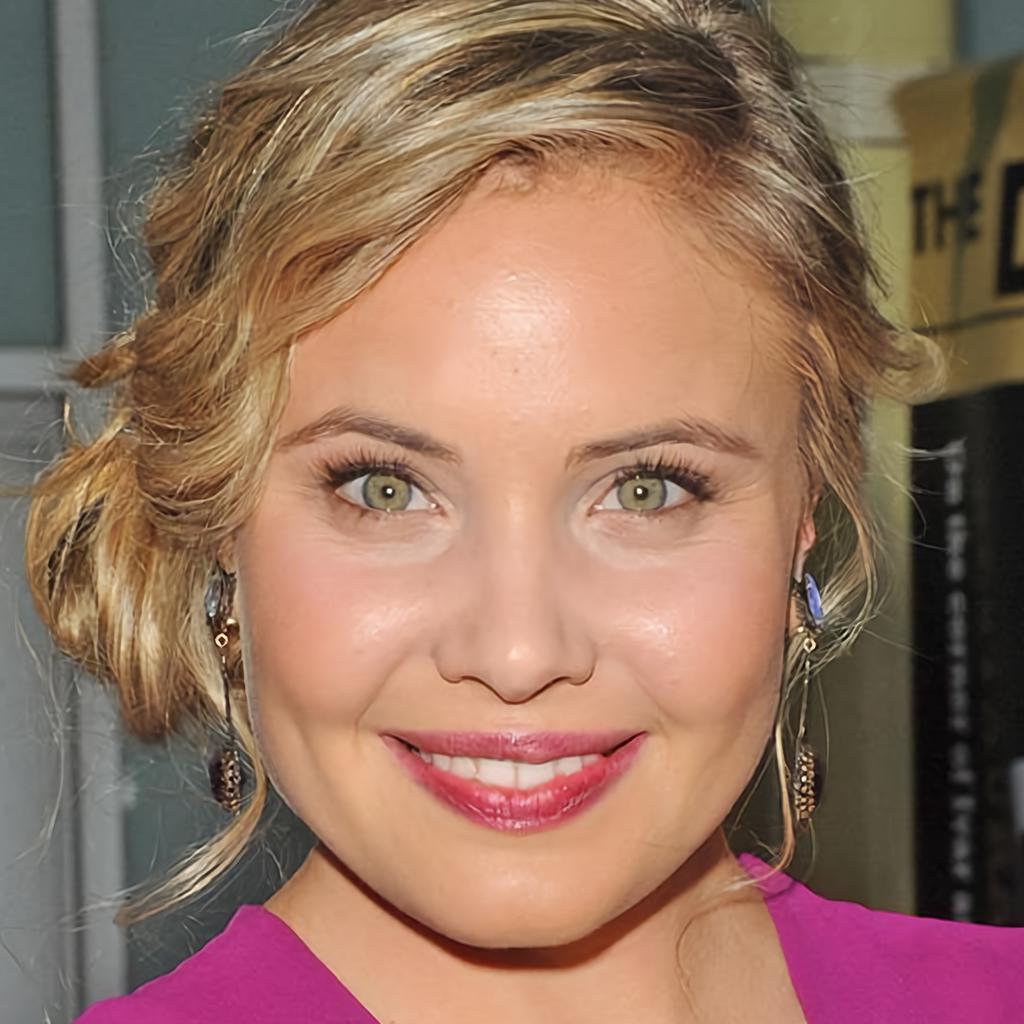}
     \includegraphics[width=\linewidth]{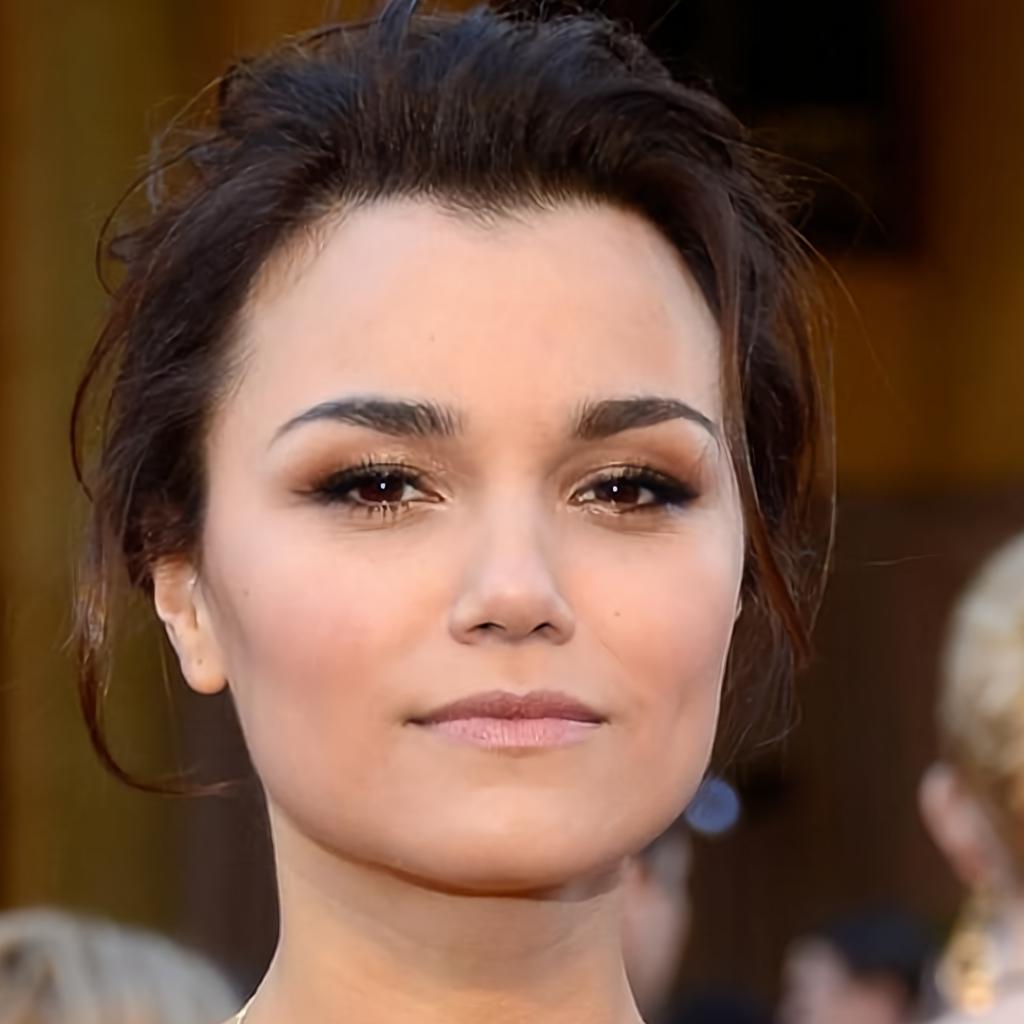}
     \caption{Target}
     \end{subfigure}
     \hspace{-2mm}
     \begin{subfigure}{.12\linewidth}
     \centering
     \includegraphics[width=\linewidth]{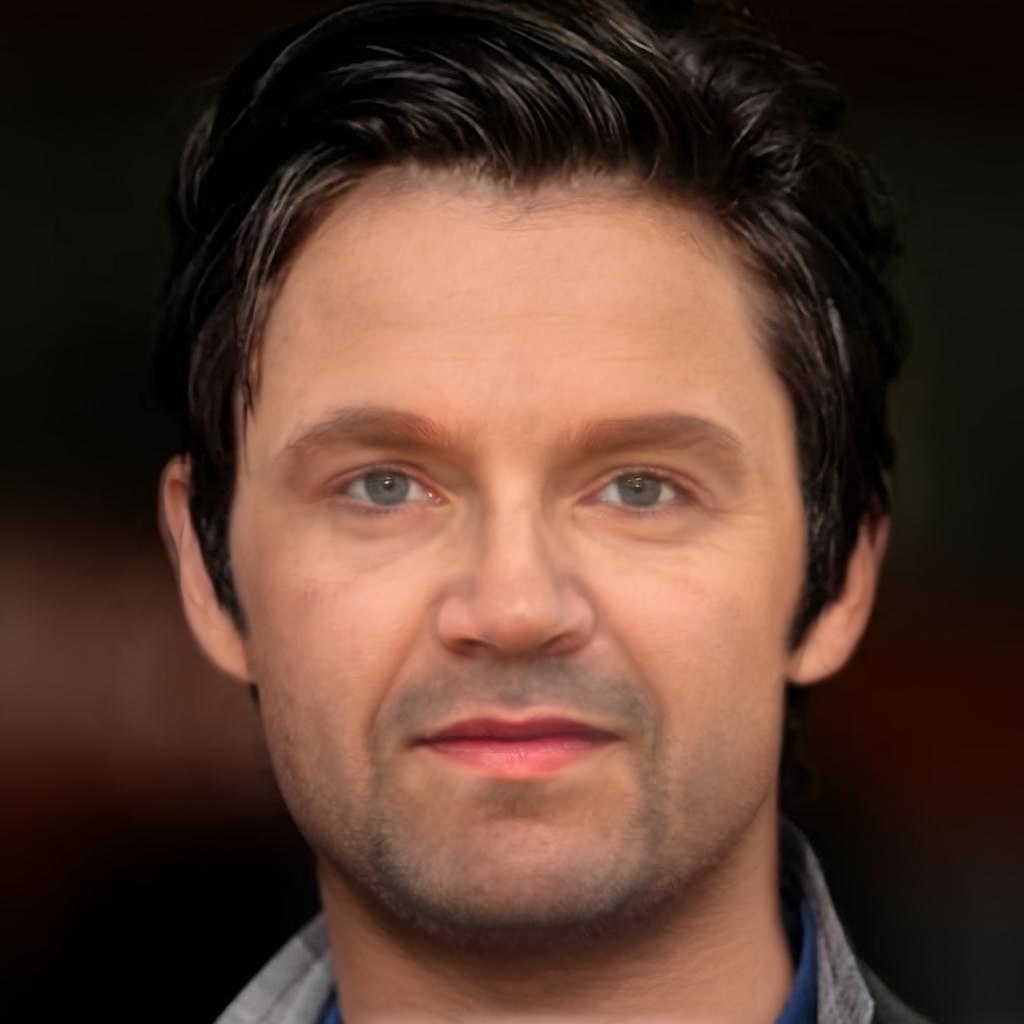}
     \includegraphics[width=\linewidth]{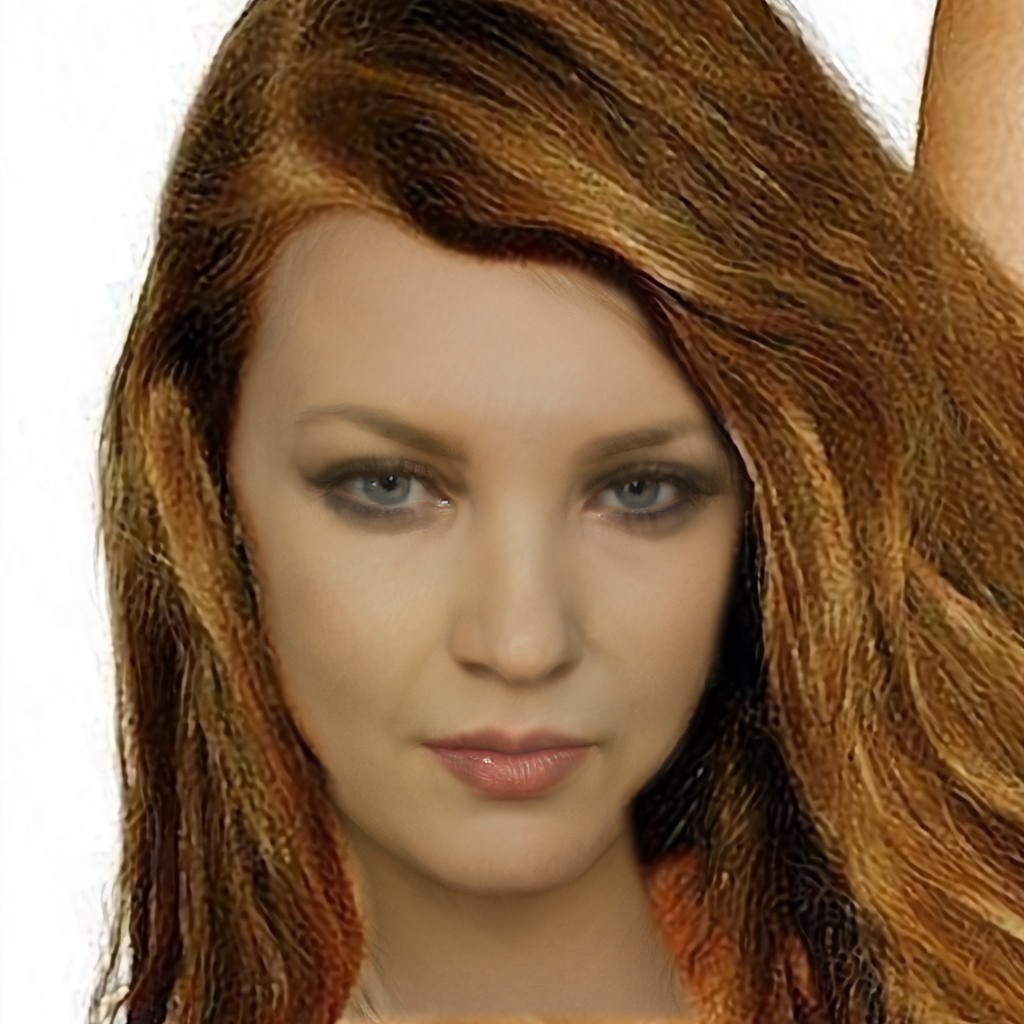}
     \includegraphics[width=\linewidth]{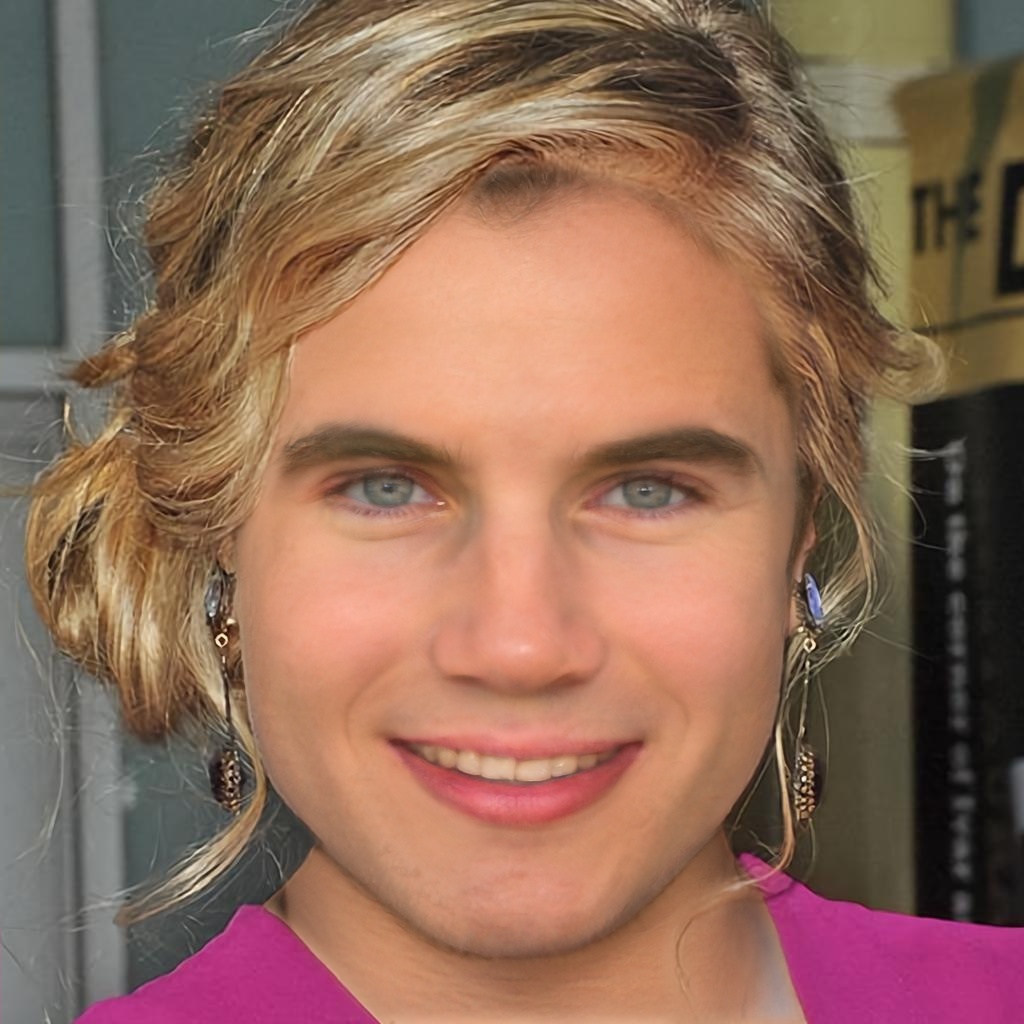}
     \includegraphics[width=\linewidth]{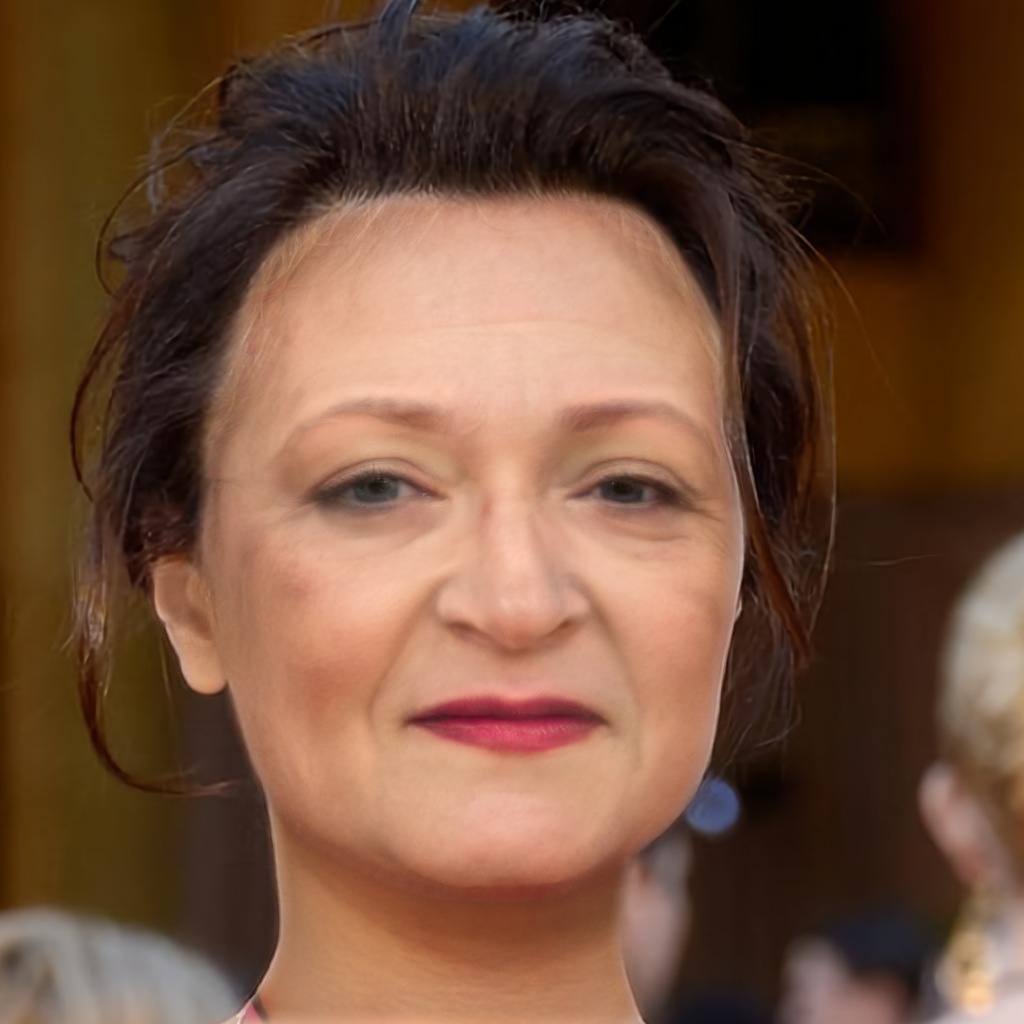}
     \caption{MegaFS}
     \end{subfigure}
     \hspace{-2mm}
     \begin{subfigure}{.12\linewidth}
     \centering
     \includegraphics[width=\linewidth]{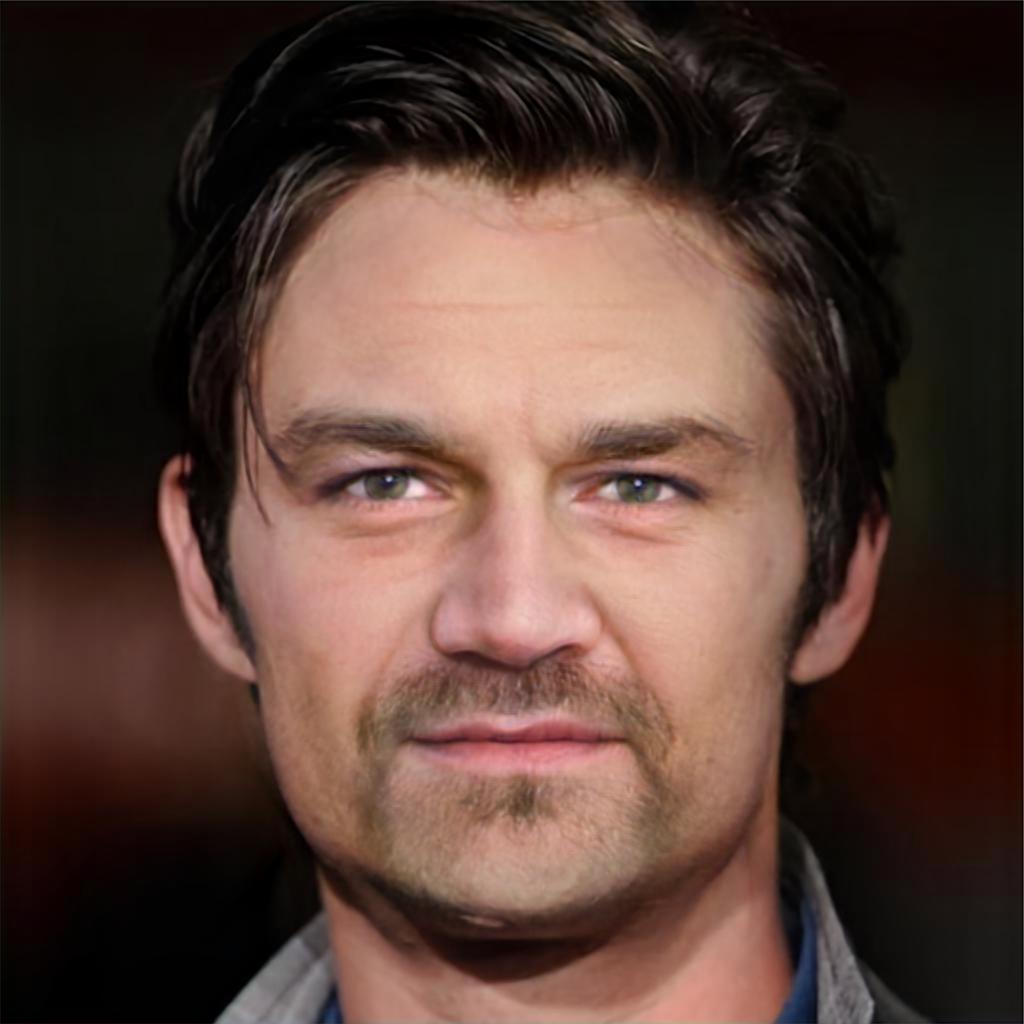}
     \includegraphics[width=\linewidth]{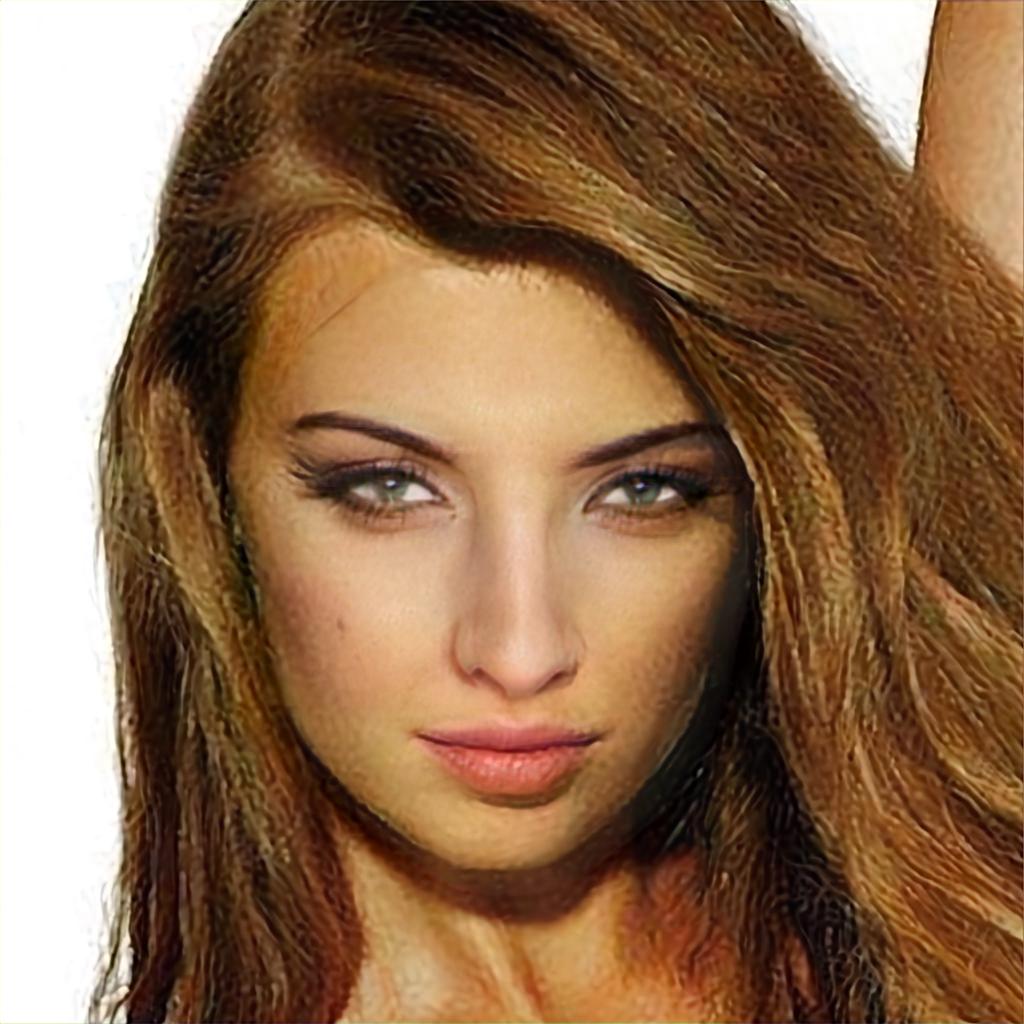}
     \includegraphics[width=\linewidth]{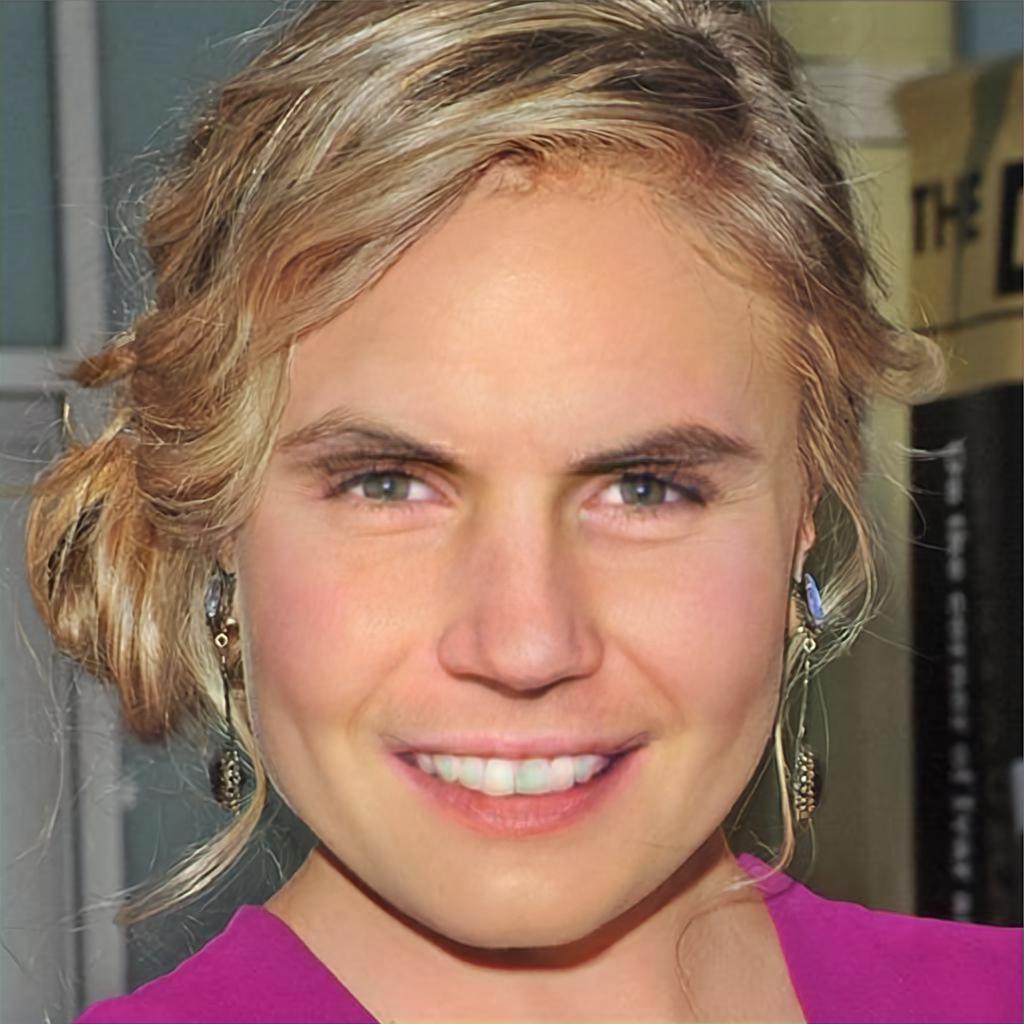}
     \includegraphics[width=\linewidth]{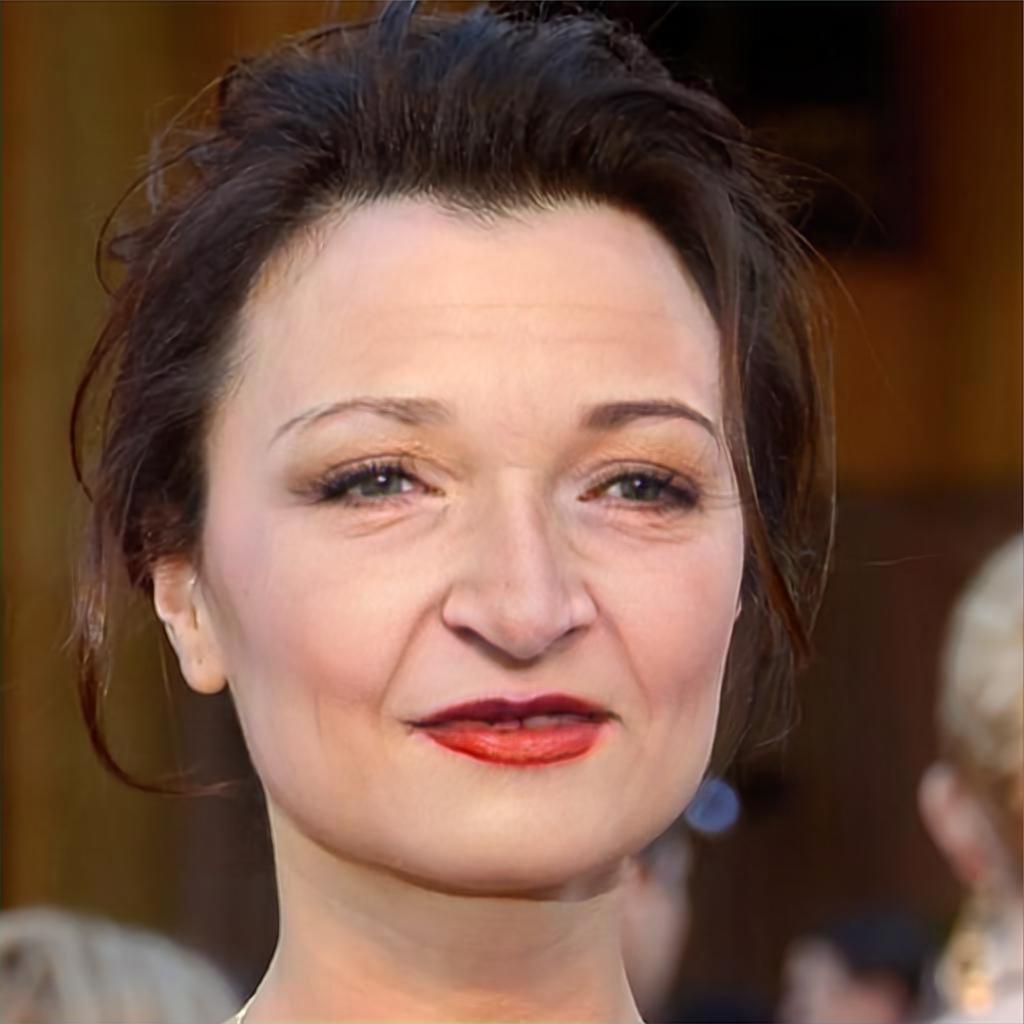}
     \caption{Ours}
     \end{subfigure}
     \capvspace
     \caption{Qualitative comparison of face swapping on the resolution of $1024$$\times$$1024$. We can see that MegaFS always present the blurry faces with a distinct boundary with the background. Besides, they cannot preserve the source identity effectively during the swapping process. Meanwhile, when the appearance attributes between source and target are huge, they cannot transfer the target to the output. In opposite, our method can transfer the structure and appearance attributes as desired with identity-preserving. Zoomed in for the best view.}%\vspace{-3mm}
     \label{fig.celeba}
     \figbottomvspace
\end{figure*}

\paraheading{Flow Trajectory Constraint}
For appearance consistency,  we follow the local linear model~\cite{liu2017video,meyer2015phase} that assumes uniform changes between neighboring frames~\footnote{Here we follow the local linear assumption for simplicity, we can also follow the acceleration-aware assumption~\cite{xu2019quadratic,liu2020enhanced} for better approximating real-world scene motion.}.
% and $y_f^{n+2}$ (limited by the GPU's memory size, we only take the relation among three frames into consideration),
Specifically, we denote the optical flow from the frame $y_f^{i}$ to a nearby frame $y_f^{j}$ in the swapped face sequence as:
%\vspace{-2mm}
\begin{equation*}
f_{i \Rightarrow j} = \Phi(y_f^{i},y_f^{j}),
\end{equation*}
where $\Phi(\cdot,\cdot)$ is the pre-trained PWC-Net~\cite{sun2018pwc} for the flow prediction. %Based on the above equation, we also can obtain the backward
From the local linear assumption, the dense correspondence between two neighboring frames in the swapped face sequence can be approximated by interpolating the forward flow and backward flow across the two frames, which leads to the following loss:
\begin{equation*}
{L}_{ft} = \sum\nolimits_{k} \| (f_{k \Rightarrow k+2} + f_{k+2 \Rightarrow k})/2 - f_{k \Rightarrow k+1} \|_2.
\end{equation*}
We note that ${L}_{ft}$ is effectively the $\ell_1/\ell_2$ norm for the temporal Laplacians of the optical flows. This promotes sequences that have small Laplacians at most frames while allowing for large Laplacians at some frames~\cite{Bach2012}, which allows for motion sequences that are piecewise smooth temporally.
%This constraint is applied on the image space.

% However, this assumption cannot approximate the complex motion in real world well and often leads to inaccurately interpolated results. To solve this problem, we propose a quadratic interpolation method for predicting more accurate intermediate frames

% \rev{The term $(f_{k \Rightarrow k+2} + f_{k+2 \Rightarrow k})/2 - f_{k \Rightarrow k+1}$ is effectively the Laplacian of the optical flow at frame $k$, and $L_{ft}$ is an $\ell_1$ regularizer of the Laplacian norms for all frames. As $\ell_1$ minimization induces sparsity, $L_{ft}$ promotes sequences that have small Laplacian norms at most frames while allowing for large Laplacian norms at some frames \cite{Bach2012}. This allows for motions that are \emph{piecewise} smooth, which is the case for most face videos. Similar $\ell_1$ formulation has also been used in \cite{Grundmann2011} to compute camera paths consisting of smooth segments.}
% We further note that the Laplacian terms impose a \emph{local linear} regularization on the sequence, which is widely used for video frame interpolation (see references [28,30] of our paper).

% !TeX root = main.tex
\section{Experiments}
\subsection{Implementation Details}

We utilize the StyleGAN2 generator~\cite{karras2020analyzing} pre-trained on the FFHQ dataset~\cite{karras2019style} with the resolution of $1024\times 1024$. The pSp encoder is also pre-trained on this dataset. We use a modification of pSp as our landmark (more details of the structure can be seen in the supplementary materials). We use the Adam optimizer~\cite{kingma2014adam} to train the model with a learning rate of $1 \times 10^{-4}$, and the exponential decay rates for the $1_{st}$ and $2_{nd}$ moment estimate being ${\beta}_1 = 0.9$ and ${\beta}_2 = 0.999$ respectively, and ${\epsilon} = 1 \times 10^{-8}$. The batch size is set to be 8 and the model is trained with 500,000 iterations. We empirically set the weights in Eq.~\eqref{eq14} as $\lambda_{1}=1$, $\lambda_{2}=2$, $\lambda_{3}=0.1$, $\lambda_{4}=2$ and $\lambda_{5}=0.2$. We implement our method in Pytorch with four Tesla V100 GPUs. It takes about two days to train the whole model.

\subsection{Datasets}

\topparaheading{Datasets} We evaluate our model on three datasets:
\begin{itemize}[leftmargin=*,nosep]
\item \textbf{CelebA-HQ} contains 30,000 celebrities faces with a resolution of $1024$$\times$$1024$~\cite{karras2018progressive}. Due to its high quality, it has been widely used in many face editing works.
\item \textbf{FaceForensics++} consists of 1,000 original talking videos downloaded from YouTube and manipulated with 5 face swapping methods~\cite{rossler2019faceforensics}. This dataset serves as a benchmark in many face swapping works. %Though its resolution is not that high, we still evaluate our method on this dataset.
\end{itemize}

% \item \textbf{RAVDESS} contains 2,452 videos with 24 subjects speaking and singing with various semantic expressions~\cite{livingstone2018ryerson}. This dataset has a relatively high resolution, hence, we evaluate our video-based face swapping method on this dataset.

\paraheading{Evaluation Metrics} We use several metrics in our quantitative experiments. The \emph{ID retrieval rate},
%or ID similarity, pose error, expression error and Fr\'{e}chet Inception Distance (FID)~\cite{heusel2017gans}. The ID retrieval rate,
measured by the top-1 identity matching rate based on the cosine similarity, indicates the identity preservation ability.
For some experiments, we follow MegaFS~\cite{zhu2021one} and compute the \emph{ID similarity} instead, which is the cosine similarity between swapped faces and their corresponding sources using CosFace~\cite{wang2018cosface}, to reduce the computational cost. The \emph{pose error} and the \emph{expression error} are the $\ell_2$ distance between the pose and expression feature vectors respectively using pre-trained estimators~\cite{ruiz2018fine,chaudhuri2019joint} on the swapped and target faces, which indicates the ability to transfer structure attributes. The \emph{Fr\'{e}chet Inception Distance} (FID)~\cite{heusel2017gans} computes the Wasserstein-2 distance between the distribution of real faces and swapped images, which measures the image quality of swapped faces.

\subsection{Comparison on CelebA-HQ Dataset}

\topparaheading{Qualitative Comparison} We first conduct experiments on the high-resolution CelebA-HQ dataset~\cite{karras2018progressive}. We compare our method with MegaFS~\cite{zhu2021one} that swaps faces at $1024$$\times$$1024$ resolution, with qualitative comparison results shown in Fig.~\ref{fig.celeba}. We can see the faces produced by MegaFS tend to have a blurry appearance without vivid details. This is because they perform face swapping solely on the low-dimensional latent codes without the explicit disentanglement, which easily dilutes the latent encoding of details.

Besides, there is a notable boundary between facial regions and the background in their results. In comparison, our approach utilizes multi-resolution spatial features from the StyleGAN generator and aggregates them with the background features from the target decoder, which helps to retain high-quality facial details.
Moreover, MegaFS does not transfer the target attributes to the output effectively, especially when there is a large semantic gap between the source and the target. For example, in the second row of Fig.~\ref{fig.celeba}, MegaFS cannot transfer the illumination or styles from targets, due to the large difference between the source and the target in these attributes. In comparison, thanks to our disentangled attributes transfer, our results maintain the target attributes more effectively. For the results on the left of the fourth row in Fig.~\ref{fig.celeba}, we can also see that MegaFS does not preserve the source identity, where it wrongly enlarges the eyes compared to the source. This is because the identity and the attributes are highly entangled in the latent space, and the lack of explicitly disentanglement in MegaFS can lead to unsatisfactory results. On the contrary, we transfer different levels of attributes separately, which helps to maintain the source identity and the target attributes.

\begin{figure}[t]
   \centering
    \captionsetup[subfigure]{labelformat=empty,justification=centering}
    \begin{subfigure}{.133\linewidth}
     \centering
     \includegraphics[width=\linewidth]{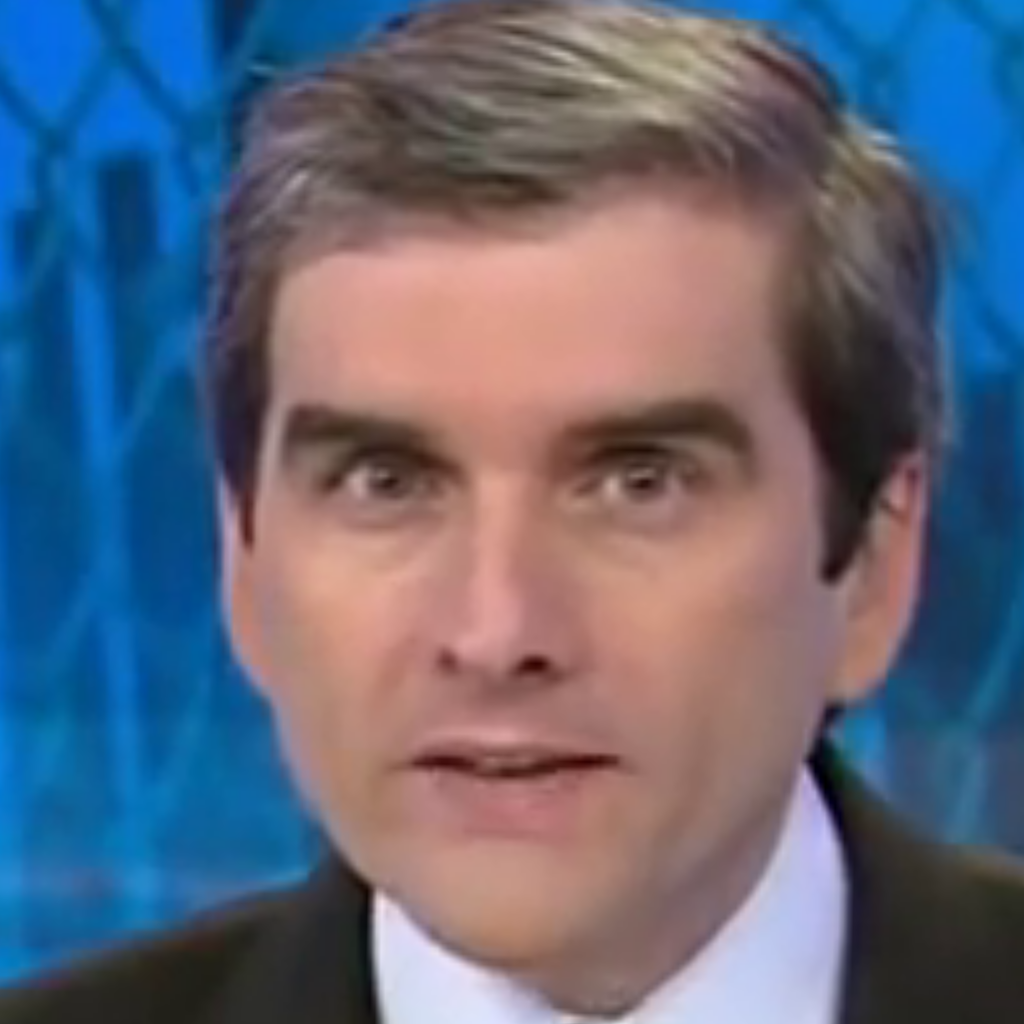}
     \includegraphics[width=\linewidth]{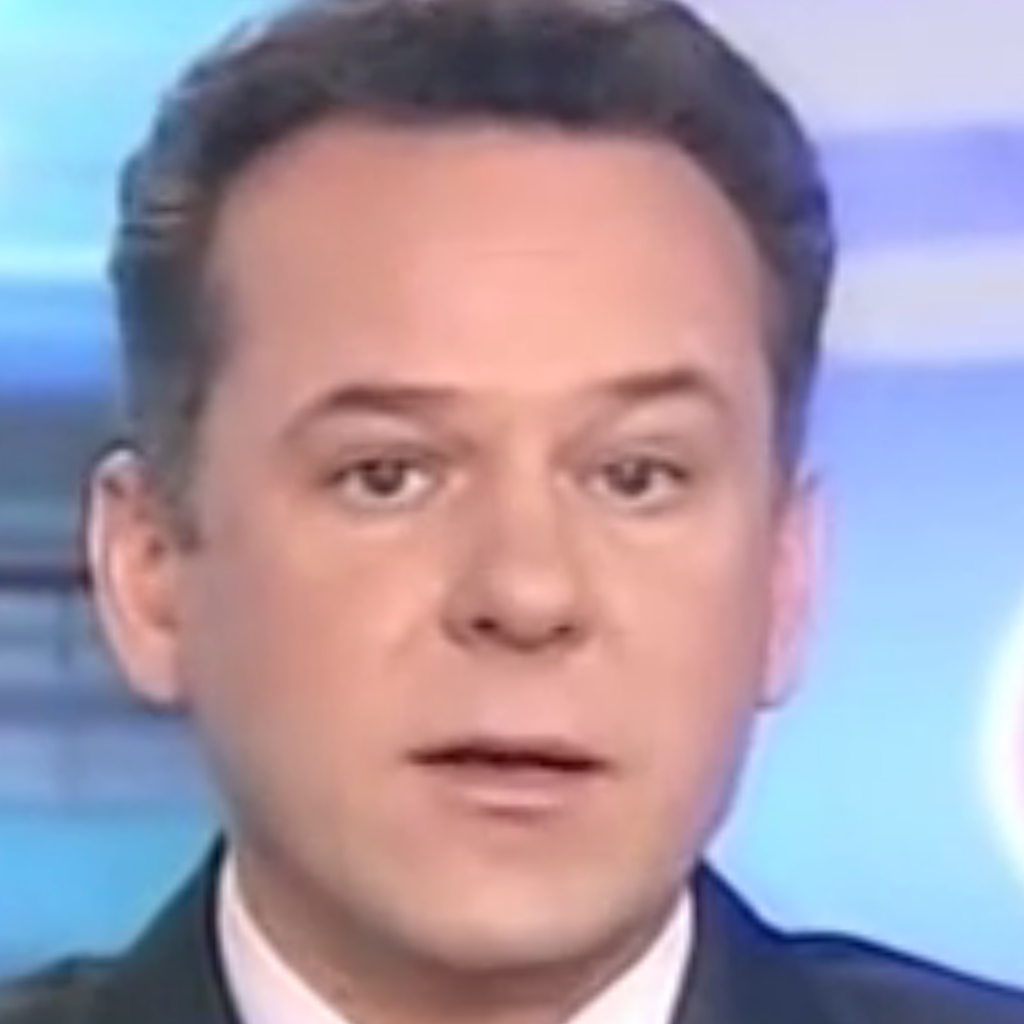}
     \includegraphics[width=\linewidth]{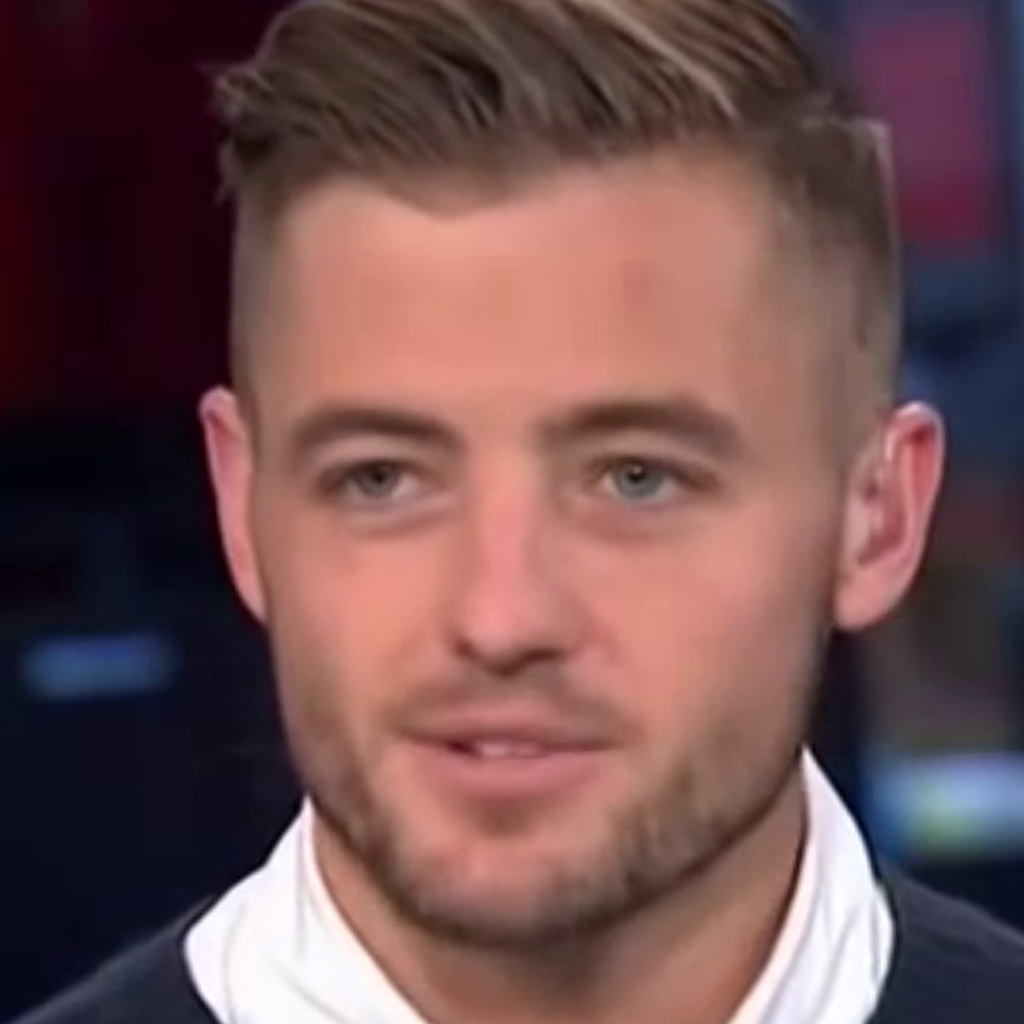}
     \includegraphics[width=\linewidth]{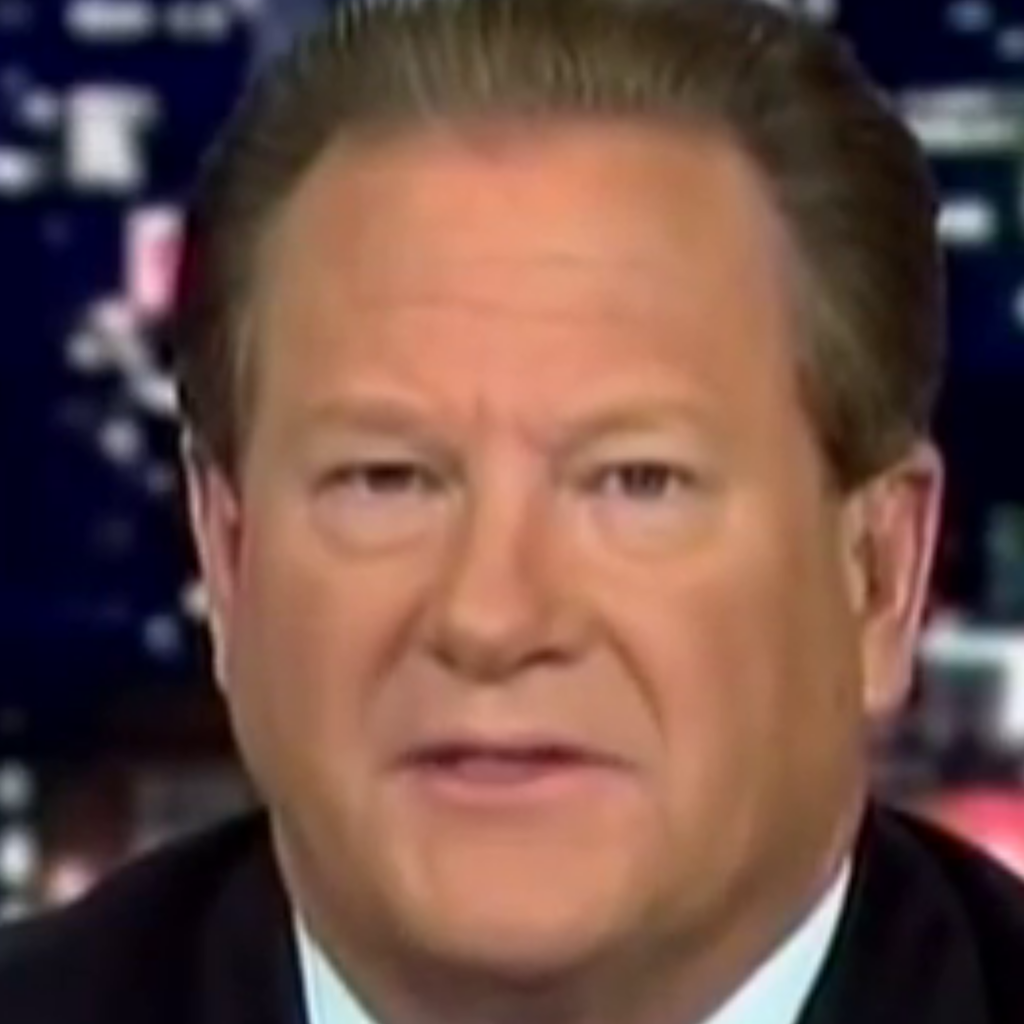}
     \caption{\subcapfont{Source}}
     \end{subfigure}
     \hspace{-1.5mm}
     \begin{subfigure}{.133\linewidth}
     \centering
     \includegraphics[width=\linewidth]{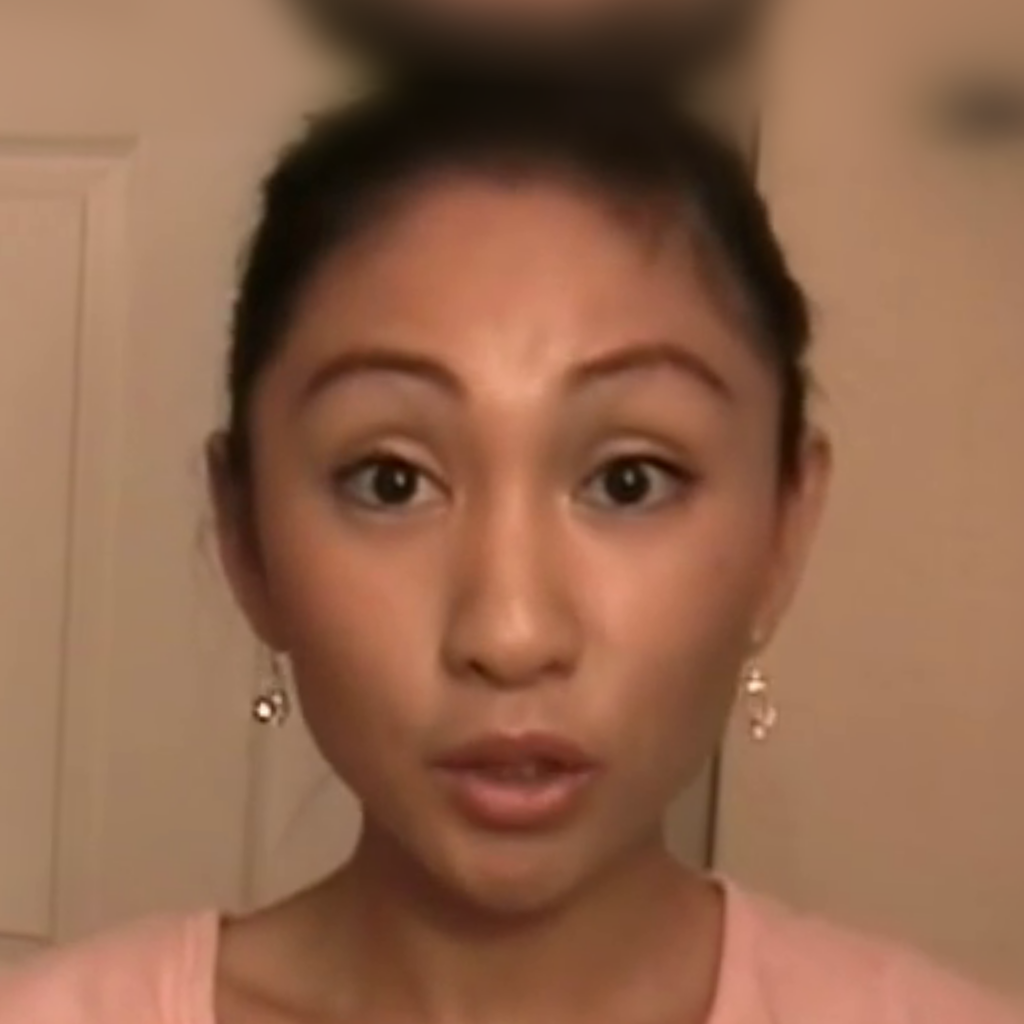}
     \includegraphics[width=\linewidth]{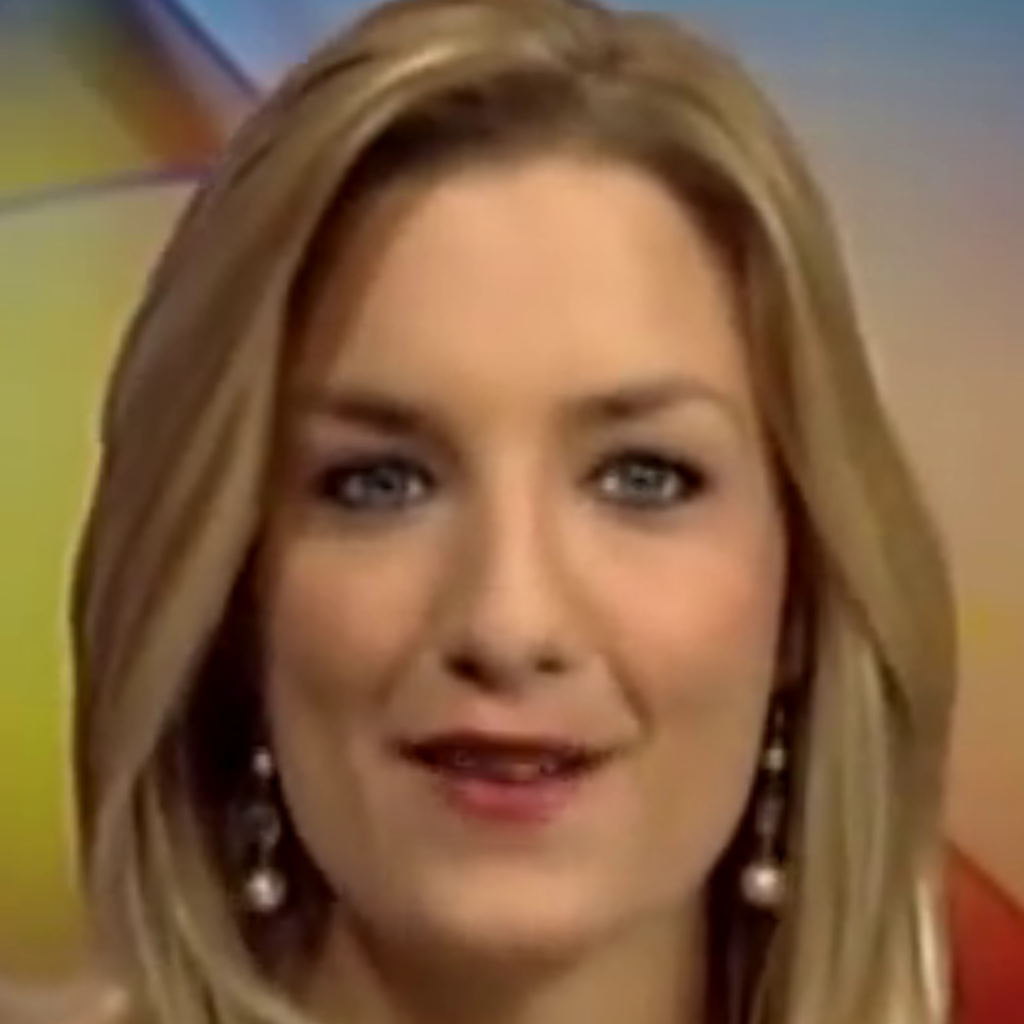}
     \includegraphics[width=\linewidth]{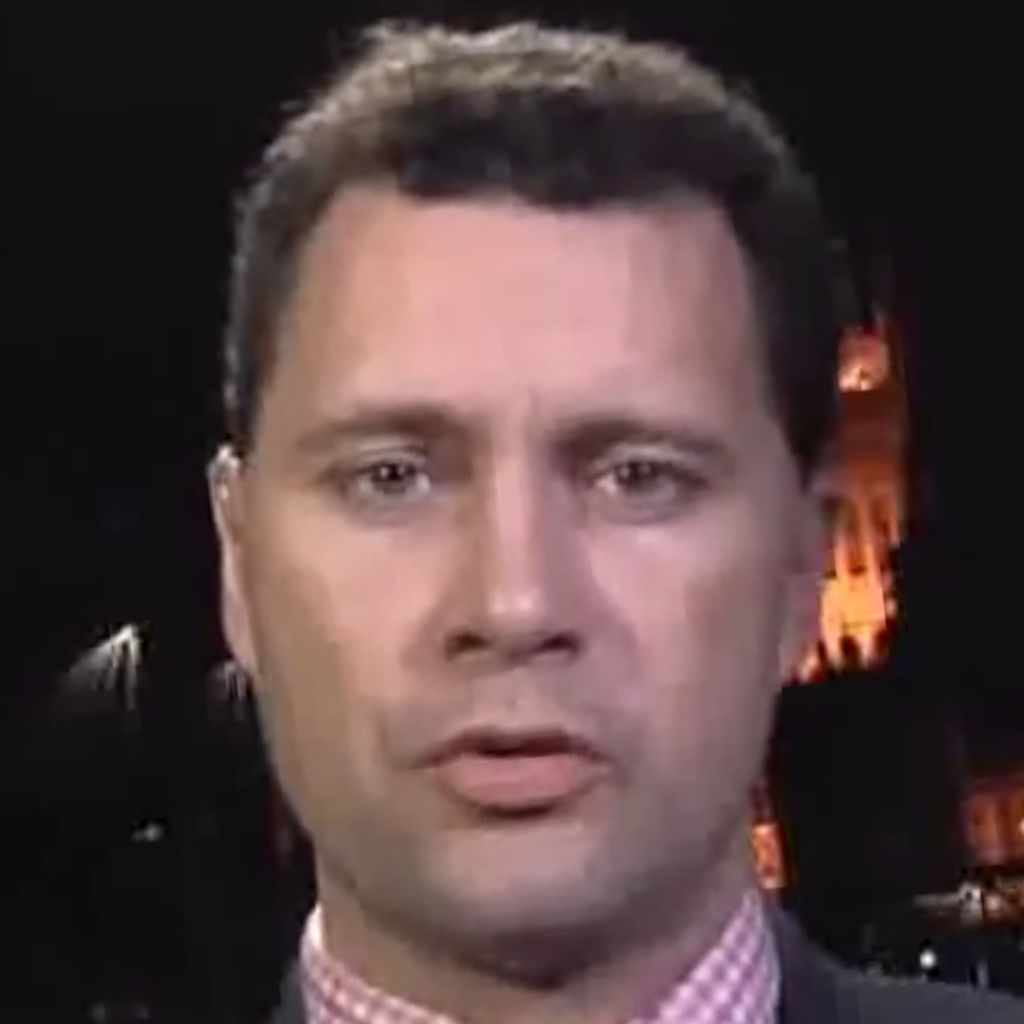}
     \includegraphics[width=\linewidth]{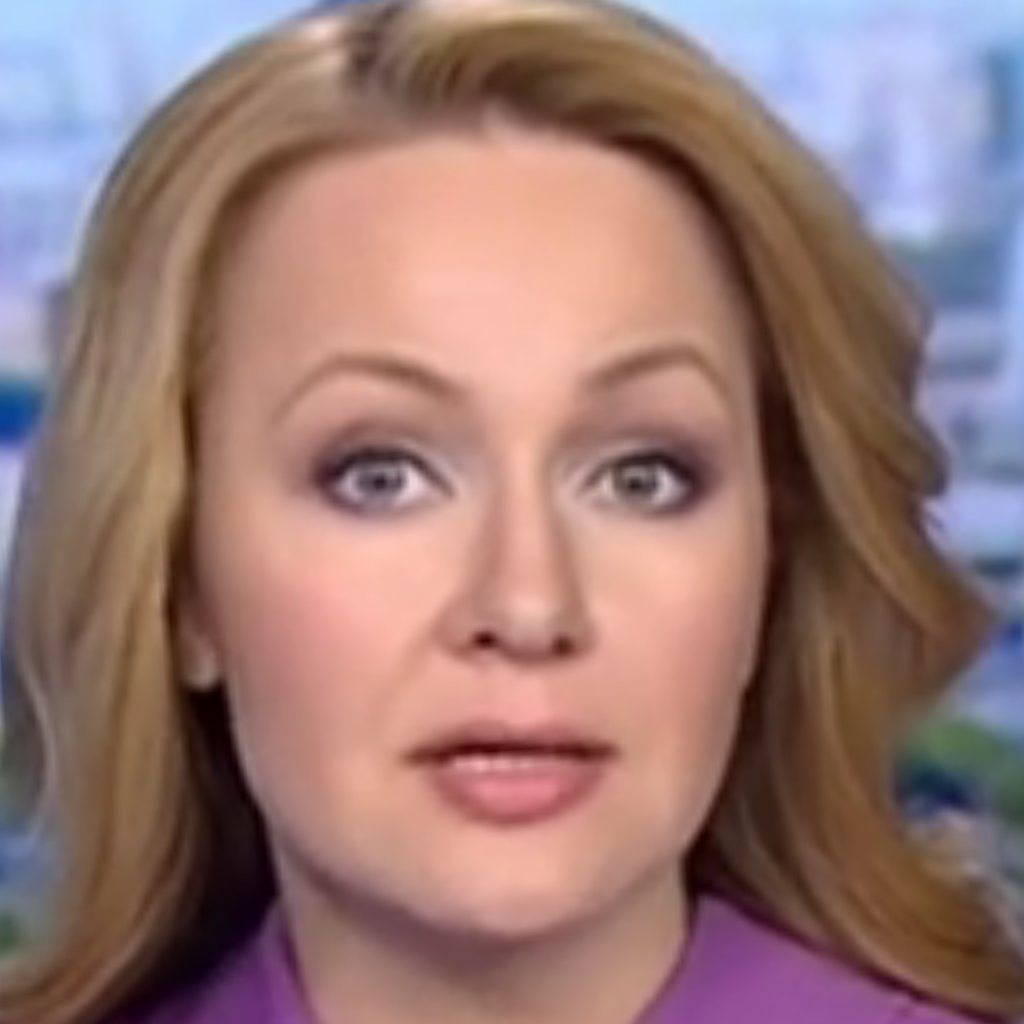}
     \caption{\subcapfont{Target}}
     \end{subfigure}
     \hspace{-1.5mm}
     \begin{subfigure}{.133\linewidth}
     \centering
     \includegraphics[width=\linewidth]{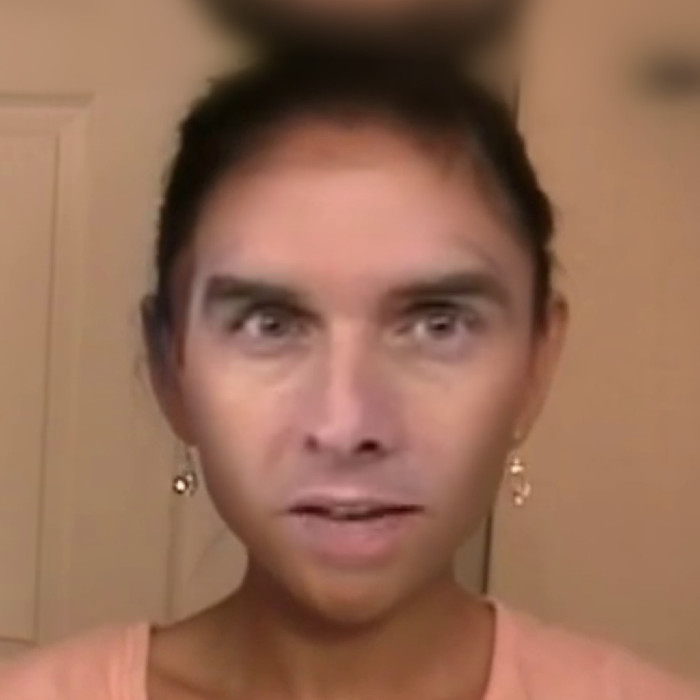}
     \includegraphics[width=\linewidth]{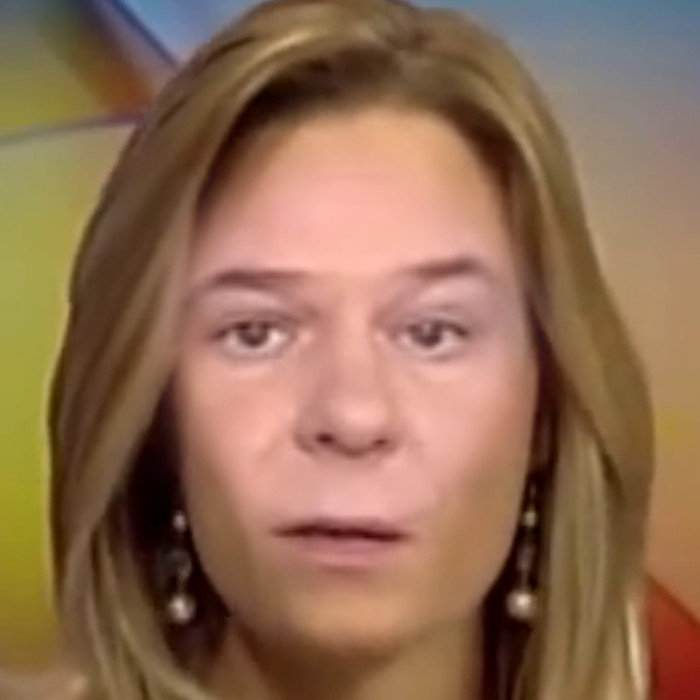}
     \includegraphics[width=\linewidth]{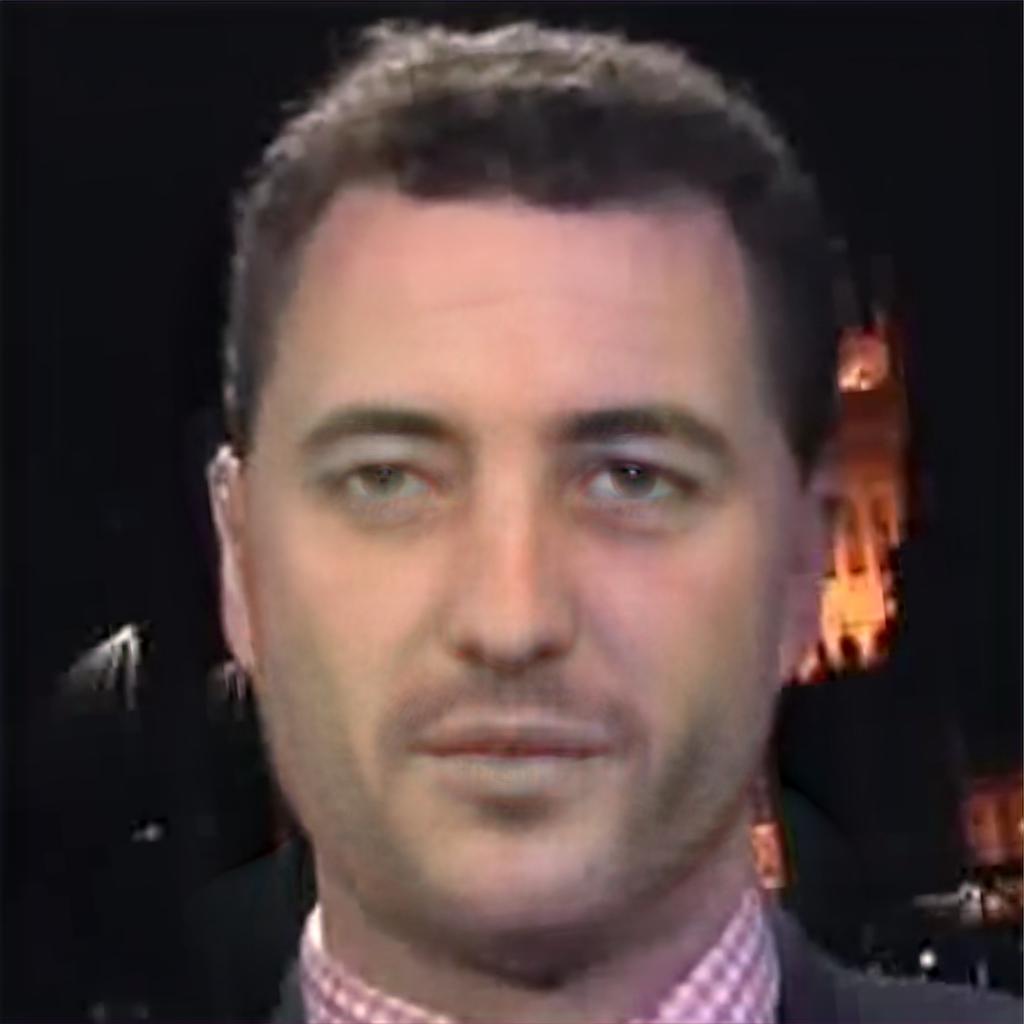}
     \includegraphics[width=\linewidth]{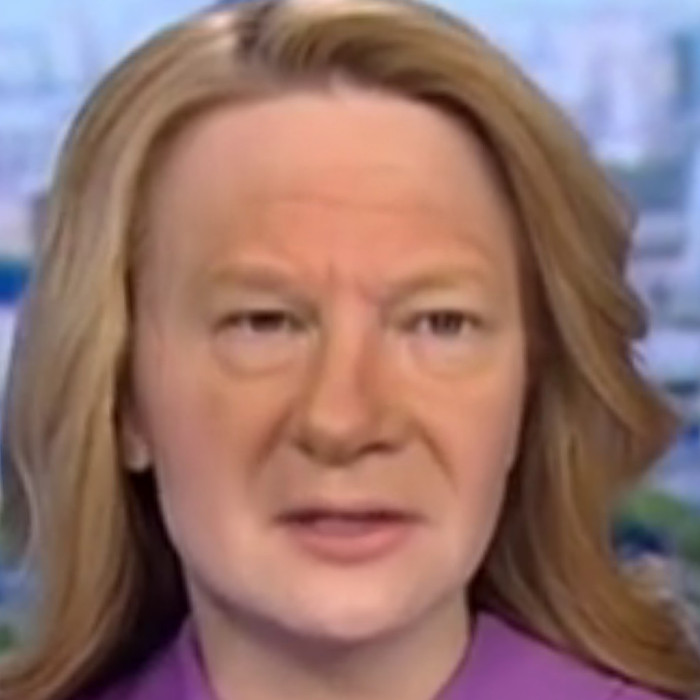}
     \caption{\subcapfont{FaceSwap}}
     \end{subfigure}
     \hspace{-1.5mm}
     \begin{subfigure}{.133\linewidth}
     \centering
     \includegraphics[width=\linewidth]{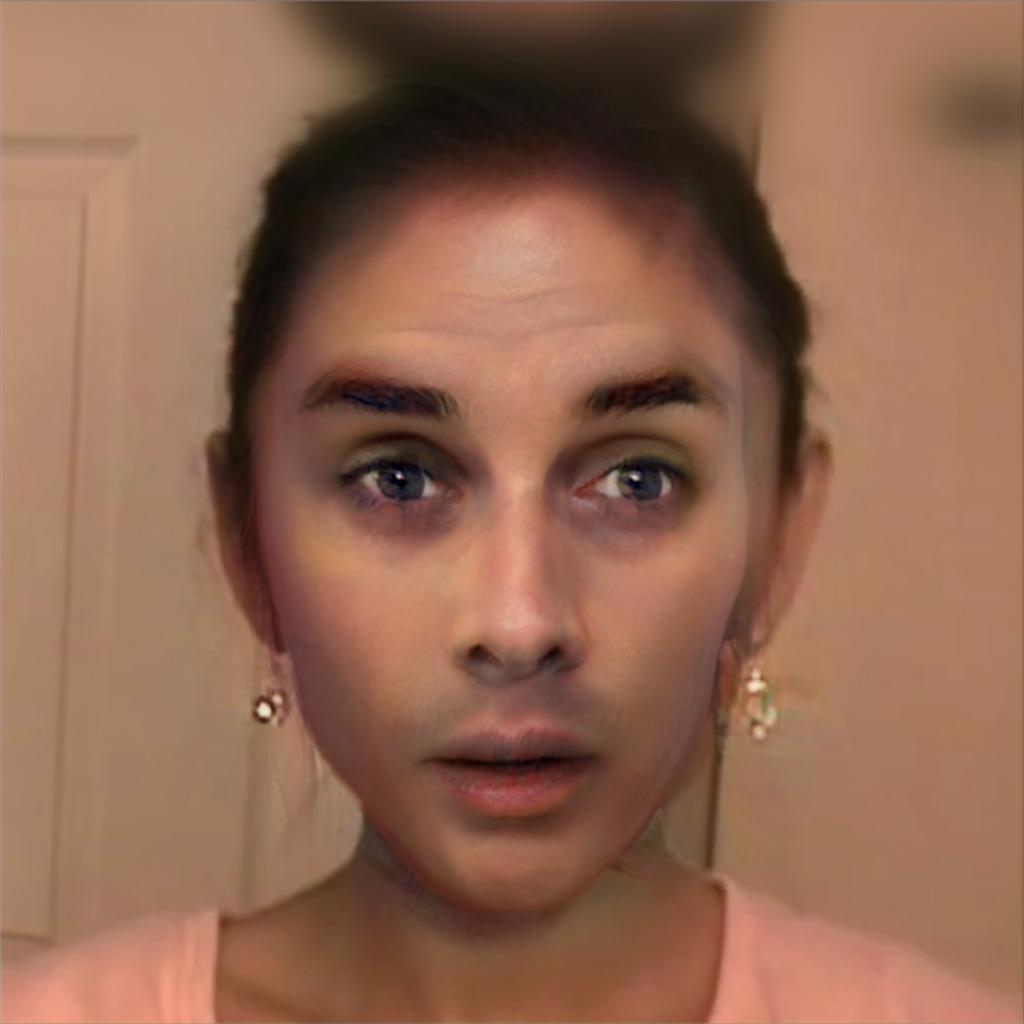}
     \includegraphics[width=\linewidth]{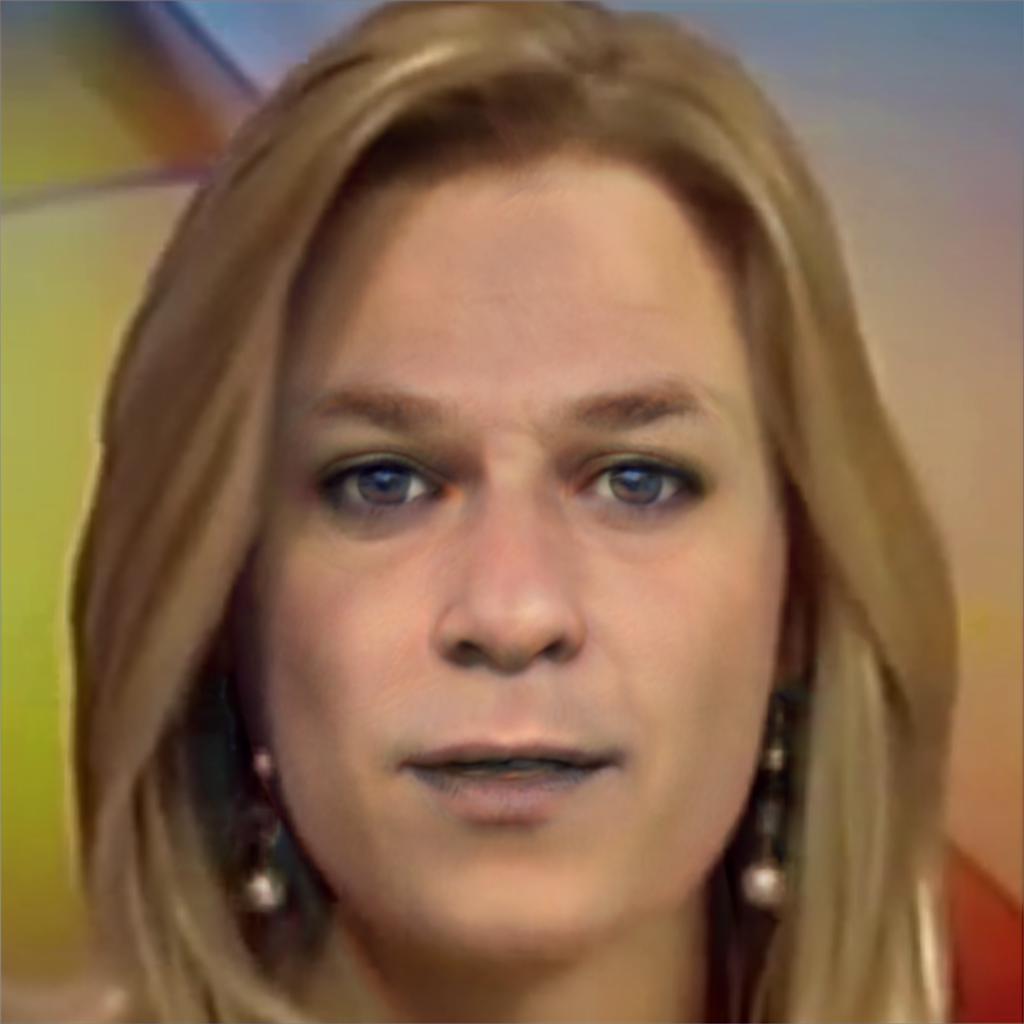}
     \includegraphics[width=\linewidth]{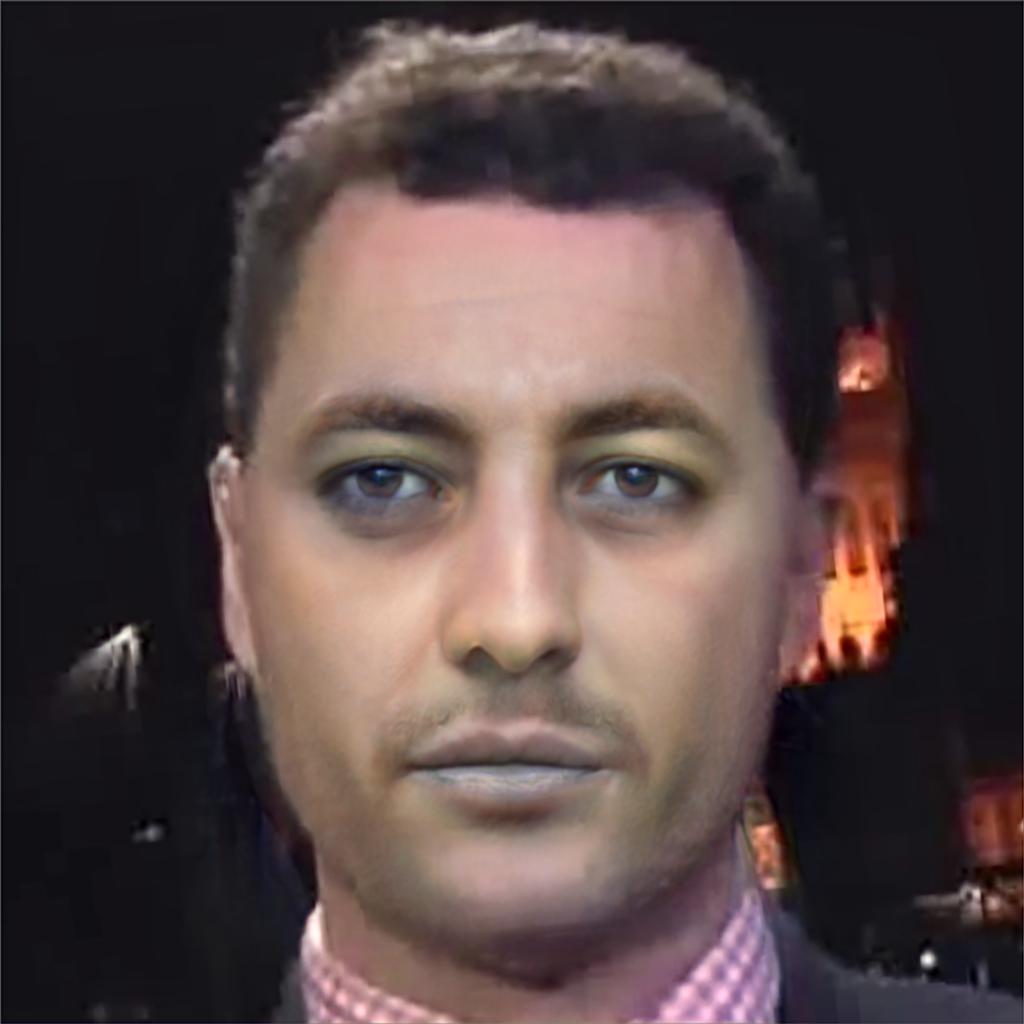}
     \includegraphics[width=\linewidth]{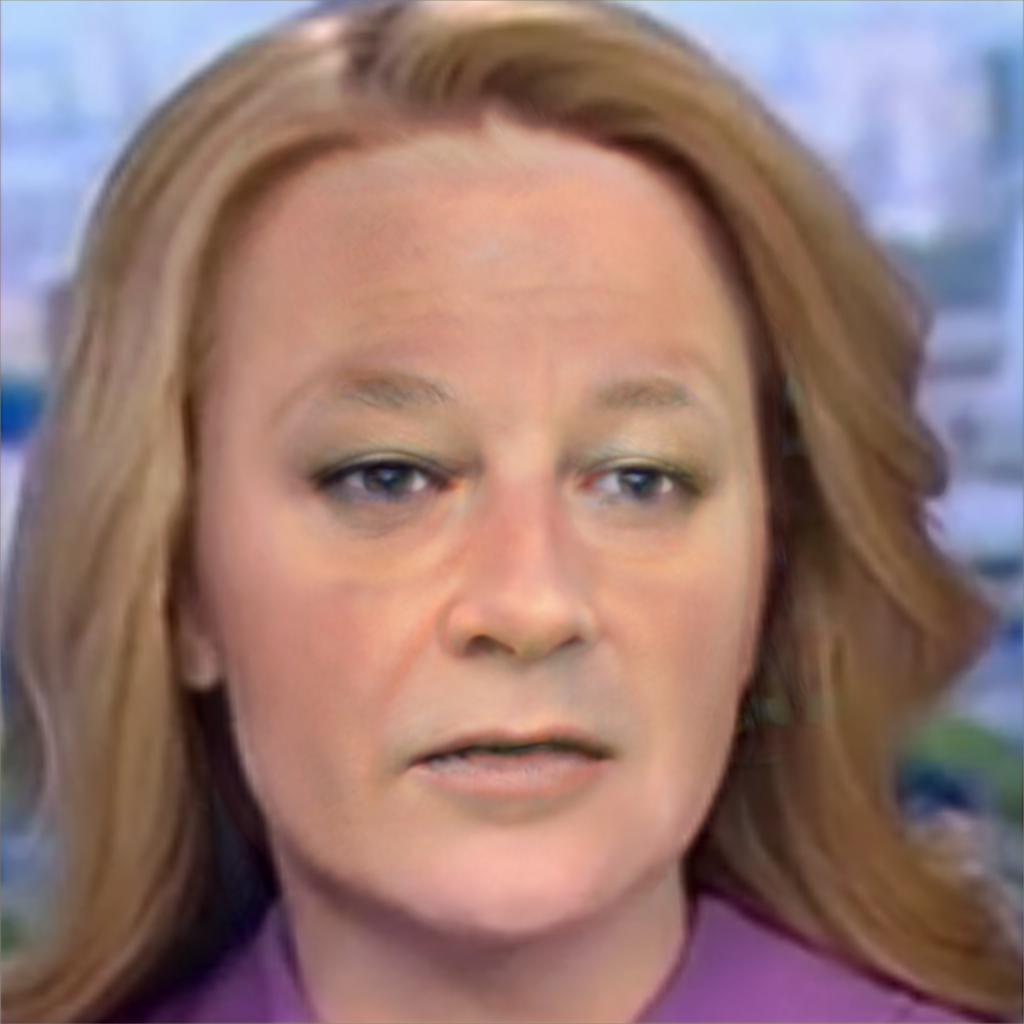}
     \caption{\subcapfont{Deepfakes}}
     \end{subfigure}
     \hspace{-1.5mm}
     \begin{subfigure}{.133\linewidth}
     \centering
     \includegraphics[width=\linewidth]{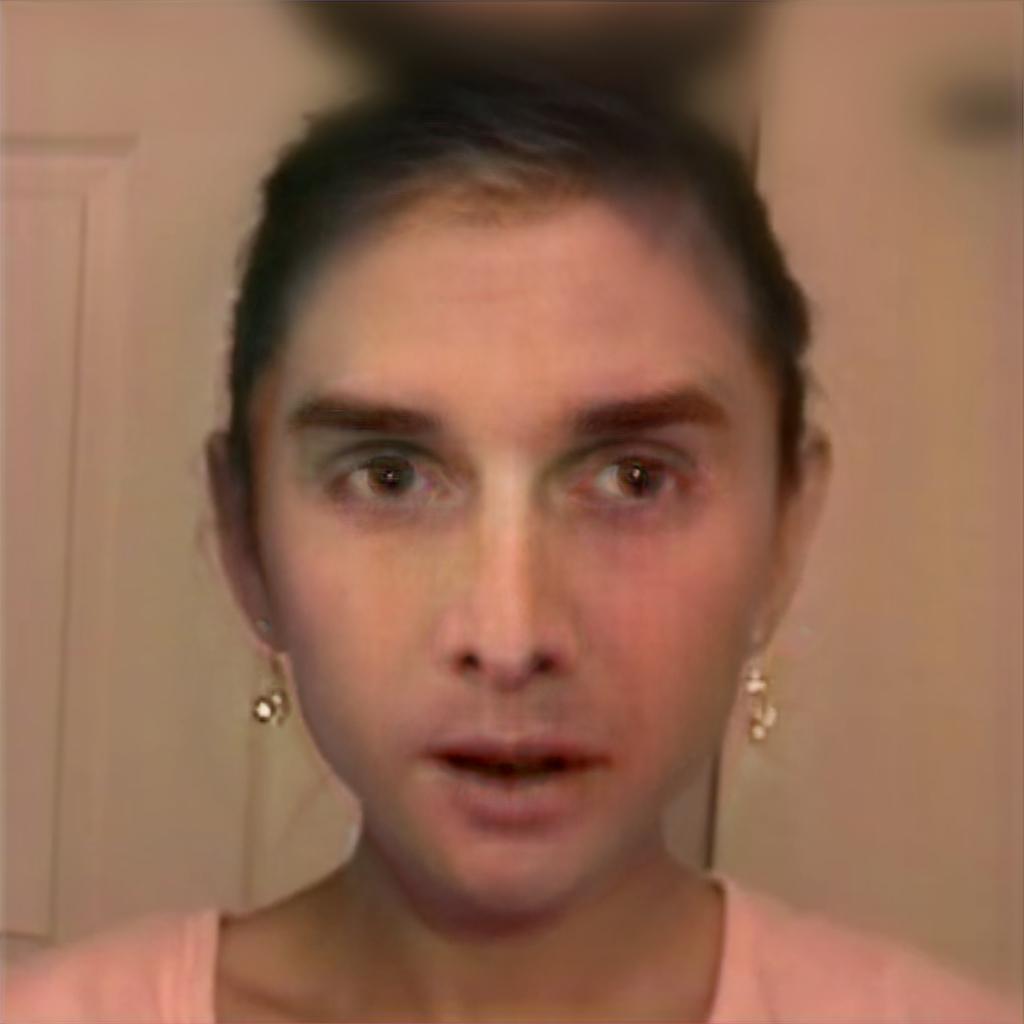}
     \includegraphics[width=\linewidth]{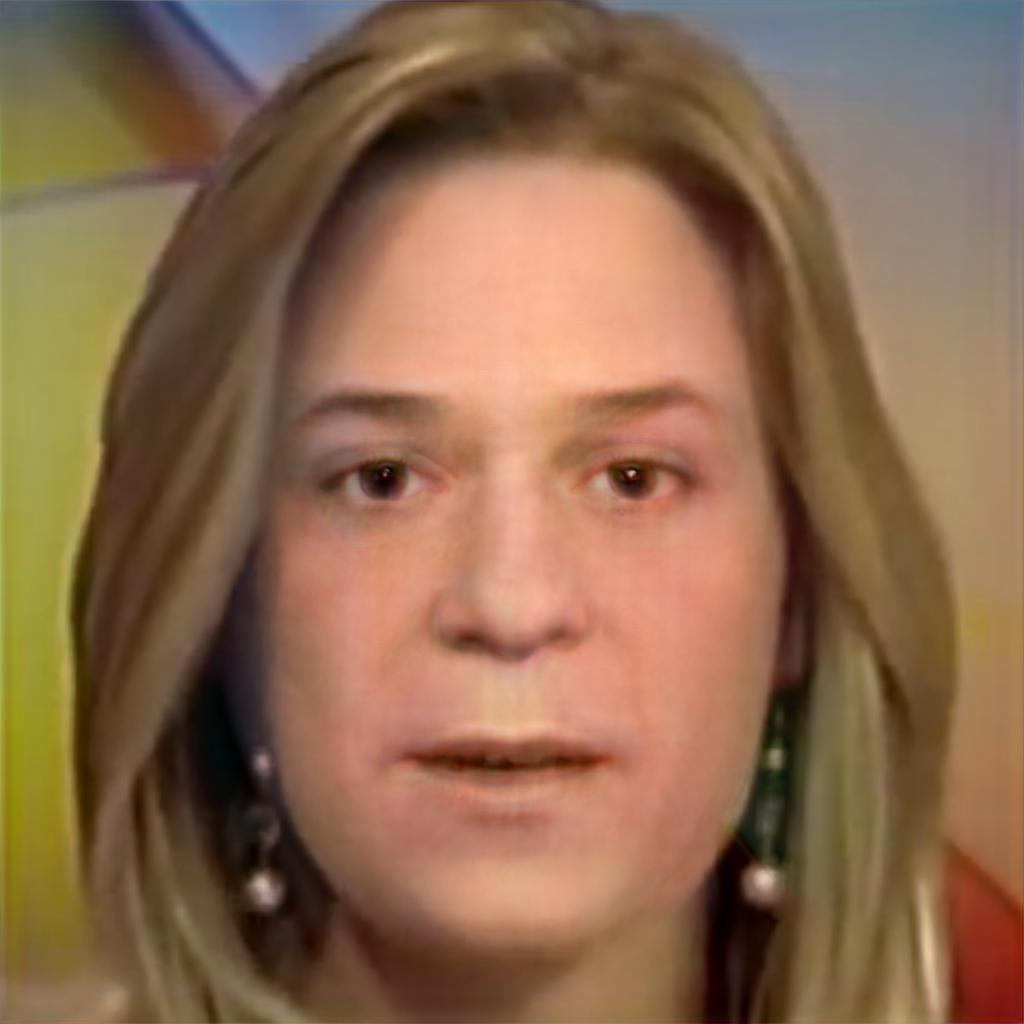}
     \includegraphics[width=\linewidth]{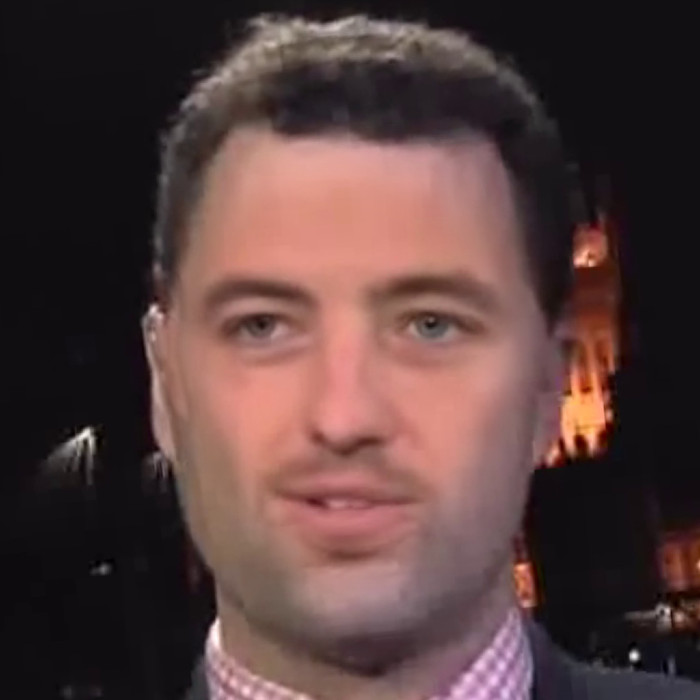}
     \includegraphics[width=\linewidth]{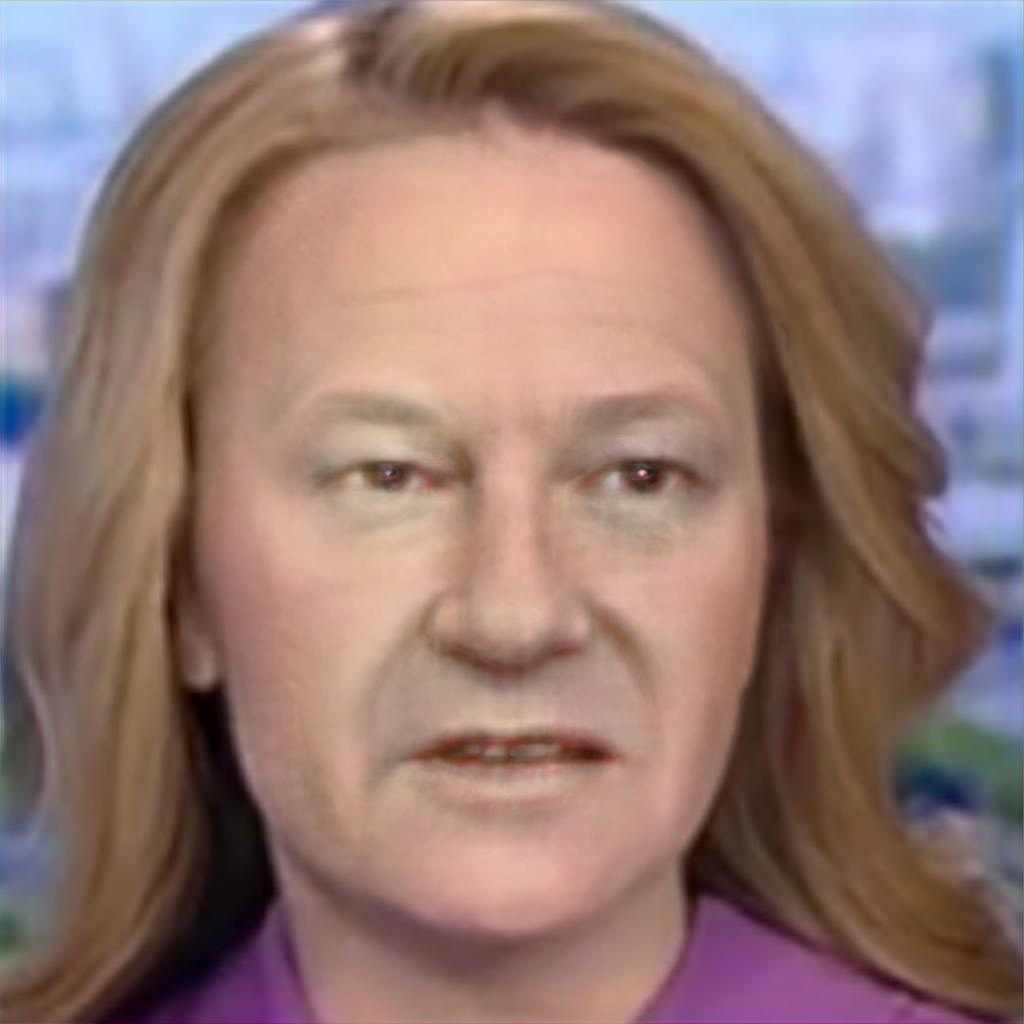}
     \caption{\subcapfont{FaceShifter}}
     \end{subfigure}
     \hspace{-1.5mm}
     \begin{subfigure}{.133\linewidth}
     \centering
     \includegraphics[width=\linewidth]{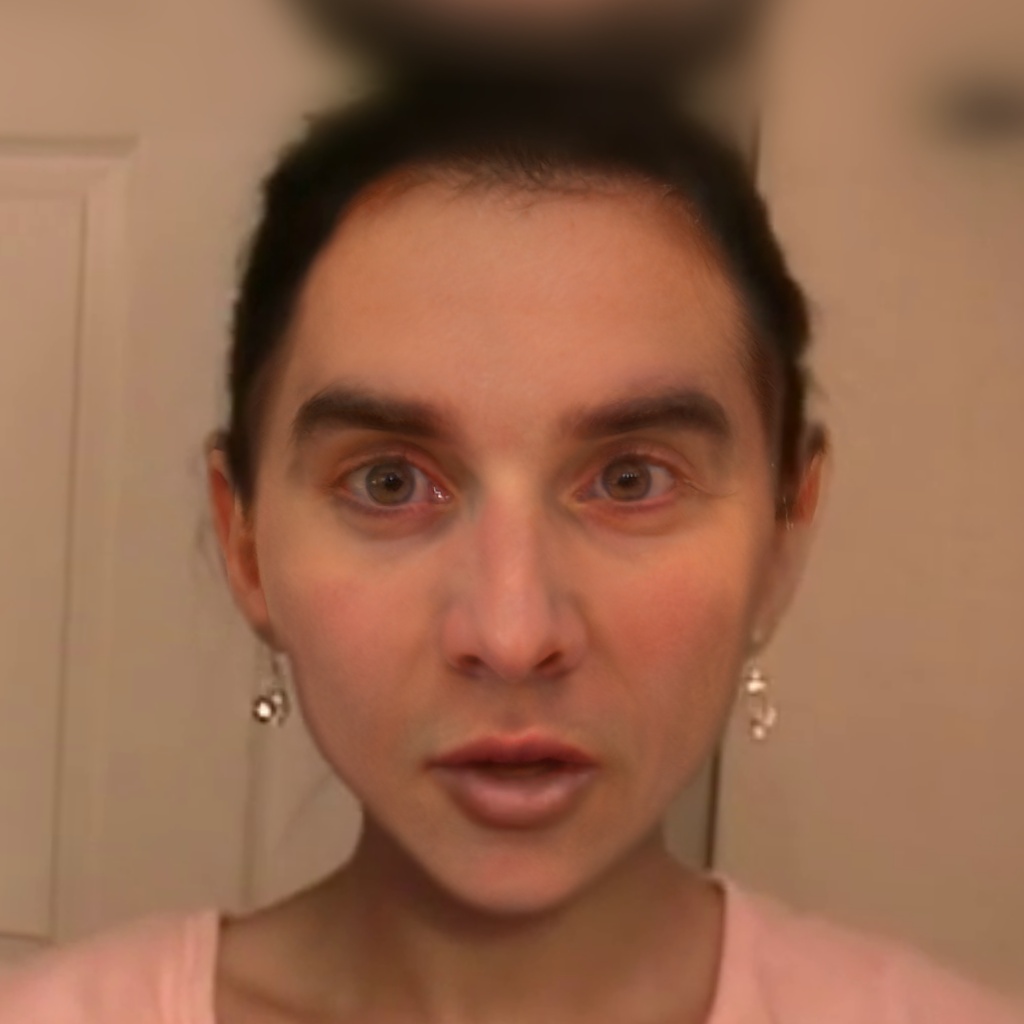}
     \includegraphics[width=\linewidth]{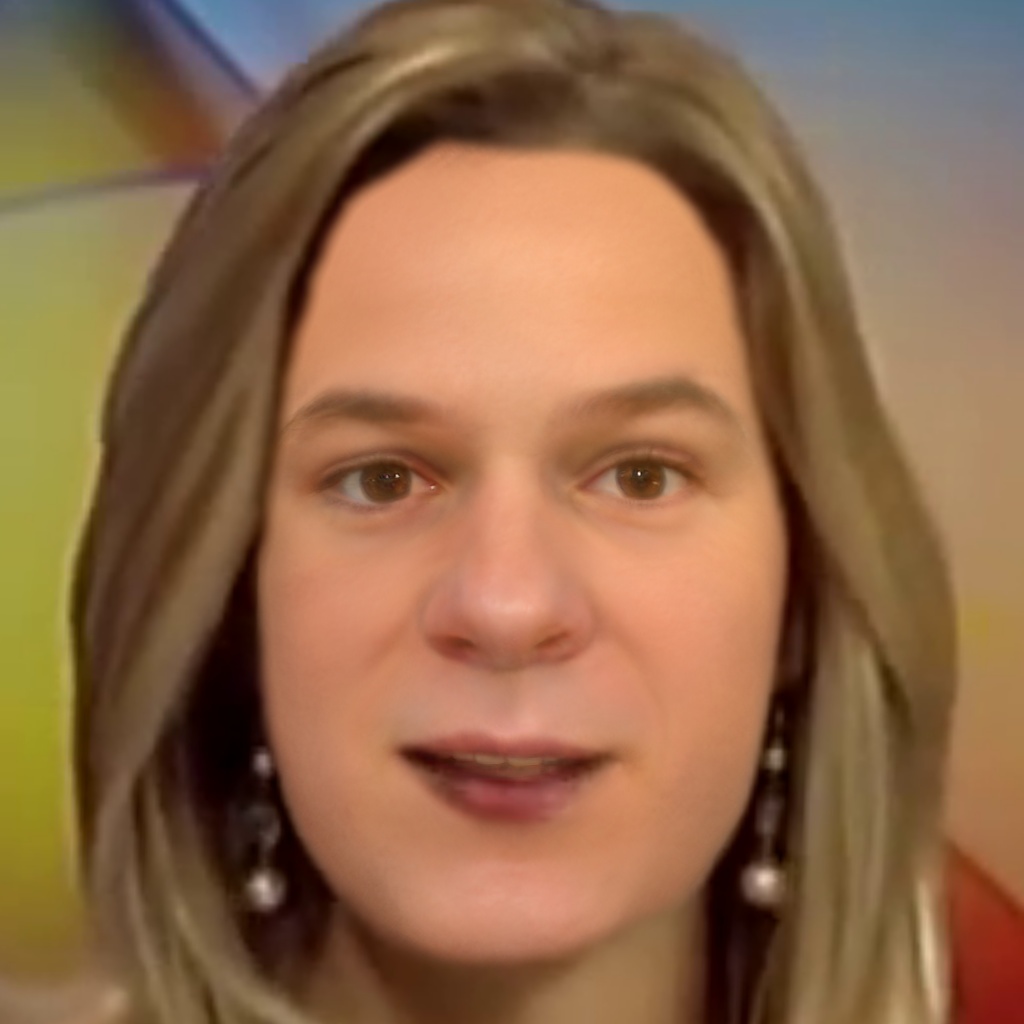}
     \includegraphics[width=\linewidth]{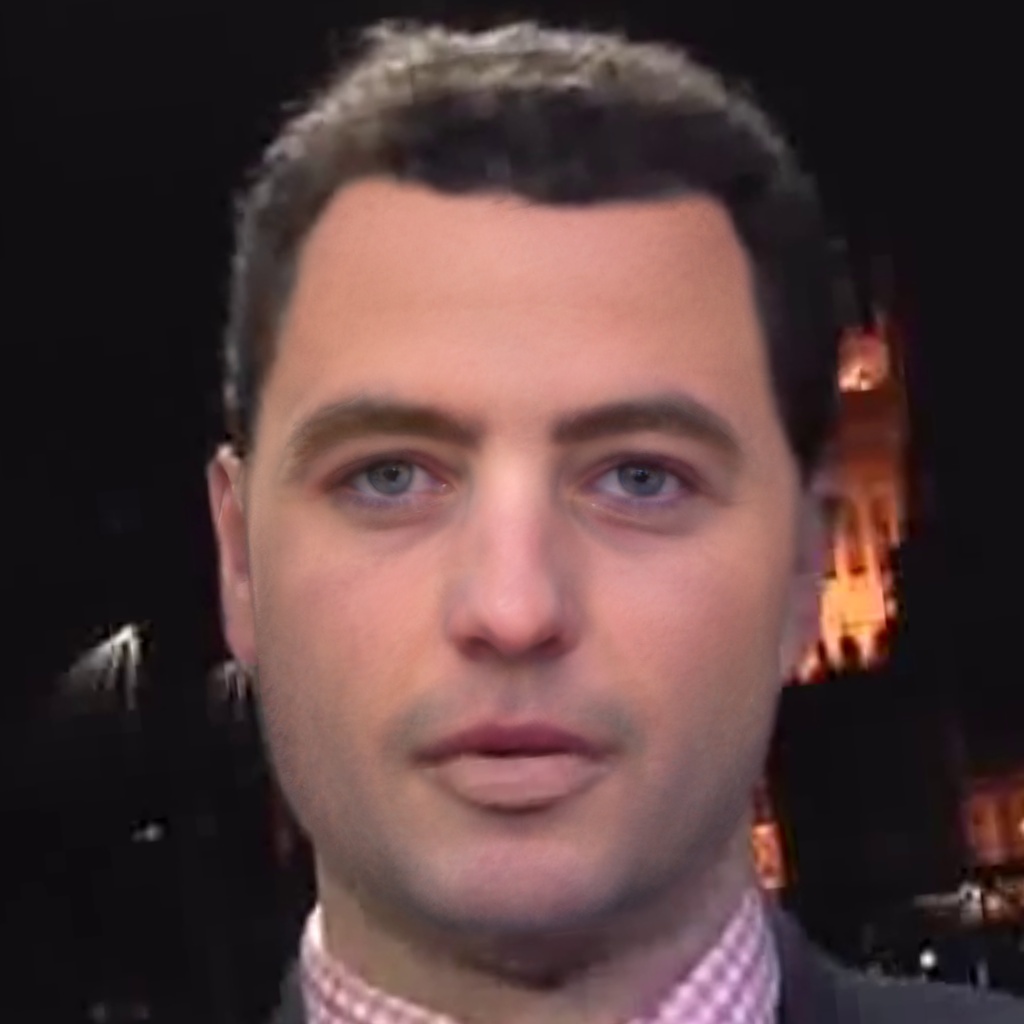}
     \includegraphics[width=\linewidth]{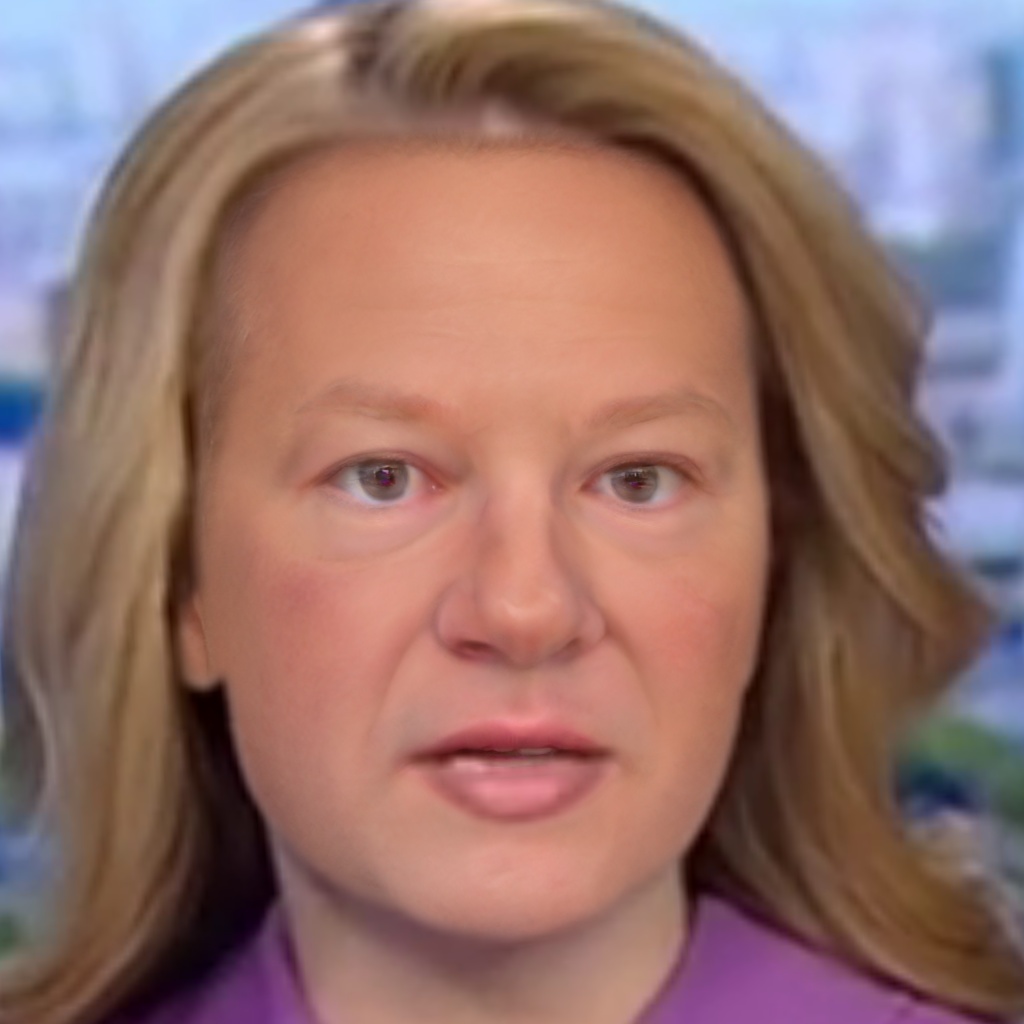}
     \caption{\subcapfont{MegaFS}}
     \end{subfigure}
     \hspace{-1.5mm}
     \begin{subfigure}{.133\linewidth}
     \centering
     \includegraphics[width=\linewidth]{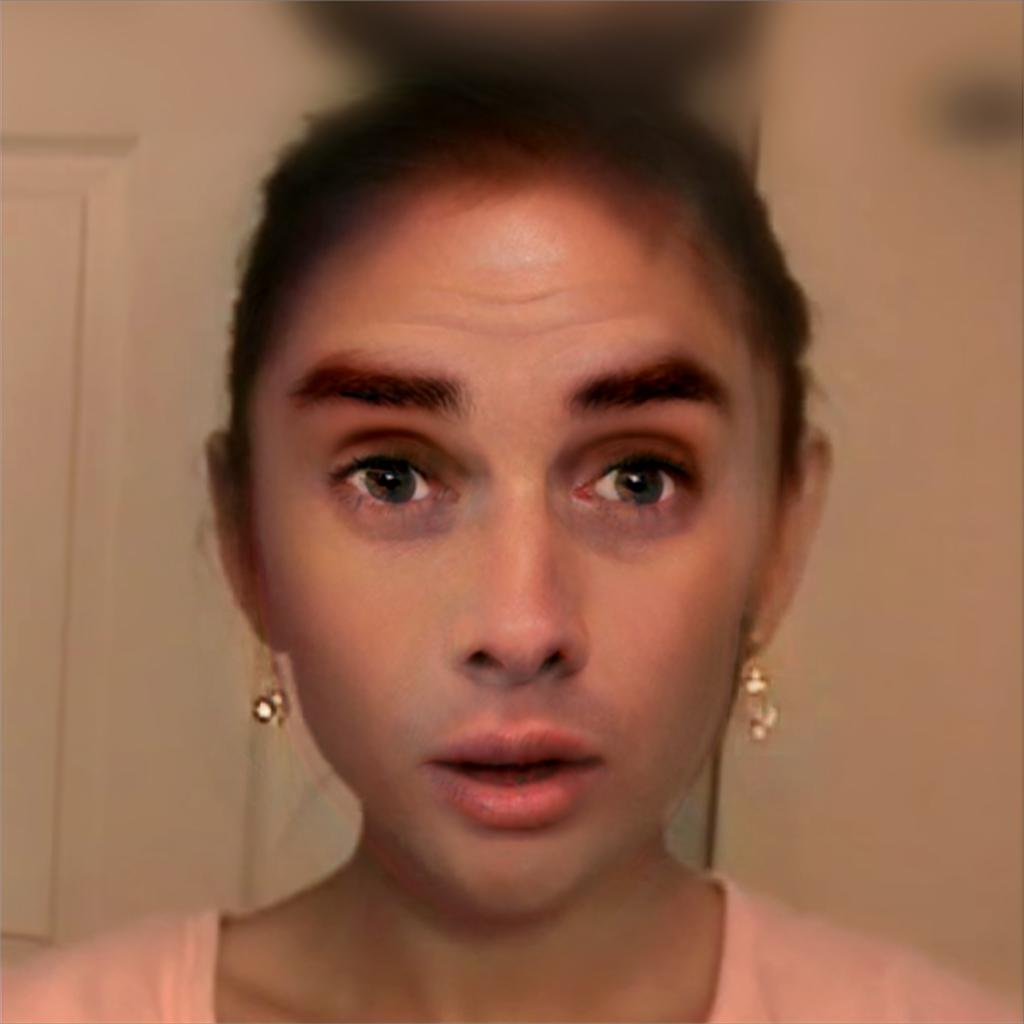}
     \includegraphics[width=\linewidth]{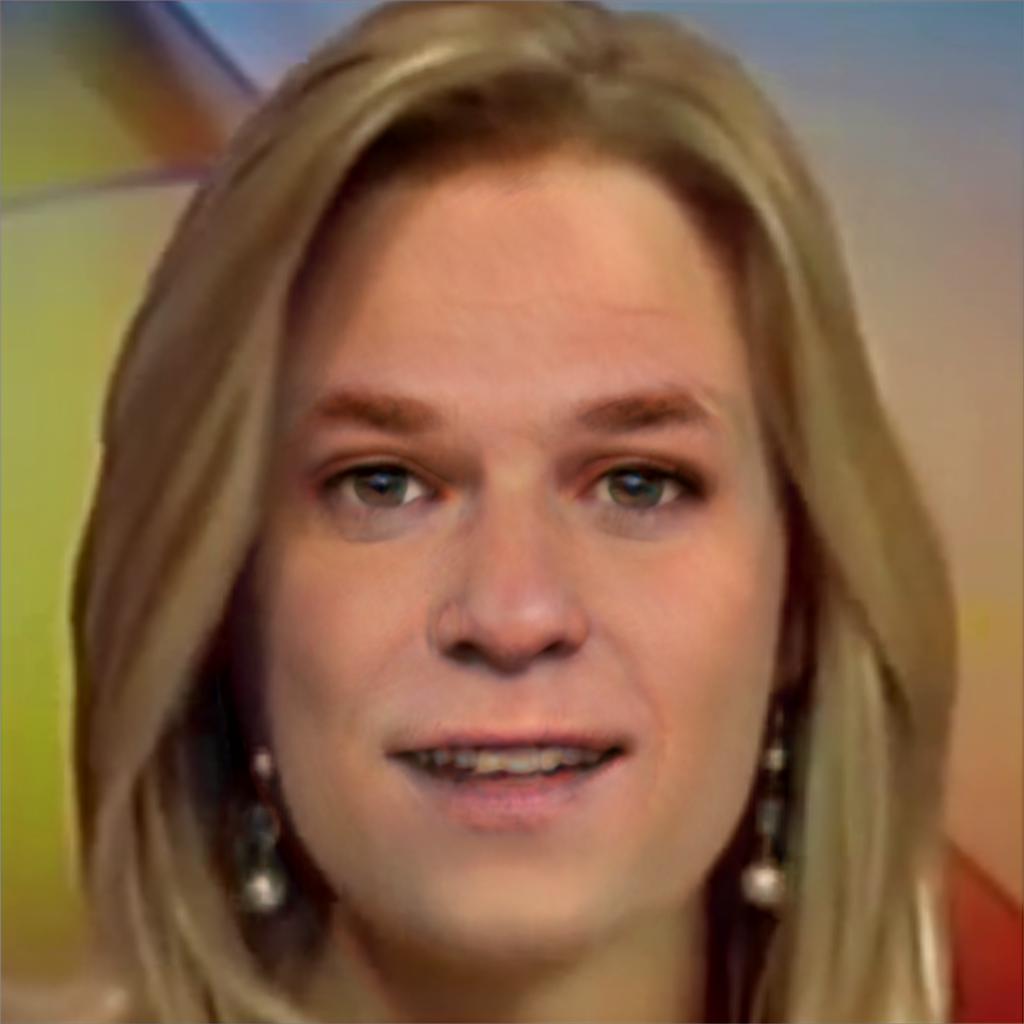}
     \includegraphics[width=\linewidth]{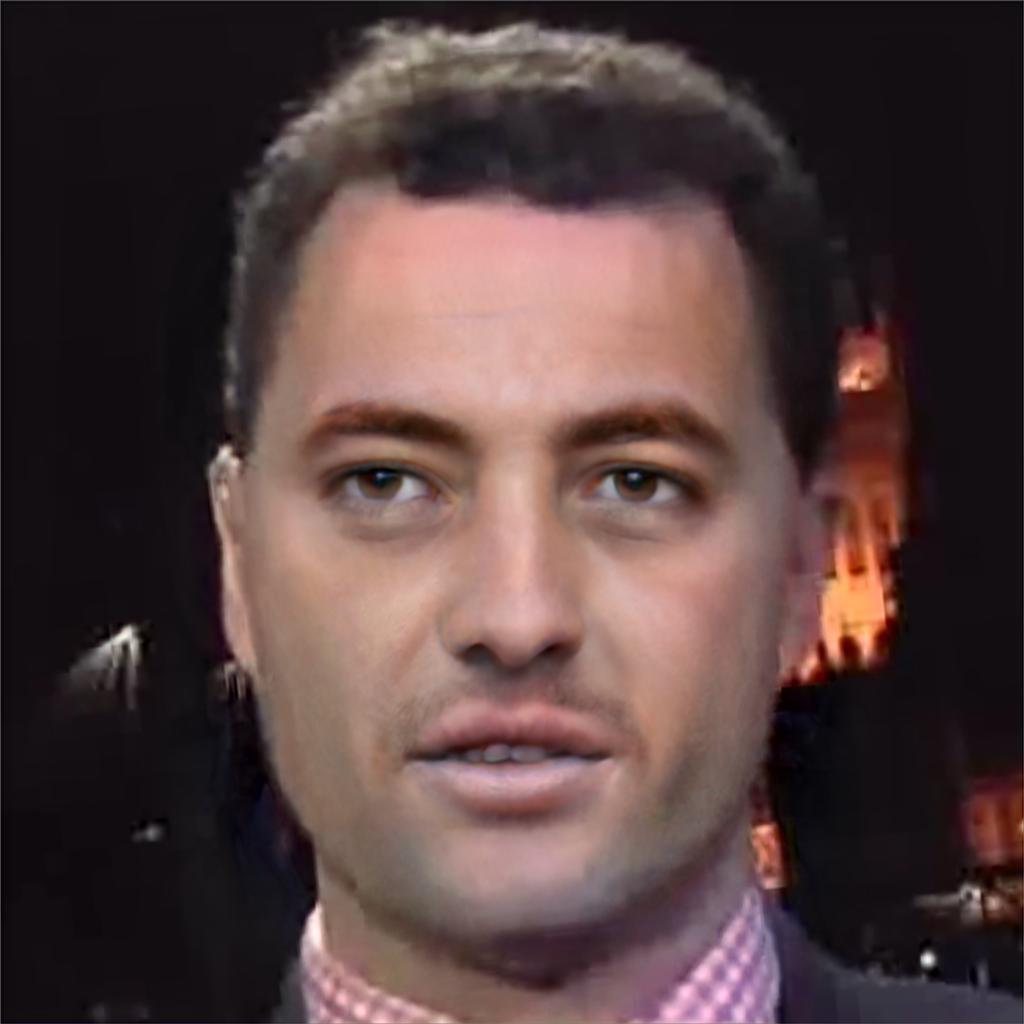}
     \includegraphics[width=\linewidth]{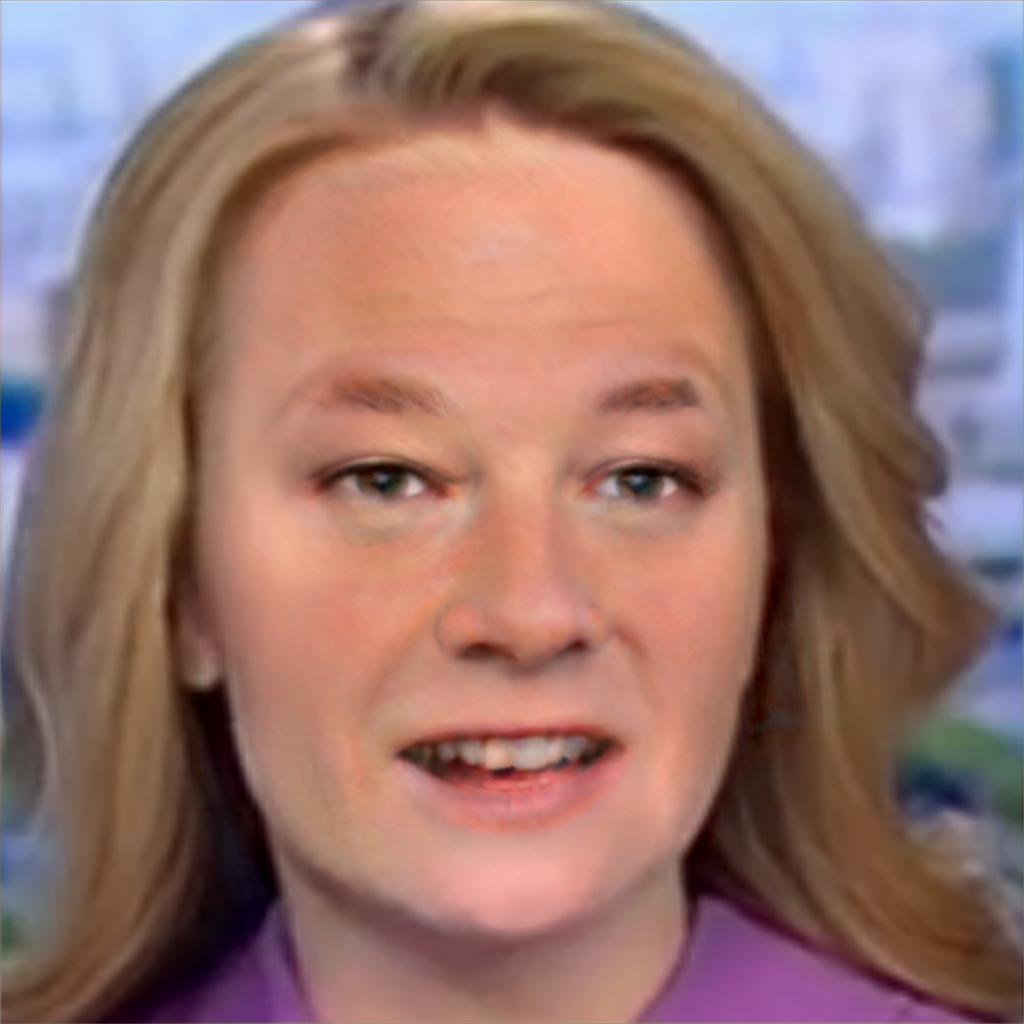}
     \caption{\subcapfont{Ours}}
     \end{subfigure}
     \capvspace
     \caption{Qualitative comparison with MegaFS and other low-resolution non-GAN priors face swapping methods.}
     \label{fig.ff}
     \figbottomvspace
\end{figure}

\paraheading{Quantitative Comparison} We also perform quantitative comparison on the CelebA-HQ dataset. We follow MegaFS and conduct a comparison on 300,000 swapped faces for a fair comparison. Tab.~\ref{table:celeba} shows the average values of evaluation metrics for each method. Our method has a stronger ability both on identity preservation and attributes transfer. Furthermore, our result has a lower FID, which indicates that our swapped faces are more realistic than those of MegaFS.

\subsection{Comparison on FaceForensics++ Dataset}
To allow comparing with other face swapping methods that can be only applied to low-resolution images, we further evaluate our method on the Face-Forensics++ dataset. Fig.~\ref{fig.ff} presents qualitative comparison results against MegaFS as well as three non-GAN prior based and low-resolution methods: FaceSwap~\cite{Faceswap}, Deepfakes~\cite{Deepfakes}, and FaceShifter~\cite{li2020advancing}. We can see that FaceSwap and MegaFS can lead to noticeable blending boundaries or artifacts, while our method can eliminate them due to our background transfer. However, the quality of our swapped results is not that high as high-resolution face swapping in the Fig.~\ref{fig.celeba}. This is potentially due to domain gap: the styleGAN generator and the pSp encoder used in our method are both pre-trained on high-resolution data, while the majority of the data from FaceForensics++ are of lower resolutions. The large domain gap between low- and high-resolution datasets can potentially decrease the performance of our method.

We also perform quantitative evaluation on this dataset. In particular, we follow MegaFS~\cite{zhu2021one} that samples 10 frames from each video evenly then processed by MTCNN~\cite{zhang2016joint}. After filtering the repeated identities, we obtain 885 videos with 88500 frames in total.
% \bailin{Need more information, for example the amount of data used for testing.}
Tab.~\ref{table:ff} shows the average values of evaluation metrics for different methods. We can see that our method preserves the target poses and expressions more effectively, thanks to our structure attributes transfer that takes the landmarks as input and provides a strong guidance signal to the final swapped faces. Meanwhile, our model is outperformed by FaceShifter and MegaFS on ID retrieval rate. We conjecture that this is due to the domain gap between the low- and high-resolution images as mentioned above, which makes our inversion model less effective in preserving the identity information from low-resolution images.

\begin{table}[t]
	\caption{Quantitative evaluation of face swapping on CelebA-HQ dataset with four metrics. $\downarrow$ denotes the lower the better and vice versa, the best results are marked in \textbf{bold}.}
	\tabcapvspace
	\begin{center}
		\setlength{\tabcolsep}{0.12cm}{
			\begin{tabular}{c|c|c|c|c}
				\hline
				Methods       &{ID Simi.$\uparrow$}   &{Pose Err.$\downarrow$} &{Exp. Err.$\downarrow$}   &FID$\downarrow$ \\
				\hline
				MegaFS~\cite{zhu2021one}                         &0.5214              &3.498            &2.95                     &11.645                                \\
				Ours            &\textbf{0.5688}         &\textbf{2.997}   &\textbf{2.74}            &\textbf{9.987}       \\
				\hline
			\end{tabular}
		}
	\end{center}
	\label{table:celeba}
	\tabbottomvspace
\end{table}

\begin{table}[t]
   \caption
   {Quantitative evaluation of face swapping on FaceForensics++ dataset with four metrics. $\downarrow$ denotes the lower the better and the better and vice versa, the best results are marked in \textbf{bold}.}
       \tabcapvspace
       \begin{center}
       \setlength{\tabcolsep}{0.15cm}{
       \begin{tabular}{c|c|c|c}
           \hline
           Methods       &{ID Retri.(\%)$\uparrow$}   &{Pose Err.$\downarrow$} &{Exp. Err.$\downarrow$}   \\
           \hline
           FaceSwap~\cite{Faceswap}           &72.69   &2.58   &2.89      \\
           Deepfakes~\cite{Deepfakes}          &88.39   &4.64   &3.33      \\
           FaceShifter~\cite{li2020advancing} &90.68   &2.55   &2.82   \\
           MegaFS~\cite{zhu2021one}           &\textbf{90.83}   &2.64   &2.92        \\
           Ours                               &90.05   &\textbf{2.46}  & \textbf{2.79}    \\
           \hline
       \end{tabular}
       }
      \end{center}
      \label{table:ff}
      \tabbottomvspace
\end{table}
% \vspace{-3mm}
\subsection{Ablation Study}
\begin{figure}[t]
   \centering
    % \rotatebox[origin=c]{90}{Ours \hspace{15mm} InD \hspace{15mm} pSp \hspace{15mm} I2S}\
    \captionsetup[subfigure]{labelformat=empty,justification=centering}
    \begin{subfigure}{.166\linewidth}
     \centering
     \includegraphics[width=\linewidth]{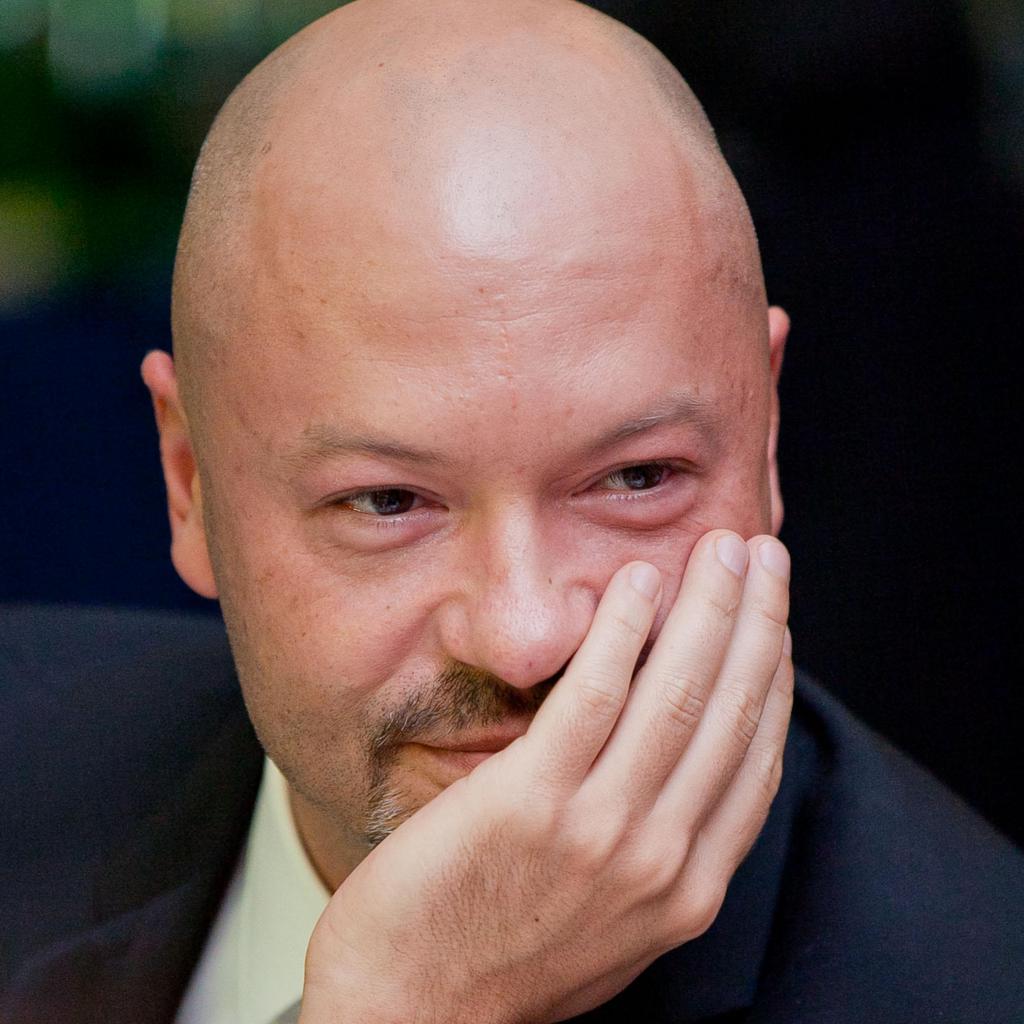}
     \includegraphics[width=\linewidth]{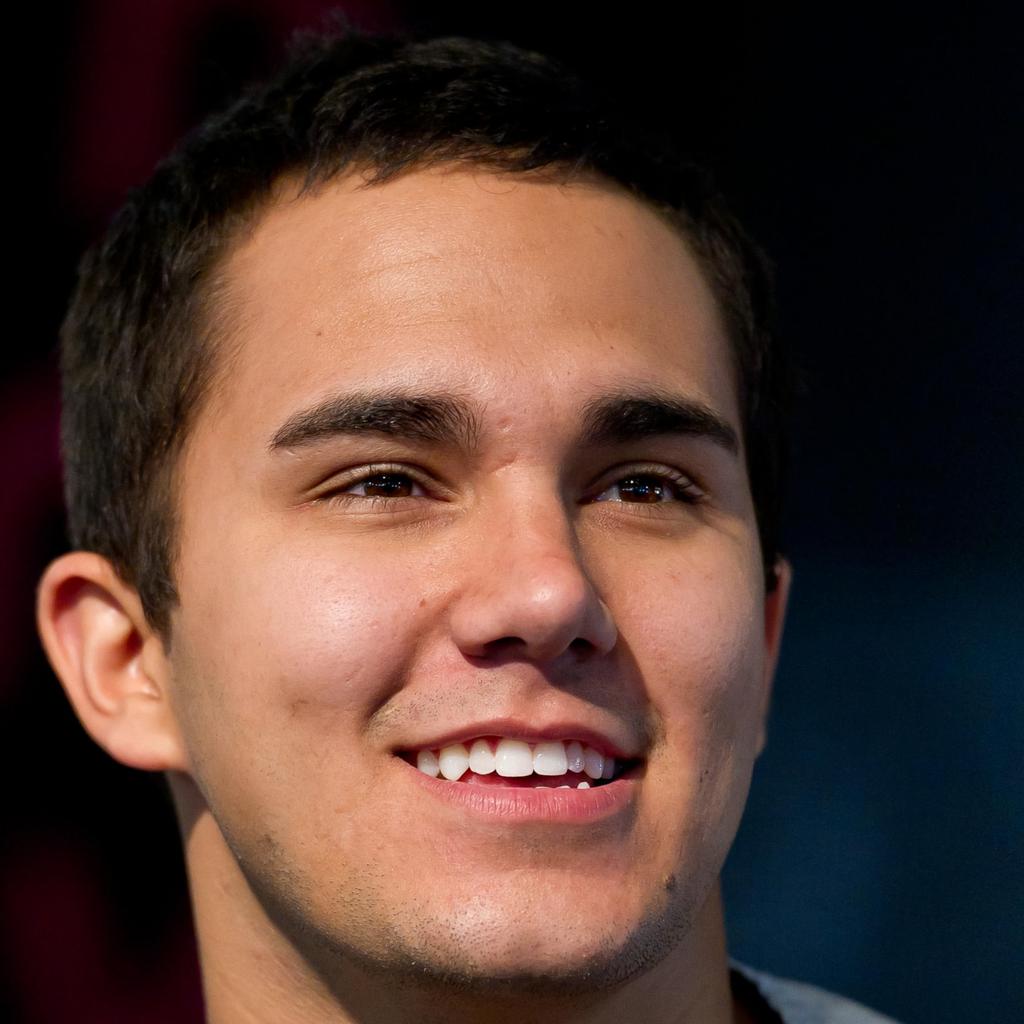}
     \includegraphics[width=\linewidth]{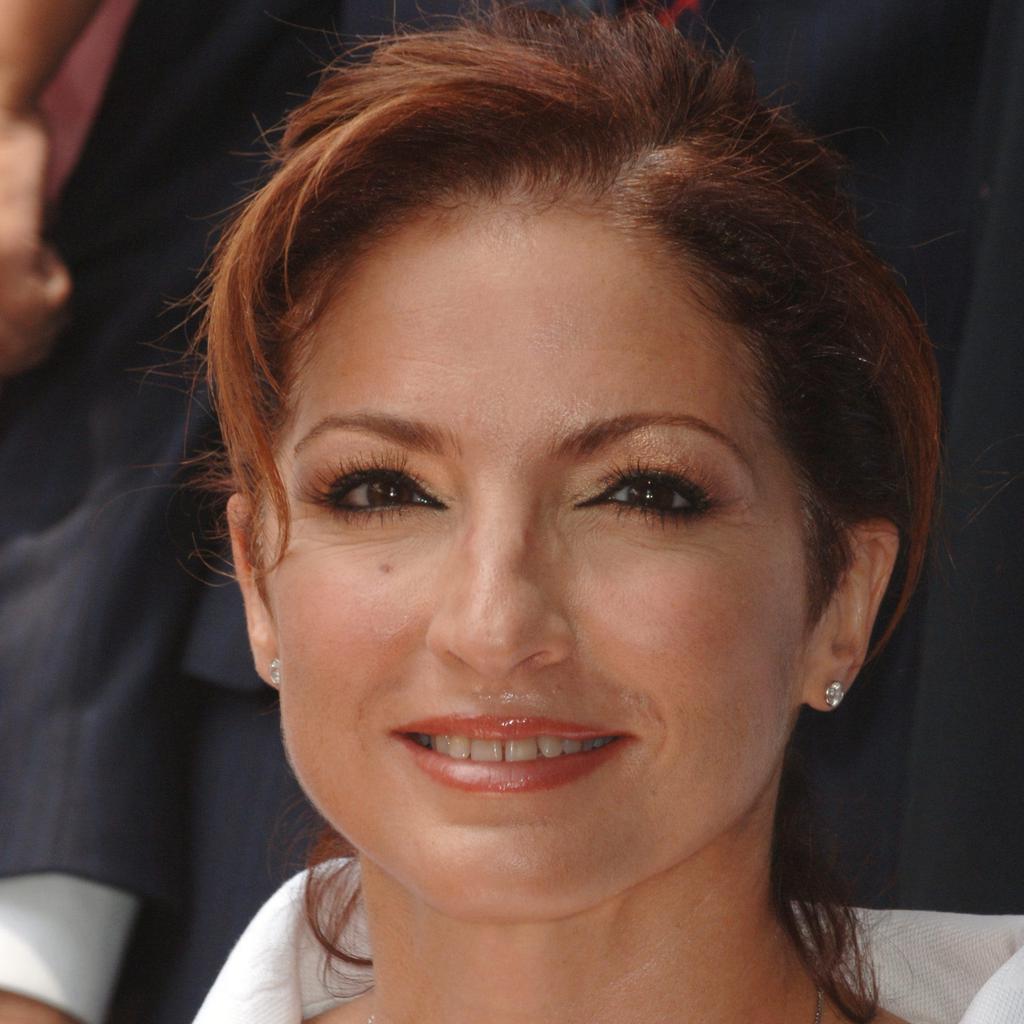}
     \caption{Source}
     \end{subfigure}
     \hspace{-2mm}
     \begin{subfigure}{.166\linewidth}
     \centering
     \includegraphics[width=\linewidth]{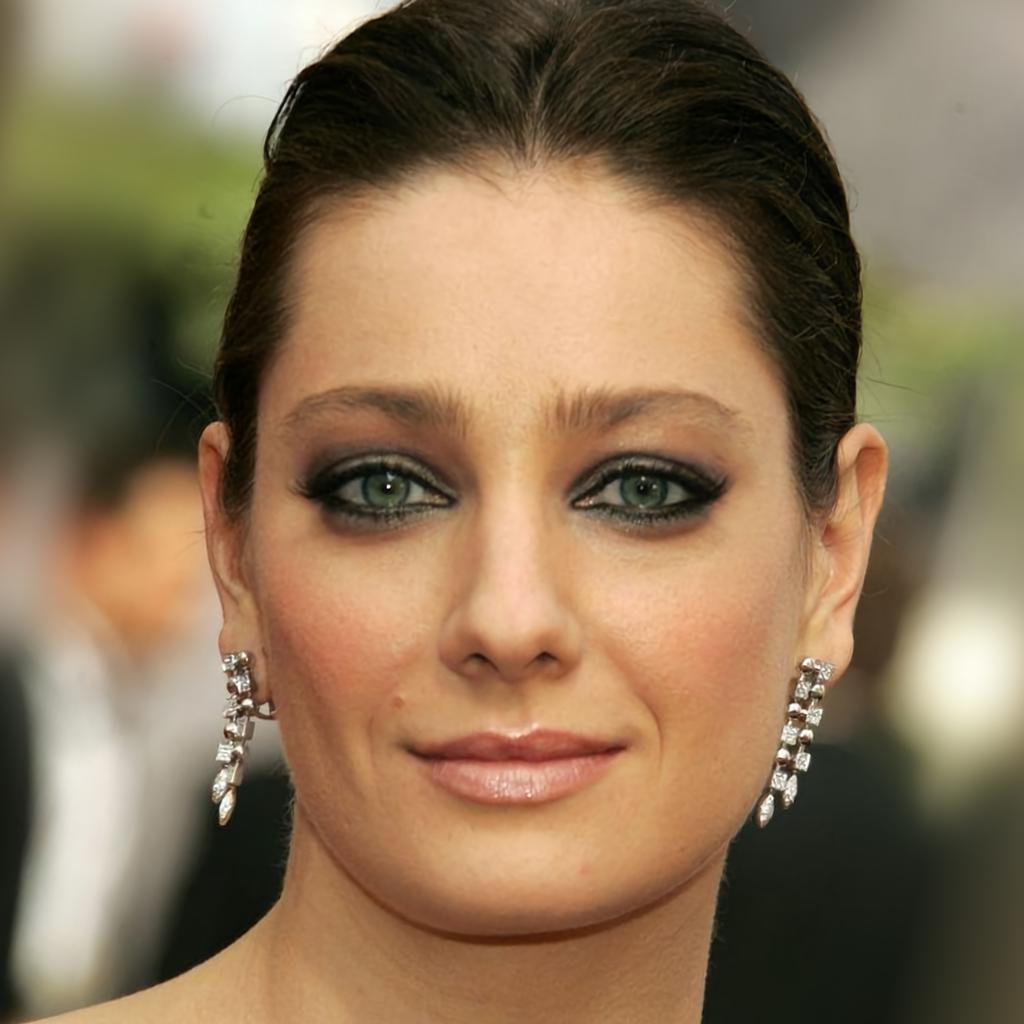}
     \includegraphics[width=\linewidth]{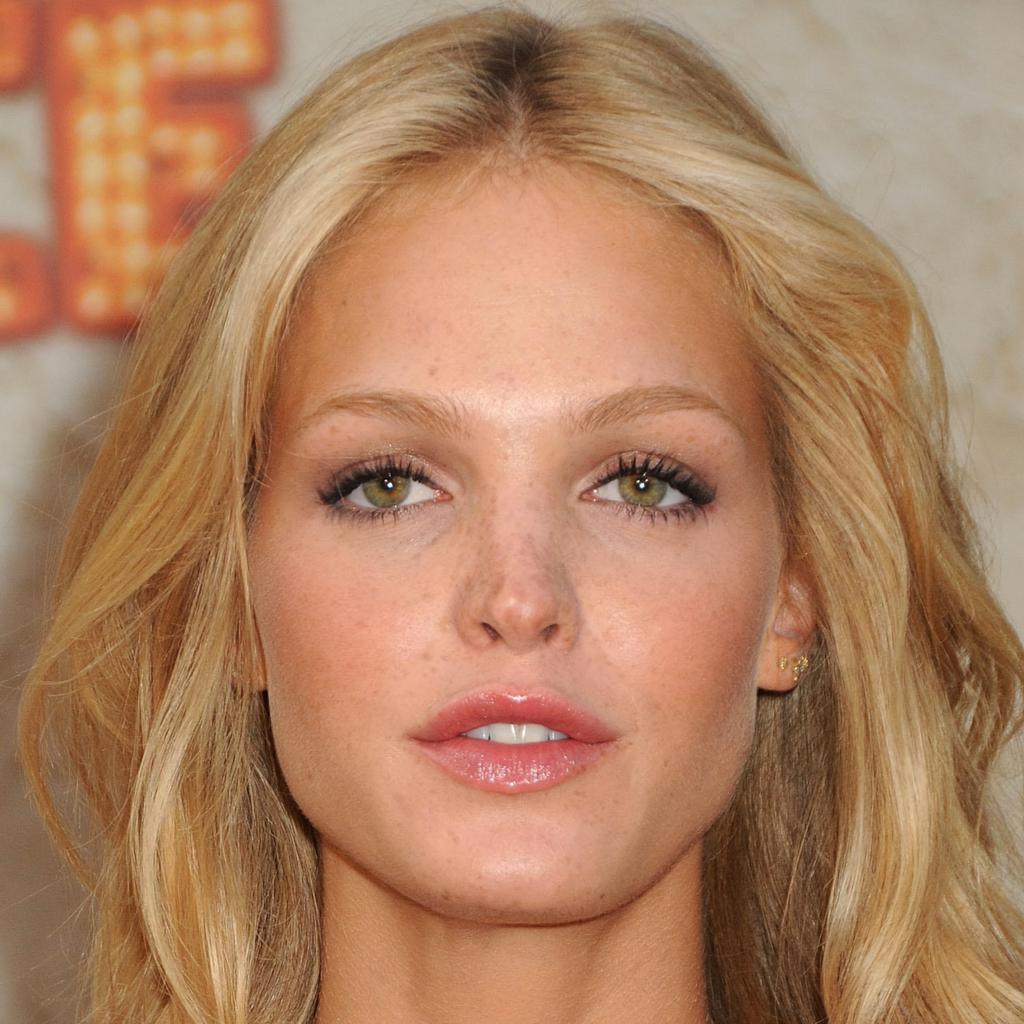}
     \includegraphics[width=\linewidth]{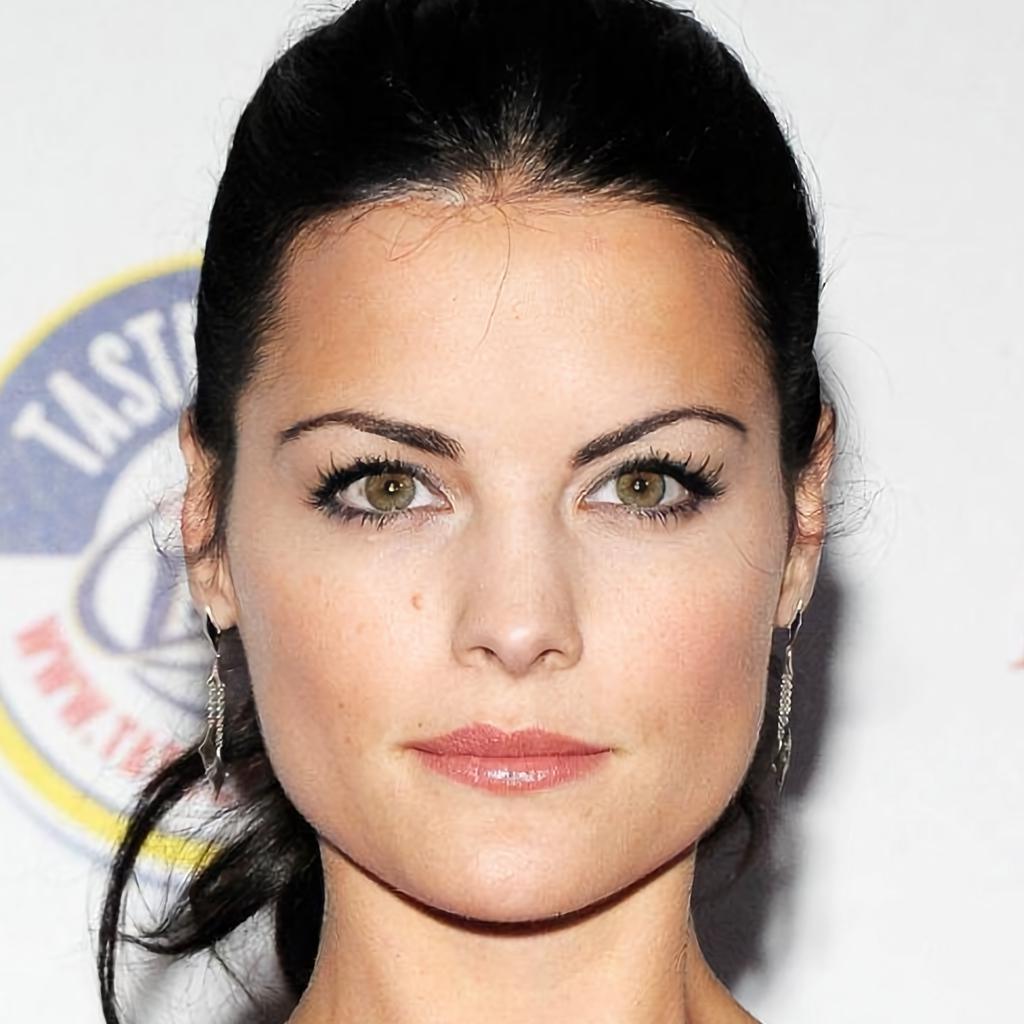}
     \caption{Target}
     \end{subfigure}
     \hspace{-2mm}
     \begin{subfigure}{.166\linewidth}
     \centering
     \includegraphics[width=\linewidth]{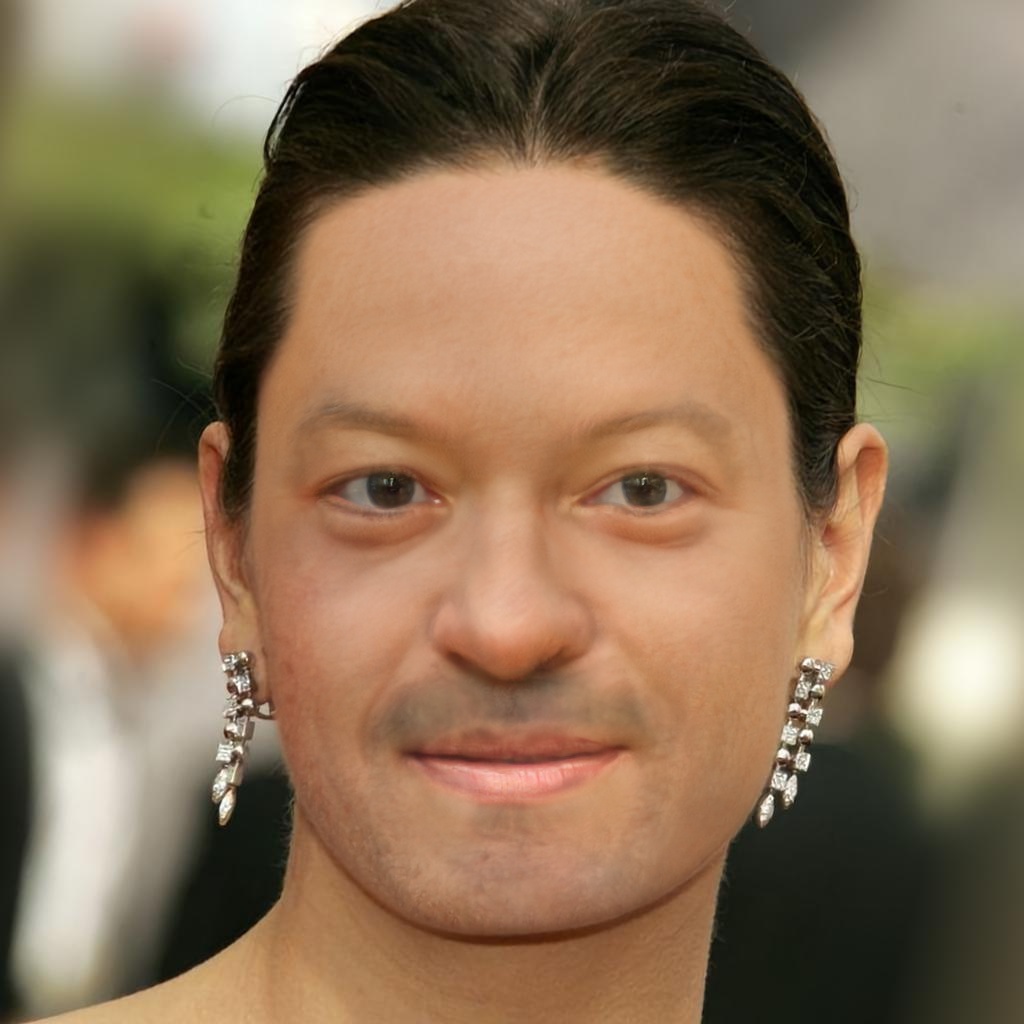}
     \includegraphics[width=\linewidth]{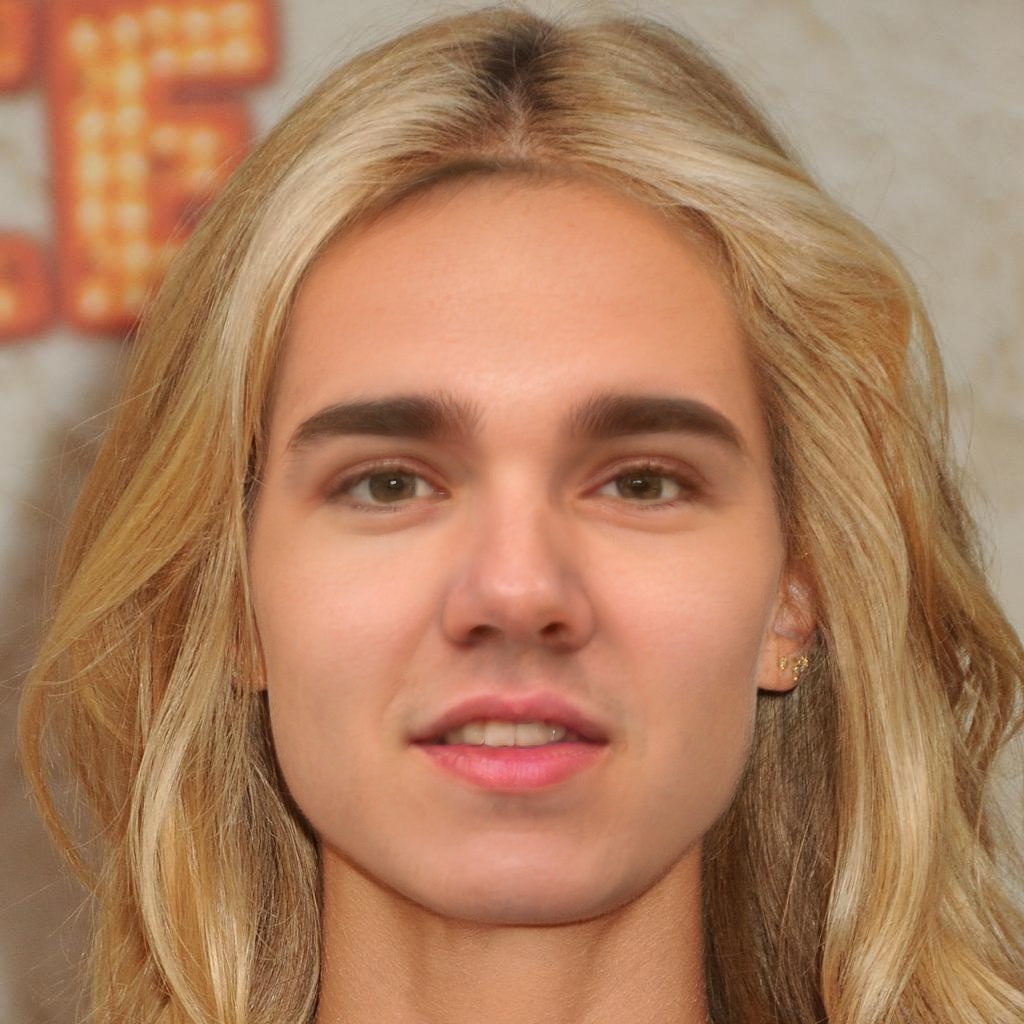}
     \includegraphics[width=\linewidth]{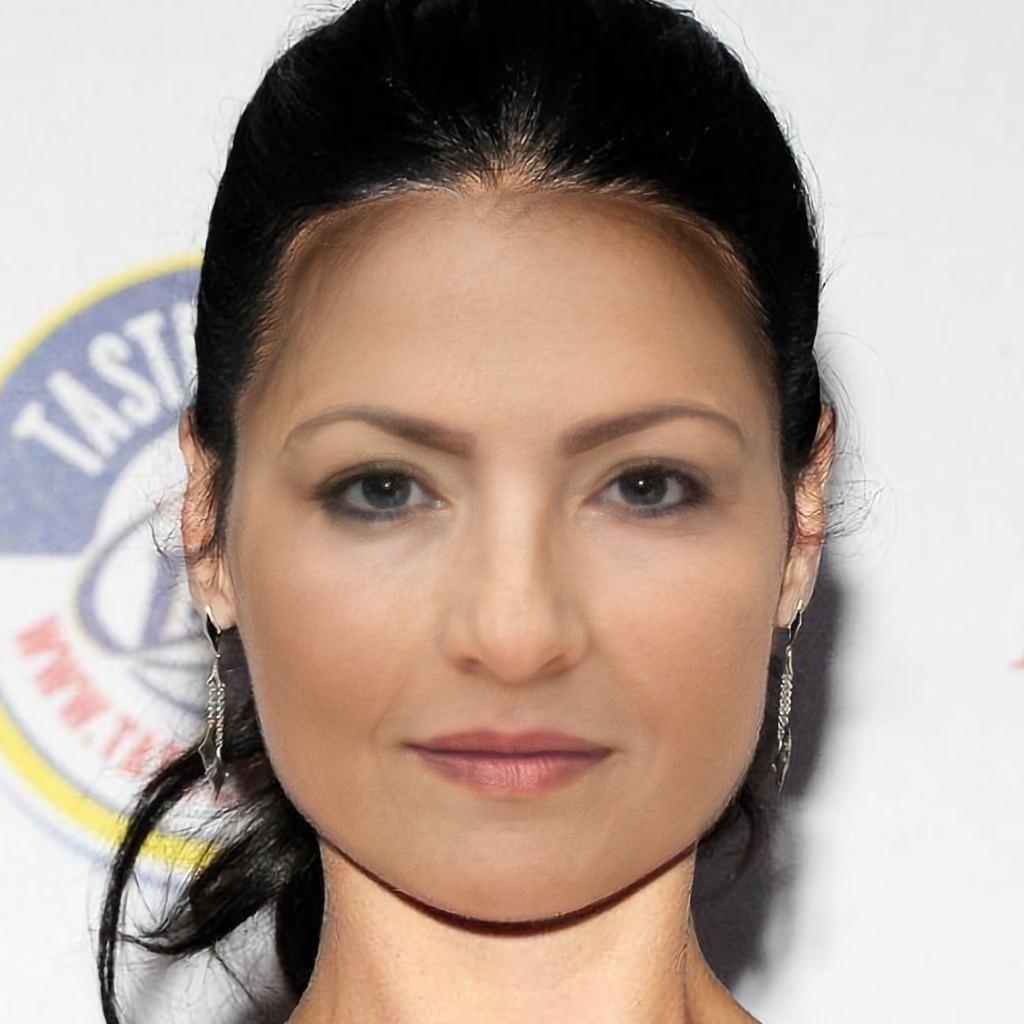}
     \caption{Baseline}
     \end{subfigure}
     \hspace{-2mm}
     \begin{subfigure}{.166\linewidth}
     \centering
     \includegraphics[width=\linewidth]{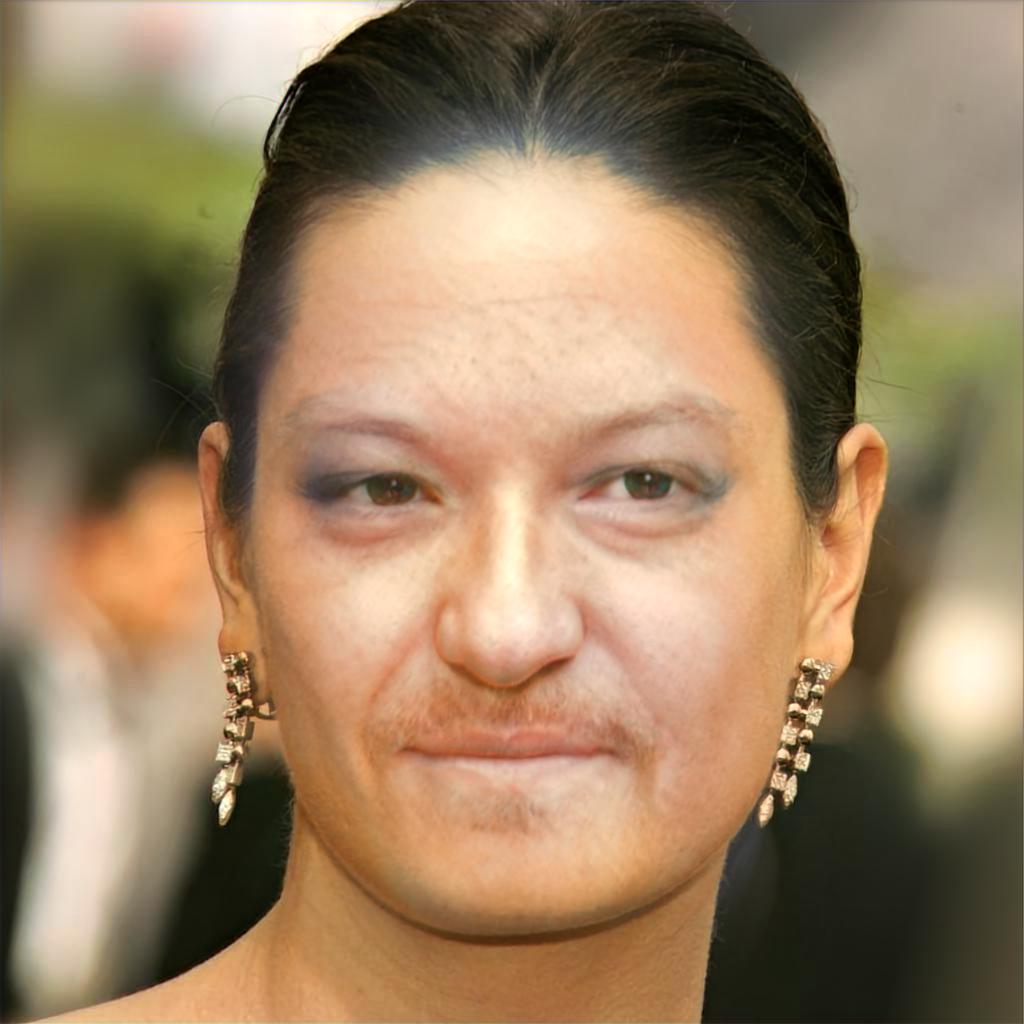}
     \includegraphics[width=\linewidth]{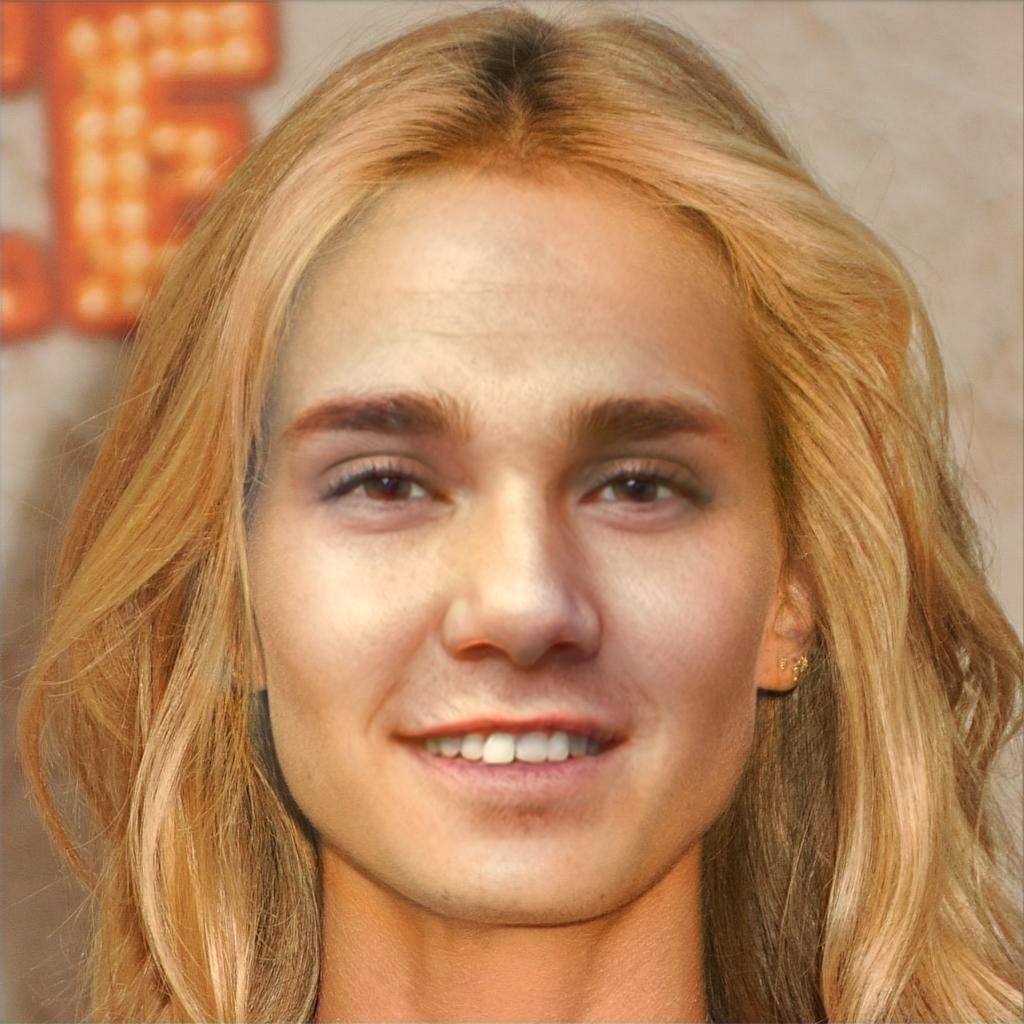}
     \includegraphics[width=\linewidth]{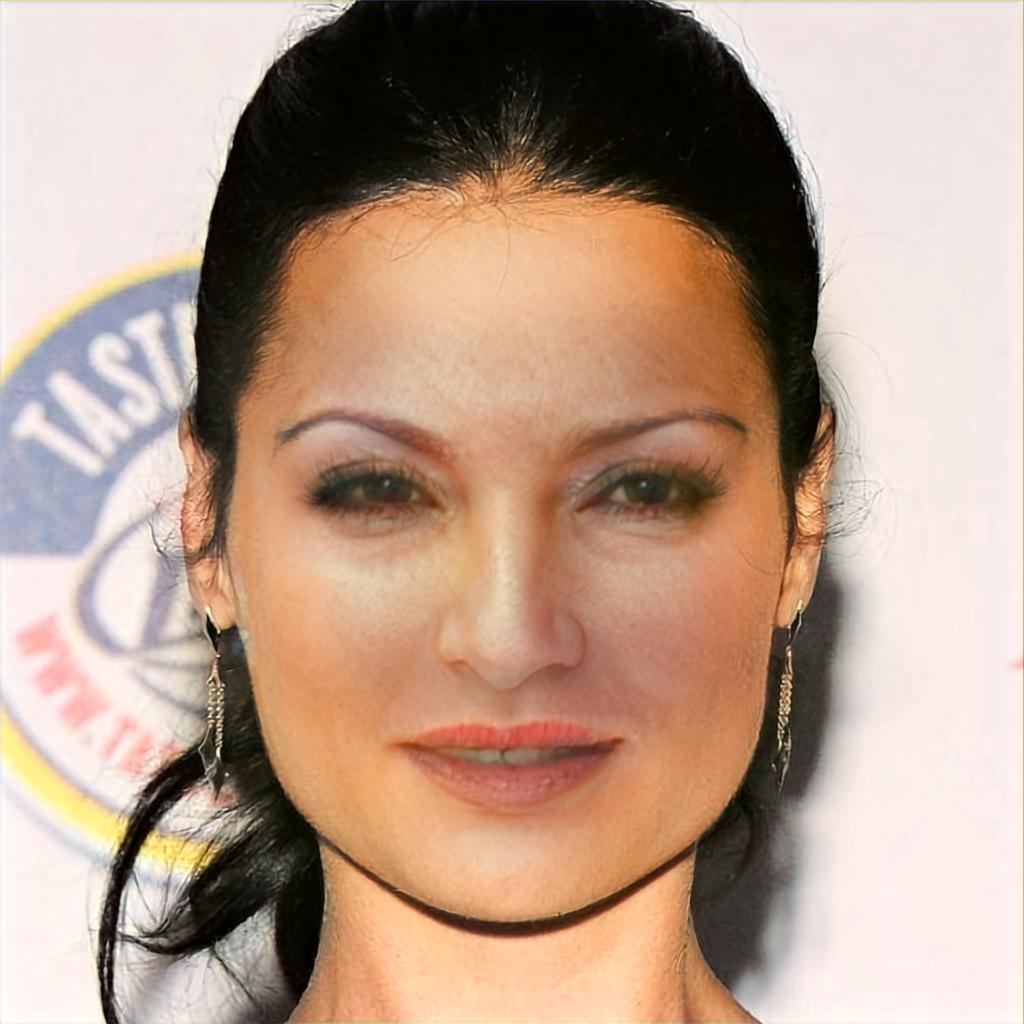}
     \caption{Var.1}
     \end{subfigure}
     \hspace{-2mm}
     \begin{subfigure}{.166\linewidth}
     \centering
     \includegraphics[width=\linewidth]{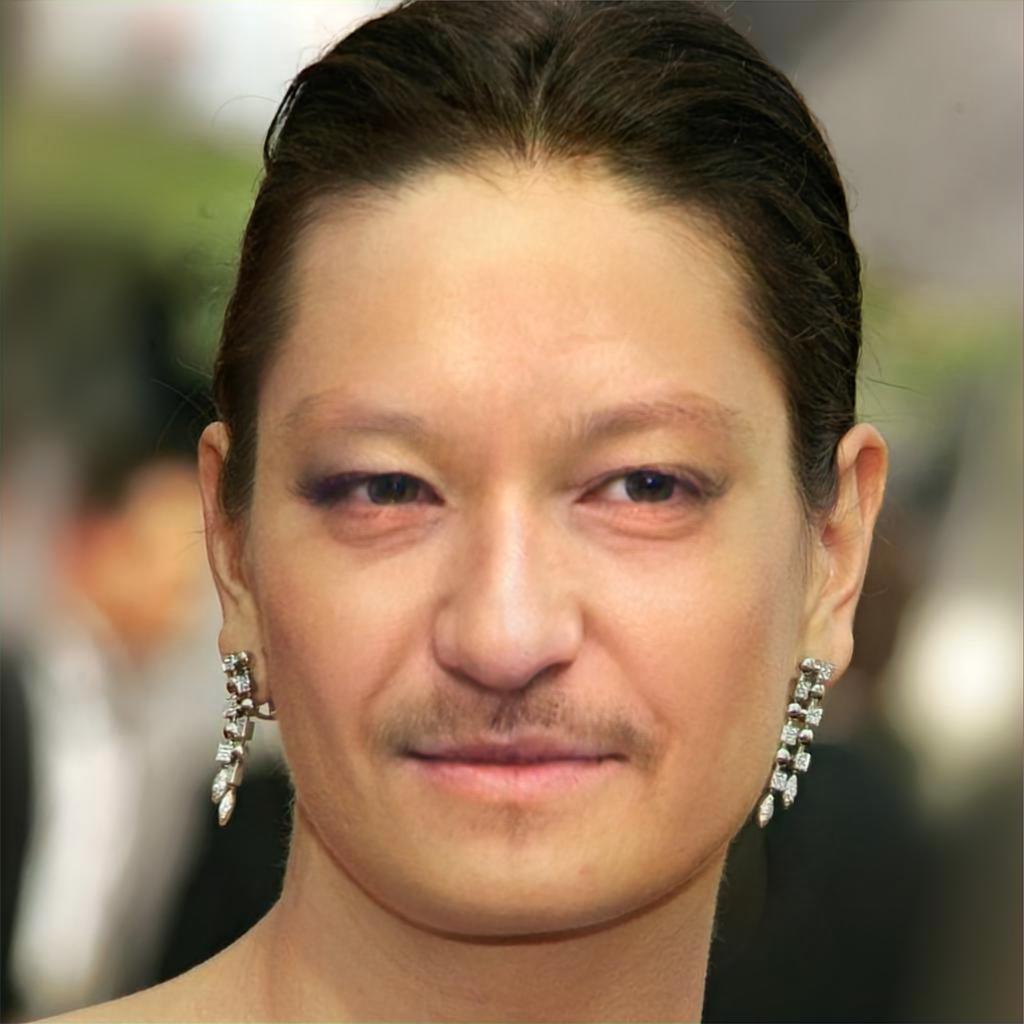}
     \includegraphics[width=\linewidth]{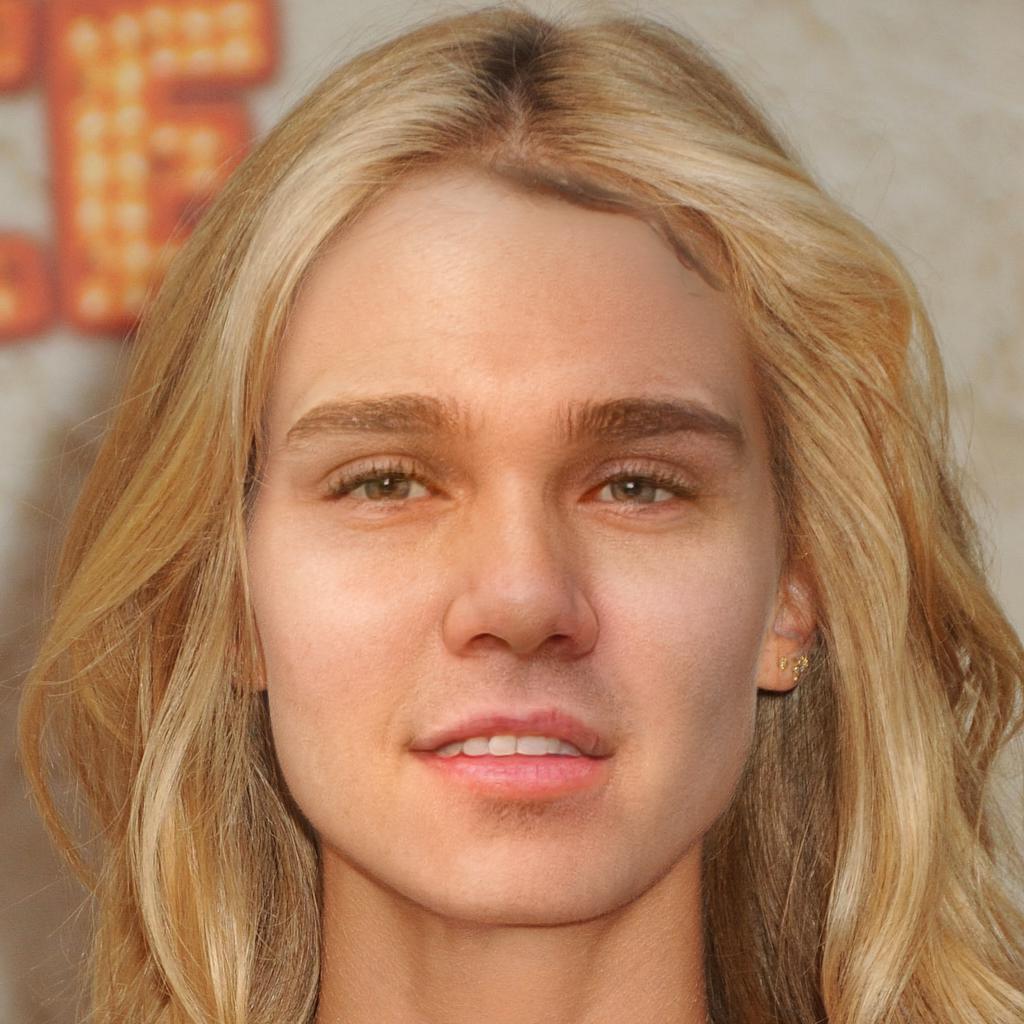}
     \includegraphics[width=\linewidth]{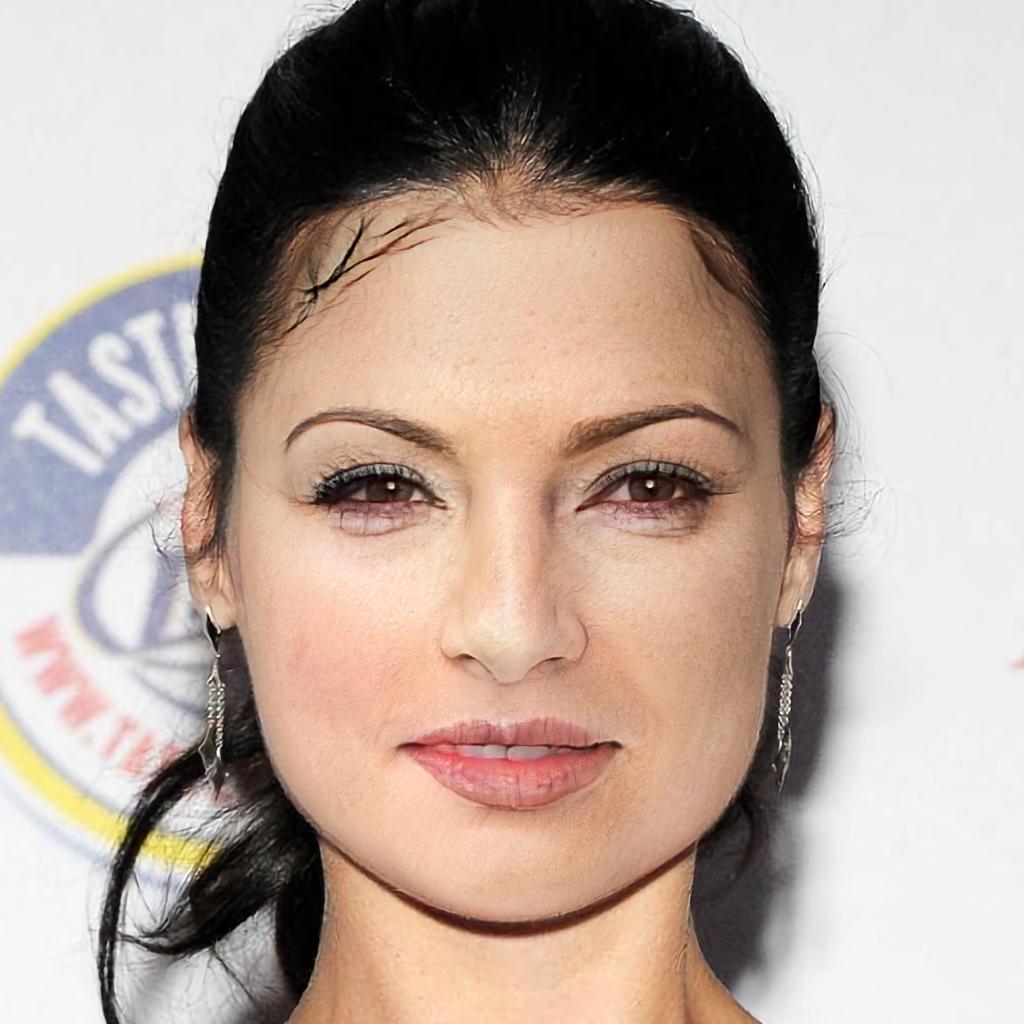}
     \caption{Var.2}
     \end{subfigure}
     \hspace{-2mm}
     \begin{subfigure}{.166\linewidth}
     \centering
     \includegraphics[width=\linewidth]{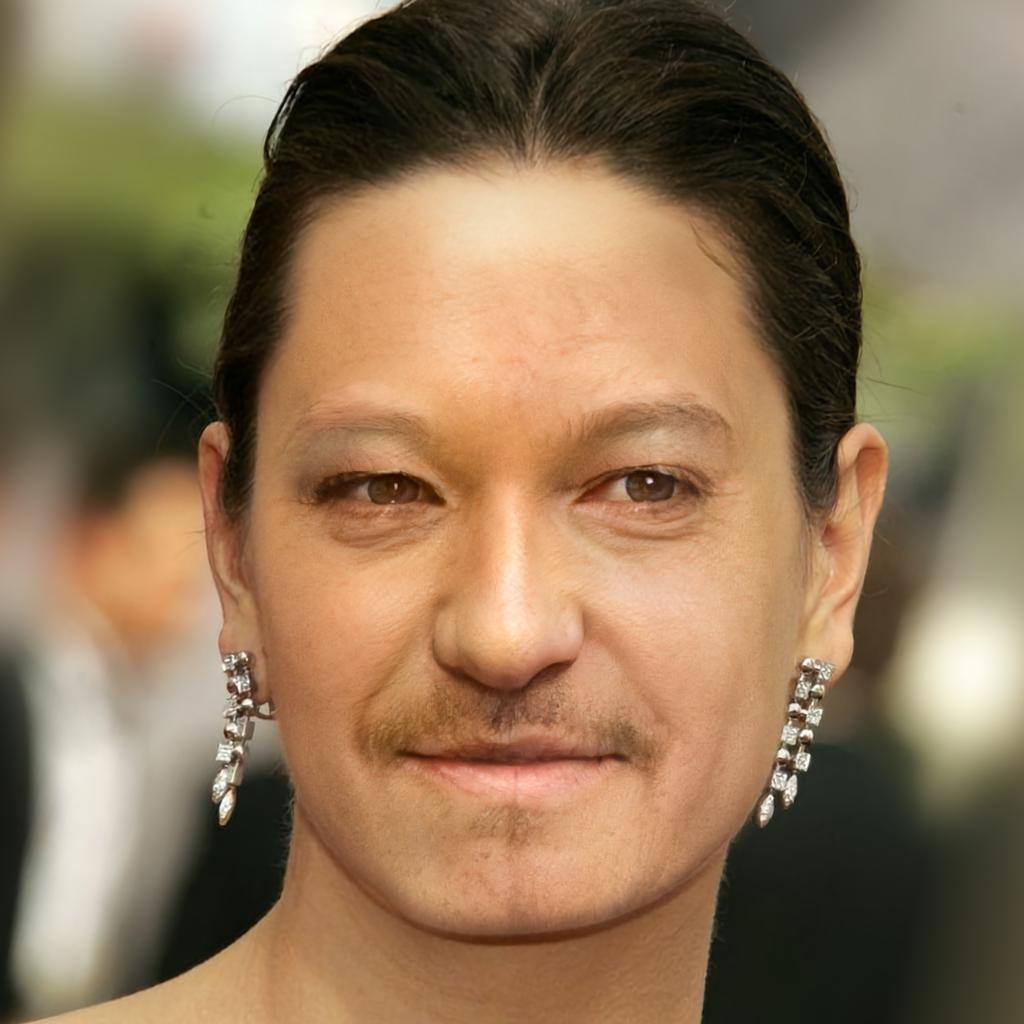}
     \includegraphics[width=\linewidth]{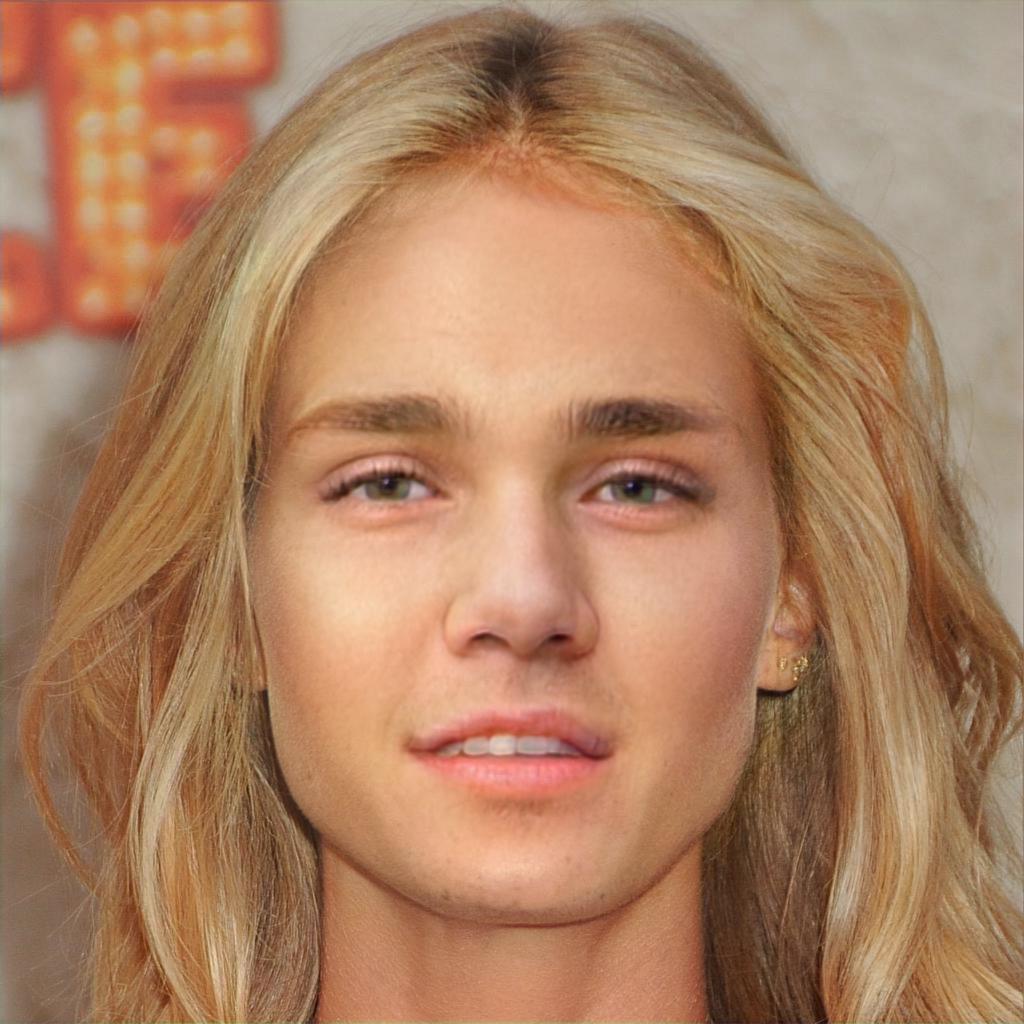}
     \includegraphics[width=\linewidth]{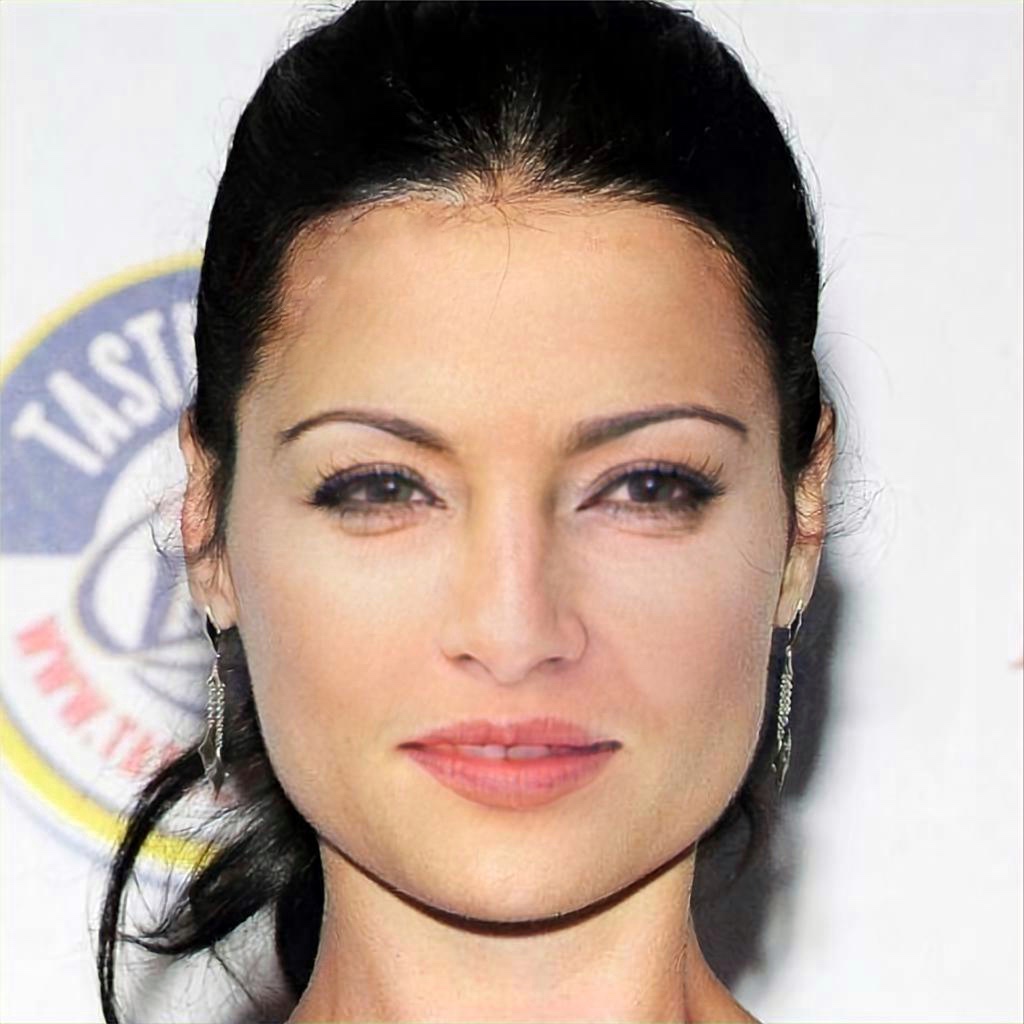}
     \caption{Ours}
     \end{subfigure}
     \capvspace
     \caption{Qualitative comparison of different variants. Each variant cannot transfer the target attributes effectively due to the entangled semantic transfer.}
     \label{fig.ablation}
     \figbottomvspace
\end{figure}

In this section, we perform ablation study and use the CelebA-HQ dataset to evaluate the effectiveness of our disentangled approach in transferring the attributes. We use MegaFS as a baseline, since they swap faces solely on two latent codes without disentanglement between attributes. We further include three variants of our approach with modifications on the modules and the loss functions. For the first variant (Var.1), we keep the appearance part of source code $w_s$ unchanged for producing the side-output face and  discard the style-transfer loss and the latent code swapping operation for appearances transfer, \ie, there is no explicit transfer guidance that contributes to the appearance attributes. For the second variant (Var.2), we discard our background transfer module and directly blend the side-output face with the target face image as the final output.
The qualitative comparison between different variants and the baseline is shown in Fig.~\ref{fig.ablation}. We can see that the MegaFS baseline leads to blurry faces and sharp boundaries. As mentioned previously, this is due to the highly entangled identity and attributes in the latent code, which can lead to information loss during the transfer.
The first variant presents the unnatural styles, which verifies the necessity of our appearance attributes transfer that provides explicit appearance guidance in the swapping process. The second variant presents a notable blending boundary between the swapped face region and the background (see the $2_{nd}$ and $3_{rd}$ samples), due to the lack of the background transfer module which fuses multi-resolution features from the source and the target for a natural blending. Meanwhile, our target encoder-decoder structure also can tolerance the structure difference between source and targets, and produce the final plausible results. Thanks to our disentangled attributes transfer, our final swapped faces present the successful semantic transfer from the targets, with the source identity well preserved.
% \bailin{Need to check the discussion here.}

%\vspace{-1mm}
\subsection{Face Swapping on High-resolution Videos}
\begin{figure}[t]
   \centering
     \captionsetup[subfigure]{labelformat=empty,justification=centering}
     \vspace{-7mm}
     \begin{subfigure}{.16\linewidth}
     \centering
     \vspace{10mm}
     \includegraphics[width=\linewidth]{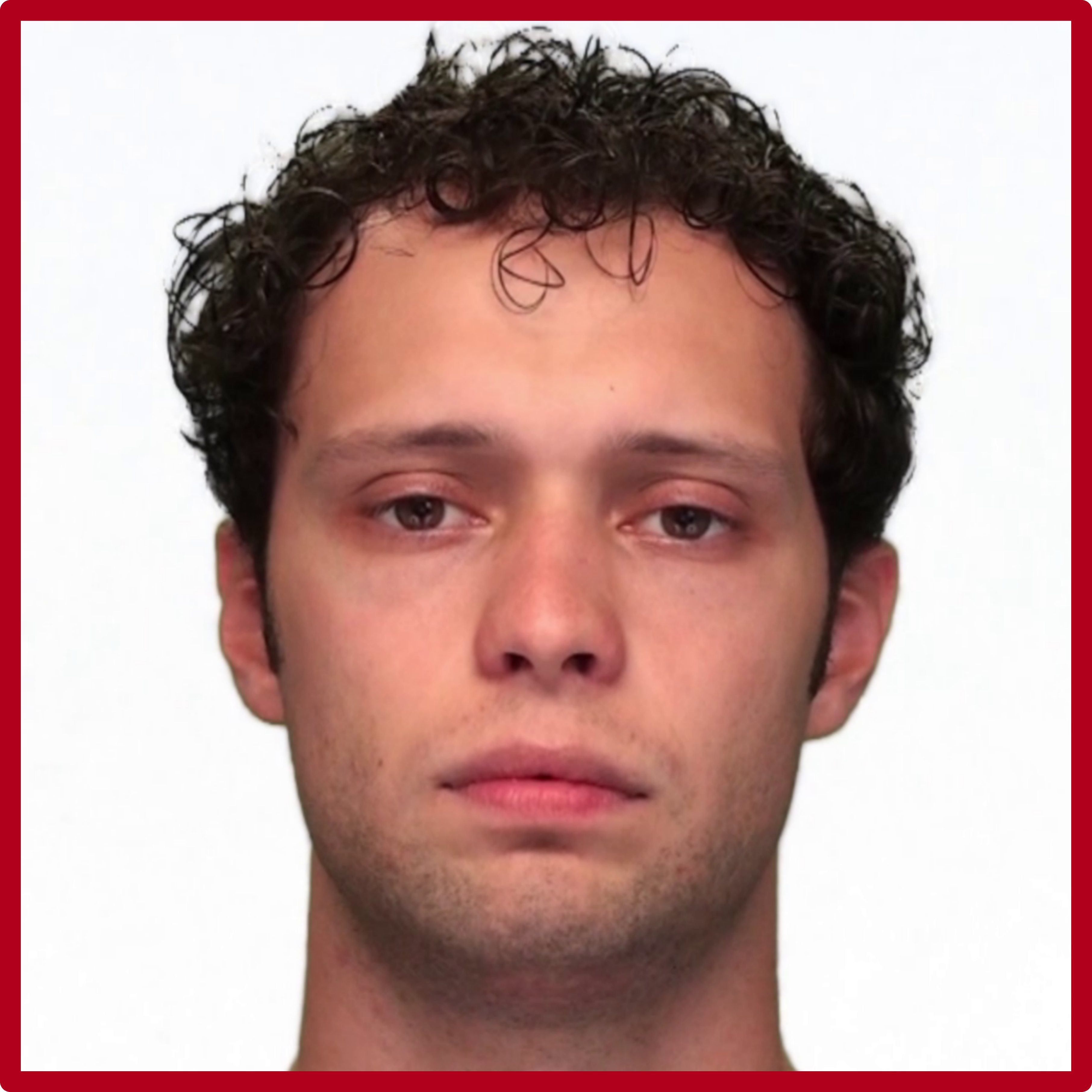}\vspace{-1mm}\caption{source}\vspace{10mm}
     \end{subfigure}
     \begin{subfigure}{.16\linewidth}
     \centering
     \includegraphics[width=\linewidth]{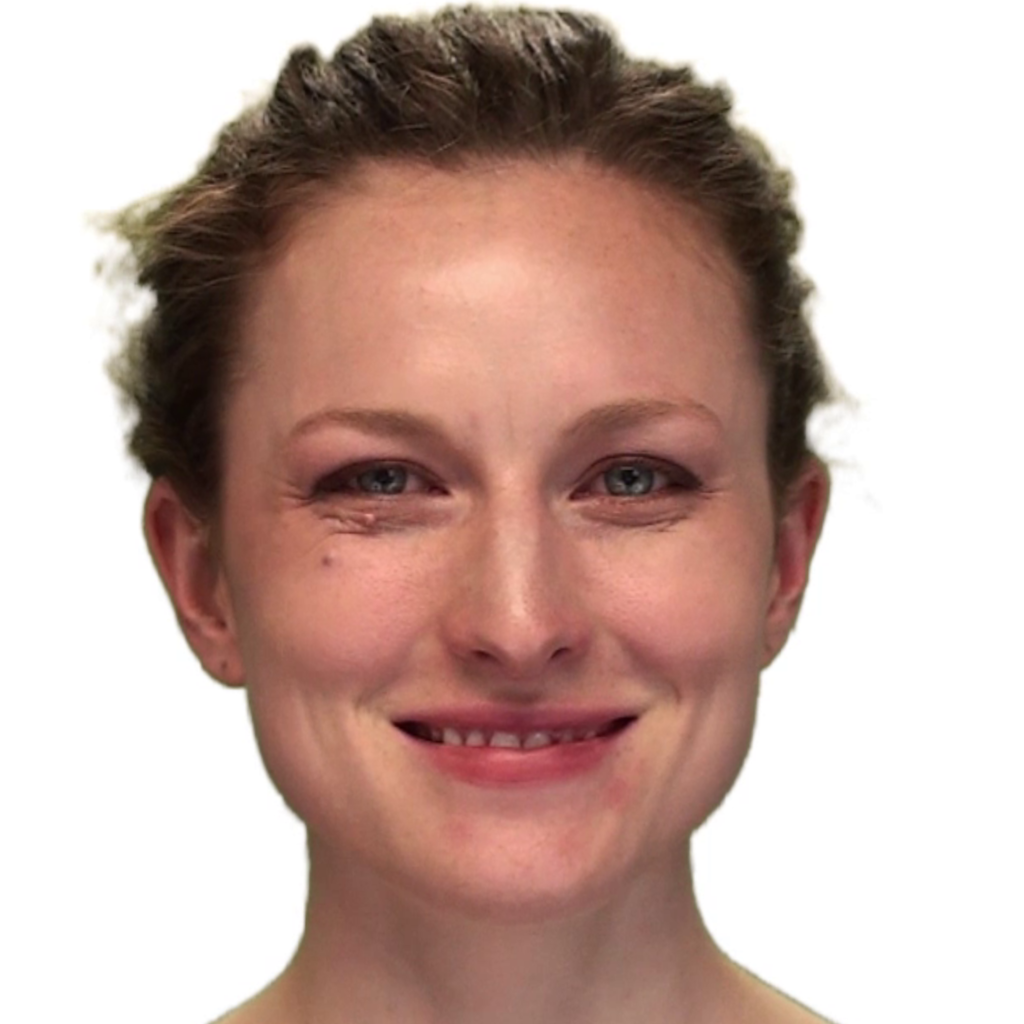}
     \includegraphics[width=\linewidth]{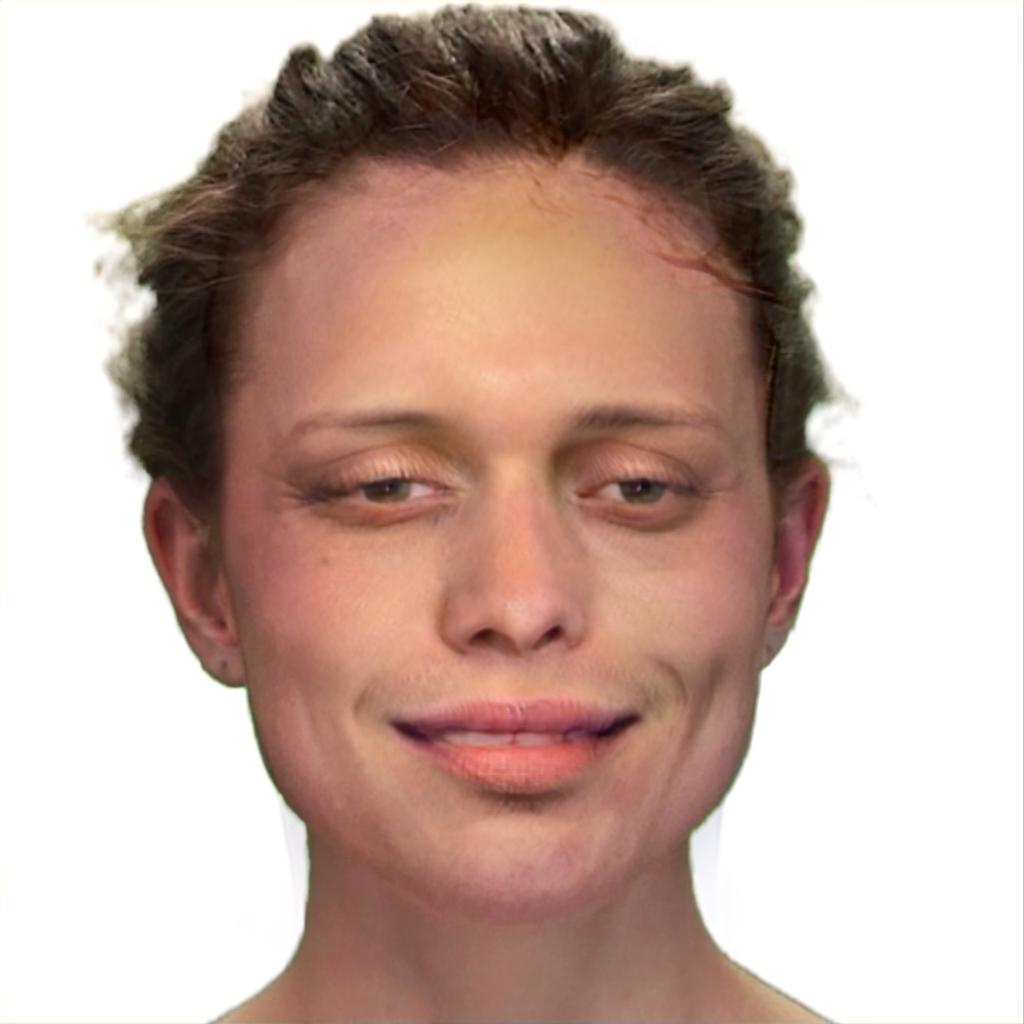}
     \includegraphics[width=\linewidth]{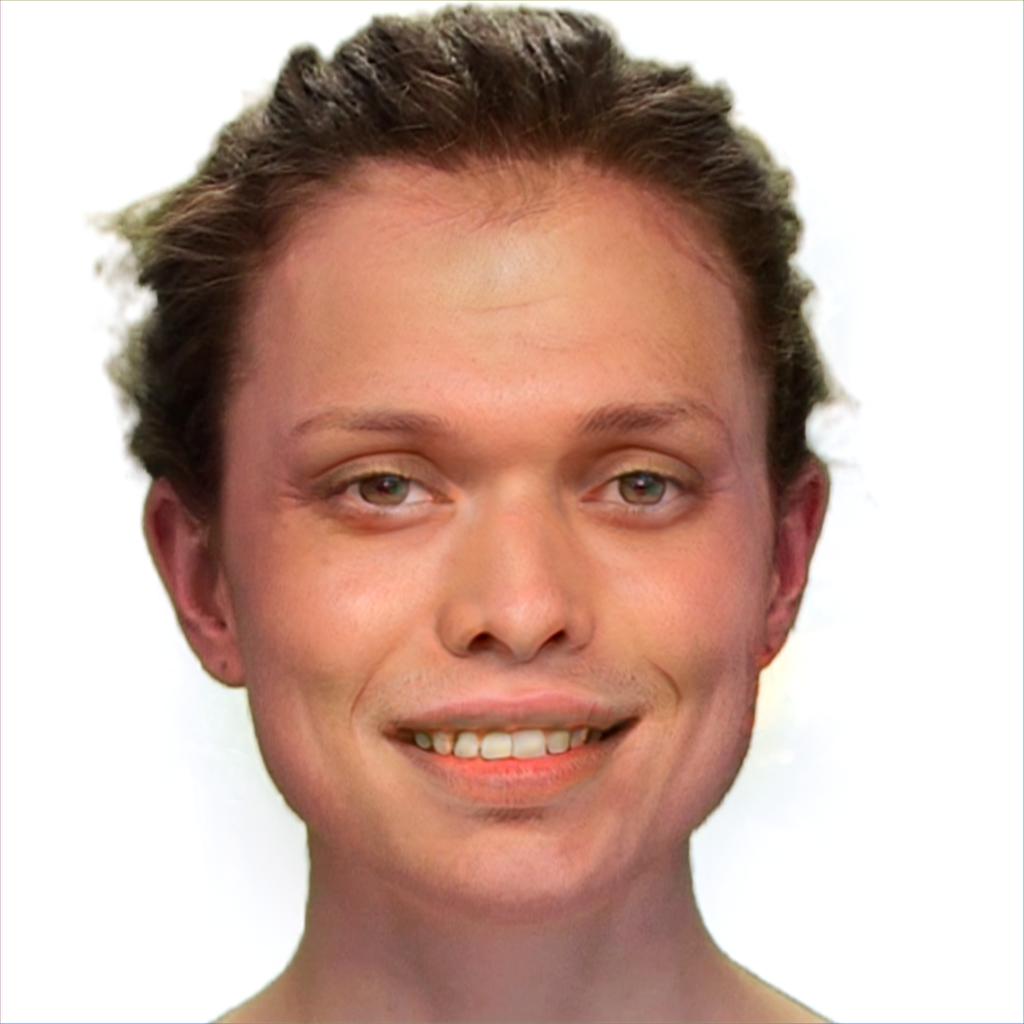}
     \end{subfigure}
     \hspace{-2mm}
     \begin{subfigure}{.16\linewidth}
     \centering
     \includegraphics[width=\linewidth]{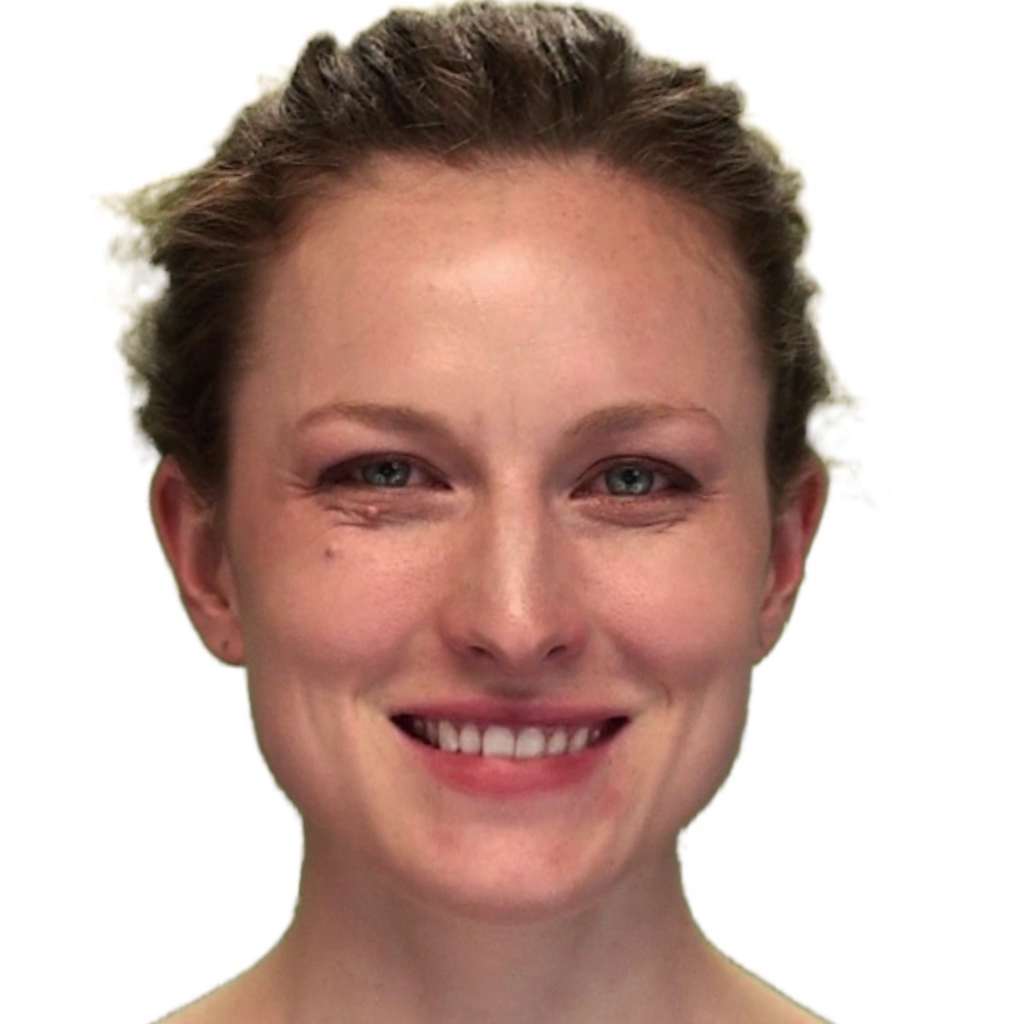}
     \includegraphics[width=\linewidth]{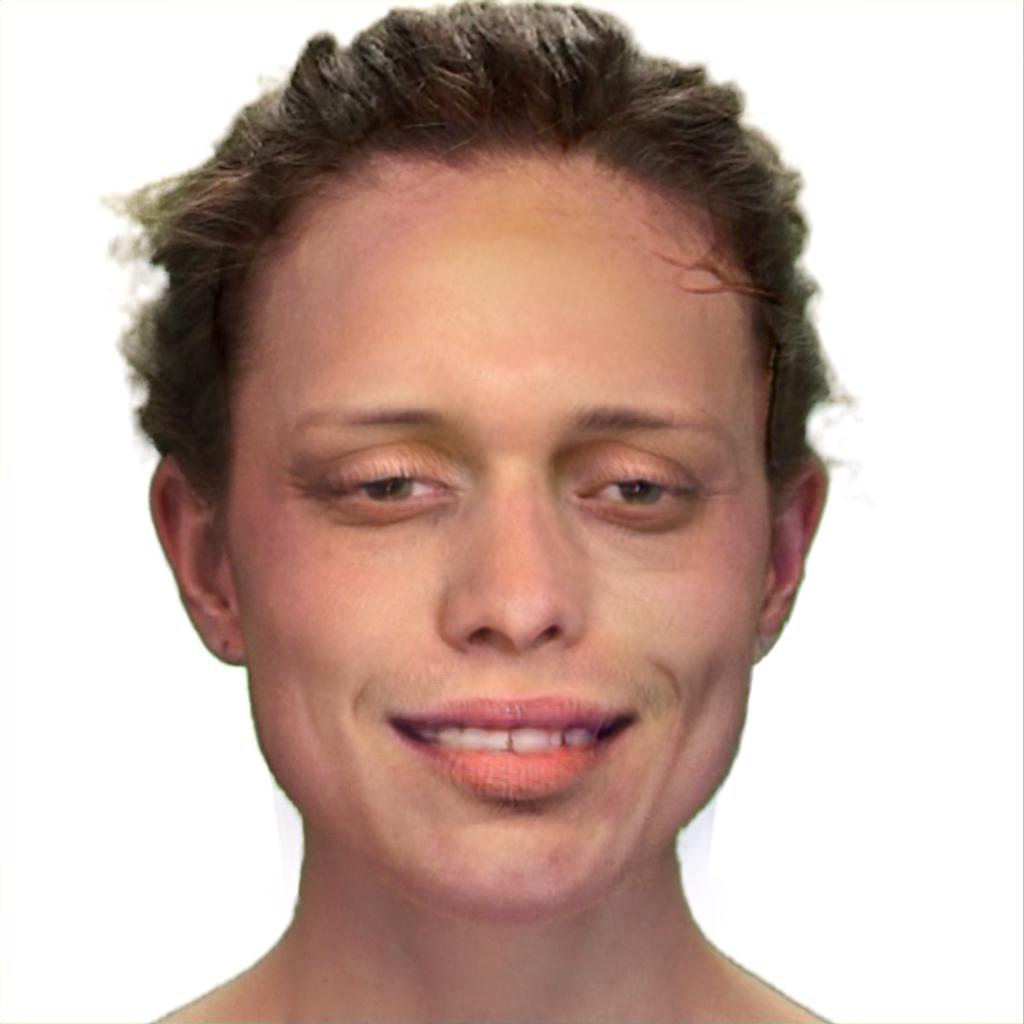}
     \includegraphics[width=\linewidth]{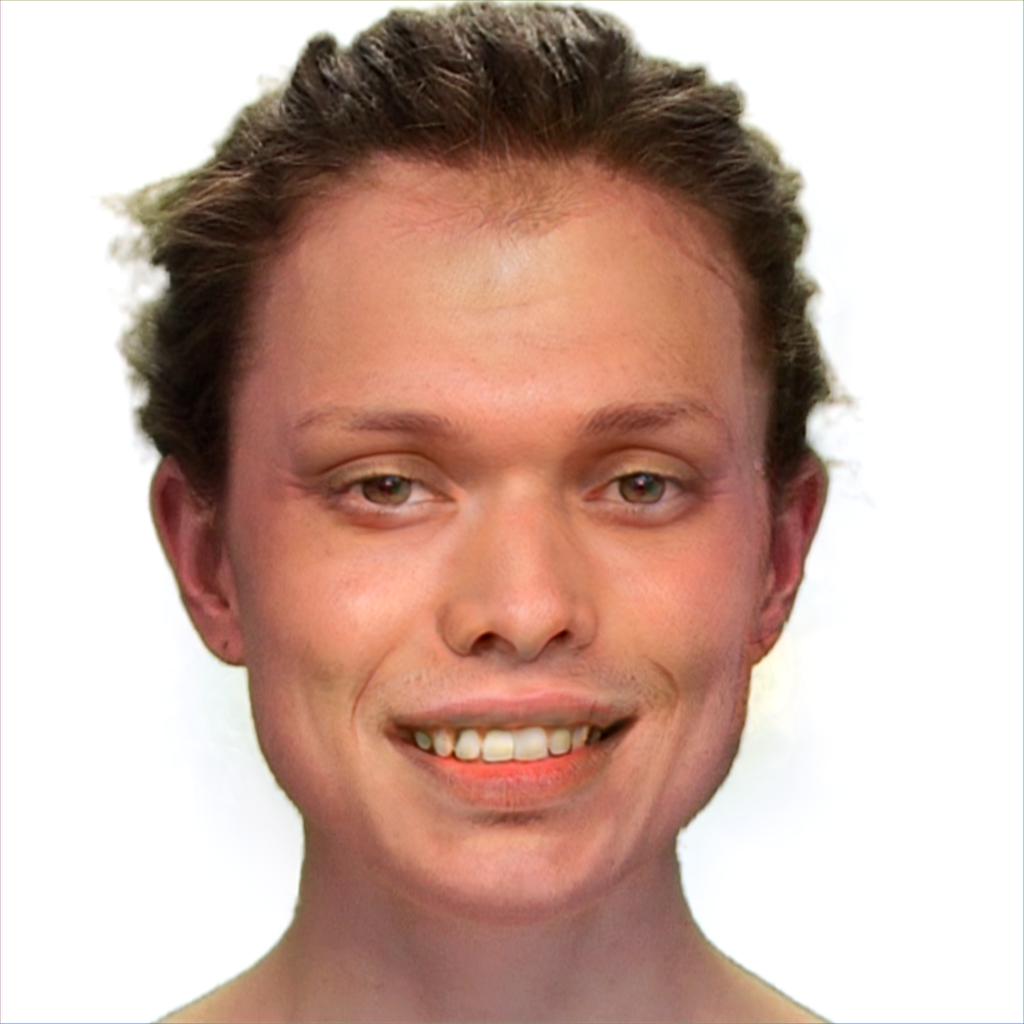}
     \end{subfigure}
     \hspace{-2mm}
     \begin{subfigure}{.16\linewidth}
     \centering
     \includegraphics[width=\linewidth]{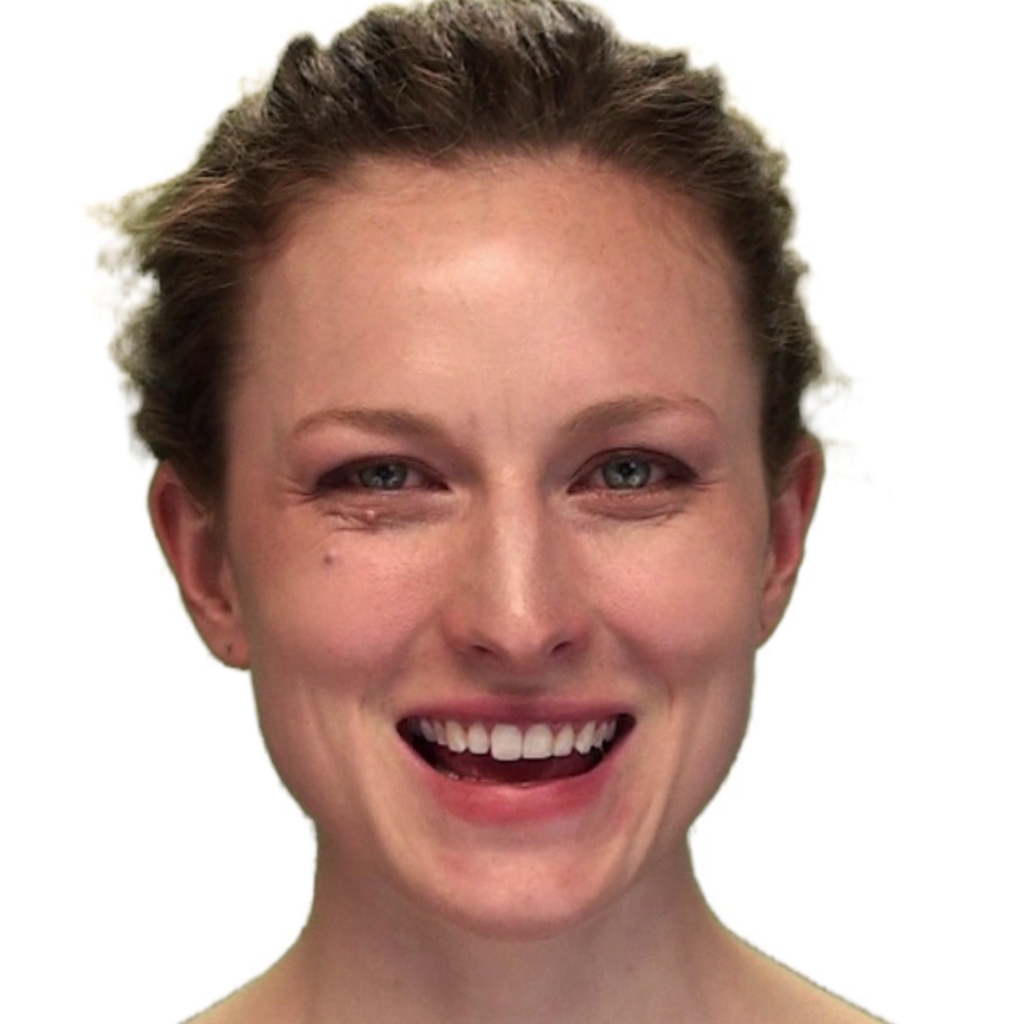}
     \includegraphics[width=\linewidth]{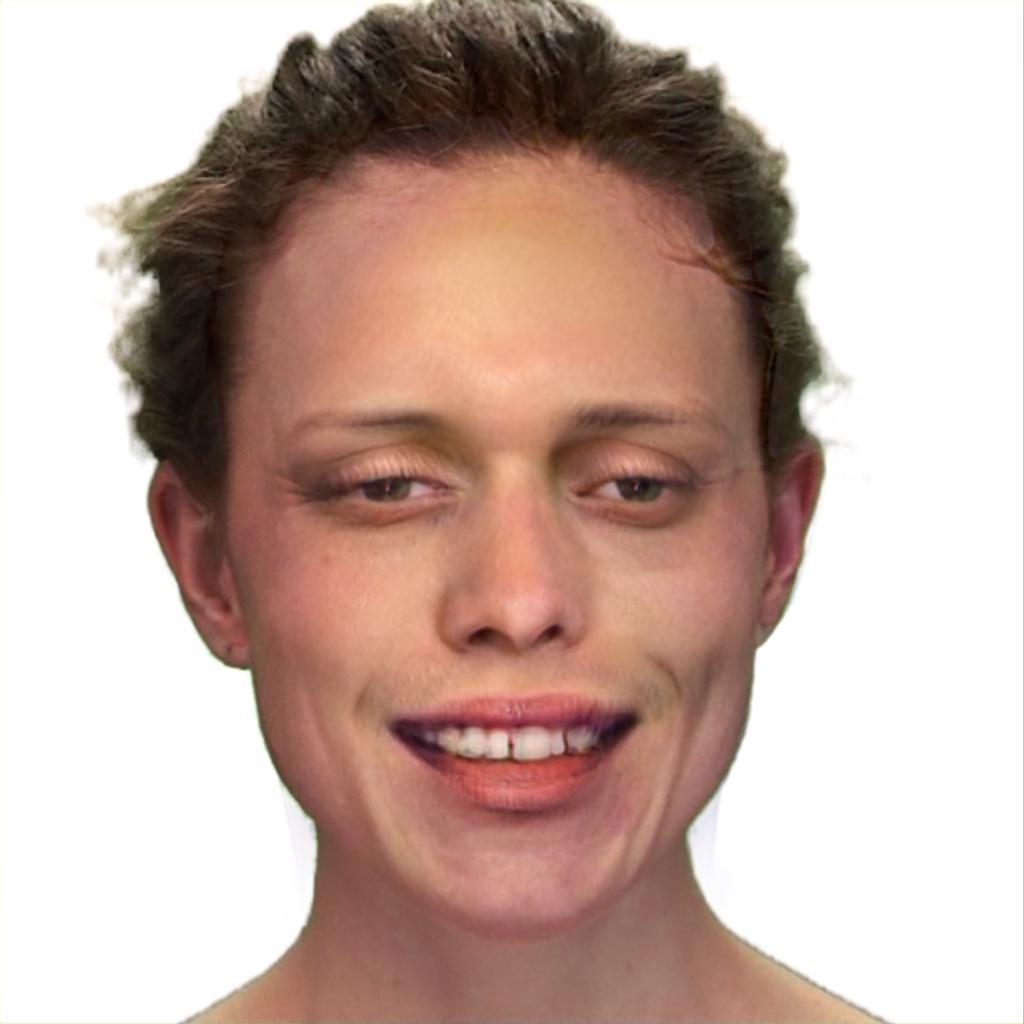}
     \includegraphics[width=\linewidth]{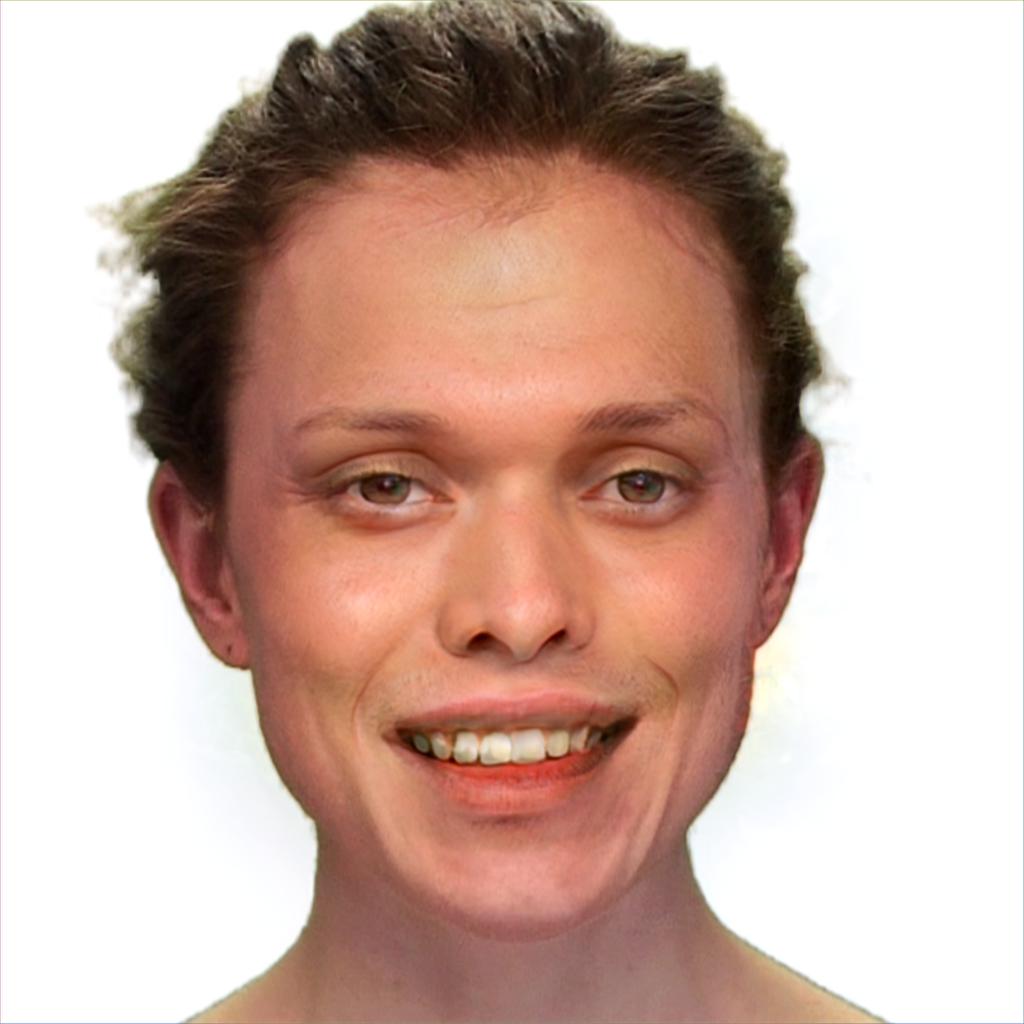}
     \end{subfigure}
     \hspace{-2mm}
     \begin{subfigure}{.16\linewidth}
     \centering
     \includegraphics[width=\linewidth]{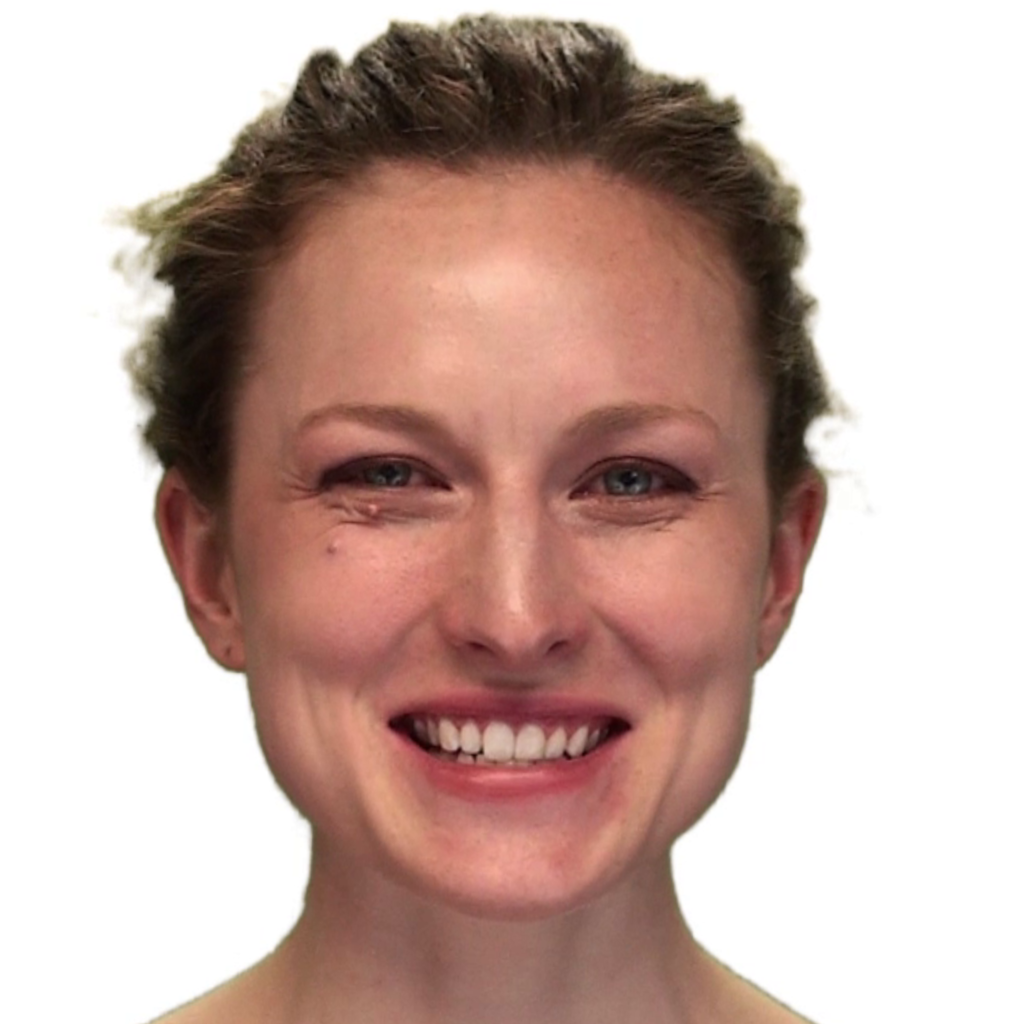}
     \includegraphics[width=\linewidth]{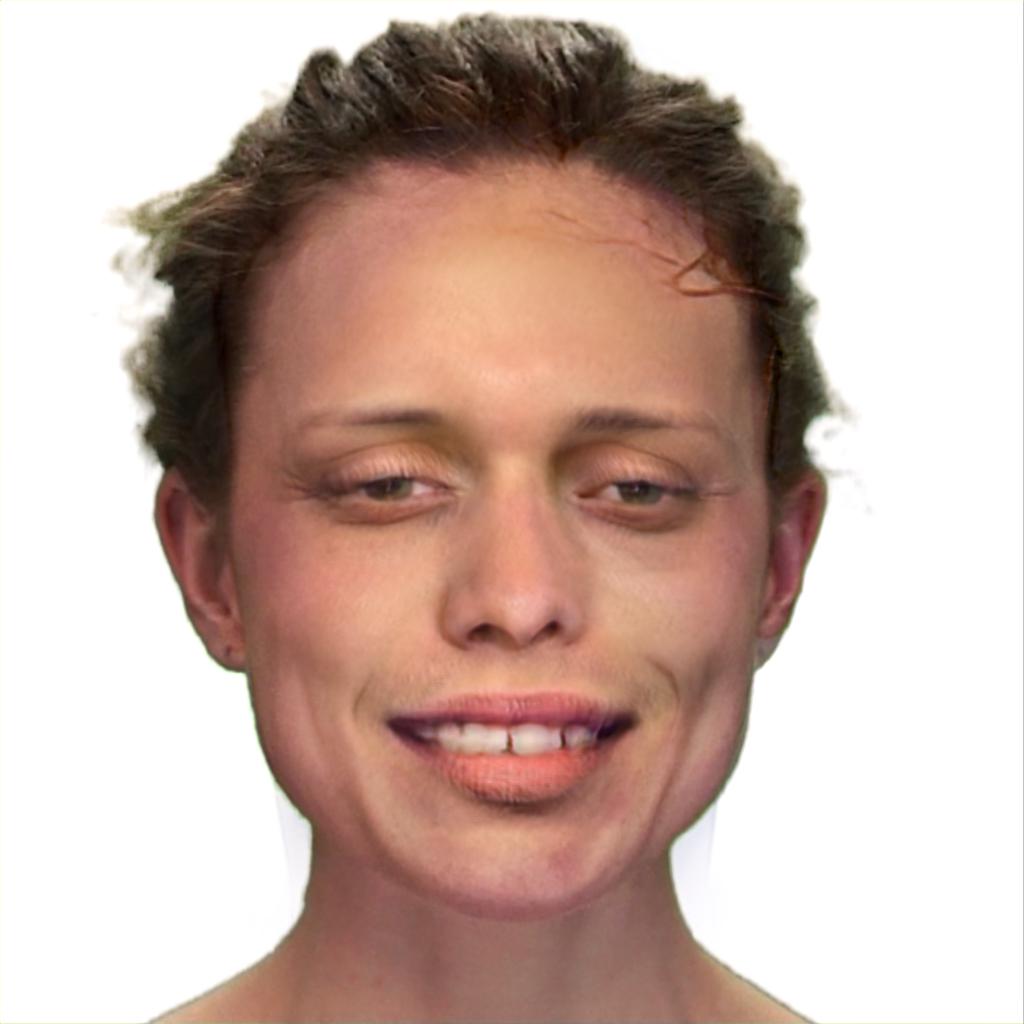}
     \includegraphics[width=\linewidth]{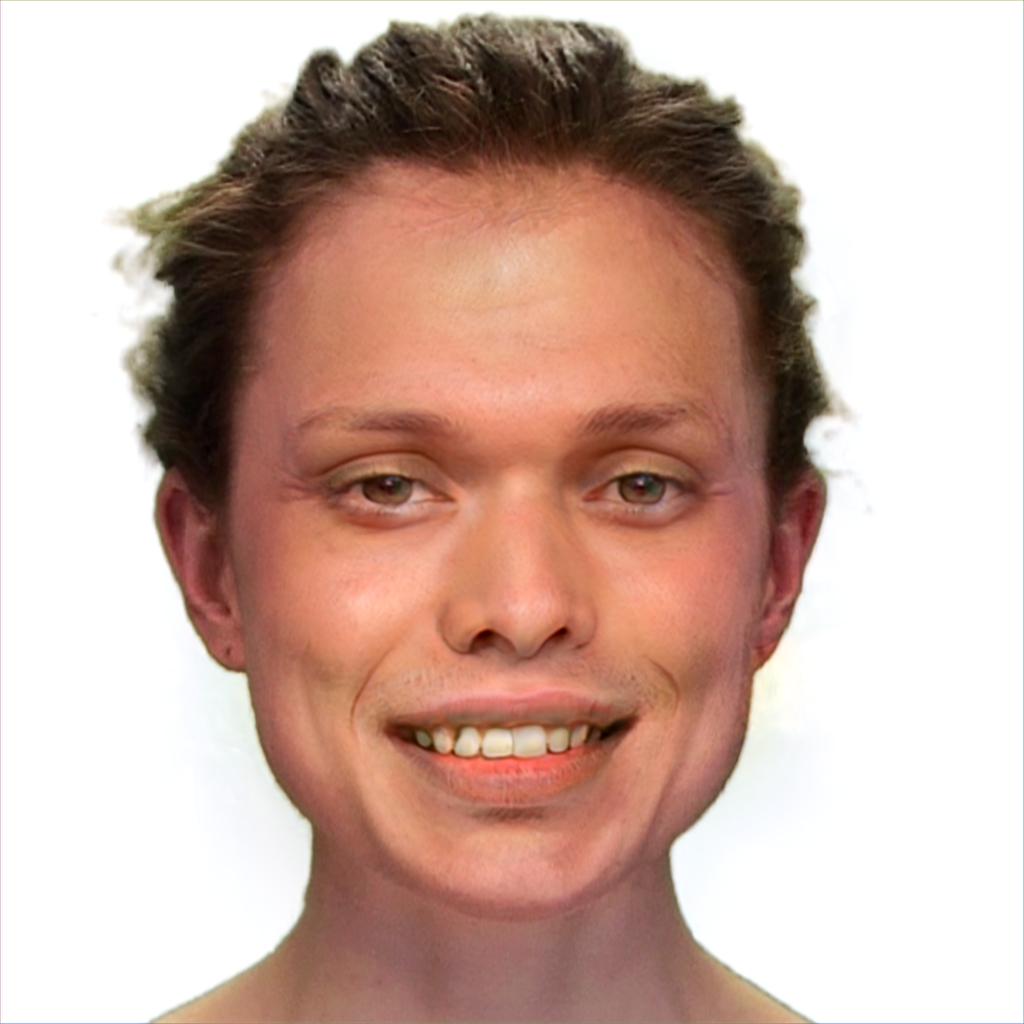}
     \end{subfigure}
     \hspace{-2mm}
     \begin{subfigure}{.16\linewidth}
     \centering
     \includegraphics[width=\linewidth]{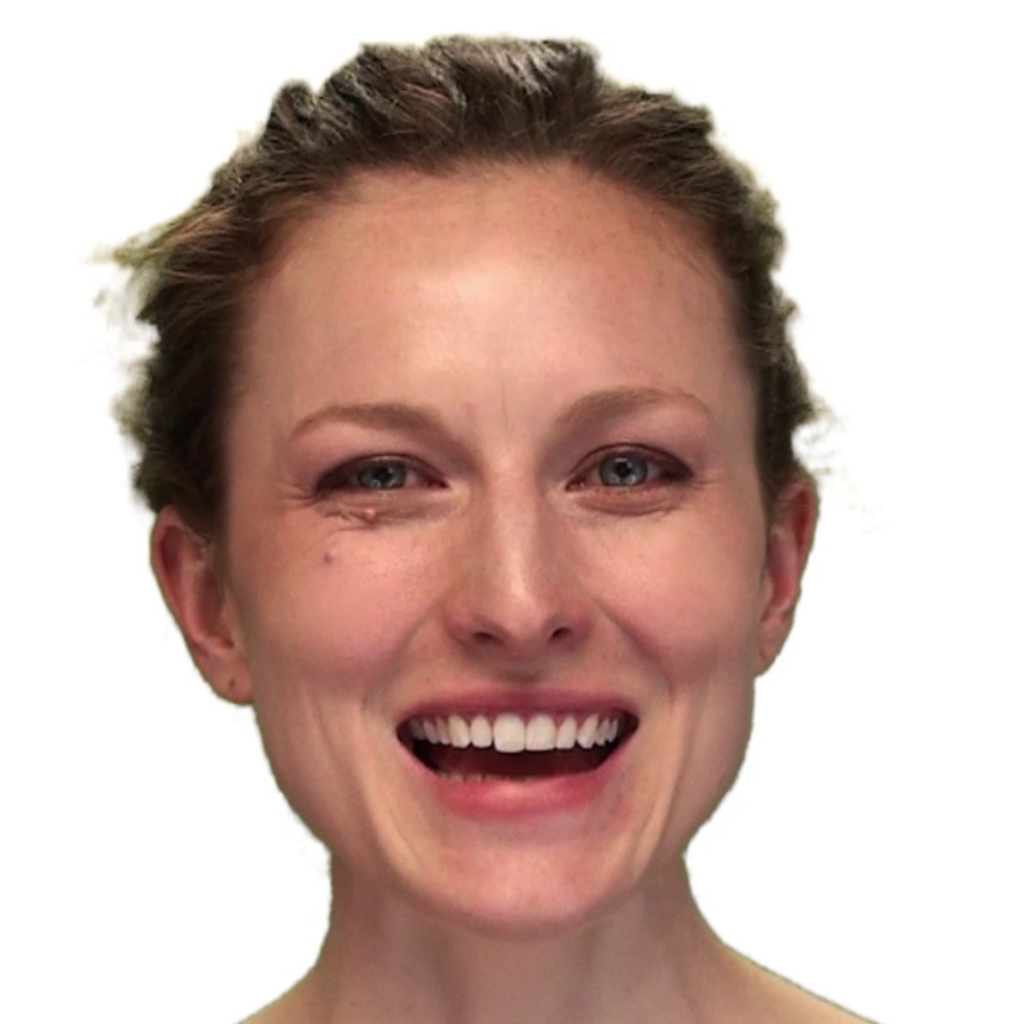}
     \includegraphics[width=\linewidth]{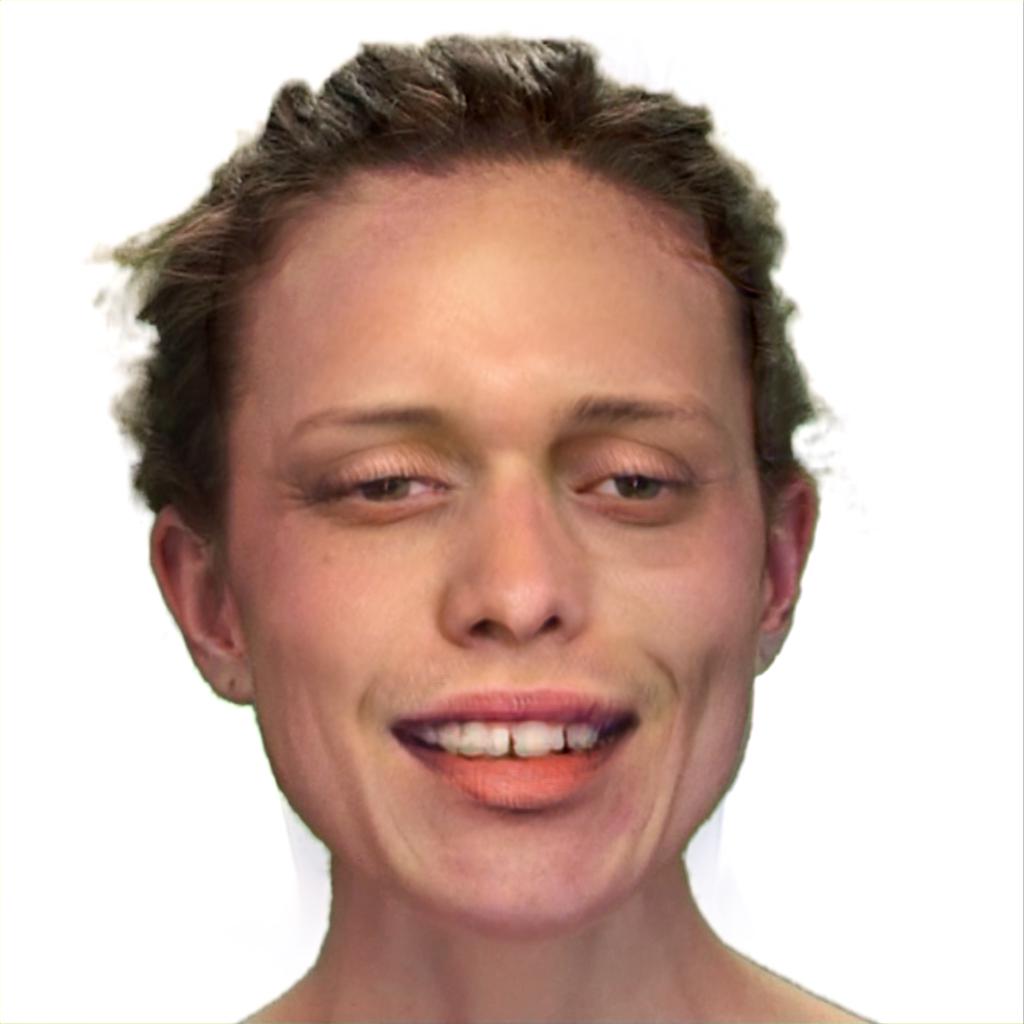}
     \includegraphics[width=\linewidth]{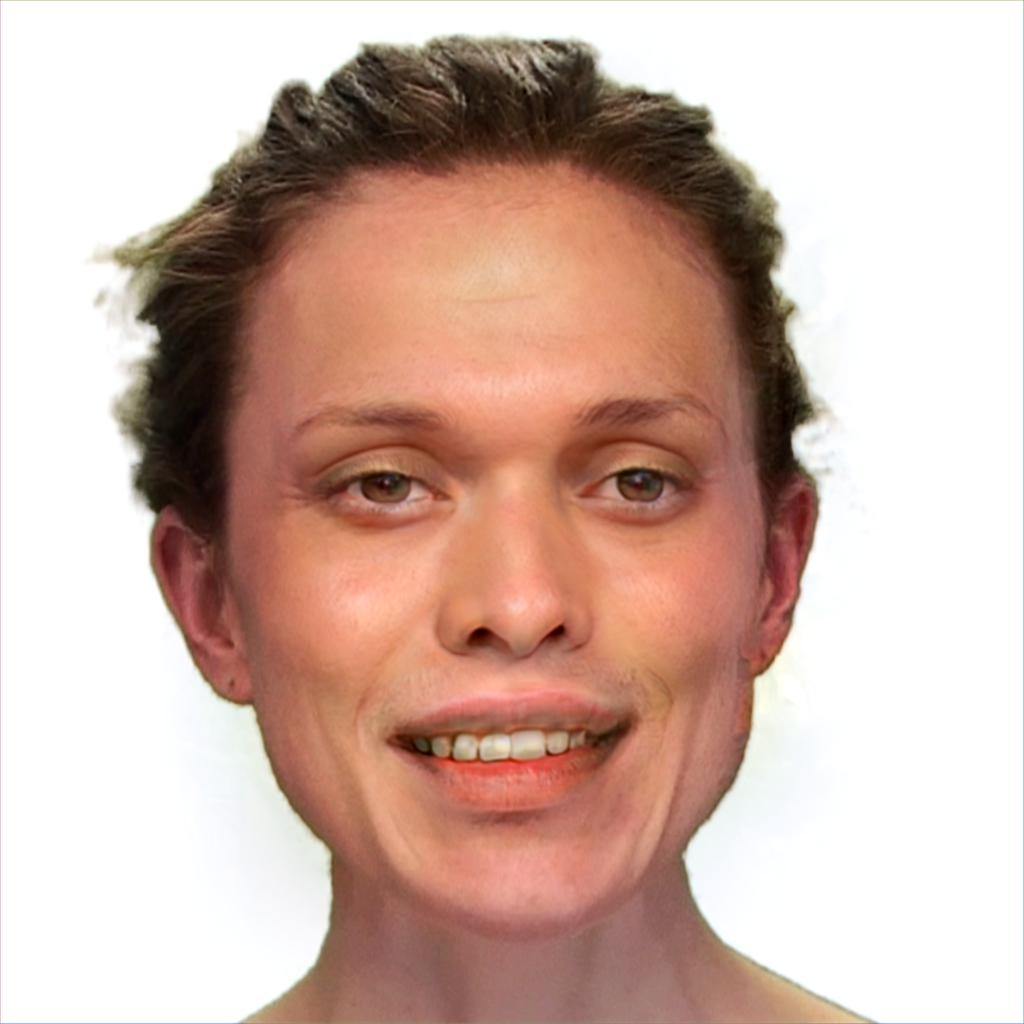}
     \end{subfigure}
     \rotatebox[origin=c]{270}{\hspace{2mm}\scriptsize{Target}\hspace{5mm} \scriptsize{Image FS} \hspace{3mm} \scriptsize{Video FS}}
     \capvspace
     \caption{Examples of video face swapping. The figure in the red box is the source face, and the top row shows the target video frames. Applying face swapping on each frame individually leads to the incoherent structure and appearance (middle row), while our trajectory constraints inhibit such incoherence (bottom row).}
     \label{fig.video_img}
     \figbottomvspace
\end{figure}

Fig.~\ref{fig.video_img} shows qualitative results of our method on high-resolution videos.
Applying our method individually to each frame leads to incoherence between neighboring frames, while our code and flow trajectory constraints improve the coherence and visual quality considerably.

% !TeX root = egpaper_for_review.tex
\section{Conclusion and Discussions}
% \paraheading{Conclusion}
We present a novel high-resolution face swapping based on the inherent prior knowledge of pre-trained StyleGAN. We categorize the attributes into structure and appearance ones, and transfer them separately in the disentangled latent space. We propose a landmark encoder that predicts a latent direction for the structure attributes transfer. The StyleGAN generative features resulting from the transferred latent code are aggregated with the multi-resolution features from a target image encoder to transfer the background information and generate a high-quality result. We further extend the method to video face swapping by enforcing two spatio-temporal constraints. Extensive experiments demonstrate the superiority of our disentangled attributes transfer in terms of hallucination quality and consistency.
\paraheading{Limitations}
Since we transfer the attributes in the latent space of StyleGAN, the quality of the result relies heavily on the GAN inversion method. In particular, if the inversion does not produce faithful latent codes for the source and target faces, the result is not guaranteed to preserve the identity of the source.
Our method transfers all the target appearance attributes to the result image and does not support selective transfer of appearance from both the source and the target. Such fine-grained control can be beneficial for some applications such as content creation, but would require further disentanglement between different categories of appearance attributes. This can be an avenue for further research. 
\paraheading{Potential Negative Impact}
Although not the purpose of this work, realistic face swapping can potentially be misused for deepfakes-related applications. The risk can be mitigated by gated release of the model and by forgery detection methods that are able to spot CNN-generated images~\cite{wang2020cnn}.
In addition, our method can be potentially used to generate new high-resolution test cases for benchmarking and further developing forgery detection techniques~\cite{rossler2019faceforensics}.

\paragraph{Acknowledgement.} This project is supported by the National Natural Science Foundation of China (No. 61972162); Guangdong International Science and Technology Cooperation Project (No. 2021A0505030009); Guangdong Natural Science Foundation (No. 2021A1515012625); Guangzhou Basic and Applied Research Project (No. 202102021074); and CCF-Tencent Open Research fund (RAGR20210114).

%%%%%%%%% REFERENCES
{\small
\bibliographystyle{ieee_fullname}
\bibliography{egbib}
}
\newpage
\section{Supplementary}

We provide more content in this supplementary, include the detail of architectures, more qualitative comparisons with MegaFS~\cite{zhu2021one} and the quantitative comparisons of different variants.
%-------------------------------------------------------------------------
% architecture detail
\section{Detail of Architectures}
\subsection{Landmark Encoder}
We use a modification of pSp~\cite{richardson2020encoding} as our landmark encoder. pSp inverts the input image to the $W+$ latent space with a feature pyramid through three levels: the coarse, medium, and fine, which controls different level of attributes. We discard pSp's higher layers and only use the coarse and medium layers to produce the semantic transfer direction $\overrightarrow{{n}}$.

\subsection{Target Encoder}

Our target encoder consists of 8 downsample blocks, each block contains a convolutional layer and a `Blur' layer~\cite{karras2020analyzing}. The target encoder produces the multi-resolution target features, which includes: $8\times8\times512$; $16\times16\times512$; $32\times32\times512$; $64\times64\times512$; $128\times128\times256$; $256\times256\times128$; $512\times512\times64$; $1024\times1024\times32$. Then they blend with the source features produced by StyleGAN generator with the corresponding resolution.

\subsection{Decoder}
The decoder has a mirror structure with target encoder, and we use the transpose convolution for upsampling. In the first upsampling block, we take the blending results of target and source features as input. And in each rest block, we concatenate the blended features with the output of last block as input.

\section{More Ablation Studies}

\subsection{Effectiveness of the landmark encoder.}
To test the effectiveness of the landmark encoder, we implement a variant, named \emph{Var.3}, that infers the structure transfer latent direction~$\overrightarrow{n}$ from the face images directly without using the landmarks. We replace the landmark encoder with an image encoder, and feed the concatenation of the source and target face images to the encoder.

\begin{figure}[t]
    \centering
    \includegraphics[width=0.99\columnwidth]{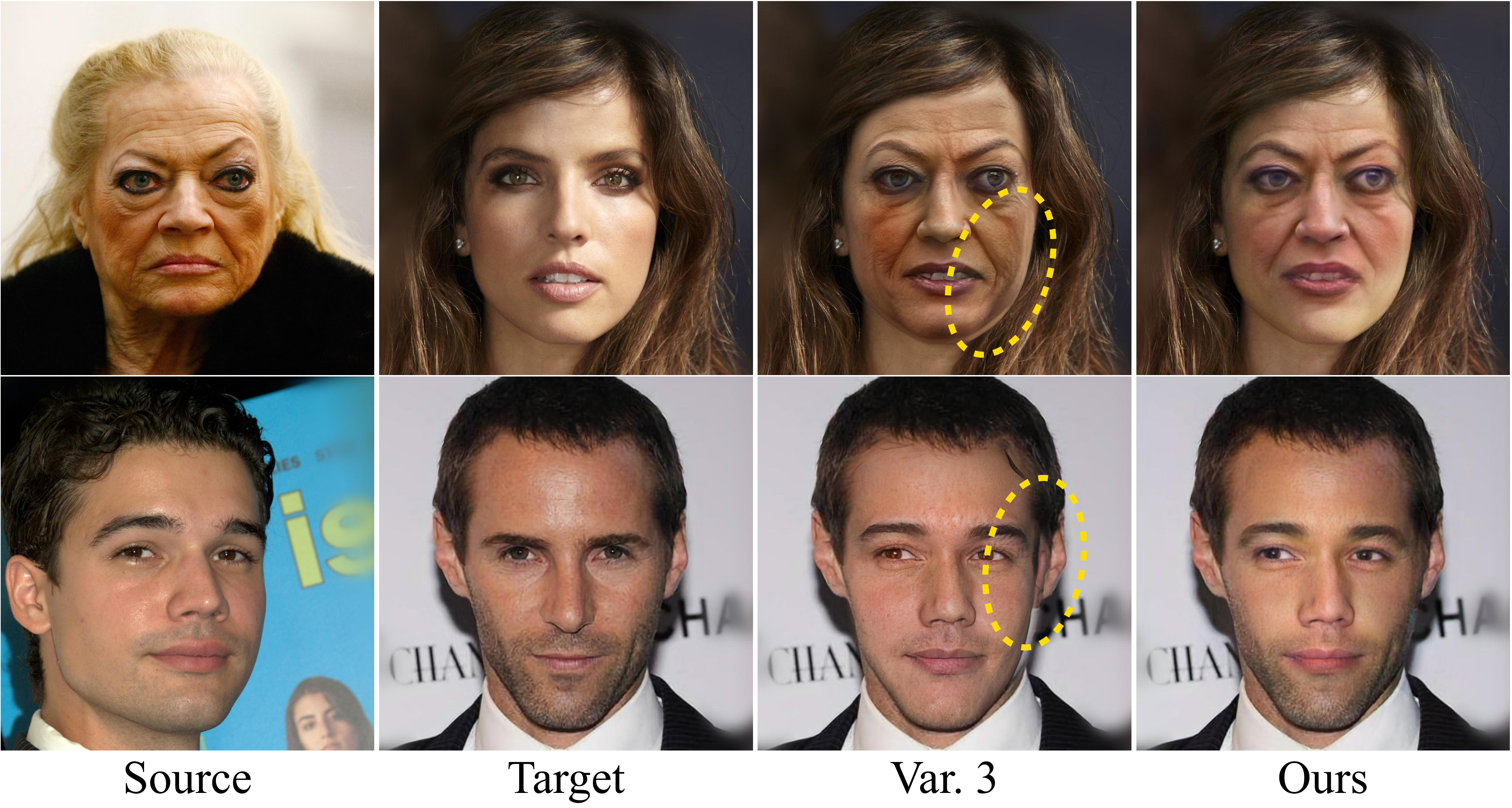}
    \vspace{-0.5em}
    \caption{Qualitative results of Var.3. Problematic regions are circled out in yellow.}
    \vspace{-0.5em}
    \label{fig:abla}
\end{figure}

Its results are shown in Tab.~\ref{table:abla} and Fig.~\ref{fig:abla}. In Tab.~\ref{table:abla}, it has worse performance for the metrics \emph{Pose Err.} and \emph{Exp. Err.}, which indicates that the landmark encoder plays a vital role in pose and expression preservation. Besides, without the landmark guidance, Var.3 cannot effectively blend the target background with the source inner face using target mask; this leads to lower quality results than our method in Fig.~\ref{fig:abla}.

\subsection{Quantitative Comparisons of Different Variants}

Tab.~\ref{table:abla} shows quantitative evaluation results on CelebA-HQ. We can see that both Var.1 and Var.2 have a higher FID value than our approach, indicating worse quality of face images from both variants. This is because our appearance transfer can narrow the color gap while our background transfer eliminates the blending artifacts, both helping to improve the quality of our results. On the other hand, Var.1 and Var.2 have similar performance as our approach on the other three metrics related to identity, pose and expression. This is because they also utilize the landmark encoder and the identity-preservation loss, which help to transfer the pose and expression while preserving the identity. The analysis of Var.3 can be seen in last subsection.

\begin{table}[t]
    \caption{Quantitative of different variants on the CelebA-HQ dataset with four metrics. $\downarrow$ denotes the lower the better and vice versa, the best results are marked in \textbf{bold}.}
    % \vspace{-7.0mm}
    \begin{center}
        \setlength{\tabcolsep}{0.12cm}{
            \begin{tabular}{c|c|c|c|c}
                \hline
                Variants  &{ID Simi.$\uparrow$} &{Pose Err.$\downarrow$} &{Exp. Err.$\downarrow$} &FID$\downarrow$ \\
                \hline
                Baseline      &0.5214      &3.498     &2.95   &11.645          \\
                Var.1         &0.5645      &3.034     &2.76   &10.986          \\
                Var.2         &0.5539      &3.023      &2.74   &10.345          \\
                Var.3         &0.5542      &3.242      &3.11   &10.642          \\

                Ours          &\textbf{0.5688}    &\textbf{2.997}   &\textbf{2.74}  &\textbf{9.987}    \\
                \hline
            \end{tabular}
        }
    \end{center}
    \label{table:abla}
    \vspace{-5mm}
\end{table}

\section{More Qualitative Comparisons}
\begin{figure*}[t]
   \centering
    % \rotatebox[origin=c]{90}{Ours \hspace{15mm} InD \hspace{15mm} pSp \hspace{15mm} I2S}
    \captionsetup[subfigure]{labelformat=empty,justification=centering}
    \begin{subfigure}{.2\linewidth}
     \centering
     \includegraphics[width=\linewidth]{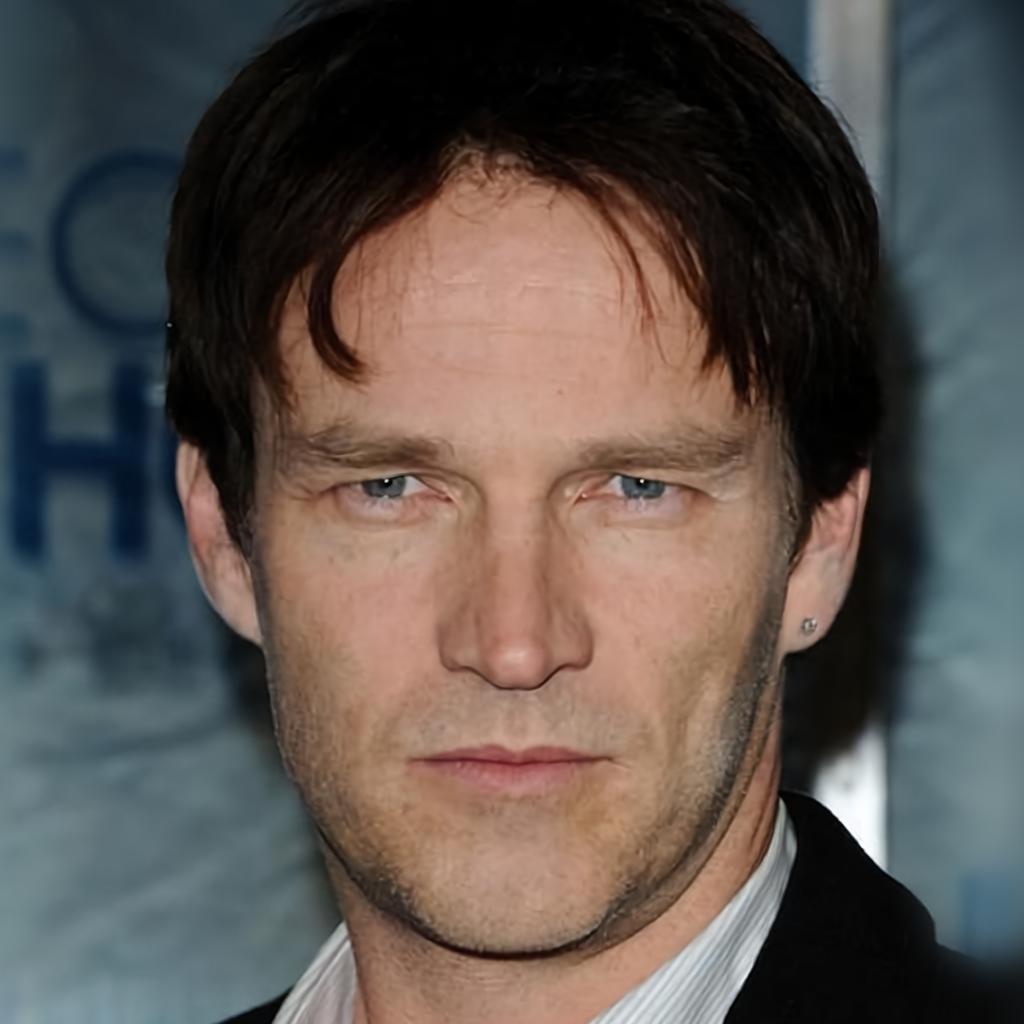}
     \includegraphics[width=\linewidth]{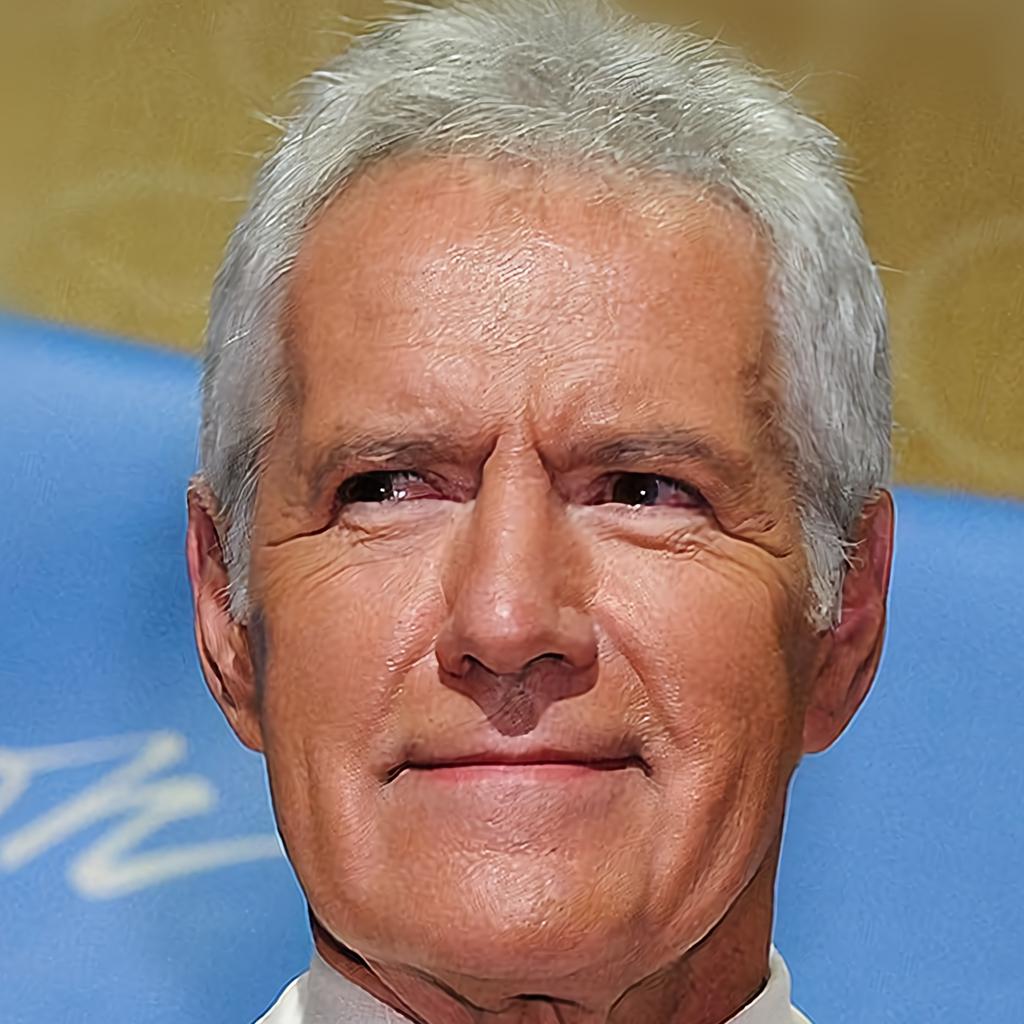}
     \includegraphics[width=\linewidth]{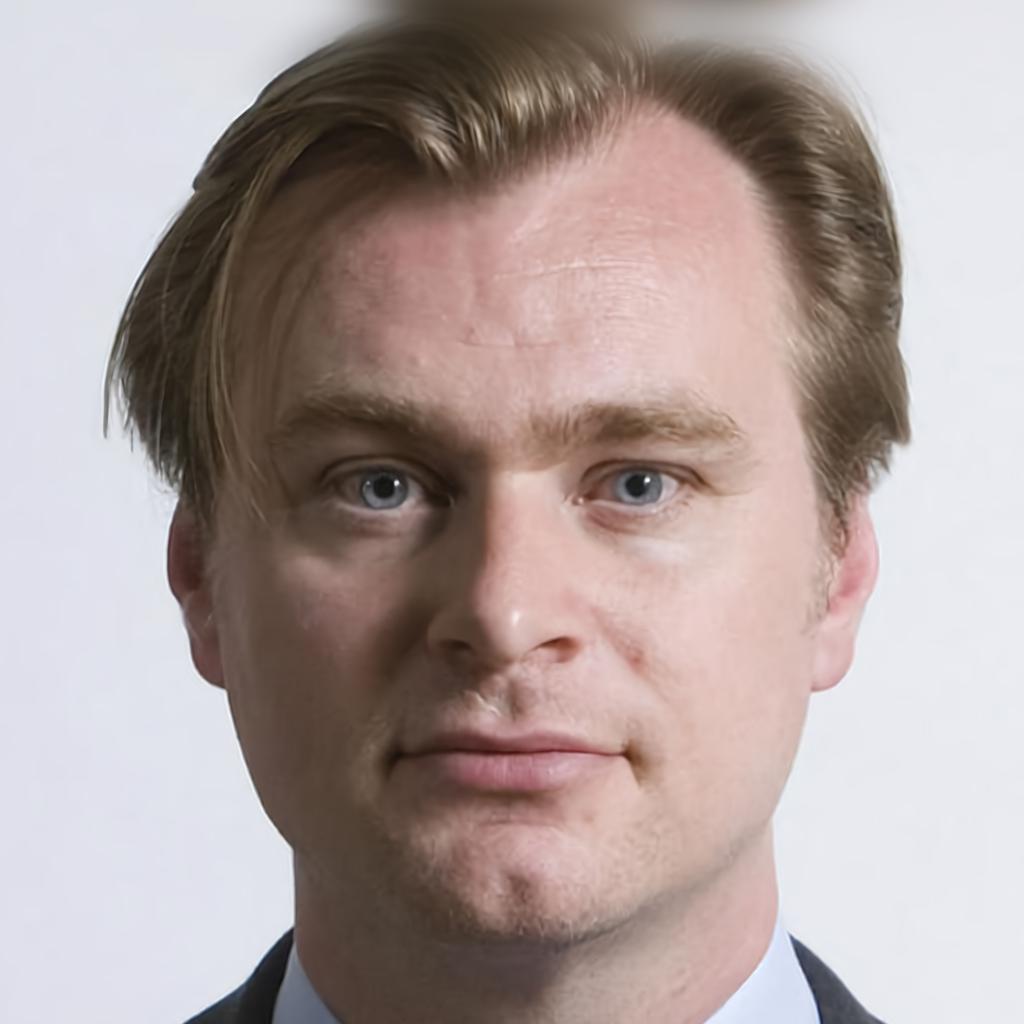}
     \includegraphics[width=\linewidth]{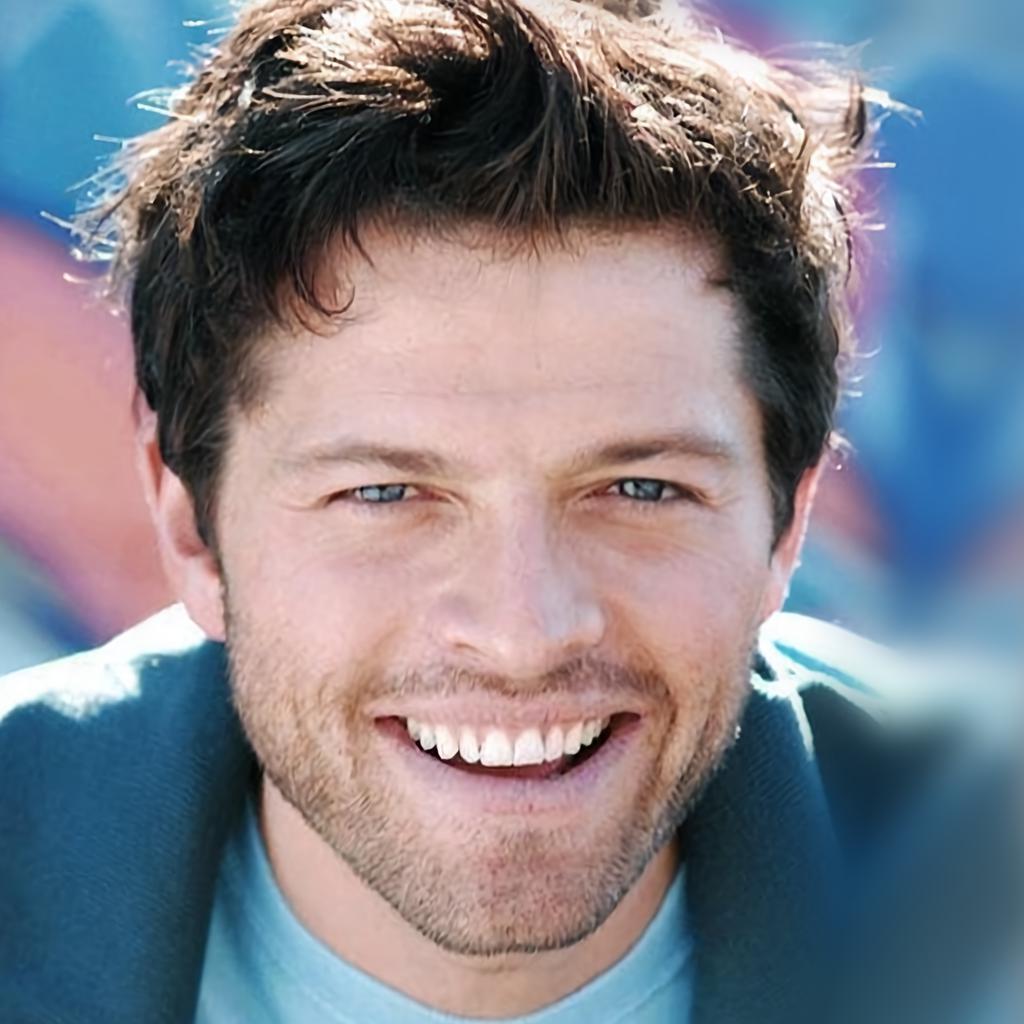}
     \includegraphics[width=\linewidth]{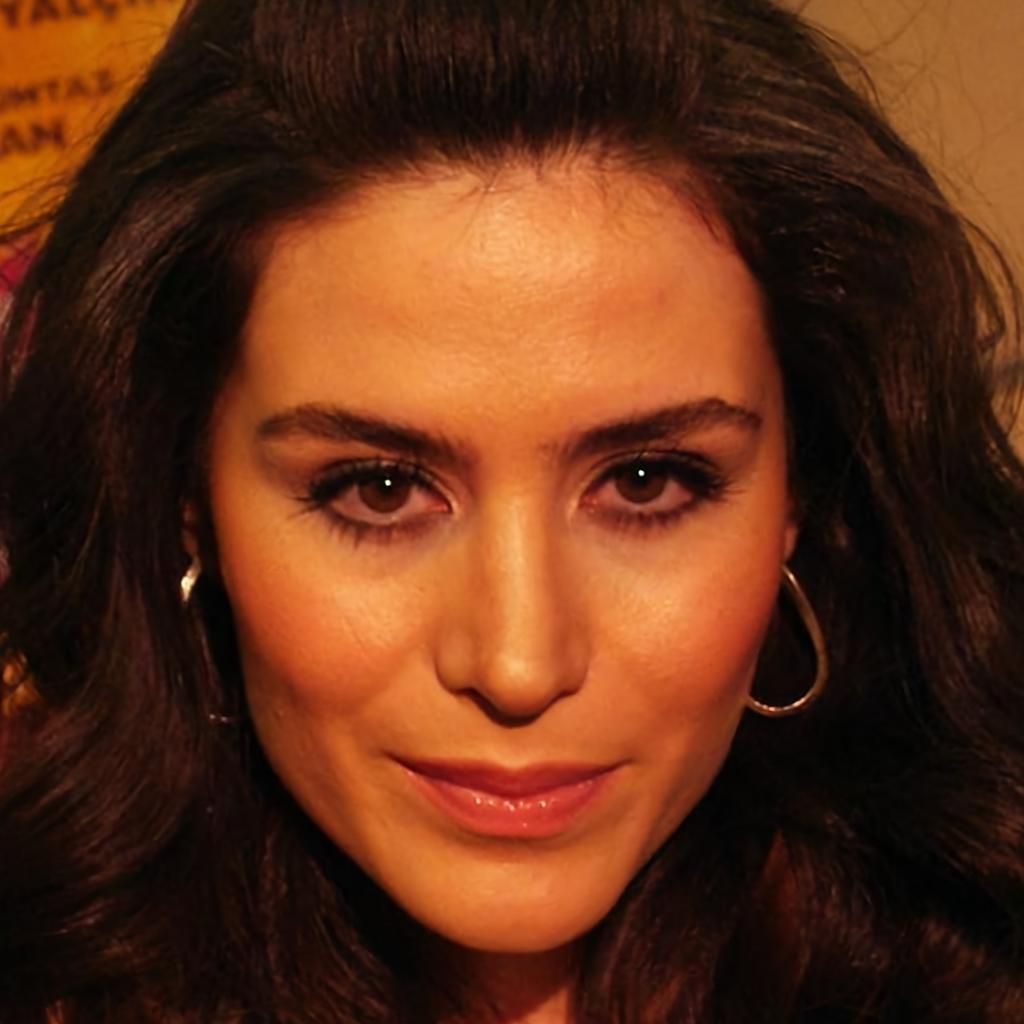}
     \includegraphics[width=\linewidth]{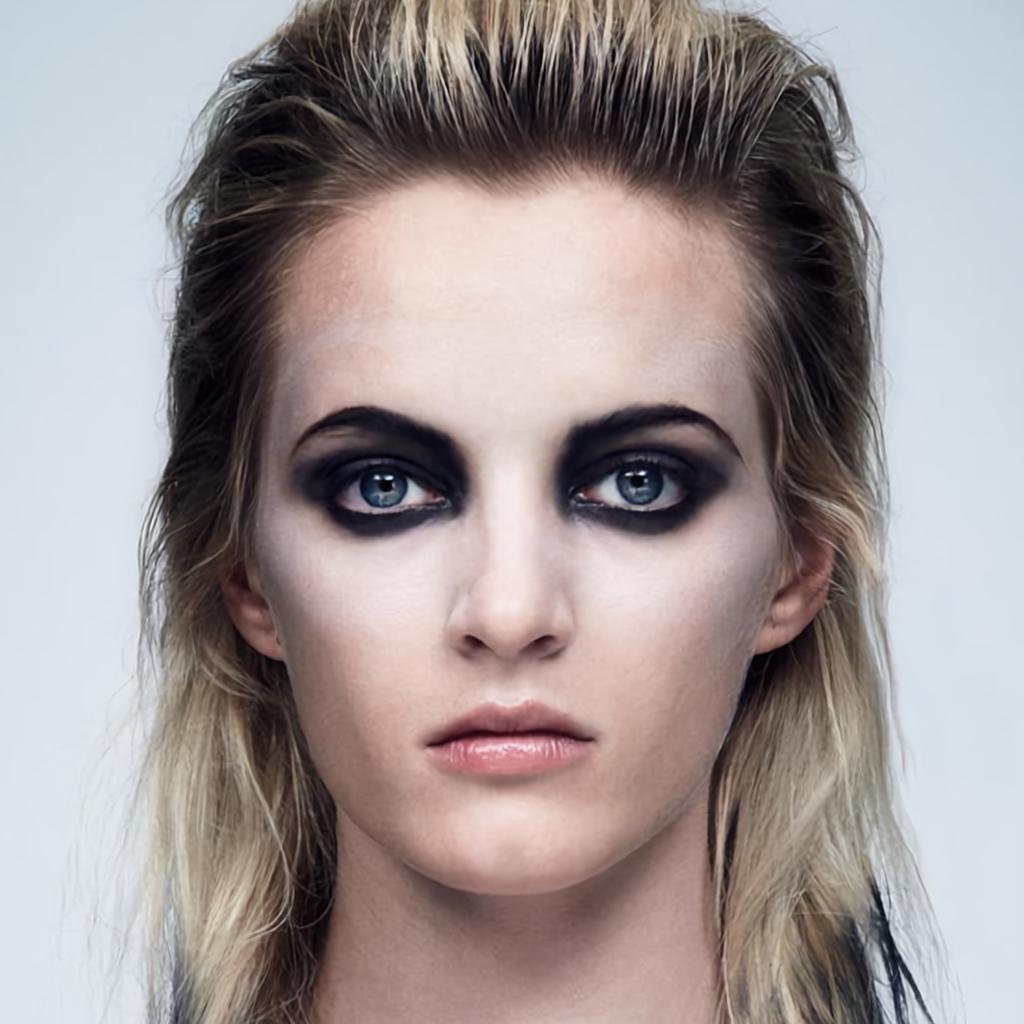}
     \caption{Source}
     \end{subfigure}
     % \hspace{-2mm}
     \begin{subfigure}{.2\linewidth}
     \centering
     \includegraphics[width=\linewidth]{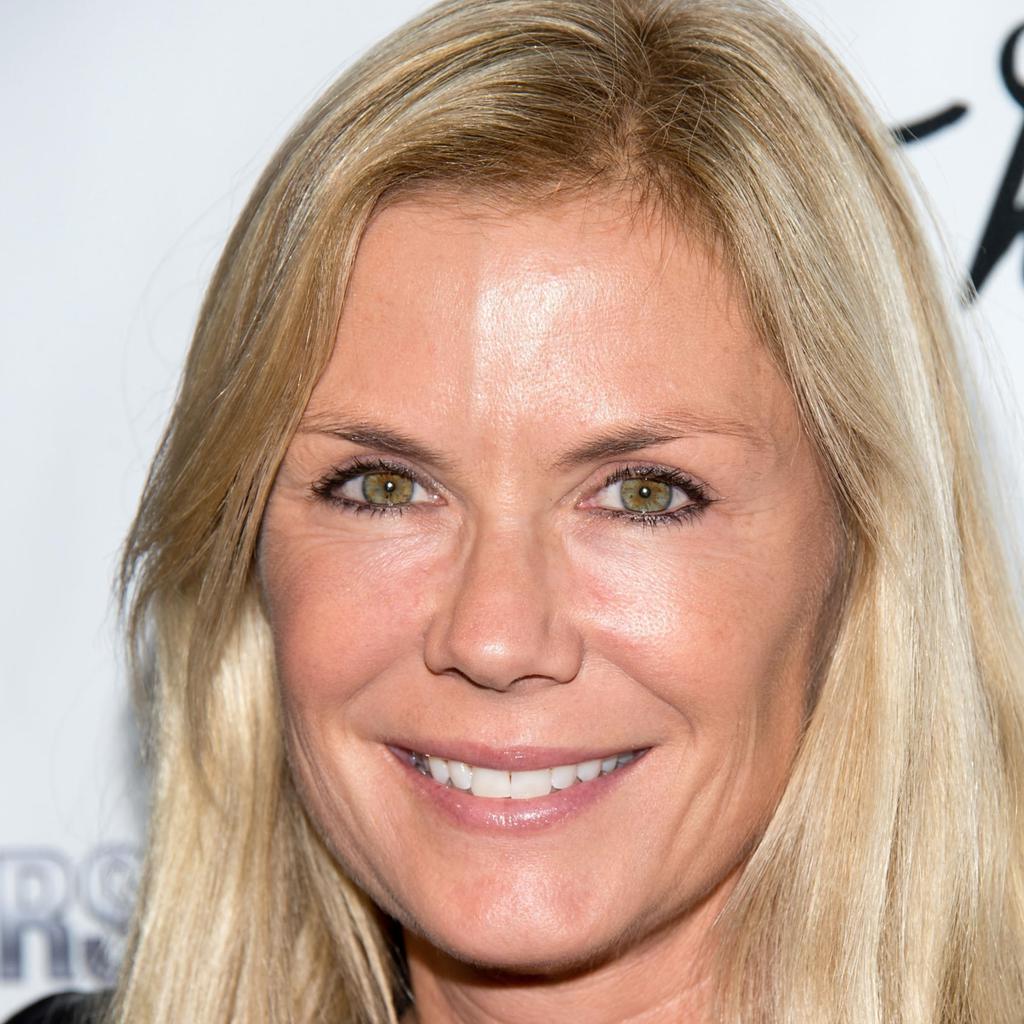}
     \includegraphics[width=\linewidth]{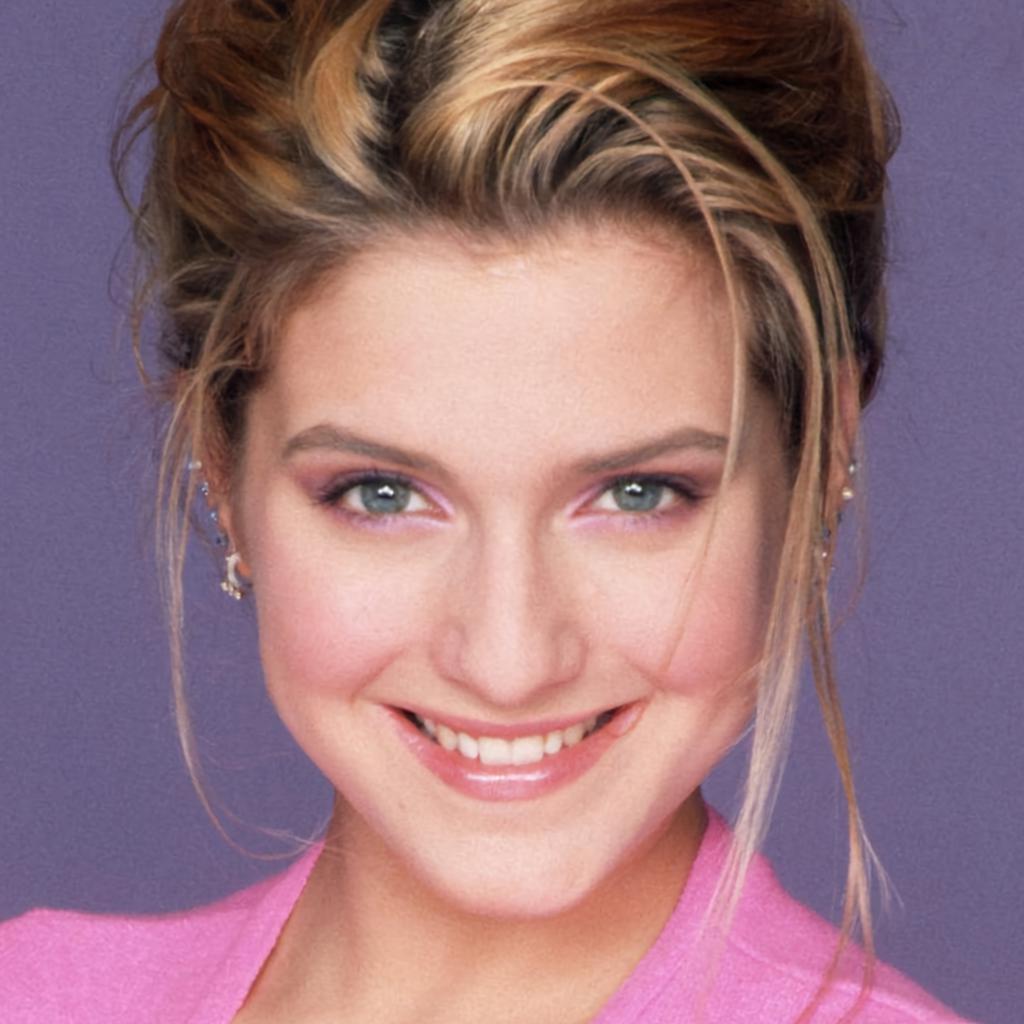}
     \includegraphics[width=\linewidth]{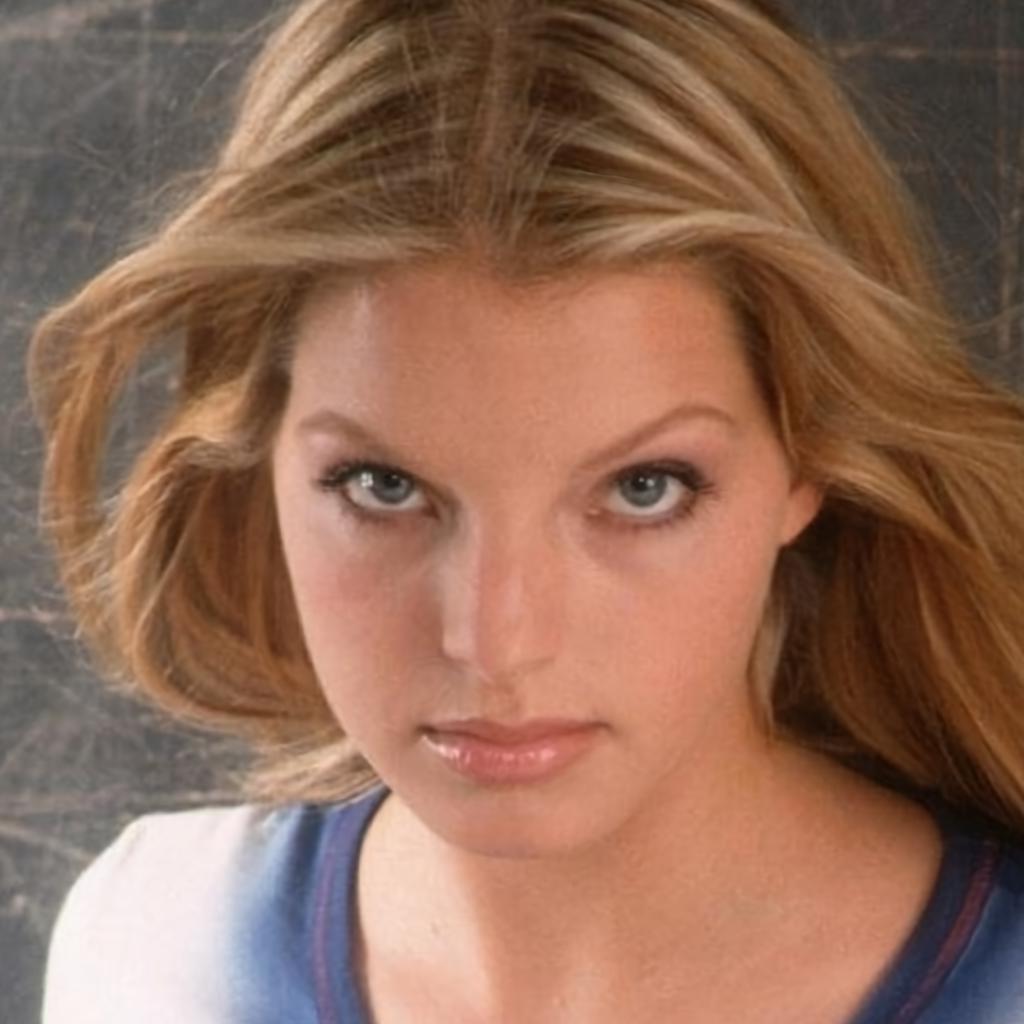}
     \includegraphics[width=\linewidth]{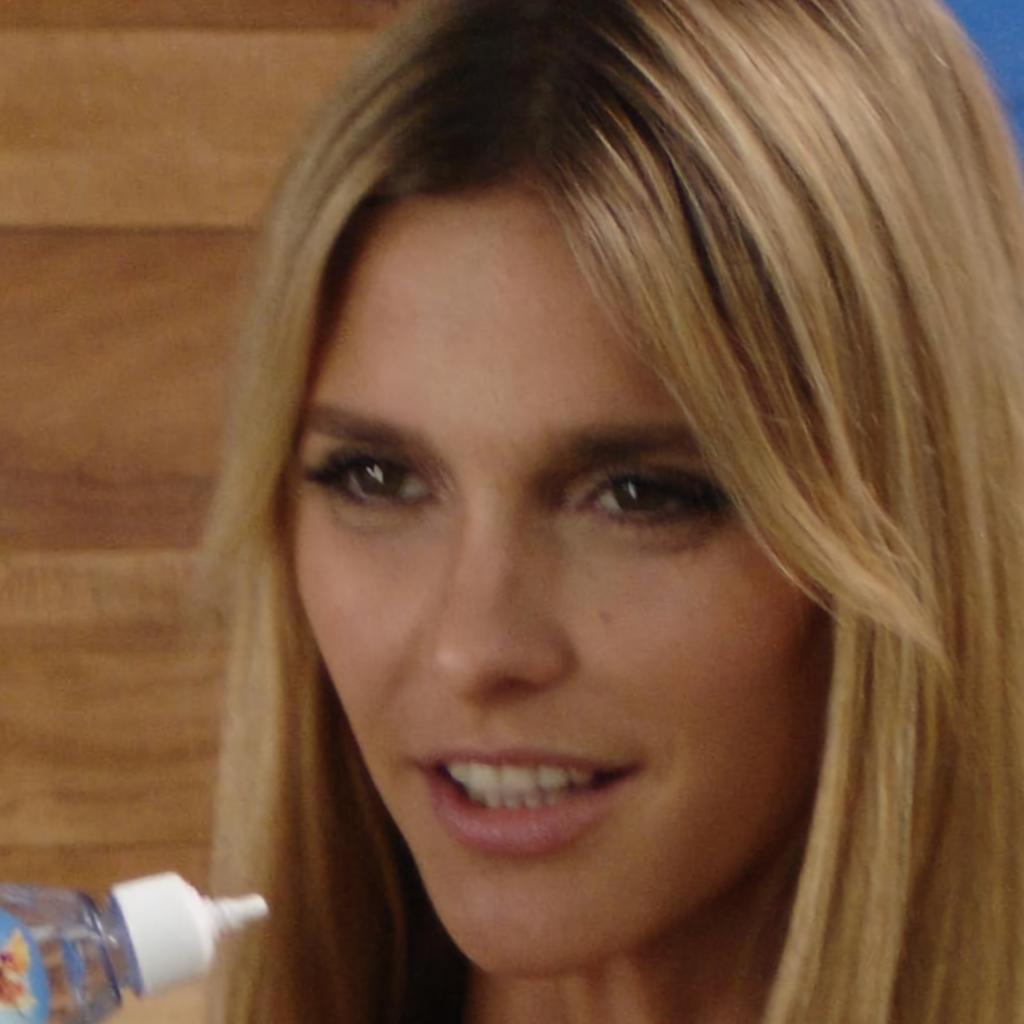}
     \includegraphics[width=\linewidth]{figure/celeba/original/27595}
     \includegraphics[width=\linewidth]{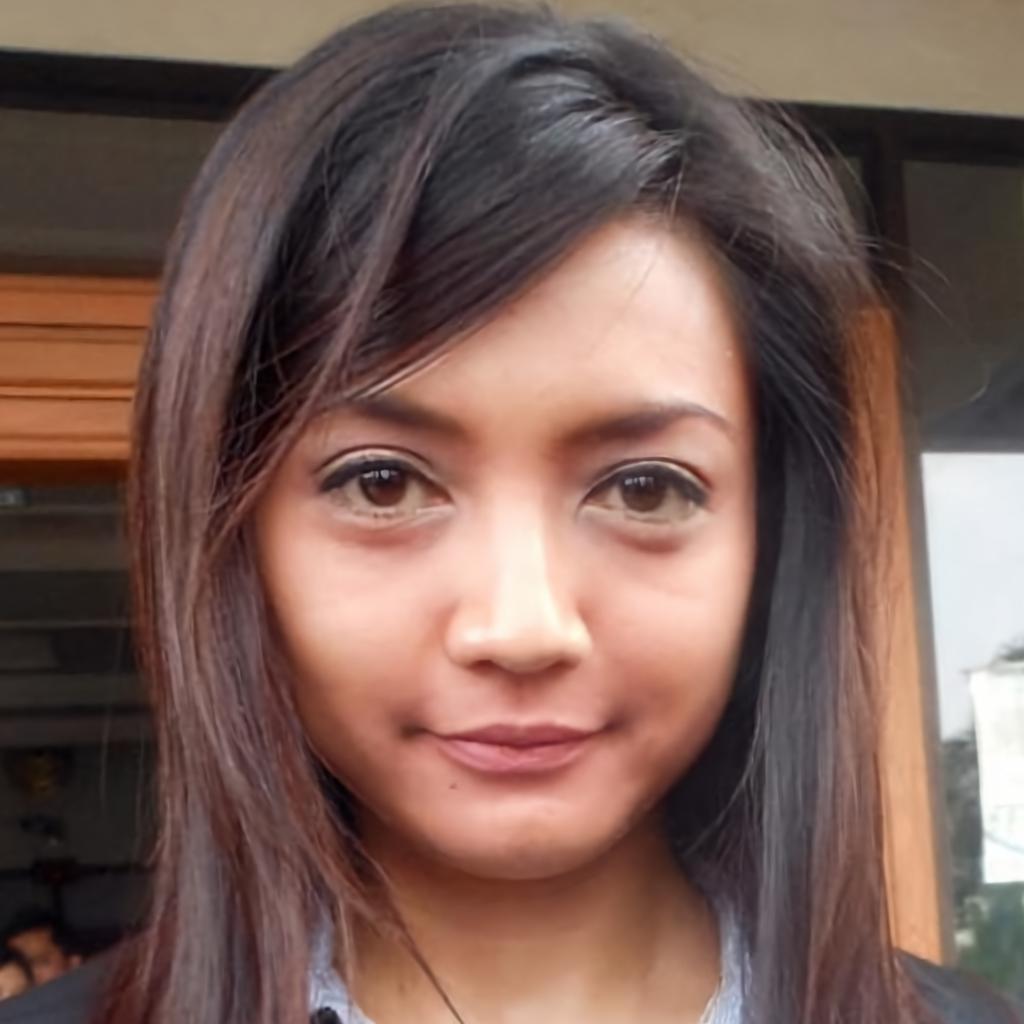}
     \caption{Target}
     \end{subfigure}
     % \hspace{-2mm}
     \begin{subfigure}{.2\linewidth}
     \centering
     \includegraphics[width=\linewidth]{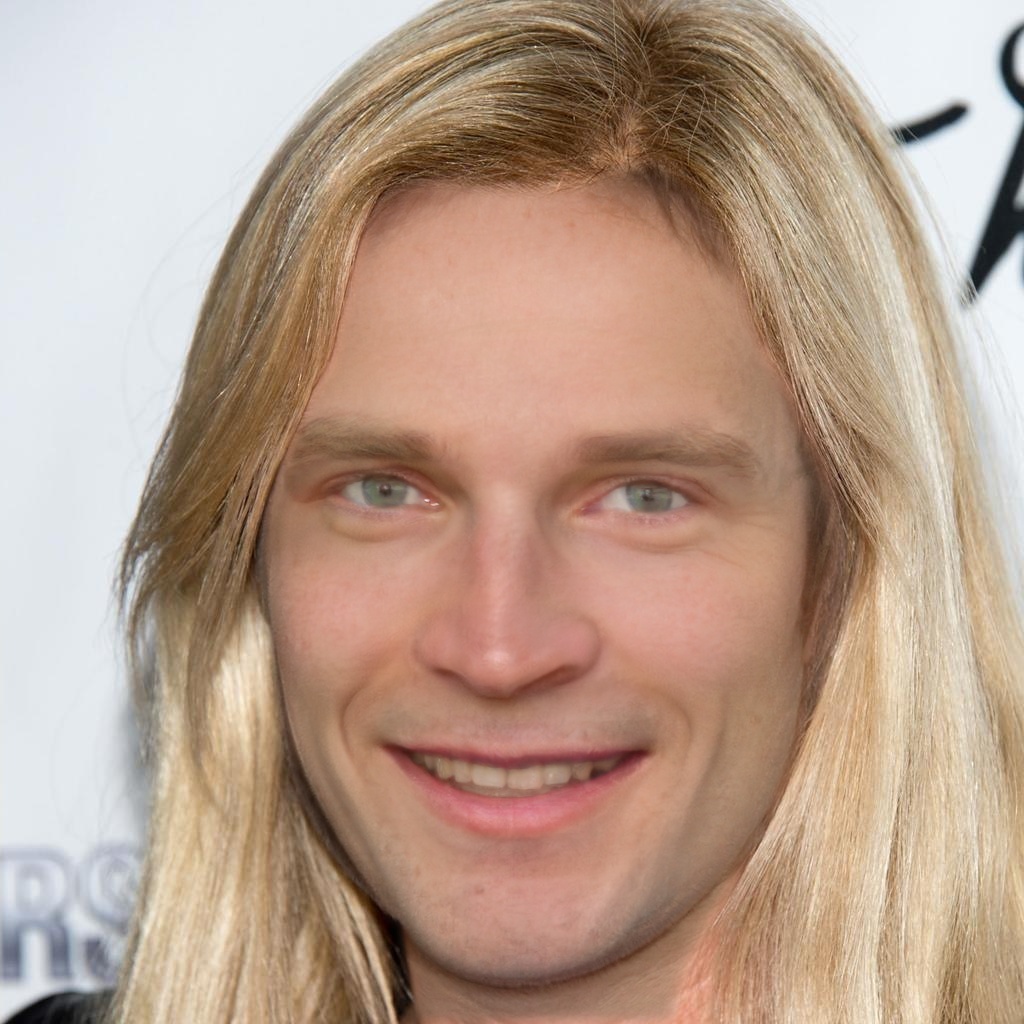}
     \includegraphics[width=\linewidth]{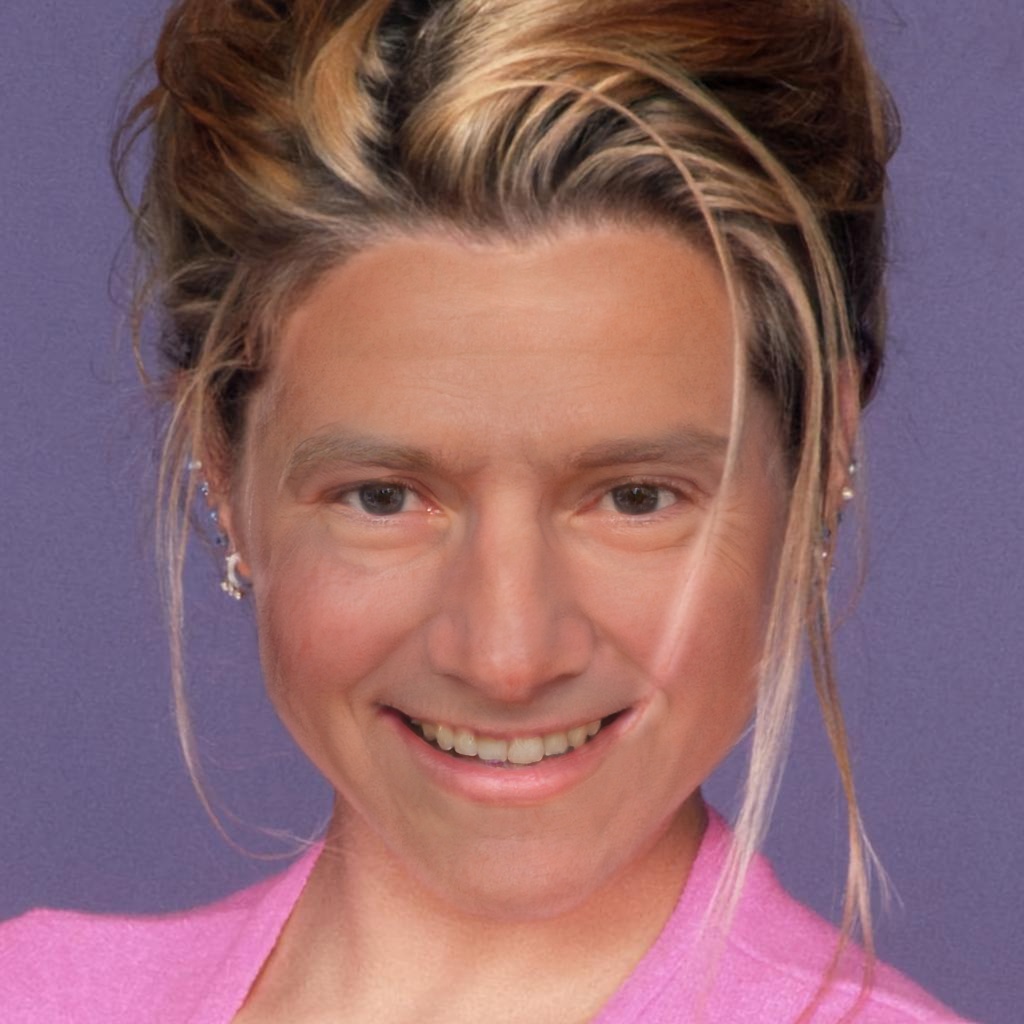}
     \includegraphics[width=\linewidth]{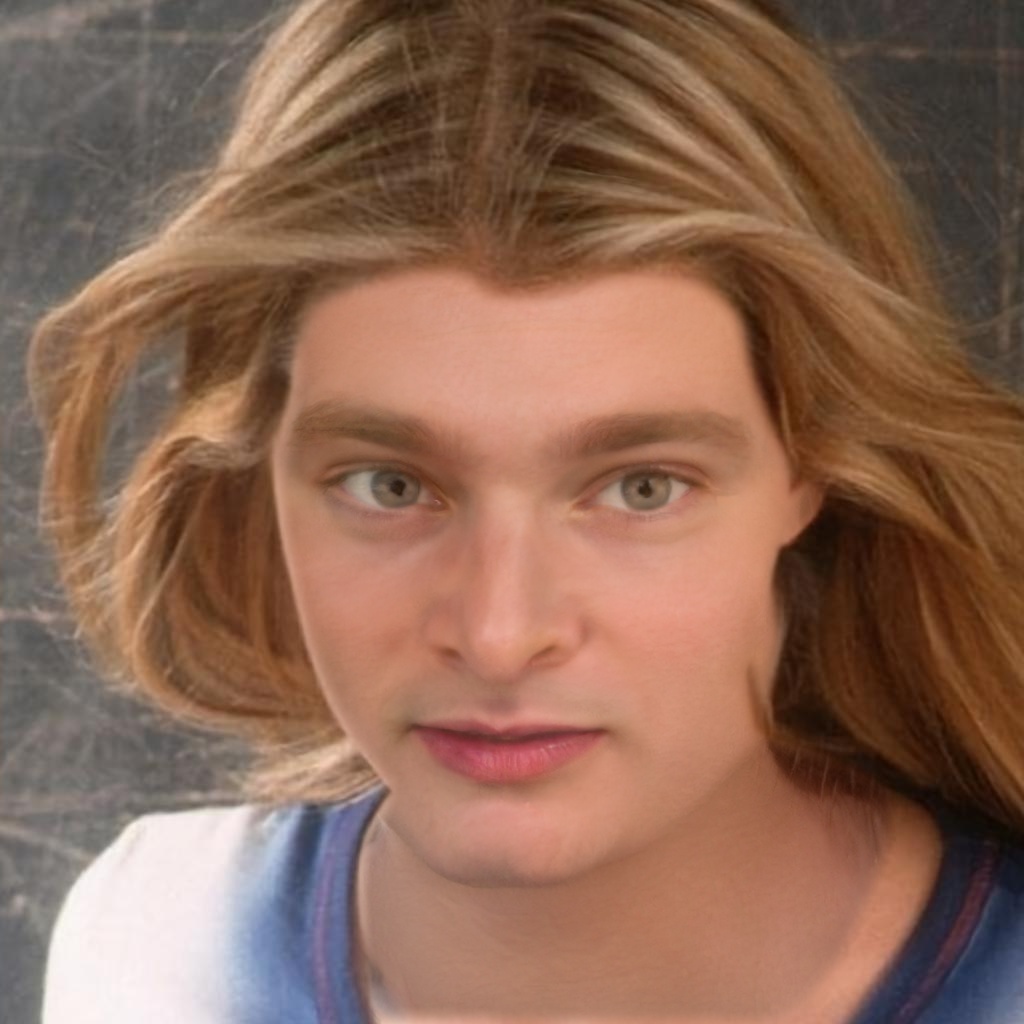}
     \includegraphics[width=\linewidth]{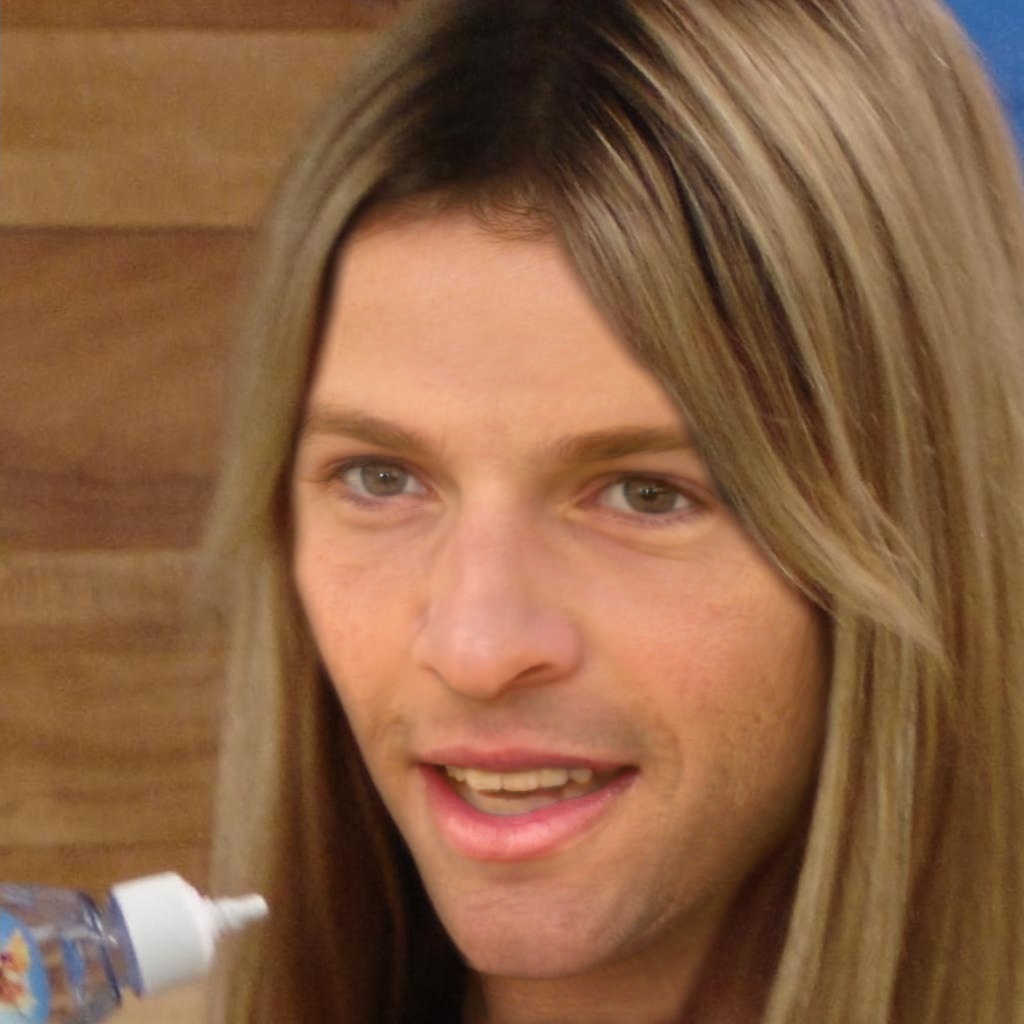}
     \includegraphics[width=\linewidth]{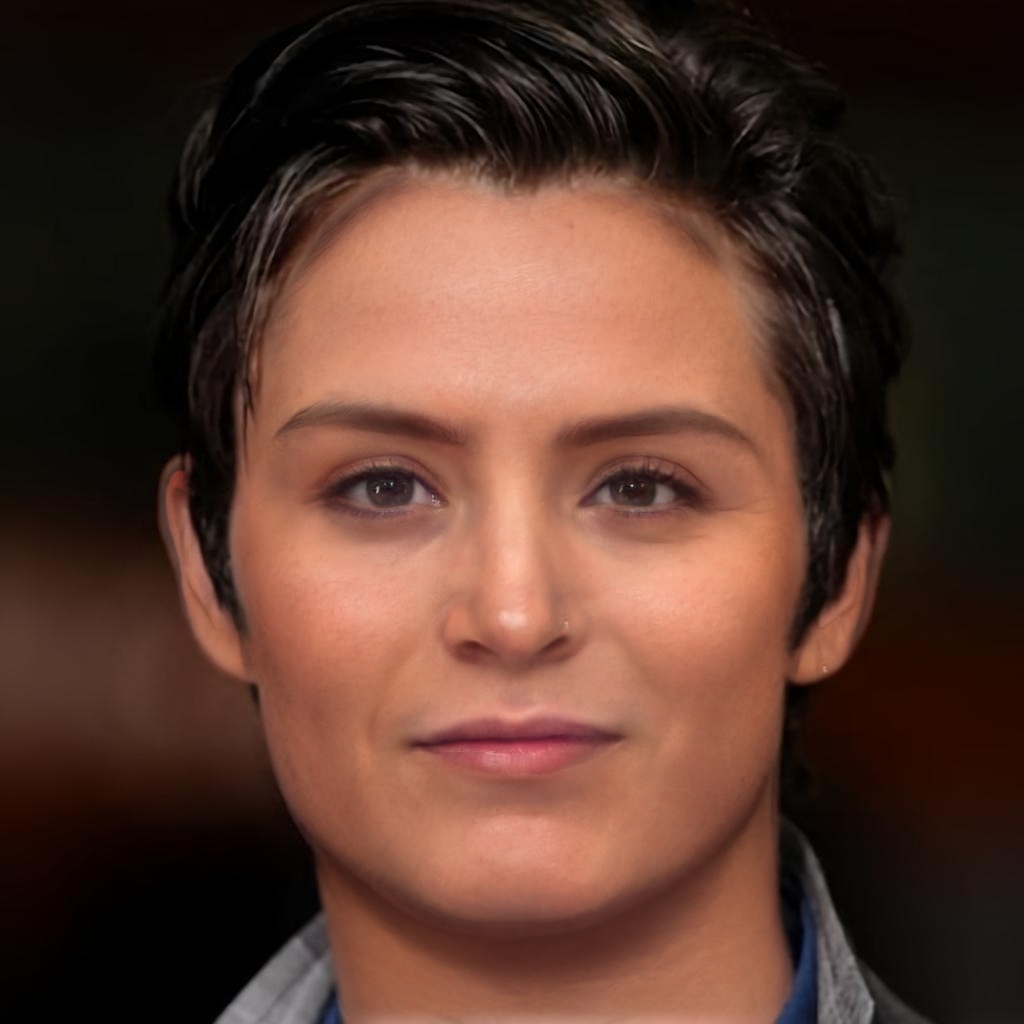}
     \includegraphics[width=\linewidth]{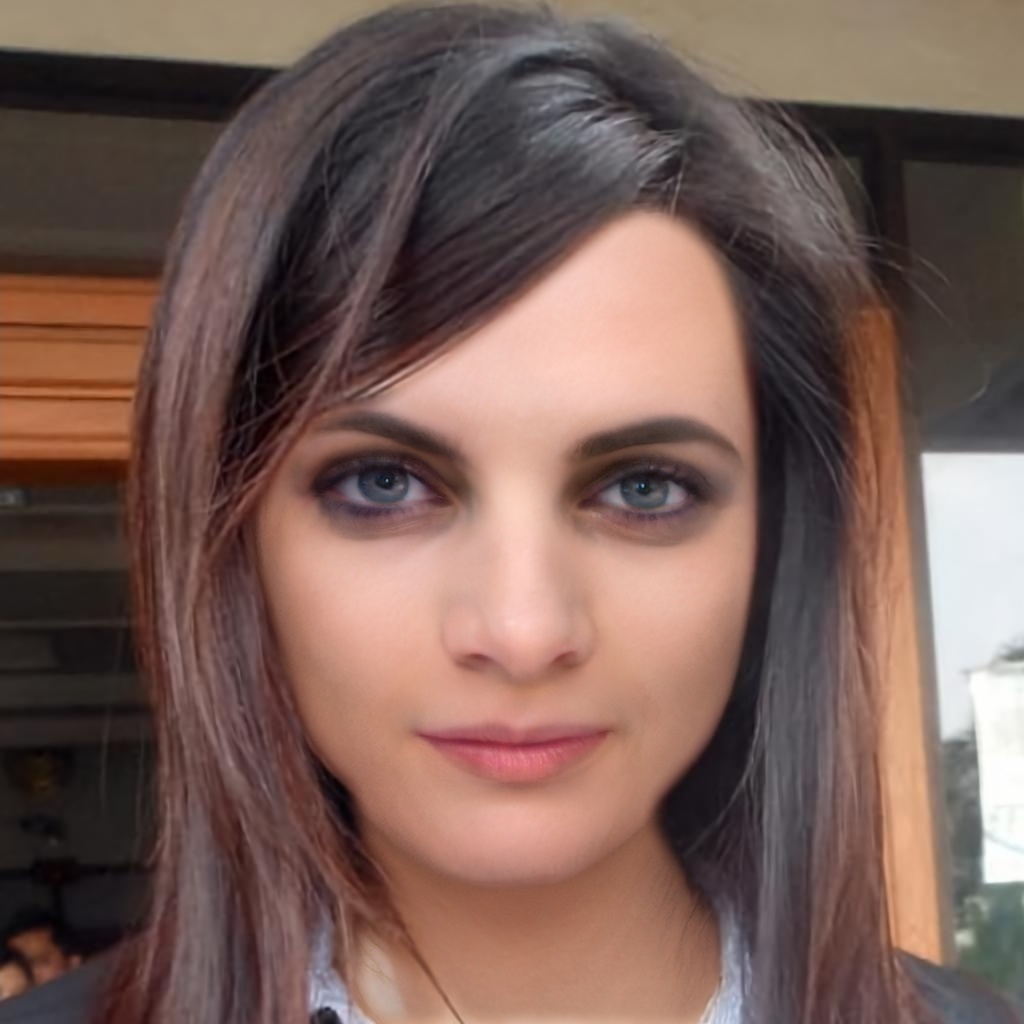}
     \caption{MegaFS}
     \end{subfigure}
     % \hspace{-2mm}
     \begin{subfigure}{.2\linewidth}
     \centering
     \includegraphics[width=\linewidth]{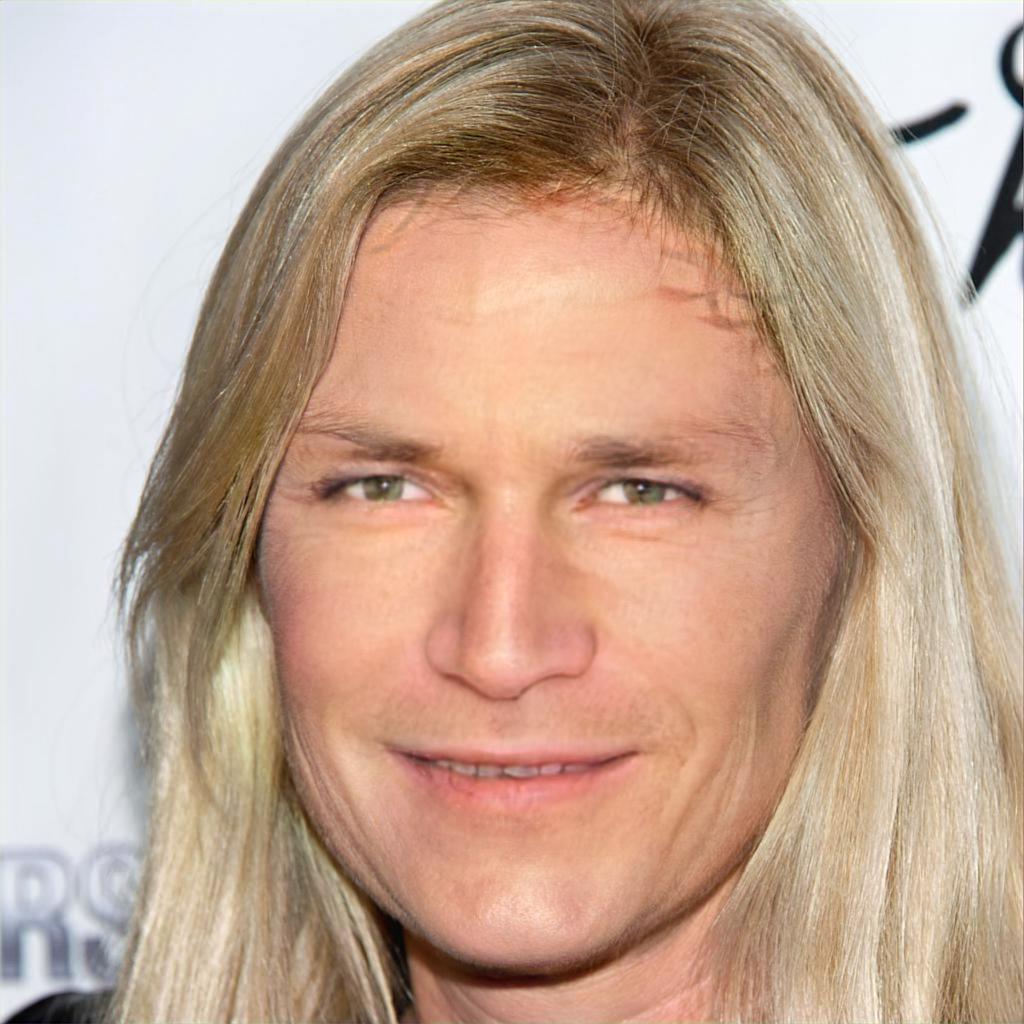}
     \includegraphics[width=\linewidth]{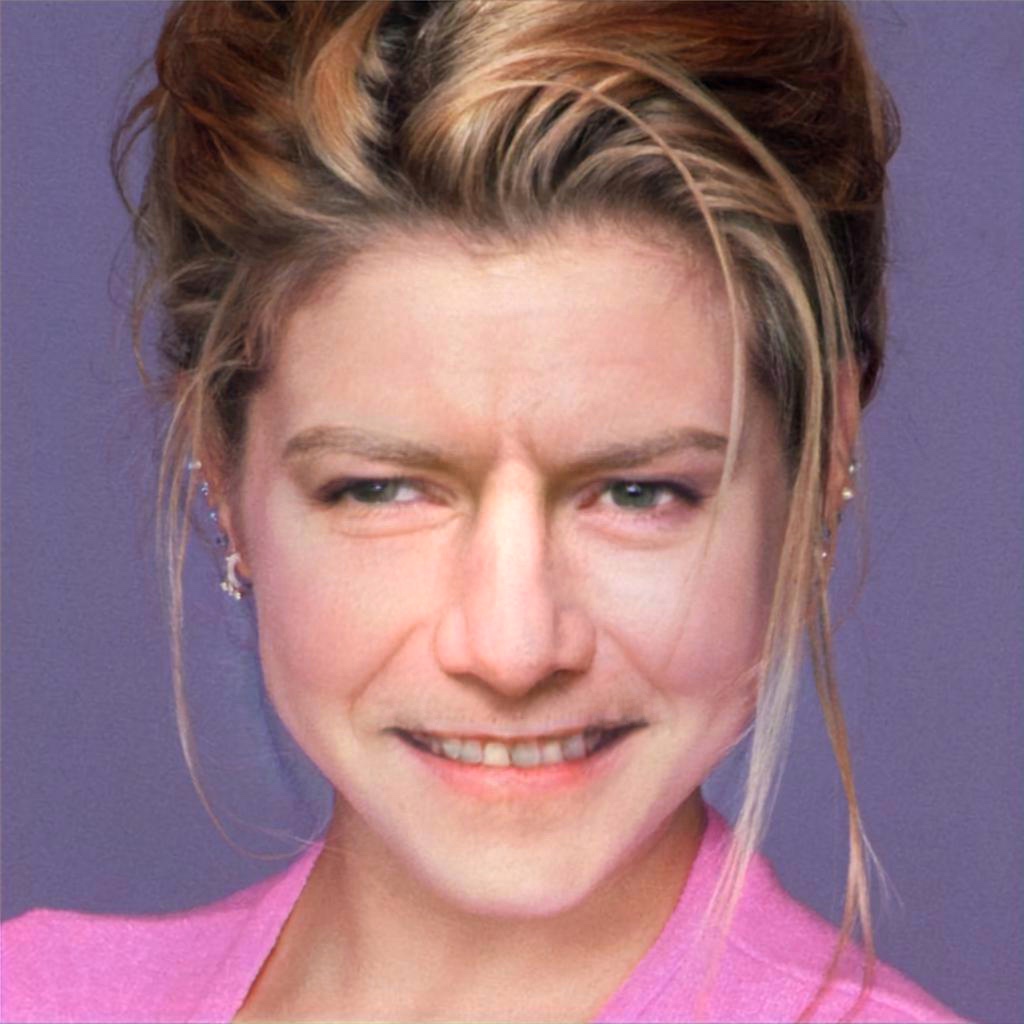}
     \includegraphics[width=\linewidth]{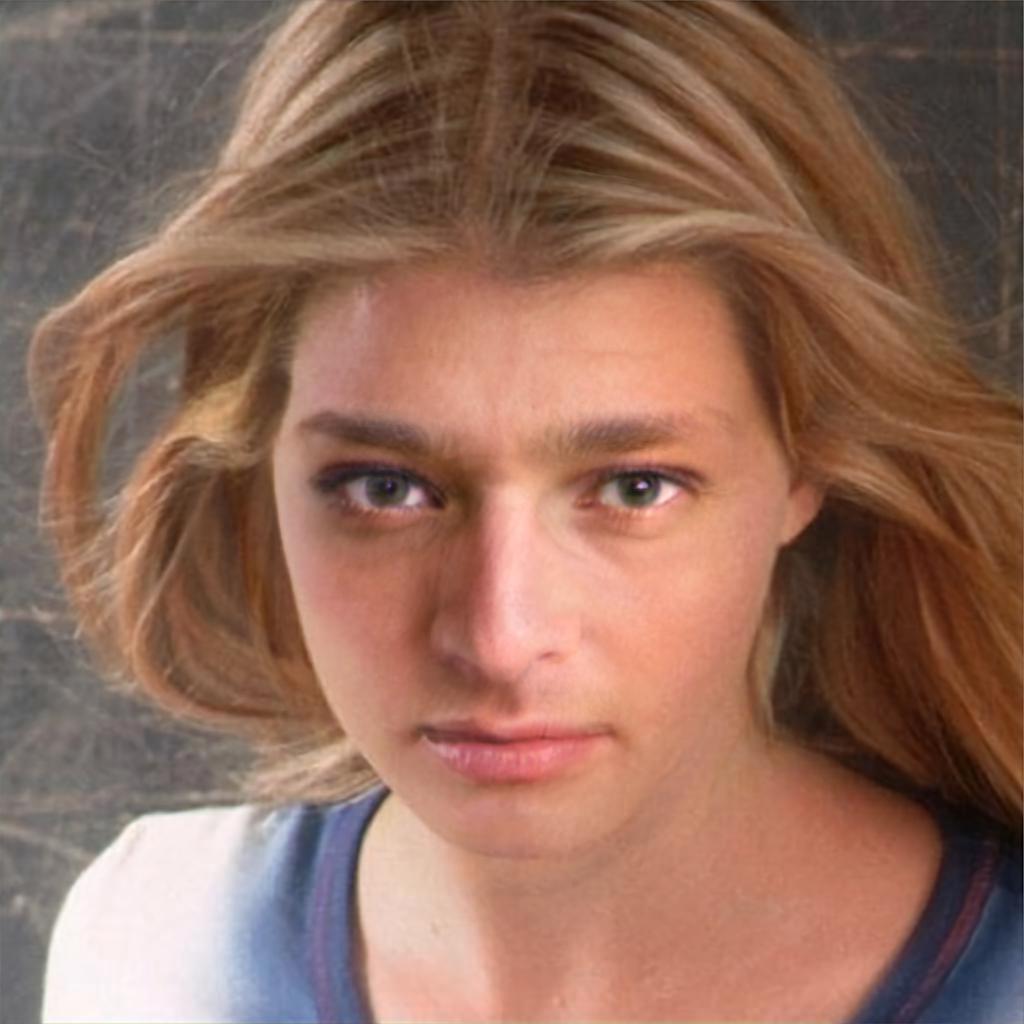}
     \includegraphics[width=\linewidth]{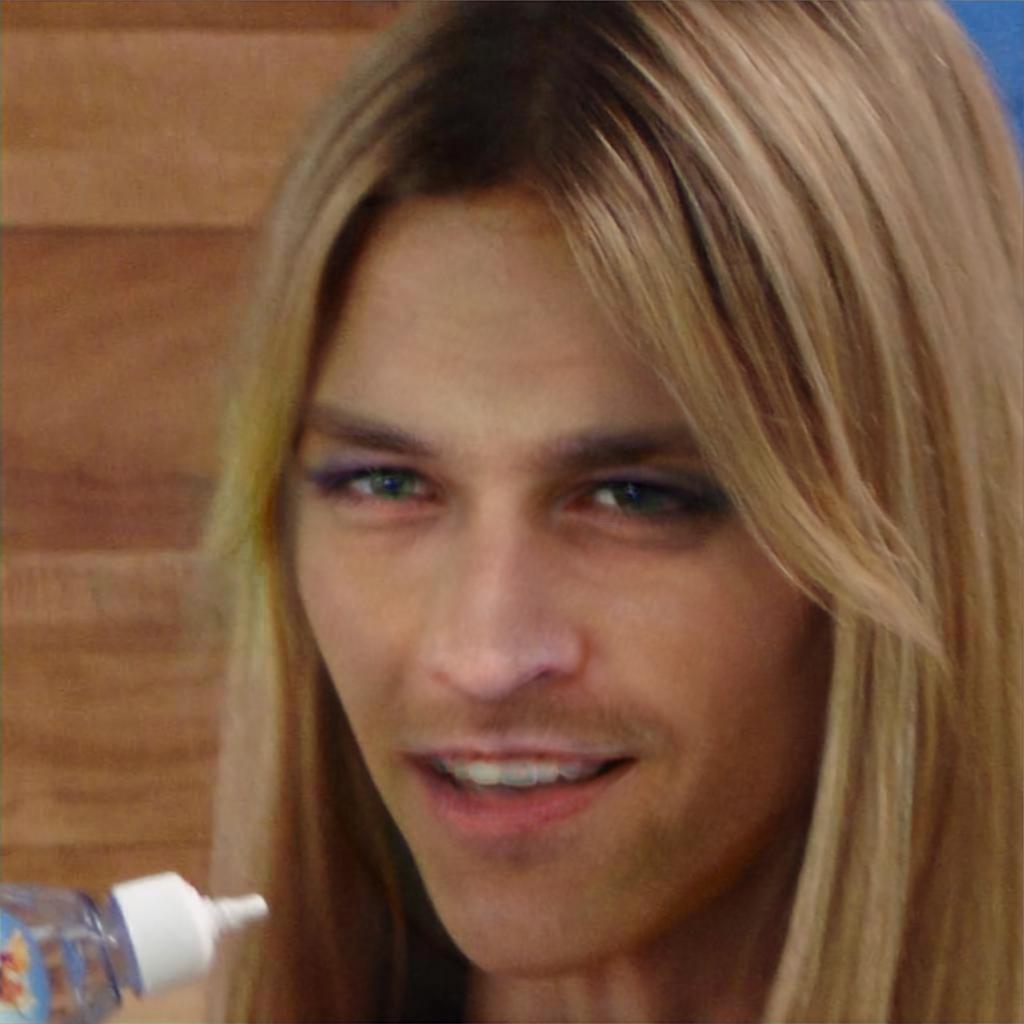}
     \includegraphics[width=\linewidth]{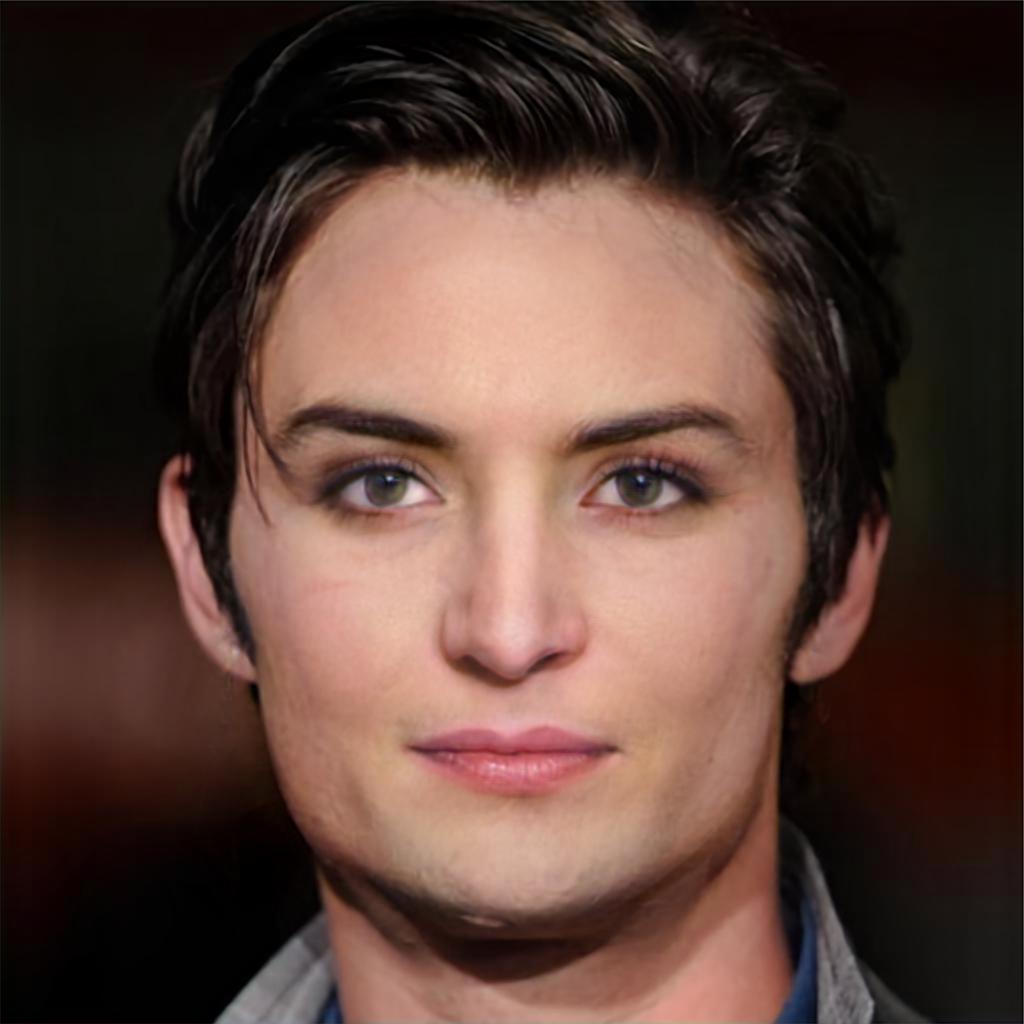}
     \includegraphics[width=\linewidth]{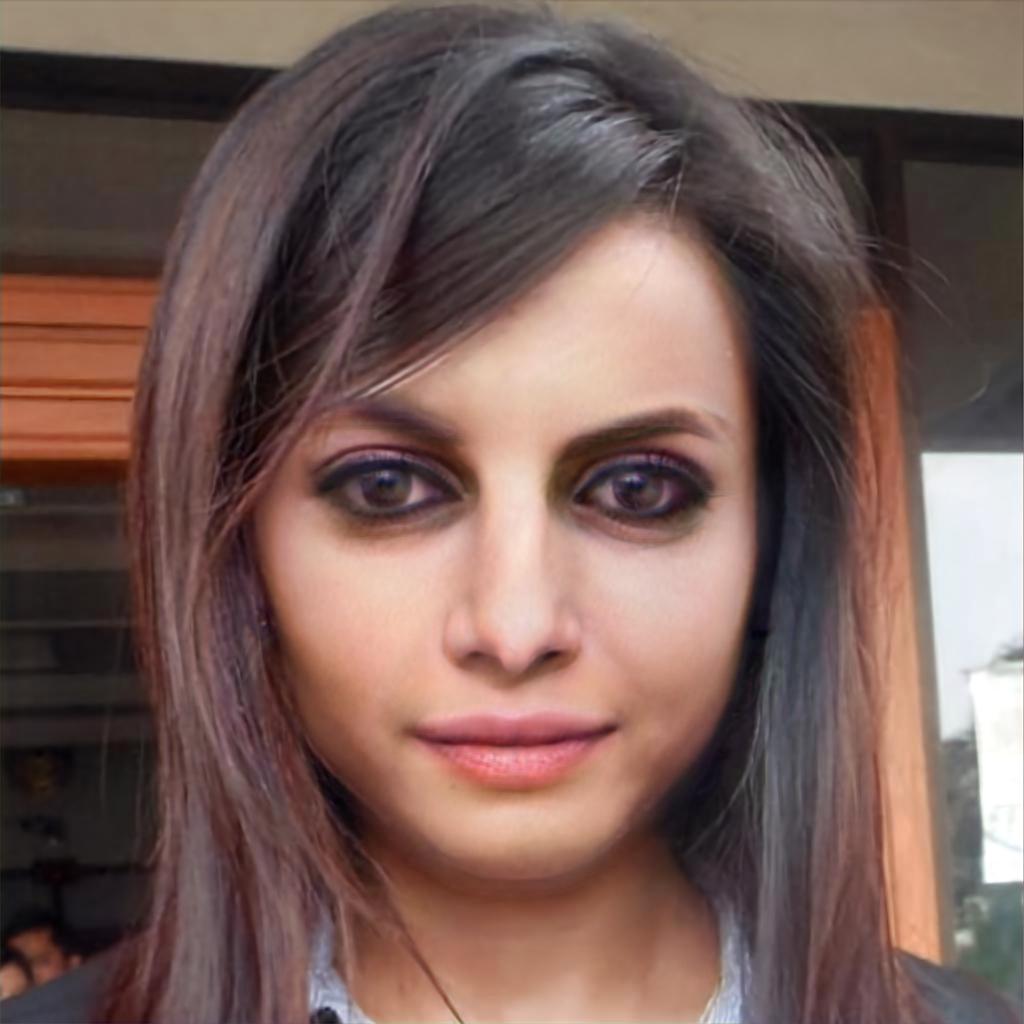}
     \caption{Ours}
     \end{subfigure}
     \caption{More qualitative comparison of face swapping on CelebA-HQ dataset.}
     \label{fig.supp}
\end{figure*}

In Fig.~\ref{fig.supp}, we present more of qualitative comparison on CelebA-HQ dataset with MegaFS~\cite{zhu2021one}, as shown in the main paper, our method can achieve favorably better face swapping results, it transfers the structure and appearance attributes as desired with identity-preserving.

% {\small
% \bibliographystyle{ieee_fullname}
% \bibliography{egbib}
% }

\end{document}